\documentclass[11pt, a4paper, twocolumn]{googledeepmind}
\pdftrailerid{redacted}

\makeatletter
\renewcommand\bibentry[1]{\nocite{#1}{\frenchspacing\@nameuse{BR@r@#1\@extra@b@citeb}}}
\makeatother

\usepackage{subcaption}
\usepackage{graphicx}
\usepackage{caption}
\usepackage{xcolor}
\usepackage[export]{adjustbox}
\usepackage{multirow}

\usepackage{kantlipsum, lipsum}
\usepackage{dsfont}
\usepackage[authoryear, sort&compress, round]{natbib}

\usepackage{xspace}
\usepackage{adjustbox}
\PassOptionsToPackage{hyphens}{url}\usepackage{hyperref}

\newcommand{\Vermeer}{{Vermeer}\xspace}
\newcommand{\SU}{{Shallow-UViT}\xspace}
\newcommand{\CC}{{\emph{core components}}\xspace}

\newcommand{\FID}{{FID}\xspace}

\newcommand{\Clip}{{CLIP$_{score}$}\xspace}
\newcommand{\PSDM}{{\emph{PSDM}}\xspace}

\title{Greedy Growing Enables High-Resolution Pixel-Based Diffusion Models}

 \author{Cristina N. Vasconcelos}
 \author{Abdullah Rashwan}
 \author{Austin Waters}
 \author{Trevor Walker}
 \author{Keyang Xu}
 \author{Jimmy Yan}
 \author{Rui Qian}
 \author{Shixin Luo}
 \author{Zarana Parekh}
 \author{Andrew Bunner}
 \author{Hongliang Fei}
 \author{Roopal Garg}
 \author{Mandy Guo}
 \author{Ivana Kajic}
 \author{Yeqing Li}
 \author{Henna Nandwani}
 \author{Jordi Pont-Tuset}
 \author{Yasumasa Onoe}
 \author{Sarah Rosston}
 \author{Su Wang}
 \author{Wenlei Zhou}
 \author{Kevin Swersky} 
 \author{David J. Fleet}
 \author{Jason M. Baldridge}
 \author{Oliver Wang}

\affil[1]{Google}

\begin{abstract}
\vspace*{-0.25cm}
We address the long-standing problem of how to learn effective pixel-based image diffusion models at scale, introducing a remarkably simple greedy growing method for stable training of large-scale, high-resolution models.
without the needs for cascaded super-resolution components.
The key insight stems from careful pre-training of core components, 
namely, those responsible for text-to-image alignment {\it vs.} high-resolution rendering.
We first demonstrate the benefits of scaling a {\it Shallow UNet},
with no down(up)-sampling enc(dec)oder.
Scaling its deep core layers is shown to improve alignment, object structure, and composition.
Building on this core model, we propose a greedy algorithm that grows the architecture into high-resolution end-to-end models, 
while preserving the integrity of the pre-trained representation, stabilizing training, and reducing the need for large high-resolution datasets.
This enables a single stage model capable of  generating high-resolution images without the need of a super-resolution cascade.
Our key results rely on public datasets and show that we are able to train non-cascaded models up to 8B parameters with no further regularization schemes.
Vermeer, our full pipeline model trained with internal datasets to produce $1024\times1024$ images, without cascades, is  preferred by $44.0\%$ {\it vs.} $21.4\%$ human evaluators over SDXL.
\end{abstract}

\begin{document}
\maketitle

% % Include paper content from external files
% \input{template_content}

\section{Introduction}
\label{sec:intro}

Training large-scale \emph{Pixel-Space text-to-image Diffusion Models} (\PSDM) to generate high-resolution images has been challenging due to optimization instabilities arising when growing model size and/or target image resolution,
and due to the increasing demand for computational resources and high resolution training corpora.
The predominant alternatives include \emph{cascaded models},
comprising a sequence of diffusion models each targeting a progressively higher resolution and trained independently \citep{Ho2022,saharia2022photorealistic,nichol2021glide}, and {\em latent diffusion models} (LDMs), where generation is performed in a low-dimensional latent representation, from which high resolution images are generated via a pre-trained latent decoder \citep{rombach2022high}.
% \df{more citations here?}

In the development of cascaded models, it is challenging to identify sources of quality degradation and distortion resulting from design decisions at specific stages of the model. 
One well-known issue of cascades is the distribution shift between training and inference, where  inputs to super-resolution or decoder models during training are obtained by down-sampling or encoding training images, but during inference they are generated from other models, and hence may deviate from the training  distribution.  This can cause amplification of unnatural distortions produced by models early in the cascade. 
The generation of realistic small objects such as faces or hands is one such challenge that has been difficult to diagnose in such models.
% \df{are there citations available here?}
% (e.g., small faces and hands), is a clear practical case whose progress is affected by the aforementioned attribution and distribution shifts problems.

% \df{why so many citations here vs elsewhere. Perhaps shorten citations to reduce emphasis on this point.}
Beyond image generation {\it per se}, diffusion models serve as image priors for myriad downstream tasks, including  inverse problems \citep{NEURIPS2021_7d6044e9,NEURIPS2021_6e289439,kawar2022denoising,song2023pseudoinverseguided,chung2023diffusion,graikos2022diffusion,tang2023emergent,jaini2023,zhan2023,song2024solving}, or other generative tasks \citep{ho2022video,controllable-music,poole2023dreamfusion,Tan2023,bartal2024lumiere, single-stage-diffusion, tewari2023diffusion}.
Cascaded diffusion models are not readily applicable to such tasks, and as a consequence, many such applications rely solely on the score function from the base model of a cascade, often at a relatively low resolution.
A high resolution end-to-end model would alleviate these issues, but model development and effective training procedures have been elusive.

Key barriers to training high resolution models include prohibitive resource requirements in both memory and computation.
Existent recipes require large batch sizes during training to avoid instabilities,
and as a consequence, intractably large amounts of memory for high-resolution images.
% given the dimensionality of model activations and their gradients.
% This also makes meaningful studies of different architectures and scaling laws, and the emergent properties of end-to-end models, prohibitive. %Kevin - I'm not sure this sentence adds much.
Another issue concerns the need for high quality, high resolution training data.  Existing training methods require large, diverse corpora of text-to-image pairs at the target resolution, while in practice, such data are not readily available at high resolution.

% \df{Before writing last couple of paragraphs, wait until results are concrete.}

This paper introduces a framework for training high resolution, large-scale text-to-image diffusion models without the use of cascades. 
% enables training high resolution large scale non-cascaded models.  
To that end we explore the extent to which one can decouple the training of 'visual concepts' associated with textual prompts, from the resolution at which one aims to render the image.
% In other words, we investigate whether \emph{is it feasible to split the training of the two tasks to some degree, while still targeting an end-to-end model at inference time?}
Such disentanglement has two goals. It aims at a better understanding of alignment, composition and image fidelity (especially for well-known hard cases like generating consistent hands, text rendering, scene composition, etc.) as a function of model scaling  (e.g., see \autoref{fig:core_qualitative}).
% , independent of myriad design choices in the composition of the UNet architecture. 
Second, and of equal importance, our framework yields a robust and stable recipe for training large-scale, non-cascaded pixel-based models targeting high-resolution generation. A bonus is that our recipe allows us to jointly train a single model with data comprising multiple resolutions, even if high-resolution text-image pairs are relatively scarce.
% while benefiting from the available image datasets at different resolutions with long tailed distributions.

The contributions of this paper can be summarized as follows:
\begin{itemize}
    % Its initial phase leverages large volumes of  text-image data at low image resolutions for robust pretraining layers identified as core for learning text-to-image alignment (\autoref{sec:method}).
    %   The algorithm then grows the model to higher resolutions from those core layers with considerably smaller batch sizes (as small as 256).
    % The algorithm then grows the model to produce higher resolutions images. The training at target resolution exhibits stable behaviour even at batch sizes (256 versus the typical 2k adopted in previous solutions).
    % ]Second, the final architecture seamlessly integrates the learned representation of those core layers, while the grown model training exhibits stable behaviour even with considerably smaller batch sizes (256 versus the typical 2k adopted in previous solutions).
    % As a consequence, one can stably train large-scale models with image resolutions as high as \df{XXX}.
    % high-resolution outputs.

    \item We introduce a novel architecture, \SU, 
    which allows one to pretrain  the \PSDM's core layers on huge datasets of text-image data (\autoref{sec:shallow}), eliminating the need to train at the entire model with high resolution images.
    % That is, it eliminates the constraint that such corpora must only contain images in the model's final resolution.
    %
    This also allows us to investigate the emergent properties of PSDM  representation scaling in isolation from layers targeting generation at the  final resolution.
    
    \item We present a \emph{greedy algorithm} for training the \SU architecture that allows us to successfully train a high-resolution text-to-image model with small batch sizes (256 versus the typical 2k used in end-to-end solutions) (\autoref{sec:method}).

    \item We show that one can significantly improve different image quality metrics by leveraging the representation pretrained at low-resolution, while growing model resolution in a greedy fashion.
    % on the full corpora into the final model using greedy growing.
    Scaling the core components of the \SU architecture alone leads to significant improvements in image distribution, quality and text alignment (\autoref{sec:experiments}).
    % across four broad semantic categories (entity, attribute, relation, and global) and 14 fine-grained subcategories (\autoref{sec:experiments}).
    
    % \item Aiming for replicability, our key results leverage a public dataset, CC12M \citep{changpinyo2021cc12m}, containing 12M text-to-image pairs, of which only 8M have an image resolution of 512 pixels or more
    % % when restricting to its subset with minimum resolution of 512 pixels
    % (\autoref{sec:exp_settings}). 
    
    % \item We propose the use of a closed interval for model's guidance weights tuning, that better correlates the trade-off between natural color, text consistency and with human perception of shapes and object's parts (\autoref{sec:guidance}).

    \item We demonstrate that these principles work at scale by presenting {\bf Vermeer}, a model trained with our greedy algorithm on large-scale corpora, in conjunction with other well-known methods like asymmetric aspect ratio finetuning, prompt preemption and style tuning (\autoref{sec:vermeer}).
    Vermeer is shown to surpass previous cascaded and auto-regressive models across different metrics. In a human evaluation study with 500 challenging prompts and 25 annotators per image, Vermeer is preferred over SDXL \citep{sdxl} by a 2 to 1 margin.

\end{itemize}

\medskip
\section{Related work}
\label{sec:related}

% \df{
% % Also see \cite{karras-NeurIPS22} and  \cite{NicolICML2021} for improved training.

% }

Current high-resolution image generation with diffusion models presents a trade-off between architectural complexity and efficiency. 
Cascaded diffusion models \citep{nichol2021glide,DMbeatGANs,Saharia2022,ramesh2022hierarchical,balaji2022ediffi} were originally introduced to circumvent the difficulty of training a single stage, end-to-end model.
Cascaded models employ a multi-stage architecture that progressively up-scales lower-resolution images to address the computational challenges of generating high-resolution images directly. 
Nevertheless, they entail significant complexity and training overhead, as the stages of the cascade are trained independently.

Simple Diffusion
\citep{Hoogeboom23} sought to simplify the process by targeting the high resolution generation with a single stage model, introducing a novel UViT architecture and several useful modifications to training methods that improve stability. 
While this approach is shown to be effective, stability issues remain when  targeting large-scale models, and high resolution images, due in part to their dependence on large batch sizes.
In this work we adopt a similar UViT architecture, and some of their techniques for scaling, extending the model to much higher resolutions through greedy training. Through scaling the core backbone of the model, and with our greedy training procedure, we find with can scale to much high resolution models
($2\times$ to $8\times$ higher than Simple Diffusion), with excellent alignment, and much smaller batches when training high resolution layers of the model.

Another line of work proposed Matryoshka Diffusion Models (MDM) \citep{gu2023matryoshka} that denoises multiple resolutions using a proposed Nested UNet architecture. They progressively train the network to preserve the representation at higher resolutions. We show in this work an alternate and simpler approach where denoising multiple resolutions is not required, but instead it is crucial to preserve the representation by freezing the pretrained weights as we grow the architecture up to its final design.

On another front, latent diffusion models (LDMs) \citep{rombach2022high,jabri2022scalable,betker2023improving} reduce computational costs by operating within a compressed latent representation. However, LDMs still require separate super-resolution or latent decoder networks to produce final high-resolution images.
% This introduces an additional layer in the generation pipeline.

The model we introduce also resembles progressive GAN training \citep{karras2018progressive} in which layers of increasing resolution are added at each stage. 
Our work can be thought of as an extension of progressive growing for diffusion models, where we evaluate different growing configurations, and come up with a two-step recipe that arrives at a good trade-off of training efficiency, robustness, and generation quality.  Specifically, while all layers remain trainable in progressive GANs, and a sequence of growing operations is performed before reaching the final architecture, we pretrain a core representation that remains frozen when training all grown layers at once up to the target resolution. We find that this is crucial to preserve the quality of the representation learned at lower resolutions. 
% While their proposed technique requires multiple training stages, we point that at each stage there is a risk of loss of representation if large number of layers are introduced. 
% To overcome these challenges, we use a shallow encoder/decoder network, while greedily growing the network once at target resolution, and to avoid loss of representation, we fix the pretraining weights of the transformer blocks.

\section{Method}
\label{sec:method}

Our goal is to create a straightforward, stable methodology for training large scale pixel-space diffusion models that operate as a single stage model, i.e., non-cascaded, at inference time.
To this end, we first revisit the UNet architecture, aiming to decouple layers that have a major impact on text-to-image alignment (\emph{core components}) from those responsible for rendering at the target image resolution (\emph{encoder-decoder} or \emph{super-resolution components}).
% (\autoref{sec:core}).
%
Next, we focus on pre-training  the  core components pretraining and on 
% supporting the study of its 
representation scaling (\autoref{sec:shallow}).
Finally, we present a greedy algorithm to grow the initial architecture core by adding encoder-decoder layers while protecting core layers' representation.
This yields a single-stage model at inference time (\autoref{sec:greedy}).

\subsection{Text-to-image core components}
\label{sec:core}
UNet is the architecture of choice for diffusion models. Two  architecture families are common. In one, convolutional networks comprise a stack of convolutional blocks alternated with pooling or downsampling layers in the encoder, and upsampling layers in the decoder. More recently, the UViT family emerged~\citep{Emiel2023}, in which convolutional blocks are used at the higher layers of the encoder and decoder but augmented with transformer layers at the bottom of the UNet.  
In both architectural families, text conditioning is accomplished via cross-attention layers, also at the bottom, low-resolution layers of the UNet.
In doing so, these layers are responsible for conditioning the models' deepest representation on the textual and/or multi-modal inputs.
At these low-resolution layers, the text conditioning signal is able to influence the global image composition while the computational cost of attention is kept relatively low.

Our search for a  methodology that allows stable training of large models starts by identifying and isolating \emph{core layers} responsible for text-to-image alignment.
Our main conjecture is that it is possible to reduce the instability typically observed during training  large-scale PSDMs by warming up layers responsible for text-to-image alignment in isolation from layers responsible for target resolution encoding/decoding.

Specifically, we define the \CC as those that directly interface with text conditioning signals and those that are crucial in the diffusion process. They can be described as:  
\begin{itemize}
\setlength\itemsep{0em}
    \item {\em Text encoding layers}  combine one or more textual, character, and/or multimodal pretrained representations (such as those from \cite{t5_embs, byt5_embs,Char_embs,clip_emb}), and project them into the embedding space of the UNet. Typically composed of MLP on top of pooling layers.

    \item {\em Core representation layers} comprise hidden layers in the main backbone interfacing with cross-attention layers. They include the bottom layers of the UNet architecture whose features are directly combined with the embedded text by the cross attention operation and layers between them. 
    
    \item {\em Time encoding layers} map the diffusion time step into the model's embedding space. 
    Typically designed as a sinusoidal positional encoder, followed by a shallow MLP. Despite not participating directly in the cross-attention operation, it is a core component of the diffusion process.
\end{itemize}

We isolate these core components of a \PSDM text-to-image model in order to study their effect on the final model's properties. Next, we propose an architecture that enables the pretraining of these layers, and also supports the study of the properties emerging from scaling them. 

\subsection{\SU} 
\label{sec:shallow}
\begin{figure*}[hbt!]
% \vskip 0.2in
\begin{center}
\centerline{\includegraphics[width=0.65\linewidth]{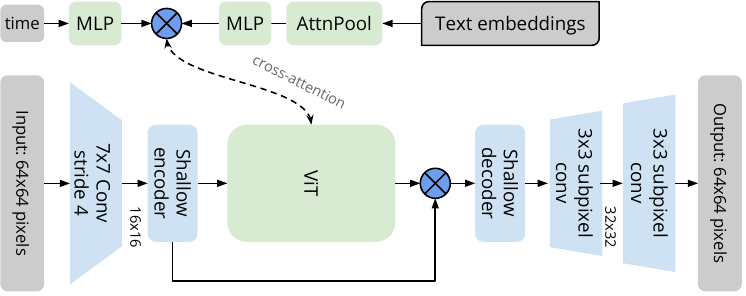}}
\caption{\SU~architecture: The input image grid is quickly reduced at the entry convolution, while a single residual block with no subsampling layers is used as a shallow encoder and decoder. The layers within the \CC  (in light green) are reused in the final end-to-end architecture, increasing its training stability, while remaining layers are discarded.}
\label{fig:shallow}
\end{center}
\vskip -0.2in
\end{figure*}

% We employ a few simplifications about the final model in order to assist the pretraining of its \CC and, at the same time investigate the emerging properties from their scaling.
% %
% Firstly, we isolate the \CC training and scaling from the confounding factors emerging from the specification of the UNet's encoder-decoder layers. 
%
To assist the pretraining of the \CC and, at the same time investigate the emerging properties from their scaling, we isolate the \CC training and scaling from other confounding factors in the specification of the UNet's encoder-decoder layers. 
To that end, we simplify the UNet’s conventional hierarchical structure, which operates on multiple resolutions, and define the \SU (SU), a simplified architecture comprising a shallow encoder and decoder operating on a fixed spatial grid  (\autoref{fig:shallow}). Its encoder and decoder have a single residual block each, containing two layers of $3 \times 3$ convolutions with swish activations~\cite{ramachandran2017searching}, and no upsampling or downsampling layers. 
As a result, they share the same spatial grid as the \emph{core representation layers} at the bottom.
The first convolutional layer at the entry of the architecture projects the input image into the fixed size grid used by its core layers. 
A corresponding upsampling head at the model's output reverses this operation.
These input/output layers facilitate quickly projecting input images with larger resolution into the core representation with fixed and lower resolution.
% Note that there are no skip connection between the STEM and output layers, differentiating this shallow architecture from typical encoder-decoder structures.

As a second simplification, we restrict our investigation to the \CC from the UViT model family owing to the uniform structure of its \emph{core representation layers}.
In contrast, the corresponding layers of convolutional UNets present a broader spectrum of
design and hyperparameter choices, 
owing to their non-uniform yet hierarchical structure,
rendering their analysis more complex.
%
%design choices owing to their non-uniform yet hierarchical structure, rendering their analysis more complex.

% As a third simplification of our study, we keep the \SU design fixed other than changing its capacity uniformly by increasing its hidden size and \emph{core representation layers} as described in \autoref{tab:shallow_params}.
% %
% We stress that we do not claim that the architecture adopted or that the hyper-parameter selection in our study is optimal. It is widely recognized for instance
% that larger text encoders increase the generated images quality \citep{Saharia2022,balaji2022ediffi,sdxl}.
% %
% Investigating the optimal design of each core component is beyond the scope of this work. Instead, we focus on analyzing their training interdependencies from the components that will constitute the final model.
%
% Alternatively to the \SU proposed, is to train the \CC as directly as a VIT, as previous explored in latent diffusion models \citep{DIT2023}. 

An alternative to the proposed use of the \SU architecture,
might be to train the \CC directly as an augmented ViT, 
as previously explored in latent diffusion models \citep{DIT2023}.
Our attempt to explore this approach proved not to be straightforward.
A crucial difference between PSDM and LDM becomes highly relevant here.
In the case of LDM, the transformer operates on latent tokens, and the diffusion model captures the latent token distribution.
Our task, on the other hand, is to pretrain a rich representation directly from the raw pixels, for subsequent reuse as deep features within a higher-resolution pixel-space model.
We conjecture that in such approaches the initial layers that are closer to the raw data do not transfer as well when reused within the final model. 

Instead, our \SU includes proxy additional layers that help with closing the gap between \CC feature pretraining and their later use. That is, the auxiliary, yet shallow, input (output) and encoding (decoding) layers help adding expressiveness to the transformations between the input (output) and the models' hidden representation.
Across the variations explored, the input convolution expands the number input channels up to 256 (we observed no improvement with more channels).

Beyond ablations on scaling (see  \autoref{sec:experiments}), we also found that certain variations for the \SU composition tend to degrade performance in comparison to our best architecture.
In particular, these include the removal of the shallow encoder/decoder blocks;
the use of smaller/larger filters ($4\times4, 5\times5, .., 9\times9$) and strides (from 1 up to 8) at the entry convolution;  
and the use of a single output head with a subpixel convolution upsampling by a factor of $4$. 
We also experimented with convolutional \emph{core representation layers}, but like \cite{dosovitskiy2021an}, we find they under-perform their transformed-based counterparts. 

\subsection{Greedy growing}
\label{sec:greedy}

Here we describe a greedy approach to learn \PSDM{s} for high-resolution images. Our process consists of two distinct stages, where we first pretrain the \emph{core representation layers} at a low resolution using a \SU architecture. Then, in the second phase, we replace the encoder/decoder layers with a more expressive set of UNet layers and train at the target resolution. This two-stage process is in contrast to progressive growing, which seeks to add one layer at a time. With this approach, we aim to mitigate the well-known instabilities observed during training of large models \citep{Saharia2022,Hoogeboom23}, while making the best use of the available training corpora.

% In order to learn high resolution pixel-space models on top of the core \SU layers, we propose a greedy procedure.

% We now concentrate on the task of building a pixel-space large-scale non-cascaded model targeting the generation of high-resolution images from text. 
% The \emph{greedy growing} algorithm aims to mitigate the well known instabilities observed during training of such models \citep{Saharia2022,Hoogeboom23} while maximizing the use of available training corpora.

% Given the architecture of the PSDM to be trained, the
The \emph{greedy growing} algorithm can be described as follows.
\paragraph{Phase 1} In this phase, the \CC of the chosen architecture are identified (see  \autoref{sec:core}), and a \SU model is build on top of them. The \SU is trained on the entire training collection of text-image pairs, as it is not limited to high resolution training images.
\paragraph{Phase 2} The second phase greedily grows the \SU's encoder/decoder (namely, throwing away the lower-resolution blocks and adding higher-resolution blocks) to obtain the final model.
More specifically, this phase adds encoder and decoder layers at different resolutions, while preserving the \emph{core representation layers} at the spatial resolution used during the first phase. In other words, the \CC continue operating on a $16 \times 16$ grid.
The added layers are randomly initialized, while the \CC are initialized with the weights obtained on the first phase. The remaining components of the \SU model are discarded.
% and do not integrate the final architecture.

% Finally, the complete architecture is trained. 
%
Next, the grown model is trained. As it is a common practice for the generation of high fidelity images, at this point we filter the training data to remove text-image pairs
with either image dimension is lower than the final model's target resolution.
The \emph{text encoding layers} and the \emph{core representation layers} are kept frozen, to preserve the richness of the pretrained representation.
The \emph{time encoding layers}, on the other hand, are further tuned, jointly with the new encoder and decoder layers introduced in the second phase, 
% end-to-end architecture, 
which allows it to adapt to changes in the diffusion noise schedule. 
We adjusted the diffusion logSNR shift for high resolution images as suggested by \cite{Hoogeboom23}, by a factor of $2 \log(64/d)$.
An optional third defrosting phase, may be applied in which all layers are jointly tuned, and seeks to benefit from the full capacity of the end-to-end architecture, but in practice we find that the first two phases are sufficient to obtain a good \PSDM. 

% (i.e., the \emph{text encoding layers},
% the \emph{core representation layers}, and
% the \emph{time encoding layers},)

We empirically investigate the behaviour of the proposed algorithm in models of increasing size in \autoref{sec:exp_greedy}. 
We investigate the effects of splitting the training of the two tasks
in phase one and phase two (i.e., for text-alignment and high-resolution generation), and we compare with models jointly trained from scratch, end-to-end.
During these ablations, we constrain the greedy growing phase to use considerably smaller batch sizes than previous work, with no further regularization to demonstrate the optimization stability. 
\section{Experimental settings}
\label{sec:exp_settings}

\paragraph{\SU:}
The proposed \SU provides a proxy architecture for pre-training the \CC of a larger PSDM. 
The ablation studies below us a specific instantiation of the model, but we expect \SU to be flexible enough to be used with other component parts.
In particular we adopt a combination of two pretrained text encoders for text conditioning:
T5-XXL~\citep{2020t5} with 128 sequence length and CLIP (VIT-H14)~\citep{clipradford21a} with 77 sequence length. 
Given a text prompt, 
we first tokenize and encode the text using the two encoders independently, and then concatenate the embeddings,
% upon token dimension, 
yielding a final embedding with sequence length of 205.
They are projected into model's \emph{hidden size} by the \emph{text encoding layers}.
We keep the \SU design fixed, except for changing the capacity by increasing its width (hidden size) and depth (number of transformer's blocks), as detailed in \autoref{tab:shallow_params}.
That produces a set of models varying from 672M up to 7.7B trainable parameters, mostly dedicated to the \CC.

\begin{table*}[tbh!]
% \vspace*{\baselineskip}
\centering
\resizebox{0.87\linewidth}{!}{
\begin{tabular}{lcccccc} 
\hline
 \hline
 model & transf. blocks &  hidden size & MLP channels & heads %& embedding
 & params\footnotemark\\
\hline 
\small{\SU Small} &
6 & 1536 & 6144 & 12 % & 768 
& 672M
\\
\small{\SU Large} &
8 & 2048 & 8182 & 16 % & 768 
& 1.3B
\\
\small{\SU Huge} &
12 & 3072 & 12288 & 24 % & 768
& 3.5B
\\
\small{\SU XHuge} &
16 & 4096 & 16384 & 32 % & 1024 
&  7.7B
\\
\hline
\end{tabular}
}
%\vspace*{0.25cm}
\caption{\SU variants explored. Transformer layers operating at a $16\times16$ grid. The components within the shallow encoder and decoder block operate at same spatial resolution and hidden size.
}
\label{tab:shallow_params}
\end{table*}
\footnotetext{Number of trainable parameters after ignoring text encoders.}
\begin{table}[tbh!]
\vspace*{-\baselineskip}
\centering
\resizebox{\linewidth}{!}{
\begin{tabular}{lccccc} 
% \multicolumn{5}{c}{ 
% Models grown up to $512\times512$ pixels:
% }
\\
\hline
 \hline
 \small{End-to-end model} & 
 \small{channels per layer} & \small{residual blocks} %& \small{stem-stride} 
 & \small{*params} \\
\hline 
\small{UViT Small} 
& 256-384-768-1536 & 1-1-1-1  %& 2
& 707M
\\
\small{UViT Large} &
256-512-1024-2048 & 1-1-1-1  %& 2
& 1.4B
\\
\small{UViT Huge} &
384-768-1536-3072 & 1-1-1-1  %& 2 
& 3.6B
\\
\small{UViT XHuge} &
512-1024-2048-4096 & 1-1-1-1  %& 2 
& 7.9B
\\
\hline \hline 
% \\
% \multicolumn{5}{c}{ 
% Models grown up to $1024\times1024$ pixels:
% }
% \\
%  \hline \hline
%  \small{End2end model} & 
%  \small{channels per layer} & \small{(enc/dec) blocks} & \small{stem-stride} &
%  \small{*params} \\
% \hline
% \small{UViT-XHuge} &
% 256-512-1024-2048-4096 & 1-1-1-1-1  & 2 & 8.0B
% \\
% \hline \hline 
% \\
% \multicolumn{5}{c}{ 
% Model grown up to $2048\times2048$ pixels:
% }
% \\
%  \hline \hline
%  \small{End2end model} & 
%  \small{channels per layer} & \small{(enc/dec) blocks} & \small{stem-stride} &
%  \small{*params} \\
% \hline
% \small{UViT-XHuge} &
% 256-512-1024-2048-4096 & 1-1-1-1-1  & 4 & 8.0B
% \\
% \hline \hline 
% \\
% \multicolumn{5}{c}{ 
% Model grown up to $4096\times4096$ pixels:
% }
% \\
%  \hline \hline
%  \small{End2end model} & 
%  \small{channels per layer} & \small{(enc/dec) blocks} & \small{stem-stride} &
%  \small{*params} \\
% \hline
% \small{UViT-XHuge} &
% 256-256-512-1024-2048-4096 & 1-1-1-1-1-1  & 4 & 8.0B
% \\\hline \hline
\end{tabular}}
 \vspace*{0.25cm}
\caption{Composition of the encoder-decoder layers grown on top of corresponding \SU variants. \CC identical to the corresponding shallow variant.
% and operating at $16\times16$ grid when targeting  $512\times512$ pixels, and at $32\times32$ when targeting $1024\times1024$ pixels. 
}
\label{tab:uvit_final_params}
\end{table}
%%%%%%%%%%%%%%%%%%%%%%%%%%%%%%%%%%%%

We stress that we do not claim that these specific \CC are optimal. 
For instance, it is widely recognized that larger pretrained text encoders and longer token sequence lengths increase image quality \citep{Saharia2022,balaji2022ediffi,sdxl}.
Investigating the optimal design of each core component is beyond the scope of this work. 
Instead, the variations of the \SU were intentionally designed to explore the performance benefits gained by increasing \CC's capacity independent of the remaining model components.
% We cover multi-aspects in this evaluation as described in \autoref{sec:metrics}.

%%%%%%%%%%%%%%%%%%%%%%%%%%%%%
%%%%%%%%%%%%%%%%%%%%%%%%%%%%%
%%%%%%%%%%%%%%%%%%%%%%%%%%%%%
\paragraph{Greedy growing:} In the experiments that follow we consider several different model sizes.
\autoref{tab:shallow_params} specifies the \SU variants, while 
\autoref{tab:uvit_final_params} specifies encoder/decoder parameterizations.
% For every \SU variant in \autoref{tab:shallow_params},
% we experiment with growing its \CC into a non-cascaded model targeting producing high resolution images (\autoref{sec:exp_greedy}). 
% The grown encoder/decoder are detailed in \autoref{tab:uvit_final_params}.

To ablate our hypothesis that greedy growing helps the model learn strong representations with larger, diverse corpora, we also train the full model on a high resolution subset of data used to train the \SU; i.e., we simply removed all samples with resolution lower than the target model resolution.
To that end, beyond greedy growing, we explore the three training baselines: 1) We create a baseline with all layers trained from scratch on this subset; 2) As an  alternative to the frozen phase in the greedy growing, we fine-tune the \CC on this smaller high resolution subset jointly with the  grown components (randomly initialized); and 3) A third baseline adds the optional phase of unfreezing the \CC after warming up the random weights for 500k steps. 
% This last modality and seeks to benefit from the full capacity of the end-to-end architecture.
Models are trained for 2M steps in total.

The greedy growing algorithm aims to make training large-scale PSDMs at high resolutions more stable.
In the case of Simple Diffusion \citep{Hoogeboom23}, large batch sizes and regularizers like dropout and multi-scale losses enable end-to-end training from scratch.
To stress test the stability and convergence of our greedy growing algorithm, we restrict the batch size to 256 instead of the standard 2k, and we use no other explicit form of regularization.
Under that restriction, our largest model (UVit-XHuge) presented numerical instabilities when trained from scratch or fine-tuned, as multiple numerical issues occurred during training.
Thus, the results of this large model are presented only for the frozen, and freeze-unfreeze methods.
This behaviour confirms observations in previous work and their need for large batch sizes. 

\paragraph{Dataset:} Rigorous evaluation of generative image models is challenging when models are trained on proprietary datasets.
To avoid this issue, we first demonstrate our key findings through extensive empirical evaluations on a publicly available dataset, namely, Conceptual 12M (or CC12M) \citep{changpinyo2021cc12m}.

To evaluate the hypothesis that the greedy algorithm allows one to make good use of available corpora, we trained \SU on the entire CC12M training set, while corresponding end-to-end models were trained with CC12M's subset of $8.7M$ images whose dimensions are equal or larger than $512$ pixels. Those end-to-end models were therefore trained on $27.5\%$ less data than the corresponding \SU model. 
We do not explore more aggressive reduction of the corpora as the CC12M dataset is already a relatively small dataset for the models tested, and the variations tested already show overfitting characteristics under this setting, as discussed below.
Thus, in what follows, the \SU models were trained on $64\times64$ images, by resizing the smallest dimension of the images to $64$ and random cropping along the remaining dimension as needed. The end-to-end models are trained at a target resolution of $512\times512$ as CC12M does not contain images at resolutions above $1024$ pixels.
% (per dimension).

% Later, in \autoref{sec:vermeer} we further extend our ablations using a larger internal dataset and targetting larger resolutions.

\paragraph{Full pipeline model:} With those findings in place, we then explore the generation of larger images and train on a much larger curated datasets in order to show that the approach scales to state-of-the-art models (\autoref{sec:vermeer}). The resulting model, named Vermeer, is used to generate $1024 \times 1024$ images, well beyond the scale for which quantitative metrics are readily available.  As such, with Vermeer we reply on human evaluation, in comparison to other recent models, like SDXL.

\paragraph{Sampling:}
Unless mentioned, the images and metrics were produced using 256 steps of a DDPM sampler \citep{NEURIPS2020_4c5bcfec}  with classifier-free guidance \citep{ho2021classifierfree}. We tune the guidance hyper-parameter by a FD-Dino/Clip (VIT-L14) trade-off as described in \autoref{sec:guidance}.

% In addition to the inherent complexity of the evaluation process, unlike other research domains, where isolated methodologies are typically evaluated using publicly accessible datasets and well-defined frameworks independent of confounding variables, this practice is not commonly employed for mainstream generative models. Thus, differentiating the effectiveness of the evaluated methods from the influence of internal datasets and pre- and post-processing techniques poses a significant challenge. In light of this consideration, we present the main findings of  methodology in a publicly available dataset (Conceptual 12M dataset --CC12M \cite{changpinyo2021cc12m}) independentely from the use of larger and/or curated datasets which enhance the ultimate quality of diffusion results.
\subsection{Metrics}
\label{sec:metrics}
The evaluation of generative models poses considerable difficulties and constitutes an active research area \citep{kirstain2024pick,xu2024imagereward,hessel2021clipscore,serra2023emotions,kim2024confidenceaware,lee2023holistic}. 
 In light of its inherent complexity, we utilize a multi-faceted evaluation strategy that combines image distribution metrics, text-aligment metrics and semantic question and answering metrics to validate our intermediary results, but the overall performance of our final model evaluation, Vermeer, is delegated to human evaluators (\autoref{sec:human_eval}). The following criteria are considered:
 
\paragraph{Image distribution metrics:} 
We evaluate models on three key metrics, namely, the Fr\'echet Inception Distance (FID) \citep{Heusel17}, the Fr\'echet Distance on Dino-v2 feature space (FD-Dino) \citep{stein2023exposing,dinov2} and
the Clip Maximum Mean Discrepancy (CMMD) distance \citep{jayasumana2023rethinking}.
FID is widely used to assess generative image models and select model hyper-parameters, but our findings corroborate its known limitations: it fails to reflect model improvements through training, it
does not capture readily apparent distortions in individual images, and it does not correlate well with human perception \citep{stein2023exposing,Human_Evaluation,jayasumana2023rethinking}. 
Thus, in our study, we do not select training or sampling hyper-parameters solely on the basis of FID but, as described in Appendix \ref{sec:guidance}, we review the trade-offs between the observed set of metrics.

We also note that metrics derived from image features vary considerably with image resolution. 
In what follows we compute metrics using the same resolution as the reference papers. The exception is for CMMD on \SU outputs; the original metric taken at $336\times336$  pixels is dominated by up-sampling  effects, obscuring differences between models.
Thus, we replaced the original $ViT-L14$ operating at $336\times336$ by its version at $224\times224$ pixels.

%CLIP Score is a reference free metric that can be used to evaluate the correlation between a generated caption for an image and the actual content of the image.

\paragraph{Multimodal metrics:} We adopt CLIP Score as a metric for text-image alignment, as it is widely used, and it complements image distribution metrics above, reflecting the consistency of the generated image with the given prompt. 
Unlike the original formulation based on ViT-B with path size 32 \citep{hessel2021clipscore} and previous papers in the area \cite{saharia2022photorealistic,Hoogeboom23}, we adopt the ViT-L (patch 14) embedding due to its improved representation. 
This choice results in lower absolute values of our CLIP Scores compared to previous results, however we noticed that these scores better correlate with the presence of absence of observed distortions.

% Our results are taken using  Clip-score using  VIT-L with $14\times14$ patches embeddings instead of the original VIT-B.

\paragraph{Semantic QG/A frameworks:} 
One can also automatically generate question-answer pairs with a language model, and then compute image faithfulness by checking whether existing VQA models can answer the questions from the generated image \citep{hu2023tifa,JaeminCho2024}. 
They were intended to address the shortcomings of existing metrics.
Despite their effectiveness in evaluating color and material aspects, they often struggle in assessing counting, spatial relationships, and compositions with multiple objects.  
Such evaluation measures are naturally dependent on the quality of the underlying question generation (QG) and answering (QA) models.
Here we adopt DSG (image-text alignment metric) and its set of $1k$ prompts \citep{JaeminCho2024}. The DSG-1k test-prompts cover different challenges (e.g., counting correctly, correct color/shape/text rendering, etc.), semantic categories, and writing styles. A description of the QG, QA used, with qualitative and detailed results, are included in Appendix \ref{app:vqva_detailed}.

\section{Experiments}
\label{sec:experiments}

% We next present our ablations on the greedy growing algorithm with models trained on the CC12M dataset.
% A description our sampler hyper-parameter tuning adopted for the results described in this section is contained in Appendix \ref{sec:guidance}.

\subsection{Pretraining and scaling the \CC}

\begin{figure*}[t]
\newcommand{\imwidth}{3.0cm}
\centering
\begin{tabular}{ccccccc}
\\[-\ht\strutbox]
    \rotatebox[origin=l]{90}{{\scriptsize \SU Base}}&
    \includegraphics[width=\imwidth,raise=-\dp\strutbox]{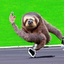}%
    \includegraphics[width=\imwidth,raise=-\dp\strutbox]{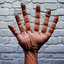}%
    \includegraphics[width=\imwidth,raise=-\dp\strutbox]{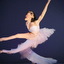}%
    \includegraphics[width=\imwidth,raise=-\dp\strutbox]{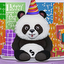}%
    \includegraphics[width=\imwidth,raise=-\dp\strutbox]{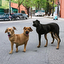}% 
    \\
    \rotatebox[origin=l]{90}{{\scriptsize \SU Large}}&
    \includegraphics[width=\imwidth,raise=-\dp\strutbox]{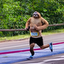}%
    \includegraphics[width=\imwidth,raise=-\dp\strutbox]{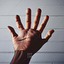}%
    \includegraphics[width=\imwidth,raise=-\dp\strutbox]{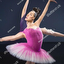}%
    \includegraphics[width=\imwidth,raise=-\dp\strutbox]{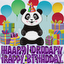}%
    \includegraphics[width=\imwidth,raise=-\dp\strutbox]{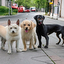}% 
    \\
    \rotatebox[origin=l]{90}{{\scriptsize \SU Huge}}&
    \includegraphics[width=\imwidth,raise=-\dp\strutbox]{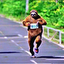}%
    \includegraphics[width=\imwidth,raise=-\dp\strutbox]{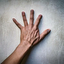}%
    \includegraphics[width=\imwidth,raise=-\dp\strutbox]{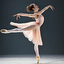}%
    \includegraphics[width=\imwidth,raise=-\dp\strutbox]{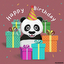}%
    \includegraphics[width=\imwidth,raise=-\dp\strutbox]{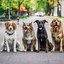}% 
    \\
    \rotatebox[origin=l]{90}{{\scriptsize \SU XHuge}}&
    \includegraphics[width=\imwidth,raise=-\dp\strutbox]{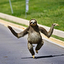}%
    \includegraphics[width=\imwidth,raise=-\dp\strutbox]{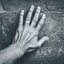}%
    \includegraphics[width=\imwidth,raise=-\dp\strutbox]{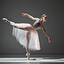}%
    \includegraphics[width=\imwidth,raise=-\dp\strutbox]{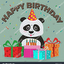}%
    \includegraphics[width=\imwidth,raise=-\dp\strutbox]{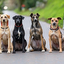}%
    \\
\end{tabular}
\caption{
\textbf{Qualitative comparison of models with \CC of increasing size} -- {\SU}s trained at $64\times 64$ pixels using CC12M dataset only. 
Prompts: {\em A sloth running a marathon, surprisingly outrunning all competitors. A hand spread out on a wall. DSLR photograph. Close-up portrait of a ballerina in mid-performance, with high motion and dramatic lighting. Word art of "happy birthday", with a smiling panda wearing a party hat, surrounded by gift boxes and a birthday cake. Four dogs on the street.}}
\label{fig:core_qualitative}
\vspace*{-0.5em}
\end{figure*}

\begin{table*}[tbh!]
% \vspace*{-\baselineskip}
\vspace*{0.2cm}
\centering
\resizebox{0.65\linewidth}{!}{
\begin{tabular}{lccc|c} 
\hline
 \hline
 models$_{@64\times64}$  & 
 FID$_{30k}\downarrow$~ &
 FD-Dino$_{30k}\downarrow$~ &
 CMMD$_{30k}\downarrow$~ & CLIP$_{score}\uparrow$
\\
\hline 
\scriptsize{\SU Base} & %\scriptsize{2K} & \scriptsize{2M} &
16.97 & 356.25 & 0.197 & 0.234
\\
\scriptsize{\SU Large}  &
14.80 & 236.24 & 0.156 & 0.240
\\
\scriptsize{\SU Huge}  & 
8.81 & 133.51 &  0.139 & 0.244
\\
\scriptsize{\SU XHuge}  & 
{\bf8.41} & {\bf116.83} & {\bf 0.136}  & {\bf 0.246}
% \\
% \multicolumn{1}{r}{\em+ data }& & &
% 9.73 & 127.14 & {\bf0.225} & 0.2984
% % \\
% \hline
% \scriptsize{Shallow UViT-Base} &   \scriptsize{8K} & \scriptsize{1M} &
% 16.63 & 357.77 &  0.264 & 0.726
% \\
% \scriptsize{Shallow UViT-Large}  &   & &
% 14.26 & 235.95 & 0.247 & 0.740
% \\
% \scriptsize{Shallow UViT-Huge}  &   & &
% 8.51 &  129.03 &  0.272 & 0.748
% \\
% \scriptsize{Shallow UViT-XHuge}  &   & &
% {\bf8.00} & {\bf111.24} & 0.255 & {\bf 0.750}
% \\
% \multicolumn{1}{r}{\em+ data } &  & & 
% 9.61 &
% 124.52 &
% {\bf0.232} &  0.748
% \\
% \hline
%  \multicolumn{1}{r}{\em+ data } & \scriptsize{8K} & \scriptsize{1.5M} & 
%  9.49 & 120.84 & 0.230  & {\bf 0.750}
% {\bf }
\\
\hline
\end{tabular}}
\vspace*{0.1cm}
\caption{\SU variants with \CC of increasing size trained on CC12M at resolution $64\times64$: Image distribution metrics evaluated on $30k$ samples from MSCOCO captions dataset.
Scaling induces performace improvements on image distribution (FID, FD-Dino, CMMD) and text-image alignment ($CLIP_{score}$) metrics simultaneously.}
% trained on CC12M dataset: 
% \emph{Training settings: batch $2k$ for $2M$ steps}: Increasing models size induces increase in performace on both $FID$,  $FD-Dino$ and $CCMD$.}.
% \emph{Training settings: batch $8k$ for $1M$ steps}: Image distribution metrics do not  present any sign of overfitting when doubling the training epochs.
% Models trained with larger datasets present worst $FID$ and  $FD-Dino$ and better $CMMD$, but take longer to converge and benefit from longer training. 
% \caption{Shallow-UVIT variants at $64\times64$ pixels trained on CC12M dataset: image distribution metrics evaluated on $30k$  samples from MSCOCO captions dataset.
% \emph{Training settings: batch $2k$ for $2M$ steps}: Increasing models size induces increase in performace on both $FID$,  $FD-Dino$ and $CCMD$.}.
% \emph{Training settings: batch $8k$ for $1M$ steps}: Image distribution metrics do not  present any sign of overfitting when doubling the training epochs.
% Models trained with larger datasets present worst $FID$ and  $FD-Dino$ and better $CMMD$, but take longer to converge and benefit from longer training. 
\label{tab:shallow_metrics}
\end{table*}
\begin{table*}[tbh!]
% \vspace*{-\baselineskip}
\centering
\resizebox{0.7\linewidth}{!}{
\begin{tabular}{lccccc} 
\hline \hline
 & \multicolumn{5}{c}{DSG - VqVa Question Types}
\\
 & 
 Entities & Relations & Attributes & Global %& mspice$_{score} (\uparrow)$
 & DSG$(\uparrow)$
 \\
 \multicolumn{1}{r}{\scriptsize{\#questions:}} &  \scriptsize{3378} & \scriptsize{1485} & \scriptsize{1722} & \scriptsize{649} & \\
 \hline 
\scriptsize{\SU Small} &  
 % ms ent, rel, att, glob, overall
 54.38 & 33.32 & 43.70 & 39.98 %& 42.19 
 & 48.08
 \\
\scriptsize{\SU Large}  & 
 % ms ent, rel, att, glob, overall
 59.93 & 39.36 & 48.75 & 43.68 %& 47.05
 & 52.54
 \\
\scriptsize{\SU Huge}  & 
 % ms ent, rel, att, glob, overall
 69.18	& 48.52	& 54.36	& 43.30	%& 53.81	
 & 60.25
 \\
\scriptsize{\SU XHuge} & 
 % ms ent, rel, att, glob, overall
 {\bf 70.66} & {\bf 51.61} & {\bf 57.38} & {\bf 44.14 %& 56.79 
 } & {\bf 61.91} 
\\
\hline
\end{tabular}
}
\vspace*{0.25cm}
\caption{Shallow-UVIT evaluated on $1k$ samples from DSG-1k dataset. Scaling \CC improves performance across all semantic categories.
Fine-grained results in Appendix \ref{app:vqva_detailed}}
\label{tab:uvit_shallow_vqva_summary}
\end{table*}

We next use \SU as a proxy architecture to investigate the effect of scaling PSDM's \CC. 
% They cover different aspects of the performance of generative models. 
We train \SU variants on $64\times64$ images from the CC12M dataset for 2k steps. Image distribution metrics and Clip-Score are obtained using $30k$ prompts from the MSCOCO-captions validation set \citep{COCO}, while the semantic metrics are extracted on the 1k prompts from DSG-1k \citep{JaeminCho2024}.
A summary of the impact of scaling the \SU model is given in Tables \ref{tab:shallow_metrics} and \ref{tab:uvit_shallow_vqva_summary}, while fine grained results on semantic categories are reported in Appendix \ref{app:vqva_detailed}.
% The three image distribution metrics adopted covering different feature spaces, the Clip-Score as well as the different categories covered by the VqVa model,
All performance measures indicate significant improvements due to model scaling. 
A smaller numerical gain is observed in the comparison of the larger two models, but the difference is reflected in qualitative comparisons of the models below. 
% We expect their performance gain to be bounded by the small training dataset adopted. 

\autoref{fig:core_qualitative}, presents a qualitative comparison of the results the \SU variants on challenging prompts. They illustrate the impact of scaling on objects structure, composition and alignment (e.g., with numbers of objects depicted). 
Despite of the small training dataset, the larger models show significant improvement in generating intricate shapes like hands, body parts and text.

\begin{figure*}[t]
\setlength{\lineskip}{0pt}
\centering
\footnotesize{
%%%%%%%%%%%%%%%%%%%%%%%%%%%%%%%%%%%%
%%%%%%%%%%%%%%%%%%%%%%%%%%%%%%%%%%%%
%%%%%%%%%%%%%%%%%%%%%%%%%%%%%%%%%%%%
%%%%%%%%%%%%%%%%%%%%%%%%%%%%%%%%%%%%
%%%%%%%%%%%%%%%%%%%%%%%%%%%%%%%%%%%%
\begin{subfigure}{\linewidth}
\setlength{\abovecaptionskip}{0pt}
    \begin{subfigure}[ht]{\linewidth}
        \begin{subfigure}[t]{0.11\linewidth}
            \includegraphics[width=\linewidth, height=0.95\linewidth]{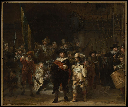}
            % \caption*{Reference}
        \end{subfigure}
        \begin{subfigure}[t]{0.11\linewidth}
            \includegraphics[width=\linewidth, height=0.95\linewidth]{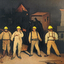}
            % \caption*{\tiny{step 250k}}
        \end{subfigure}%
        \begin{subfigure}[t]{0.11\linewidth}
            \includegraphics[width=\linewidth, height=0.95\linewidth]{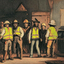}
            % \caption*{\tiny{step 500k}}
        \end{subfigure}%
        \begin{subfigure}[t]{0.11\linewidth}
            \includegraphics[width=\linewidth, height=0.95\linewidth]{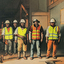}
            % \caption*{\tiny{step 750k}}
        \end{subfigure}%
        \begin{subfigure}[t]{0.11\linewidth}
            \includegraphics[width=\linewidth, height=0.95\linewidth]{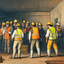}
            % \caption*{\tiny{step 1M}}
        \end{subfigure}%
        \begin{subfigure}[t]{0.11\linewidth}
            \includegraphics[width=\linewidth, height=0.95\linewidth]{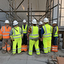}
            % \caption*{\tiny{step 1.25M}}
        \end{subfigure}%
        \begin{subfigure}[t]{0.11\linewidth}
            \includegraphics[width=\linewidth, height=0.95\linewidth]{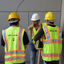}
            % \caption*{\tiny{step 1.5M}}
        \end{subfigure}%
        \begin{subfigure}[t]{0.11\linewidth}
            \includegraphics[width=\linewidth, height=0.95\linewidth]{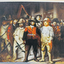}
            % \caption*{\tiny{step 1.75M}}
        \end{subfigure}%
        \begin{subfigure}[t]{0.11\linewidth}
            \includegraphics[width=\linewidth, height=0.95\linewidth]{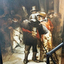}
            % \caption*{\tiny{step 2M}}
        \end{subfigure}%
    \end{subfigure}
\end{subfigure}\\
%%%%%%%%%%%%%%%%%%%%%%%%%%%%%%%%%%%%
%%%%%%%%%%%%%%%%%%%%%%%%%%%%%%%%%%%%
%%%%%%%%%%%%%%%%%%%%%%%%%%%%%%%%%%%%David
%%%%%%%%%%%%%%%%%%%%%%%%%%%%%%%%%%%%
%%%%%%%%%%%%%%%%%%%%%%%%%%%%%%%%%%%%
\begin{subfigure}{\linewidth}
\setlength{\abovecaptionskip}{0pt}
    \begin{subfigure}[ht]{\linewidth}
    
        \begin{subfigure}[t]{0.11\linewidth}
            \includegraphics[width=\linewidth, height=0.95\linewidth]{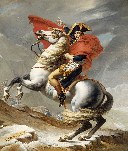}
            % \caption*{\tiny{Reference}}
        \end{subfigure}
        \begin{subfigure}[t]{0.11\linewidth}
            \includegraphics[width=\linewidth, height=0.95\linewidth]{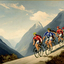}
            % \caption*{\tiny{step 250k}}
        \end{subfigure}%
        \begin{subfigure}[t]{0.11\linewidth}
            \includegraphics[width=\linewidth, height=0.95\linewidth]{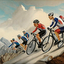}
            % \caption*{\tiny{step 500k}}
        \end{subfigure}%
        \begin{subfigure}[t]{0.11\linewidth}
            \includegraphics[width=\linewidth, height=0.95\linewidth]{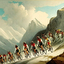}
            % \caption*{\tiny{step 750k}}
        \end{subfigure}%
        \begin{subfigure}[t]{0.11\linewidth}
            \includegraphics[width=\linewidth, height=0.95\linewidth]{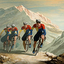}
            % \caption*{\tiny{step 1M}}
        \end{subfigure}%
        \begin{subfigure}[t]{0.11\linewidth}
            \includegraphics[width=\linewidth, height=0.95\linewidth]{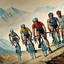}
            % \caption*{\tiny{step 1.25M}}
        \end{subfigure}%
        \begin{subfigure}[t]{0.11\linewidth}
            \includegraphics[width=\linewidth, height=0.95\linewidth]{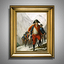}
            % \caption*{\tiny{step 1.5M}}
        \end{subfigure}%
        \begin{subfigure}[t]{0.11\linewidth}
            \includegraphics[width=\linewidth, height=0.95\linewidth]{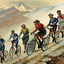}
            % \caption*{\tiny{step 1.75M}}
        \end{subfigure}%
        \begin{subfigure}[t]{0.11\linewidth}
            \includegraphics[width=\linewidth, height=0.95\linewidth]{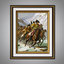}
            % \caption*{\tiny{step 2M}}
        \end{subfigure}
    \end{subfigure}
\end{subfigure}
\\
%%%%%%%%%%%%%%%%%%%%%%%%%%%%%%%%%%%%
%%%%%%%%%%%%%%%%%%%%%%%%%%%%%%%%%%%%
%%%%%%%%%%%%%%%%%%%%%%%%%%%%%%%%%%%%David
%%%%%%%%%%%%%%%%%%%%%%%%%%%%%%%%%%%%
%%%%%%%%%%%%%%%%%%%%%%%%%%%%%%%%%%%%
\begin{subfigure}{\linewidth}
\setlength{\abovecaptionskip}{0pt}
    \begin{subfigure}[ht]{\linewidth}
    
        \begin{subfigure}[t]{0.11\linewidth}
            \includegraphics[width=\linewidth, height=0.95\linewidth]{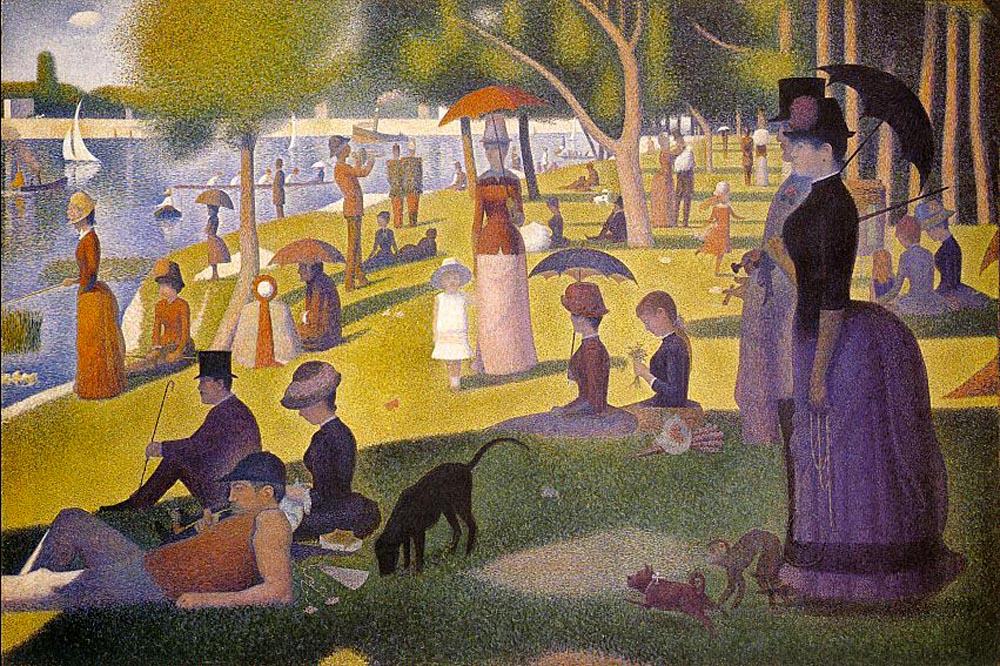}
            \caption*{\tiny{Reference}}
        \end{subfigure}
        \begin{subfigure}[t]{0.11\linewidth}
            \includegraphics[width=\linewidth, height=0.95\linewidth]{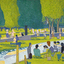}
            \caption*{\tiny{step 250k}}
        \end{subfigure}%
        \begin{subfigure}[t]{0.11\linewidth}
            \includegraphics[width=\linewidth, height=0.95\linewidth]{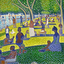}
            \caption*{\tiny{step 500k}}
        \end{subfigure}%
        \begin{subfigure}[t]{0.11\linewidth}
            \includegraphics[width=\linewidth, height=0.95\linewidth]{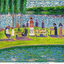}
            \caption*{\tiny{step 750k}}
        \end{subfigure}%
        \begin{subfigure}[t]{0.11\linewidth}
            \includegraphics[width=\linewidth, height=0.95\linewidth]{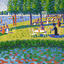}
            \caption*{\tiny{step 1M}}
        \end{subfigure}%
        \begin{subfigure}[t]{0.11\linewidth}
            \includegraphics[width=\linewidth, height=0.95\linewidth]{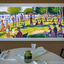}
            \caption*{\tiny{step 1.25M}}
        \end{subfigure}%
        \begin{subfigure}[t]{0.11\linewidth}
            \includegraphics[width=\linewidth, height=0.95\linewidth]{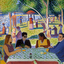}
            \caption*{\tiny{step 1.5M}}
        \end{subfigure}%
        \begin{subfigure}[t]{0.11\linewidth}
            \includegraphics[width=\linewidth, height=0.95\linewidth]{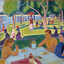}
            \caption*{\tiny{step 1.75M}}
        \end{subfigure}%
        \begin{subfigure}[t]{0.11\linewidth}
            \includegraphics[width=\linewidth, height=0.95\linewidth]{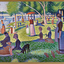}
            \caption*{\tiny{step 2M}}
        \end{subfigure}
    \end{subfigure}
\end{subfigure}
\vspace{-.5em}
\caption{Overfitting and memorization of \SU XHuge  trained on CC12M.
Prompts:
(top) {\it A group of construction workers in the style of 'The Night Watch' by Rembrandt.};
(middle) {\it A dynamic rendition of a racing cyclist leading their team through a mountain pass, rendered in the style of 'Napoleon Crossing the Alps' by Jacques-Louis David.};
(bottom) {\it 
A group of friends enjoying a summer day at a riverside restaurant in the style of 'A Sunday Afternoon on the Island of La Grande Jatte' by Georges Seurat.}
}
\label{fig:overfitting_qualitative}}
\vspace{-1em}
\end{figure*}

We observed further quantitative improvements across the metrics when training our larger models for longer (\SU-Huge and \SU-XHuge), but longer training also exhibits overfitting to the CC12 training samples.
\autoref{fig:overfitting_qualitative} illustrates images generated using the \SU XHuge model  with increasing numbers of training steps. 
As training progresses, the model diverges from the original prompt to produce images that are closer to training samples from the CC12M dataset, and/or representing parts of the prompt only.
This hidden phenomena was not associated with changes in the adopted metrics.
We conjecture that this effect is largely aggravated by the small size of the training dataset.

\begin{figure*}[bth!]
\centering
\scriptsize{
\begin{subfigure}[c]{0.35\linewidth}
    \includegraphics[width=\linewidth]{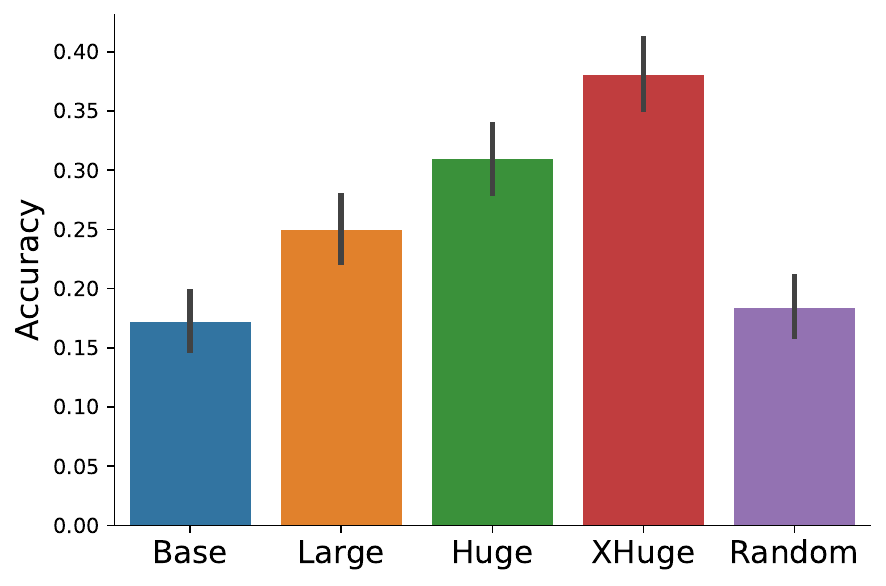}
\end{subfigure}%
\hspace*{0.75cm}
\begin{subfigure}{0.5\linewidth}
\setlength{\abovecaptionskip}{0pt}
    \begin{subfigure}[ht]{\linewidth}
        \begin{subfigure}[t]{0.24\linewidth}
            \includegraphics[width=\linewidth, height=0.95\linewidth]{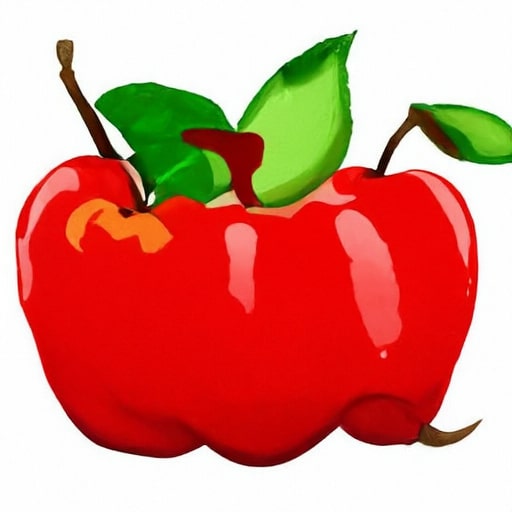}
            \caption*{{Base}}
        \end{subfigure}%
        \begin{subfigure}[t]{0.24\linewidth}
            \includegraphics[width=\linewidth, height=0.95\linewidth]{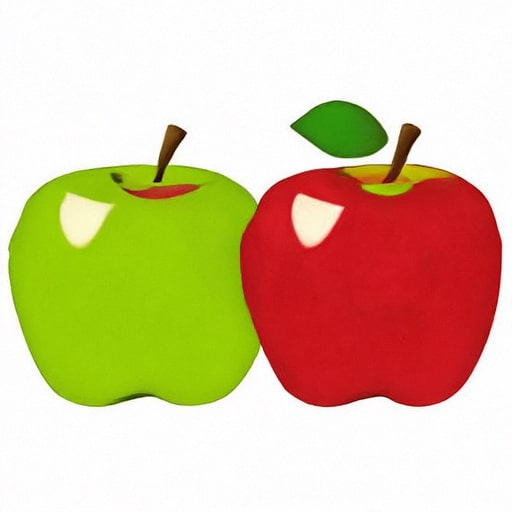}
            \caption*{{Large}}
        \end{subfigure}%
        \begin{subfigure}[t]{0.24\linewidth}
            \includegraphics[width=\linewidth, height=0.95\linewidth]{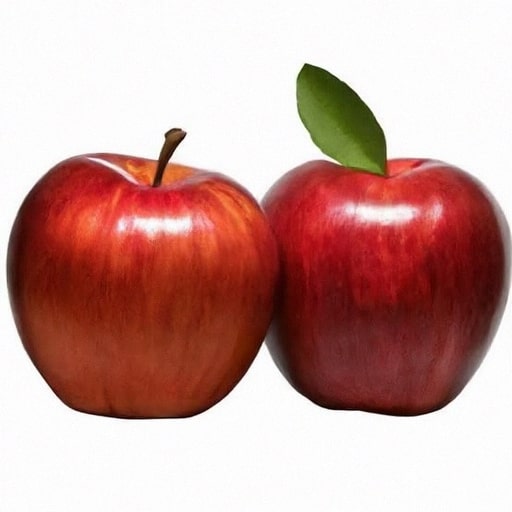}
            \caption*{{Huge}}
        \end{subfigure}%
        \begin{subfigure}[t]{0.24\linewidth}
            \includegraphics[width=\linewidth, height=0.95\linewidth]{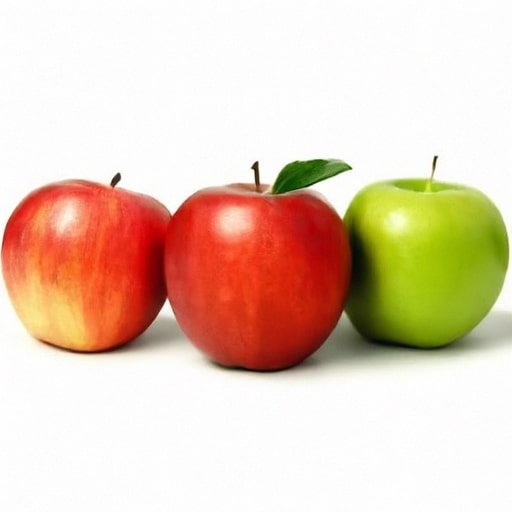}
            \caption*{{XHuge}}
        \end{subfigure}%
    \end{subfigure}
    % \\
    % \begin{subfigure}[ht]{\linewidth}
    %     \begin{subfigure}[t]{0.24\linewidth}
    %         \includegraphics[width=\linewidth, height=0.95\linewidth]{images/counting/img_0_99_base.png}
    %         \caption*{{Base}}
    %     \end{subfigure}%
    %     \begin{subfigure}[t]{0.24\linewidth}
    %         \includegraphics[width=\linewidth, height=0.95\linewidth]{images/counting/img_0_99_large.png}
    %         \caption*{{Large}}
    %     \end{subfigure}%
    %     \begin{subfigure}[t]{0.24\linewidth}
    %         \includegraphics[width=\linewidth, height=0.95\linewidth]{images/counting/img_0_99_huge.png}
    %         \caption*{{Huge}}
    %     \end{subfigure}%
    %     \begin{subfigure}[t]{0.24\linewidth}
    %         \includegraphics[width=\linewidth, height=0.95\linewidth]{images/counting/img_0_99_xhuge.png}
    %         \caption*{{XHuge}}
    %     \end{subfigure}%
    % \end{subfigure}%
\end{subfigure}%
% \vspace{-.5em}
\caption{Measuring the impact of scaling on the counting task. Using 59 systematic prompts describing 1-5 objects. Five human annotators reviewed each image (95\% bootstrapped confidence intervals are shown). Models with larger \CC are observed to perform better on counting. 
% 95\% bootstrapped confidence intervals are shown.
Sample prompt: \emph{3 apples.}}
\label{fig:counting}
\vspace{-1em}
}
\end{figure*}

Considering the complexity associated with evaluating improvements in representation and the limitations of automatic performance measures, 
we also ablate the effect of scaling the \CC under a semantic task that is evaluated by human annotators.
In this experiment  we consider a {\em simple counting task}, defined here as the task of generating images of up to 5 objects based on a subset of text prompts from the numerical split of the Gecko benchmark~\citep{gecko2024}.
We explore this task as a proxy for gauging both prompt consistency and the model's understanding of objects composition and shapes.
It allows less subjective interpretation and noise in human judgments of the model's performance than other image qualities that are influenced by individual preferences.
The task of counting under an open set would ultimately imply the ability to keep track of objects. Thus, this ablation emulates a much simpler version of the problem.
\autoref{fig:counting} shows the accuracy improvement associated with scaling observed over 59 prompts.
Random condition uses a random number between 1-5.
The detailed description of this experiment is presented on Appendix \ref{app:counting}. 

Given the shallow encoder-decoder structure of the \SU architecture, we conjecture that the performance improvements observed
here, on multiple metrics, are a direct consequence of scaling the \CC. 
This hypothesis is further investigated via the reuse of their representation in the next section.

% \df{We should have another look at the tables to more clearly describe findings.  The impact of scale and data are not as clear as text may imply. Question:  Do we need human evals here as well as Vermeer to make certains points clear?}
\subsection{Experiments on Greedy growing}
\label{sec:exp_greedy}

\begin{table*}[tbh!]
\vspace*{-\baselineskip}
\centering
\resizebox{0.9\linewidth}{!}{
\begin{tabular}{llcrccc|c} 
\hline
 \hline
 model & &\scriptsize{\emph{tr. params}} &\scriptsize{\emph{steps}} & FID$_{30k}\downarrow$ & FD-Dino$_{30k}\downarrow$ & 
 CMMD$_{30k}\downarrow$ & CLIP$_{score}\uparrow$\\
\hline 
%%% OSC 0.0:
\small{UViT-Base} & 
\emph{scratch} & 707M & 2M &
27.90 & 624.34 & 1.355 & 0.241
\\& \emph{finetuning} &  &  
% 0.5M & 27.30 &  572.80 &  1.658 &  0.237 \\
%& &  & 
2M & 23.67 & 554.99 &  1.450  & 0.241
\\
& \emph{frozen core} & 217M &
% 0.5M & 28.24 & 610.42 &   1.681 & 0.233
%\\ & & & 
2M &
24.68 & 563.35 &  1.614 &  0.235
\\
& \emph{freeze-unfreeze} & 217M/707M &2M &
{\bf 21.13} & {\bf 503.16} & {\bf 1.196} & {\bf 0.247}

\\
%%%%%%%%%%%%%%%%%%%%%%%%%%%%%%%%%
\hline \small{UViT-Large} & \emph{scratch} & 1.4B & 2M &
21.73 & 498.82 & 1.156 &  0.247
\\& \emph{finetuning} & &  
%0.5M & 20.77 & 369.58 &  1.087 & 0.254\\
%& & & 
2M &
21.89 & 414.42 & 1.160 & 0.253
\\&\emph{frozen core} & 351M & 
% 0.5M &18.58 & 246.24 &  1.087 & 0.259
% \\& & & 
2M &
 {\bf 17.68} & {\bf 195.80} & {\bf 0.752} & {\bf 0.264}
\\
& \emph{freeze-unfreeze} & 351M/1.4B & 2M &
18.37 & 362.58 & 0.952& 0.256
\\
\hline \small{UViT-Huge} & 
\emph{scratch} & 3.6B & 2M &
18.58 & 382.17 & 1.053 & 0.256
\\ & \emph{finetuning} &  & 
% 0.5M &
% 15.89 & 273.99 & 1.082 & {\bf 0.268}
% \\& &  & 
2M &
17.52 & 302.28 & 0.988 & 0.264
\\
& \emph{frozen core} & 723M & 
%0.5M & {\bf 14.86} & 163.40 & 0.719 & 0.267 
%\\& & & 
2M&
{\bf 15.21} & {\bf 156.24} & {\bf 0.663} & {\bf 0.268} 
\\
& \emph{freeze-unfreeze} & 723M/3.6B & 2M &
16.17 & 231.94 &   0.683   & 0.262
\\
\hline \small{UViT-XHuge} & \emph{freeze} &  1.2B & 2M &
{\bf 15.32} &
{\bf 152.12} &
{\bf 0.571} &
{\bf 0.269} 
\\
& \emph{freeze-unfreeze} & 
1.2B/7.9B & 2M &
16.58 &
222.38 &
0.620 &
0.267
\\
\hline
 \end{tabular}}
\vspace*{0.25cm}
\caption{End2end variants trained on CC12M dataset at $512\times512$ pixels and batch size 256: image distribution metrics (FID, FD-Dino and CMMD). Smaller models benefit from finetuning all their parameters. Larger models have more capacity in the encoder-decoder layers, and benefit from freezing the pretrained representations, under such a small batch size regime.}
\label{tab:e2e_metrics}
\end{table*}
\begin{table*}[tbh!]
\vspace*{-\baselineskip}
\centering
\resizebox{0.75\linewidth}{!}{
\begin{tabular}{lrrccccc} 
\hline \hline
& & & \multicolumn{5}{c}{DSG - VqVa Question Types}
\\
model & & \emph{steps} & Entities & Relations & Attributes & Global % & mspice$_{score}$
& DSG
 \\
\hline 
%%%%%%%%%%%%%%%%%%%%%%%%%%%%%%%%%%%%%%%%%%%%%%%%%%%
%%%%%%%%%%%%%%%%%%%%%%%%%%%%%%%%%%%%%%%%%%%%%%%%%%%
%%%%%%%%%%%%%%%%%%%%%%%%%%%%%%%%%%%%%%%%%%%%%%%%%%%
\footnotesize{UVIT-Base} &
 \emph{scratch} & \emph{2M} 
 &
 % ms ent, rel, att, glob, overall
73.16 & 53.91 & 62.31 & 55.55 % & 60.32 
& 64.83
 \\  %%%%%%%%%%%%%%%%%
 & 
 \emph{finetuning} &  
 % \emph{0.5M} &
 % ms ent, rel, att, glob, overall
% 70.03 &   & 59.73 & 53.31 % & 55.66 
% & 61.78
 
% \\  %%%%%%%%%%%%%%%%% 2M steps
% &  &
\emph{2M}  & 
 % ms ent, rel, att, glob, overall
 70.23 & 49.90 & 58.89 & 53.24 %& 57.29 
 & 62.75
 \\  %%%%%%%%%%%%%%%%%
 & 
 \emph{frozen} &
 % \emph{0.5M} & 
  % ms ent, rel, att, glob, overall
% 69.33 & 47.91 & 58.62 & 49.46 % & 55.56 
% & 59.92
%  \\  %%%%%%%%%%%%%%%%% 2M steps
% &  & 
\emph{2M} & 
  % ms ent, rel, att, glob, overall
69.57 & 49.36 & 58.22 & 53.39 % & 55.87
& 61.16
 \\  %%%%%%%%%%%%%%%%% freeze-unfreeze
& \emph{freeze-unfreeze} &  \emph{2M} &
  % ms ent, rel, att, glob, overall
{\bf 73.40} & {\bf 53.54} & {\bf 62.83} & {\bf 56.86} % & 60.23
& {\bf66.13}
 \\
 \hline 
 %%%%%%%%%%%%%%%%%%%%%%%%%%%%%%%%%%%%%%%%%%%%%%%%%%%
 %%%%%%%%%%%%%%%%%%%%%%%%%%%%%%%%%%%%%%%%%%%%%%%%%%%
 %%%%%%%%%%%%%%%%%%%%%%%%%%%%%%%%%%%%%%%%%%%%%%%%%%%
\footnotesize{UVIT-Large} &
  \emph{scratch} & \emph{2M} 
 &
 % ms ent, rel, att, glob, overall
73.31 & 52.02 & 62.95 & 58.01  %& 60.12 
& 66.02
 \\  %%%%%%%%%%%%%%%%%
 & 
 \emph{finetuning} & 
%  \emph{0.5M} 
%  &
%  % ms ent, rel, att, glob, overall
%  76.39 & 58.22 & 66.49 & 58.09 %& 64.32
%  & 69.56 
 
% \\  %%%%%%%%%%%%%%%%% 2M steps
% &  &
\emph{2M} 
 &
 % ms ent, rel, att, glob, overall
 75.01 & 54.11 & 65.82 & 57.86 %& 62.59
 & 67.39
 \\  %%%%%%%%%%%%%%%%%
 & 
 \emph{frozen} & 
%  \emph{0.5M} 
%  & 
%  % ms ent, rel, att, glob, overall
% 77.19 & 56.60 & 64.78 & 58.70 %& 63.53
% &  69.51
% \\  %%%%%%%%%%%%%%%%% 2M steps
% &  &
\emph{2M} 
 &
 % ms ent, rel, att, glob, overall
 {\bf 78.97 } & {\bf 61.55 } & {\bf 67.19 } & {\bf 61.40 } % & {\bf 65.95 }
 & {\bf 72.13 }
 \\  %%%%%%%%%%%%%%%%% freeze-unfreeze
& \emph{freeze-unfreeze} &  \emph{2M} 
 &
 % ms ent, rel, att, glob, overall
74.67 & 55.45 & 64.08 & 58.78 % & 62.83 
& 67.79
 \\
\hline 
%%%%%%%%%%%%%%%%%%%%%%%%%%%%%%%%%%%%%%%%%%%%%%%%%%%
%%%%%%%%%%%%%%%%%%%%%%%%%%%%%%%%%%%%%%%%%%%%%%%%%%%
%%%%%%%%%%%%%%%%%%%%%%%%%%%%%%%%%%%%%%%%%%%%%%%%%%%
\footnotesize{UViT-Huge} &
 \emph{scratch} & \emph{2M} 
 &
 % ms ent, rel, att, glob, overall
 % \underline{} 
74.33 & 55.02 & 62.98 & 58.63 % & 62.38 
& 66.90 
 \\  %%%%%%%%%%%%%%%%%
 & 
 \emph{finetuning} &
%  \emph{0.5M} 
%  &
%  % ms ent, rel, att, glob, overall
% 76.12 & 55.62 & 65.94 & 59.32 %& 64.20 
% & 69.26 
% \\  %%%%%%%%%%%%%%%%% 2M steps
% &  & 
\emph{2M} 
 &
 % ms ent, rel, att, glob, overall
77.29 & 56.40 & 67.13 & 62.56 % & 64.47 
& 69.67
 \\  %%%%%%%%%%%%%%%%%
 & \emph{frozen} & 
%  \emph{0.5M} 
%  &
%  % ms ent, rel, att, glob, overall
%  82.36 & 64.71 & 71.84 & 63.41 % & 69.75
%  & 75.12
%  \\  %%%%%%%%%%%%%%%%% 2M steps
% &  & 
\emph{2M} 
 &
 % ms ent, rel, att, glob, overall
{\bf 82.59} & {\bf 64.65} & {\bf 70.35} & {\bf 61.86} 
%& {\bf 68.94}
& {\bf 75.15}
 \\  %%%%%%%%%%%%%%%%% freeze-unfreeze
& \emph{freeze-unfreeze} &  \emph{2M} 
 &
 % ms ent, rel, att, glob, overall
79.04 & 58.11 & 65.97 &  60.86 %& 64.73
&71.50
\\
\hline
\footnotesize{UViT-XHuge} & \emph{frozen} & \emph{2M} &
{\bf 83.70}&
{\bf 66.77}&
{\bf 70.01}&
{\bf 62.94} &
{\bf 75.70}  
\\
& \emph{freeze-unfreeze} &  \emph{2M}  &
81.14  &
60.44  &
69.40  &
60.25  &
73.53
\\
\hline
\end{tabular}
}
\vspace*{0.25cm}
\caption{E2e variants at $512\times512$ pixels trained on CC12M dataset. Metrics evaluated on $1k$  samples from DSG-1k dataset. \emph{DSG} results are aggregated across semantic categories. Fine-grained results in Appendix \ref{app:vqva_detailed}.
% Underlining: training a large scale model end2end (Huge \emph{scratch}) struggles performing worst than it smaller model counterpart (Large \emph{frozen}). 
\label{tab:uvit_e2e_vqva_summary}
}
\end{table*}

\begin{figure*}[tbh!]
\setlength{\lineskip}{0pt}
\centering
\footnotesize{
%%%%%%%%%%%%%%%%%%%%%%%%%%%%%%%%%%%%
%%%%%%%%%%%%%%%%%%%%%%%%%%%%%%%%%%%%
%%%%%%%%%%%%%%%%%%%%%%%%%%%%%%%%%%%%Kangaroo
%%%%%%%%%%%%%%%%%%%%%%%%%%%%%%%%%%%%
%%%%%%%%%%%%%%%%%%%%%%%%%%%%%%%%%%%%
\begin{subfigure}{0.48\linewidth}
\setlength{\abovecaptionskip}{0pt}
    \begin{subfigure}[c]{0.16\linewidth}
            \includegraphics[width=\linewidth, height=0.94\linewidth,cfbox=orange 1pt 1pt]{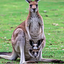}
    \end{subfigure}
    \begin{subfigure}[ht]{0.825\linewidth}
        \begin{subfigure}[t]{0.32\linewidth}
            \includegraphics[width=\linewidth, height=0.94\linewidth,cfbox=green 1pt 1pt]{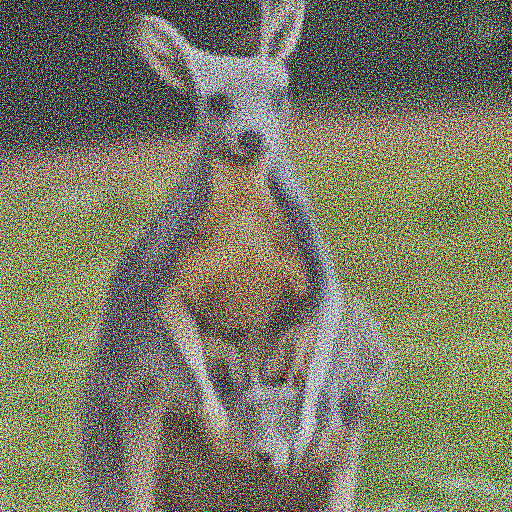}
        \end{subfigure}
        \begin{subfigure}[t]{0.32\linewidth}
            \includegraphics[width=\linewidth, height=0.94\linewidth,cfbox=green 1pt 1pt]{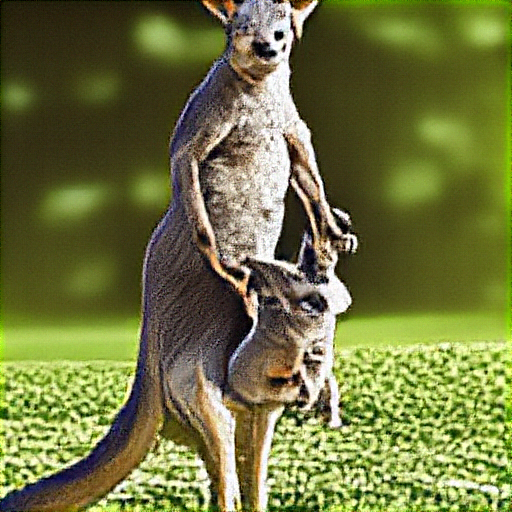}
        \end{subfigure}
        \begin{subfigure}[t]{0.32\linewidth}
            \includegraphics[width=\linewidth, height=0.94\linewidth,cfbox=green 1pt 1pt]{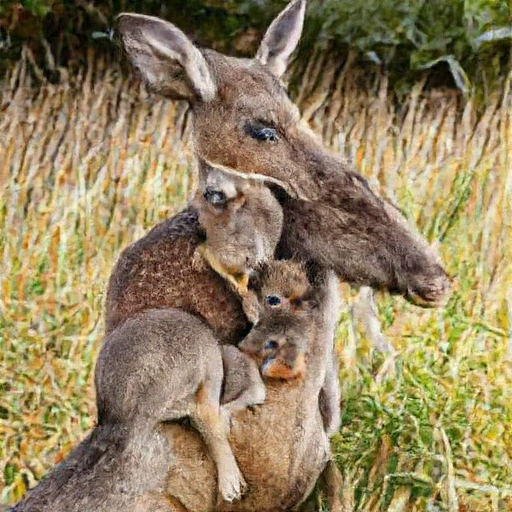}
        \end{subfigure}
        \\
        \begin{subfigure}[t]{0.32\linewidth}
            \includegraphics[width=\linewidth, height=0.94\linewidth,cfbox=blue 1pt 1pt]{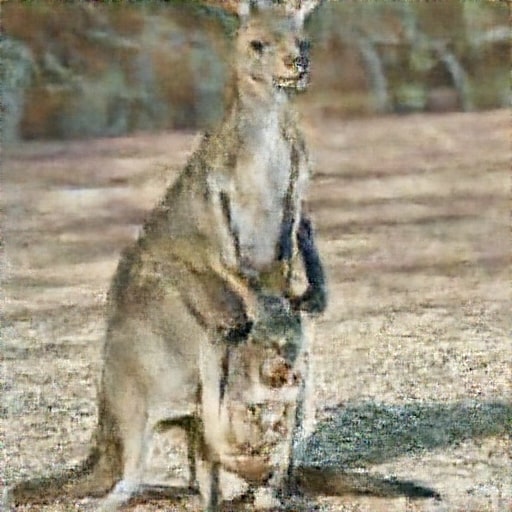}
        \end{subfigure}
        \begin{subfigure}[t]{0.32\linewidth}
            \includegraphics[width=\linewidth, height=0.94\linewidth,cfbox=blue 1pt 1pt]{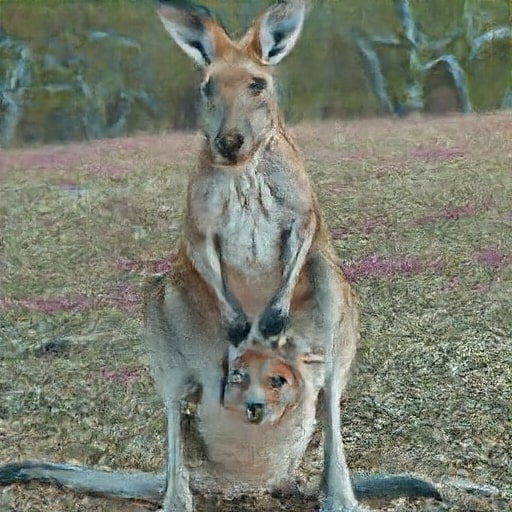}
        \end{subfigure}
        \begin{subfigure}[t]{0.32\linewidth}
            \includegraphics[width=\linewidth, height=0.94\linewidth,cfbox=blue 1pt 1pt]{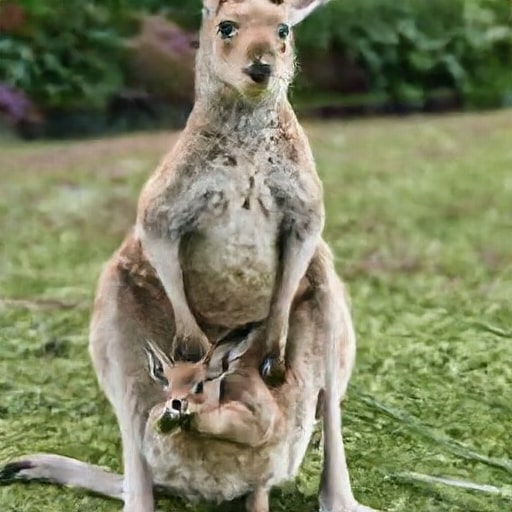}
        \end{subfigure}
    \end{subfigure}
\end{subfigure}
%%%%%%%%%%%%%%%%%%%%%%%%%%%%%%%%%%%%
%%%%%%%%%%%%%%%%%%%%%%%%%%%%%%%%%%%%
%%%%%%%%%%%%%%%%%%%%%%%%%%%%%%%%%%%% butterfly
%%%%%%%%%%%%%%%%%%%%%%%%%%%%%%%%%%%%
%%%%%%%%%%%%%%%%%%%%%%%%%%%%%%%%%%%%
\begin{subfigure}{0.48\linewidth}
\setlength{\abovecaptionskip}{0pt}
    \begin{subfigure}[c]{0.16\linewidth}
            \includegraphics[width=\linewidth, height=0.94\linewidth,cfbox=orange 1pt 1pt]{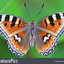} 
     \end{subfigure}
     \begin{subfigure}[ht]{0.825\linewidth}
       \begin{subfigure}[t]{0.32\linewidth}
            \includegraphics[width=\linewidth, height=0.94\linewidth,cfbox=green 1pt 1pt]{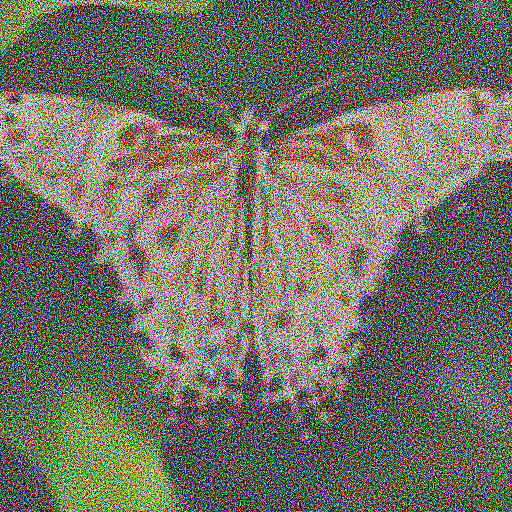} 
        \end{subfigure}
        \begin{subfigure}[t]{0.32\linewidth}
            \includegraphics[width=\linewidth, height=0.94\linewidth,cfbox=green 1pt 1pt]{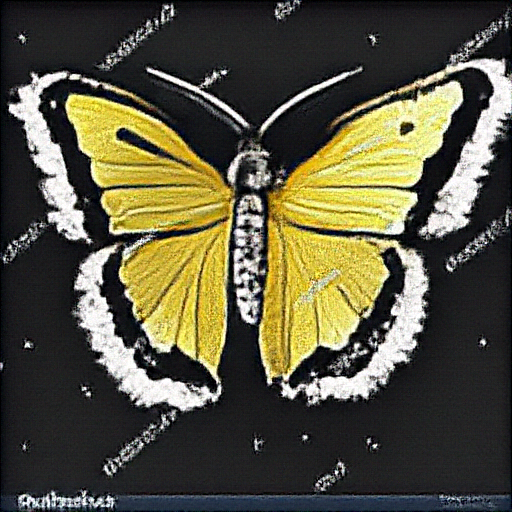} 
        \end{subfigure}
        \begin{subfigure}[t]{0.32\linewidth}
            \includegraphics[width=\linewidth, height=0.94\linewidth,cfbox=green 1pt 1pt]{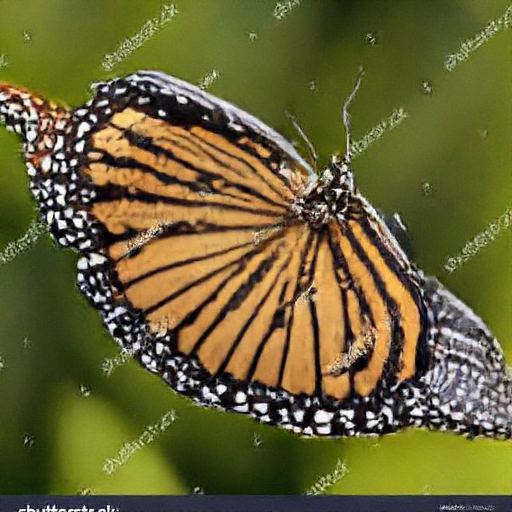} 
        \end{subfigure}
        \\
        \begin{subfigure}[t]{0.32\linewidth}
            \includegraphics[width=\linewidth, height=0.94\linewidth,cfbox=blue 1pt 1pt]{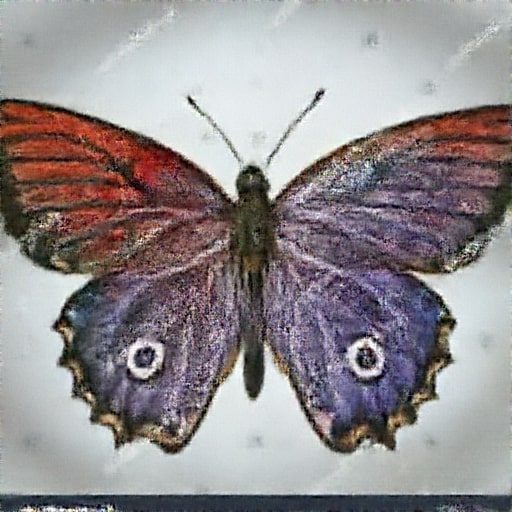} 
        \end{subfigure}
        \begin{subfigure}[t]{0.32\linewidth}
            \includegraphics[width=\linewidth, height=0.94\linewidth,cfbox=blue 1pt 1pt]{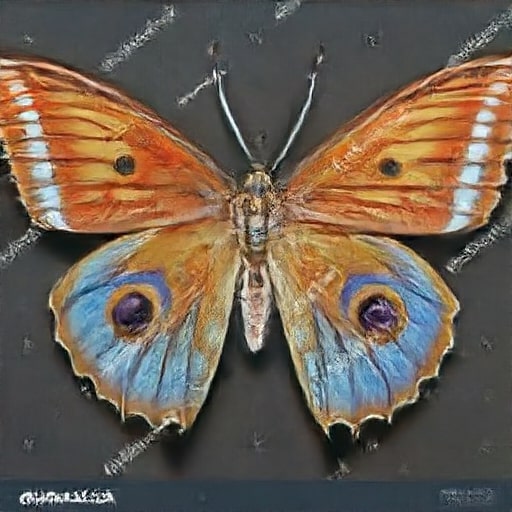} 
        \end{subfigure}
        \begin{subfigure}[t]{0.32\linewidth}
            \includegraphics[width=\linewidth, height=0.94\linewidth,cfbox=blue 1pt 1pt]{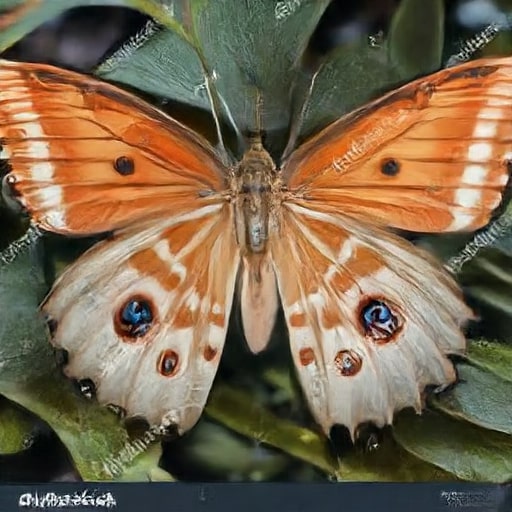} 
        \end{subfigure}
    \end{subfigure}
\end{subfigure}
\\
%%%%%%%%%%%%%%%%%%%%%%%%%%%%%%%%%%%%
%%%%%%%%%%%%%%%%%%%%%%%%%%%%%%%%%%%%
%%%%%%%%%%%%%%%%%%%%%%%%%%%%%%%%%%%%Wolf
%%%%%%%%%%%%%%%%%%%%%%%%%%%%%%%%%%%%
%%%%%%%%%%%%%%%%%%%%%%%%%%%%%%%%%%%%
\begin{subfigure}{0.48\linewidth}
\setlength{\abovecaptionskip}{0pt}
    \begin{subfigure}[c]{0.16\linewidth}
        \includegraphics[width=\linewidth, height=0.94\linewidth,cfbox=orange 1pt 1pt]{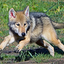} 
    \end{subfigure}
    \begin{subfigure}[ht]{0.825\linewidth}
        \begin{subfigure}[t]{0.32\linewidth}
            \includegraphics[width=\linewidth, height=0.94\linewidth,cfbox=green 1pt 1pt]{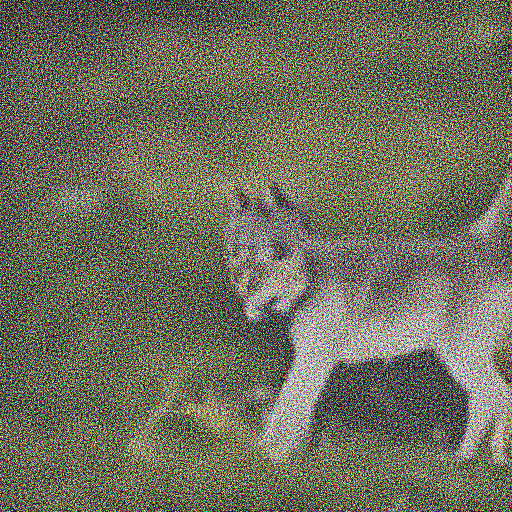} 
        \end{subfigure}
        \begin{subfigure}[t]{0.32\linewidth}
            \includegraphics[width=\linewidth, height=0.94\linewidth,cfbox=green 1pt 1pt]{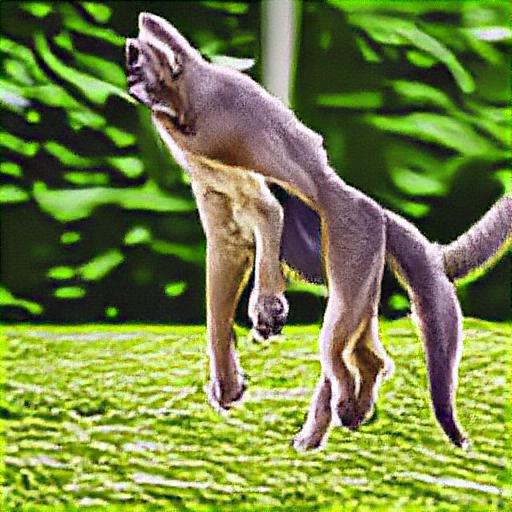} 
        \end{subfigure}
        \begin{subfigure}[t]{0.32\linewidth}
            \includegraphics[width=\linewidth, height=0.94\linewidth,cfbox=green 1pt 1pt]{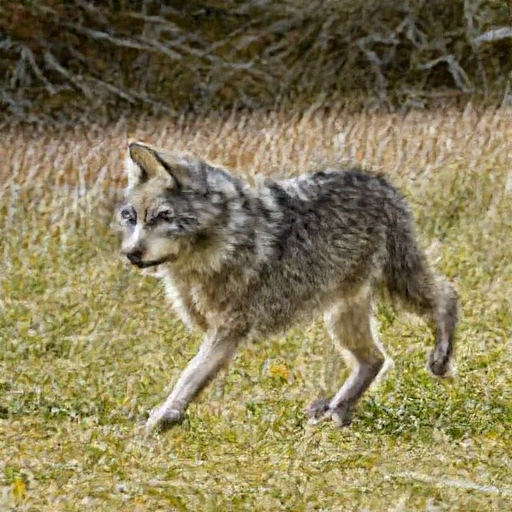} 
        \end{subfigure}
        \\
        \begin{subfigure}[t]{0.32\linewidth}
            \includegraphics[width=\linewidth, height=0.94\linewidth,cfbox=blue 1pt 1pt]{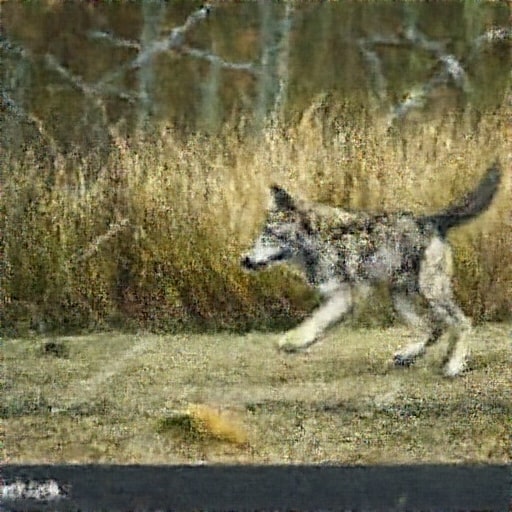} 
        \end{subfigure}
        \begin{subfigure}[t]{0.32\linewidth}
            \includegraphics[width=\linewidth, height=0.94\linewidth,cfbox=blue 1pt 1pt]{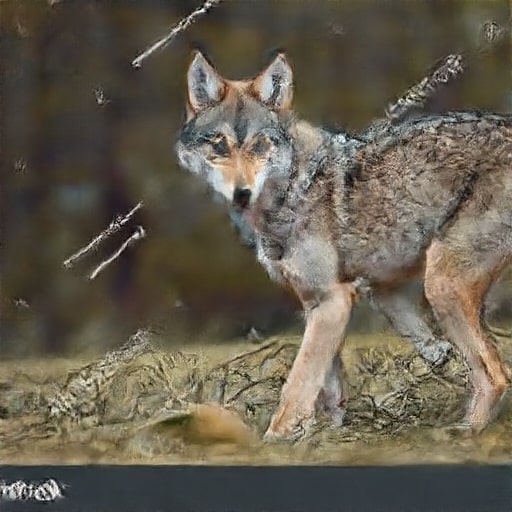} 
        \end{subfigure}
        \begin{subfigure}[t]{0.32\linewidth}
            \includegraphics[width=\linewidth, height=0.94\linewidth,cfbox=blue 1pt 1pt]{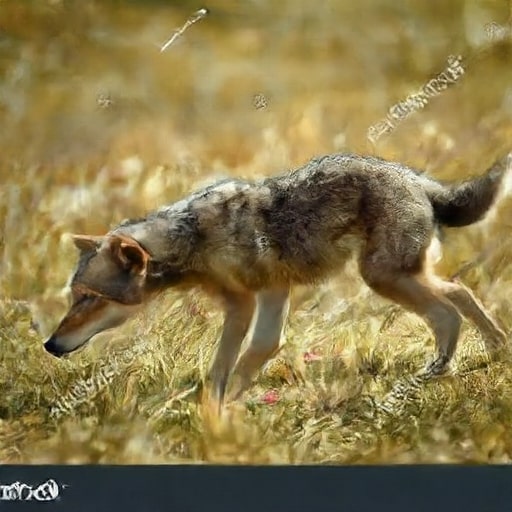} 
        \end{subfigure}
    \end{subfigure}
\end{subfigure}
%%%%%%%%%%%%%%%%%%%%%%%%%%%%%%%%%%%%
%%%%%%%%%%%%%%%%%%%%%%%%%%%%%%%%%%%%
%%%%%%%%%%%%%%humminbird%%%%%%%%%%%%
%%%%%%%%%%%%%%%%%%%%%%%%%%%%%%%%%%%%
%%%%%%%%%%%%%%%%%%%%%%%%%%%%%%%%%%%%
\begin{subfigure}{0.48\linewidth}
\setlength{\abovecaptionskip}{0pt}
    \begin{subfigure}[c]{0.16\linewidth}
        \includegraphics[width=\linewidth, height=0.94\linewidth,cfbox=orange 1pt 1pt]{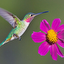} 
    \end{subfigure}
     \begin{subfigure}[ht]{0.825\linewidth}
       \begin{subfigure}[t]{0.32\linewidth}
            \includegraphics[width=\linewidth, height=0.94\linewidth,cfbox=green 1pt 1pt]{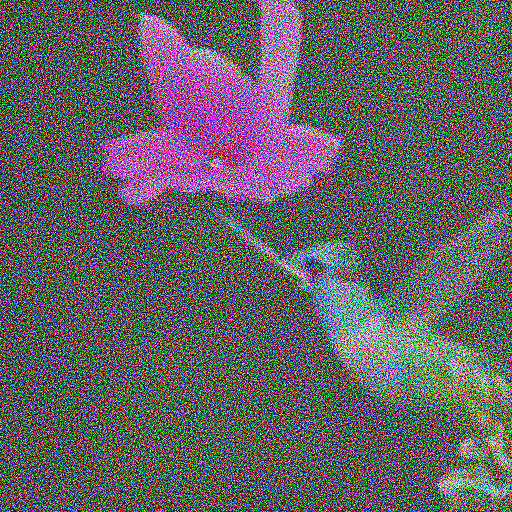} 
        \end{subfigure}
        \begin{subfigure}[t]{0.32\linewidth}
            \includegraphics[width=\linewidth, height=0.94\linewidth,cfbox=green 1pt 1pt]{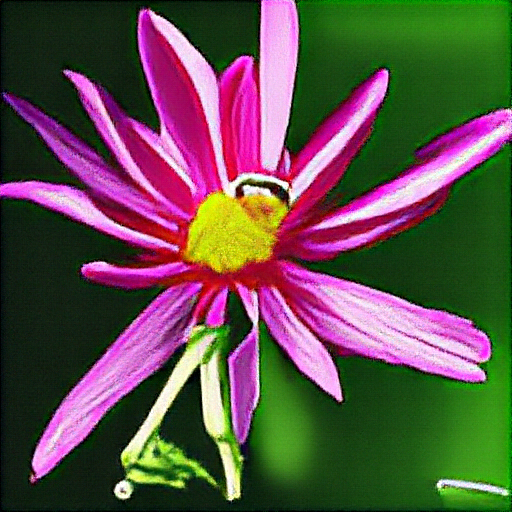} 
        \end{subfigure}
        \begin{subfigure}[t]{0.32\linewidth}
            \includegraphics[width=\linewidth, height=0.94\linewidth,cfbox=green 1pt 1pt]{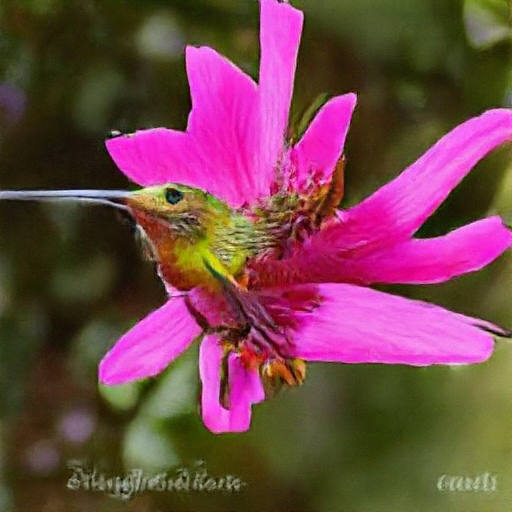} 
        \end{subfigure}
        \\
        \begin{subfigure}[t]{0.32\linewidth}
            \includegraphics[width=\linewidth, height=0.94\linewidth,cfbox=blue 1pt 1pt]{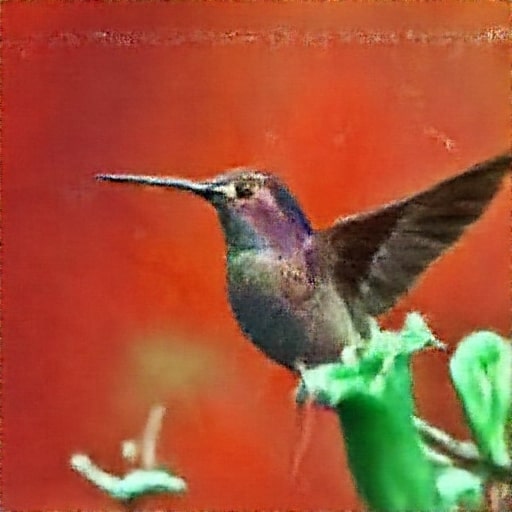} 
        \end{subfigure}
        \begin{subfigure}[t]{0.32\linewidth}
            \includegraphics[width=\linewidth, height=0.94\linewidth,cfbox=blue 1pt 1pt]{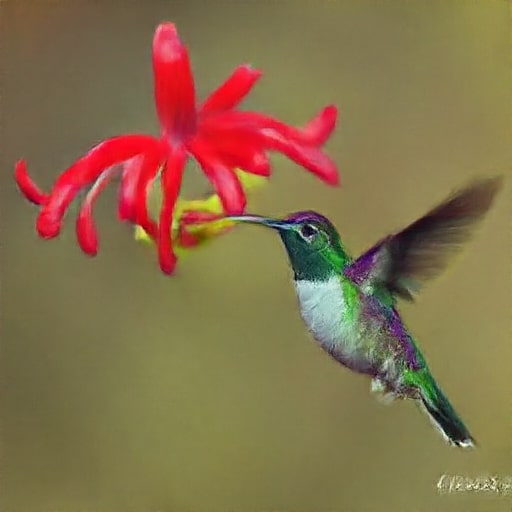} 
        \end{subfigure}
        \begin{subfigure}[t]{0.32\linewidth}
            \includegraphics[width=\linewidth, height=0.94\linewidth,cfbox=blue 1pt 1pt]{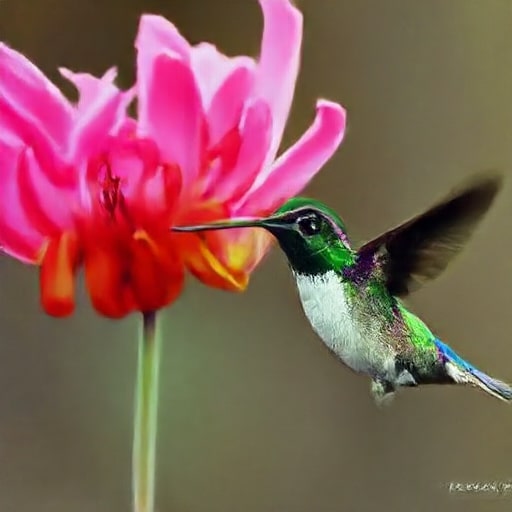} 
        \end{subfigure}
    \end{subfigure}
\end{subfigure}
\\
%%%%%%%%%%%%%%%%%%%%%%%%%%%%%%%%%%%%
%%%%%%%%%%%%%%%%%%%%%%%%%%%%%%%%%%%%
%%%%%%%%%%%%%%%%%%%%%%%%%%%%%%%%%%%%
%%%%%%%%%%%%%%%%%%%%%%%%%%%%%%%%%%%%
%%%%%%%%%%%%%%%%%%%%%%%%%%%%%%%%%%%%
\begin{subfigure}{0.485\linewidth}
    \begin{subfigure}[c]{0.16\linewidth}
            \includegraphics[width=\linewidth, height=0.94\linewidth,cfbox=orange 1pt 1pt]{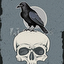} 
        \end{subfigure}
    \begin{subfigure}[ht]{0.825\linewidth}
    \begin{subfigure}[t]{0.32\linewidth}
            \includegraphics[width=\linewidth, height=0.94\linewidth,cfbox=green 1pt 1pt]{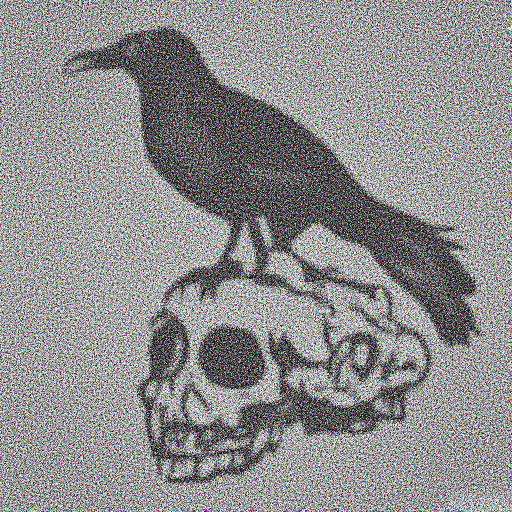} 
        \end{subfigure}
        \begin{subfigure}[t]{0.32\linewidth}
            \includegraphics[width=\linewidth, height=0.94\linewidth,cfbox=green 1pt 1pt]{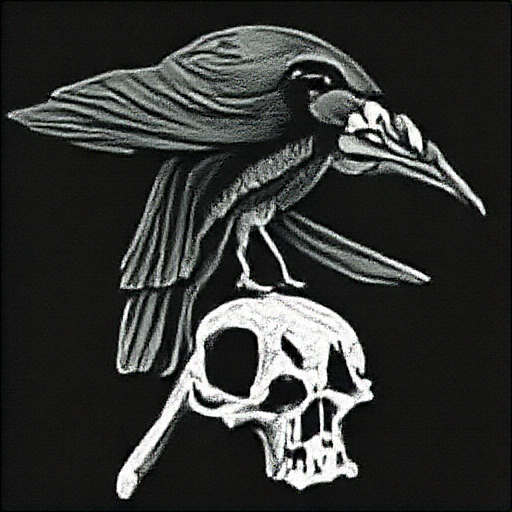} 
        \end{subfigure}
        \begin{subfigure}[t]{0.32\linewidth}
            \includegraphics[width=\linewidth, height=0.94\linewidth,cfbox=green 1pt 1pt]{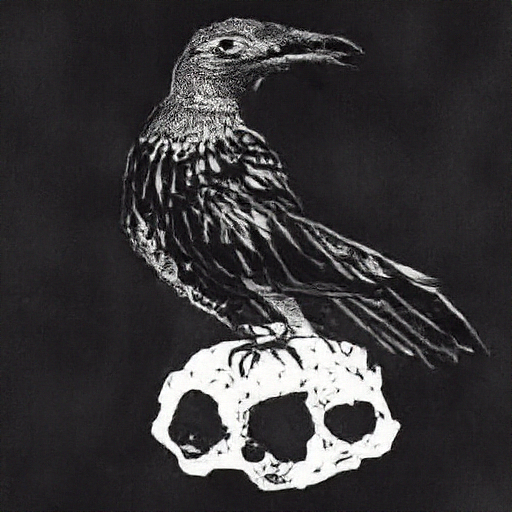} 
        \end{subfigure}
        \\
        \begin{subfigure}[t]{0.32\linewidth}
            \includegraphics[width=\linewidth, height=0.94\linewidth,cfbox=blue 1pt 1pt]{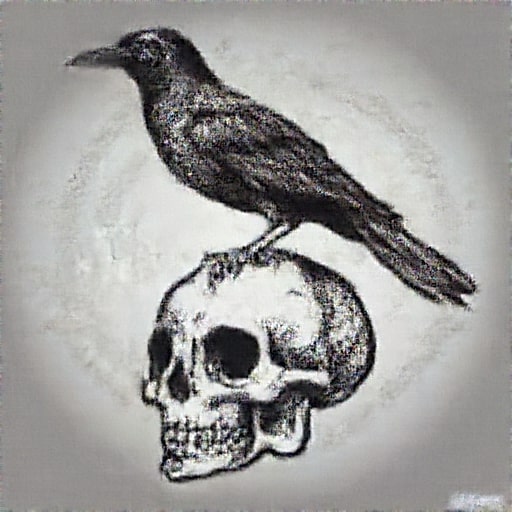} 
            \caption*{\scriptsize{20k steps}}
        \end{subfigure}
        \begin{subfigure}[t]{0.32\linewidth}
            \includegraphics[width=\linewidth, height=0.94\linewidth,cfbox=blue 1pt 1pt]{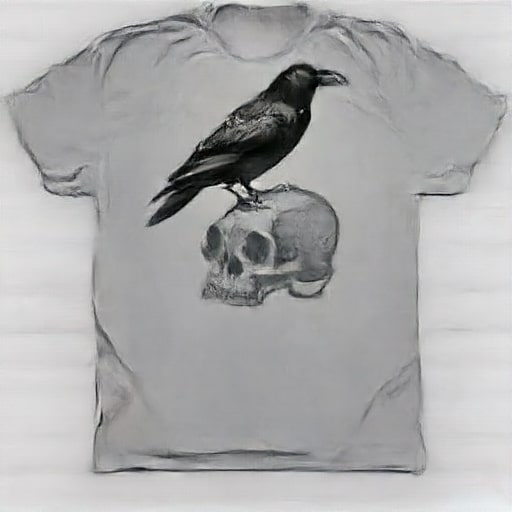} 
            \caption*{\scriptsize{50k steps}}
        \end{subfigure}
        \begin{subfigure}[t]{0.32\linewidth}
            \includegraphics[width=\linewidth, height=0.94\linewidth,cfbox=blue 1pt 1pt]{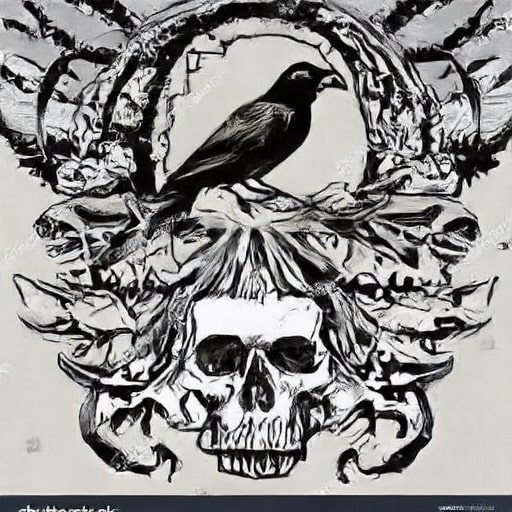} 
        \caption*{\scriptsize{100k steps}}
        \end{subfigure}
    \end{subfigure}
\end{subfigure}
%%%%%%%%%%%%%%%%%%%%%%%%%%%%%%%%%%%%
%%%%%%%%%%%%%%%%%%%%%%%%%%%%%%%%%%%%
%%%%%%%%%%%%%turtle%%%%%%%%%%%%%%%%%%
%%%%%%%%%%%%%%%%%%%%%%%%%%%%%%%%%%%%
%%%%%%%%%%%%%%%%%%%%%%%%%%%%%%%%%%%%
\begin{subfigure}{0.485\linewidth}
    \begin{subfigure}[c]{0.16\linewidth}
        \includegraphics[width=\linewidth, height=0.94\linewidth,cfbox=orange 1pt 1pt]{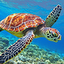}
        %\caption*{\scriptsize{Pretrained at $64x64$ pixels}}
    \end{subfigure}
    \begin{subfigure}[ht]{0.825\linewidth}
        \begin{subfigure}[t]{0.32\linewidth}
            \includegraphics[width=\linewidth, height=0.94\linewidth,cfbox=green 1pt 1pt]{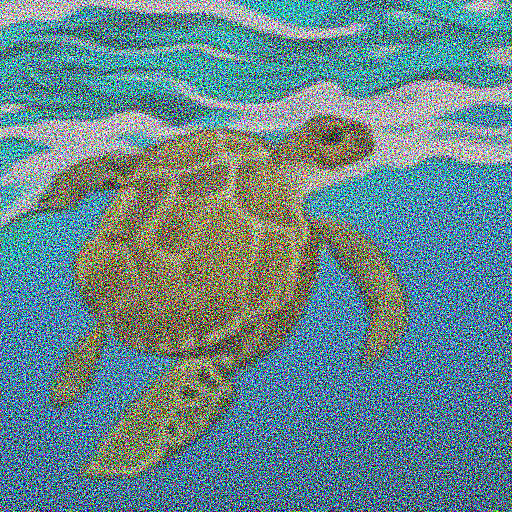}
        \end{subfigure}
        \begin{subfigure}[t]{0.32\linewidth}
            \includegraphics[width=\linewidth, height=0.94\linewidth,cfbox=green 1pt 1pt]{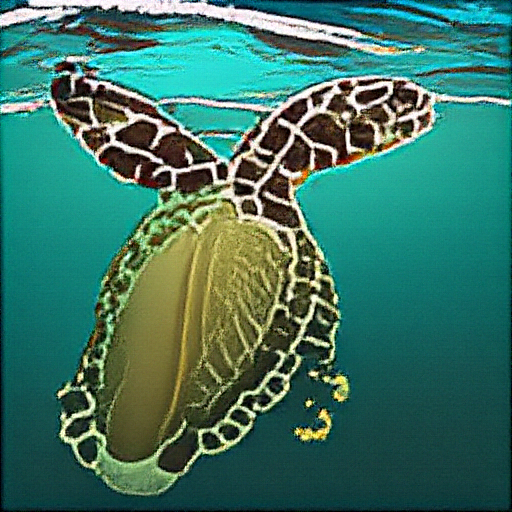}
        \end{subfigure}
        \begin{subfigure}[t]{0.32\linewidth}
            \includegraphics[width=\linewidth, height=0.94\linewidth,cfbox=green 1pt 1pt]{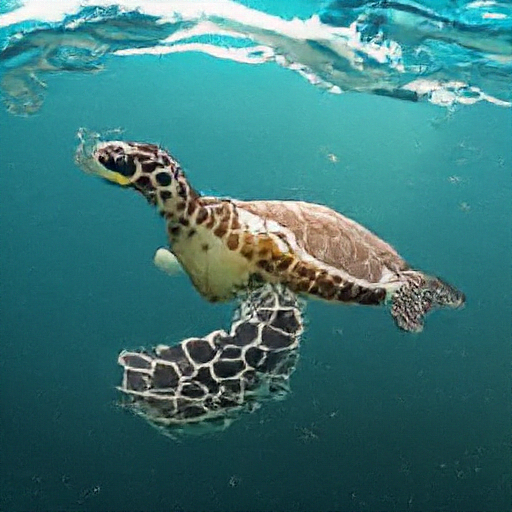}
        \end{subfigure}
        \\
        \begin{subfigure}[t]{0.32\linewidth}
            \includegraphics[width=\linewidth, height=0.94\linewidth,cfbox=blue 1pt 1pt]{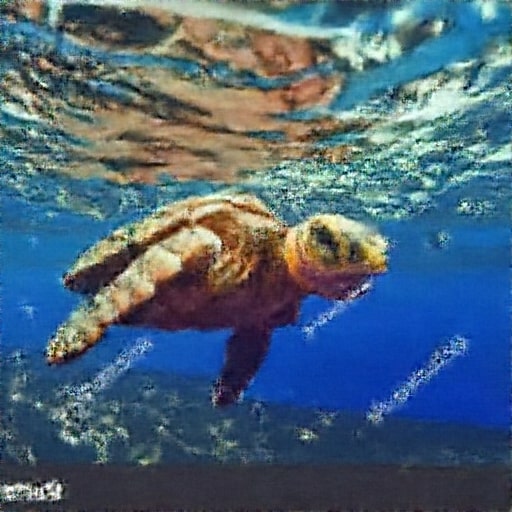}
            \caption*{\scriptsize{20k steps}}
        \end{subfigure}
        \begin{subfigure}[t]{0.32\linewidth}
            \includegraphics[width=\linewidth, height=0.94\linewidth,cfbox=blue 1pt 1pt]{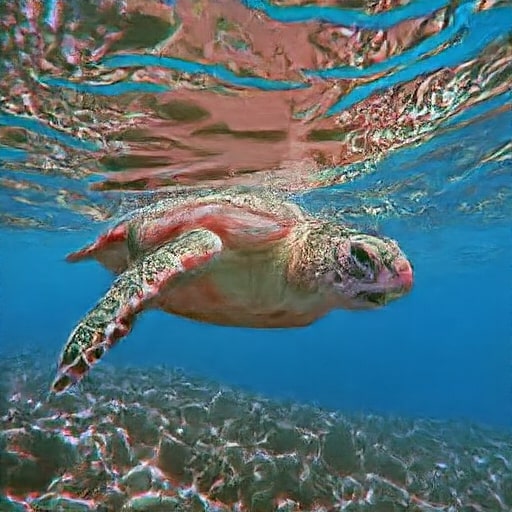}
            \caption*{\scriptsize{50k steps}}
        \end{subfigure}
        \begin{subfigure}[t]{0.32\linewidth}
            \includegraphics[width=\linewidth, height=0.94\linewidth,cfbox=blue 1pt 1pt]{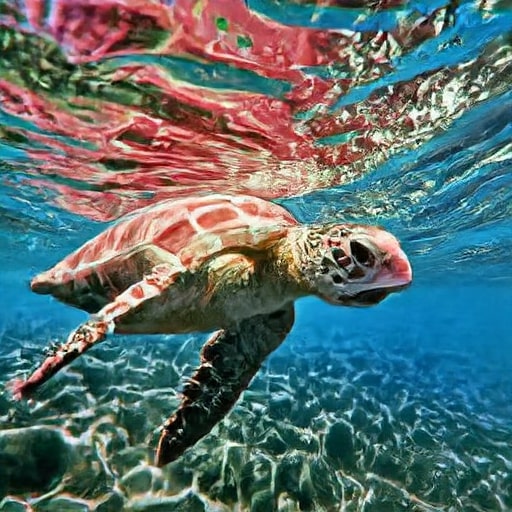}
            \caption*{\scriptsize{100k steps}}
        \end{subfigure}
    \end{subfigure}
\end{subfigure}%
}
\vspace{-.5em}
\caption{On catastrophic forgetting during early steps of finetuning: the pretrained representation quickly deteriorates due to noise introduced by the random weights from newly added layers. (from left to right) $64\times64$ image produced by the pretrained Shallow-Unet-Huge; followed by $512\times512$ images (in green)
 produced at early steps of finetuning (ft.) the core representation in an E2e model; and (in blue) freezing the core layers.
\emph{Differences better observed zooming in. Distinctions are more readily discerned when examining in closer detail.}
Prompts:
\emph{A close-up portrait of a butterfly, revealing the intricate patterns and textures on its wings in exquisite detail.}
\emph{A loving mother kangaroo carrying her joey in her pouch..}
\emph{A determined sea turtle swimming against the ocean current.}
\emph{A graceful hummingbird hovering near a bright pink flower.}
\emph{A dark and gothic illustration of a raven perched on a skull.}
\emph{A colorful macaw soaring through a lush, vibrant rainforest.}
\emph{A playful wolf pup chasing its own tail.}
}
\label{fig:forgetting}
\vspace{-1em}
\end{figure*}

We next explore greedy growing of \SU models to high resolution, non-cascaded models.
% non-cascaded UViT models targeting high-resolution images. 
We compare training models from scratch on the subset of the CC12M dataset filtered by the target resolution (512 pixels) with alternatives for reusing of the \CC pretrained on the full dataset. They validate our main intuitions behind the greedy growing algorithm, i.e., that the introduction of new, untrained layers, as well as shifts in the distribution of the training data are known causes of the catastrophic forgetting phenomena \cite{51359,kuo2023openvocabulary,YuHDSC23} possibly damaging the pre-trained representation.

Tables \ref{tab:e2e_metrics} and \ref{tab:uvit_e2e_vqva_summary} summarize performance as a function of model scale for greedy growing, along with various ablations of the training procedure.
%
% Quantitative results are presented on Tables \ref{tab:e2e_metrics} and \ref{tab:uvit_e2e_vqva_summary} taken on models of increasing size.
%
Our \emph{greedy growing} recipe with frozen \CC's and its optional defrosting phase lead to the best results across the metrics. 
The optional defrosting phase is required for improving the performance of the smallest model ablated (UViT-Base).
Its frozen counterpart showed  signs of underfitting during training, as it has a small number of trainable parameters (217M) in the added layers.
Under this low-capacity scenario, the defrosting phase offers a balance between protecting the \CC representation and the use of the model's full capacity, as it reduces the degradation of the pretrained representation by warming up the growth layers.
Other than this special case, the defrosting phase did not appear to benefit larger models.
% , being sub-optimal to their frozen counterpart according to the different metrics. 
%
These quantitative results agree with our hypothesis that the final model benefits from protecting the pretrained representation in our \emph{greedy growing} algorithm. 

\autoref{fig:forgetting} qualitatively compares generations obtained by finetuning and freezing the \CC.
Additional qualitative comparisons are shown in Appendix \ref{app:early_finetune_frozen}.
They illustrate the the benefits of protecting the \CC from the noise introduced when back-propagating through the randomly initialized growth layers. 
We observe that the low-resolution images produced by the use of the same representation under their original \SU models produce objects 
whose shapes and parts are correctly defined.

The high-resolution images generated from early steps (20k) of finetuning the \CC under the UVit architecture present objects with correct shapes superimposed with the diffusion noise.
Soon after that (around 50k-100k steps) 
the quality of object shapes and structure decays as the training backpropagates the noise introduced by the growth layers through the pretrained representation.

Under the \emph{greedy growing} regime and same number of training steps (20k steps) the frozen model is able to produce objects with correct shapes and parts, and maintain their composition as training progresses. 
Another direct side effect of maintaining the \CC representation is the fast reduction of the diffusion noise early in training. 
% The presented qualitative examples illustrate a phenomena observed on the different models ablated.

% Different from the work of \cite{Hoogeboom23}, the ablations in this section were performed under small batch size regime and do not include further regularization, aiming to stress stability of the optimization on our growing algorithm. 
% Under this setting it was not possible to train the XHuge model from scratch or finetune its layers due to numerical instability. 

% -> We investigate whether isolating the \CC pretraining provides a final indication of model performance.

\subsection{Guidance tuning}
\label{sec:guidance}
\vspace*{-0.2cm}

\begin{figure*}[htb]
\centering
\footnotesize{
\begin{subfigure}[c]{0.5\linewidth}
    \includegraphics[width=\linewidth]{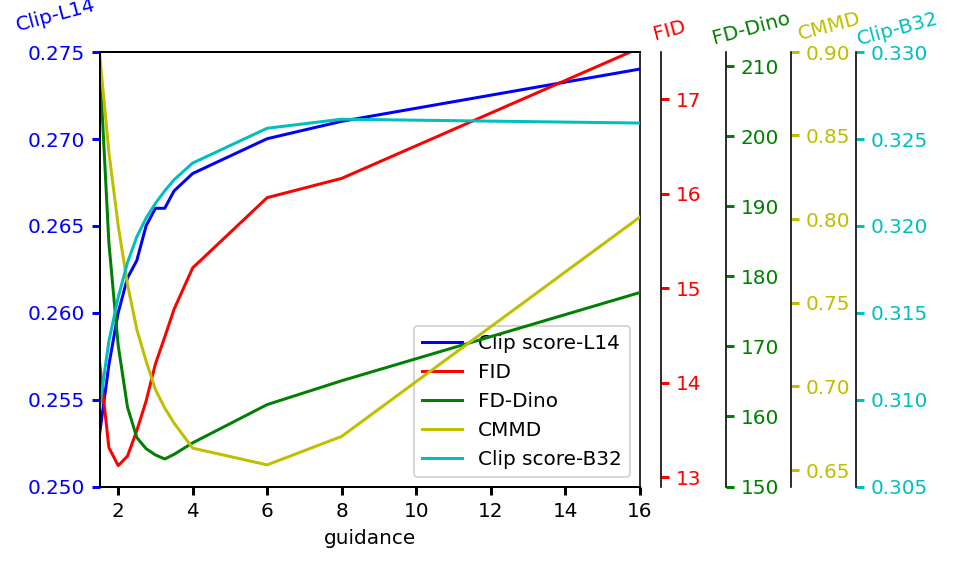}
\end{subfigure}
\begin{subfigure}{0.48\linewidth}
\setlength{\abovecaptionskip}{0pt}
    \begin{subfigure}[ht]{\linewidth}
        \begin{subfigure}[t]{0.32\linewidth}
            \includegraphics[width=\linewidth, height=0.95\linewidth,cfbox=red 1pt 1pt]{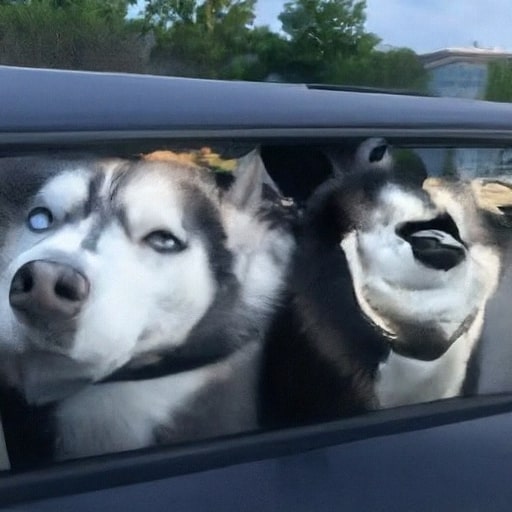}
        \end{subfigure}
        \begin{subfigure}[t]{0.32\linewidth}
            \includegraphics[width=\linewidth, height=0.95\linewidth]{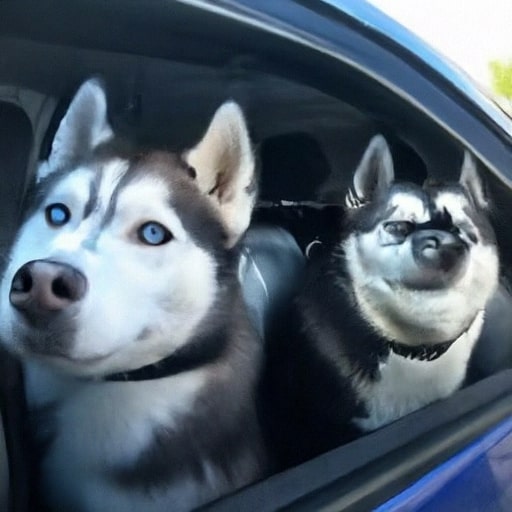}
        \end{subfigure}
        \begin{subfigure}[t]{0.32\linewidth}
            \includegraphics[width=\linewidth, height=0.95\linewidth,cfbox=green 1pt 1pt]{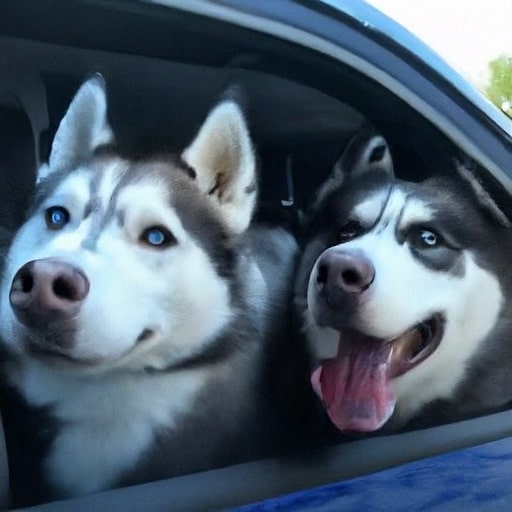}
        \end{subfigure}
        \begin{subfigure}[t]{0.32\linewidth}
            \includegraphics[width=\linewidth, height=0.95\linewidth]{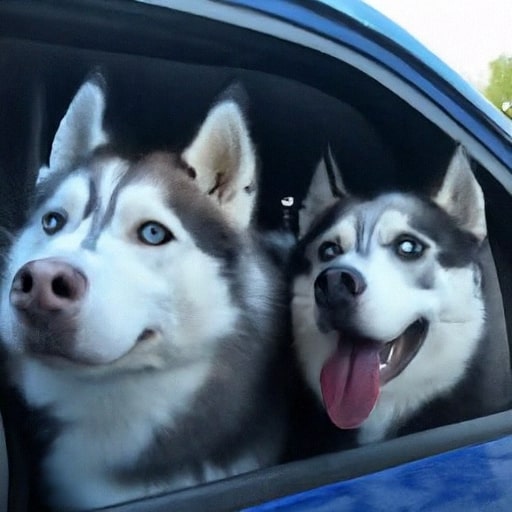}
        \end{subfigure}
        \begin{subfigure}[t]{0.32\linewidth}
            \includegraphics[width=\linewidth, height=0.95\linewidth,cfbox=yellow 1pt 1pt]{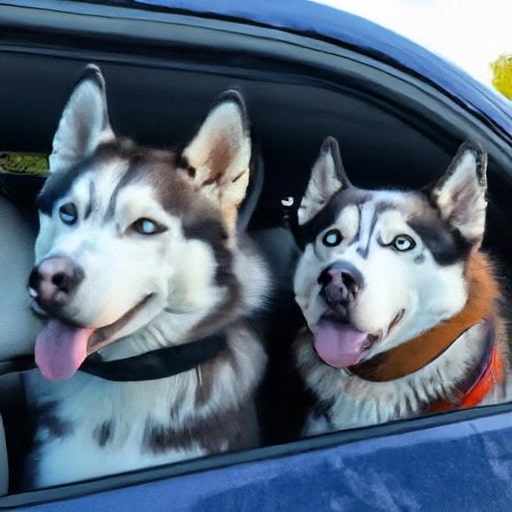}
        \end{subfigure}
        \begin{subfigure}[t]{0.32\linewidth}
          \includegraphics[width=\linewidth, height=0.95\linewidth,cfbox=cyan 1pt 1pt]{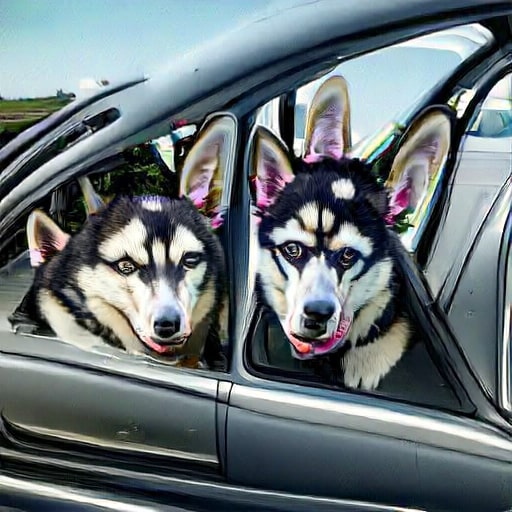}
        \end{subfigure}
    \end{subfigure}
\end{subfigure}
\vspace{-.5em}
\caption{On the FID-CLIP tradeoff and the use of SOTA feature spaces for image and text-alignment distributions.
(right) sample images with increasing guidance from left-to-right and top-to-bottom. Minimum FID in red box. Minimum FD-Dino in green. Minimum CMMD in yellow. In cyan: the saturation/cartoonish effect of  increasing CLIP score further in detriment of the other metrics.
\emph{Differences better observed zooming in. Distinctions are more readily discerned when examining in closer detail.}
Prompt (from MSCOCO captions): 
\emph{Two huskies hanging out of the car windows.}
% \df{let's try to make colors on the right the same as colors in the plot.}
% \CV{I thought they were. I plan to make them closer when I have a final version to the appearance of this plot.}
}
\label{fig:tradeoff}
\vspace{-1em}
}
\end{figure*}

\begin{figure*}[htb]
\centering
\footnotesize{
\begin{subfigure}{\linewidth}
\setlength{\lineskip}{0pt}
\setlength{\abovecaptionskip}{0pt}
\raisebox{+0.1in}{
\subcaptionbox*{}{\rotatebox[origin=b]{90}{{\small FID}}}}
\begin{subfigure}[t]{0.185\linewidth}
\includegraphics[width=\linewidth, height=0.95\linewidth]{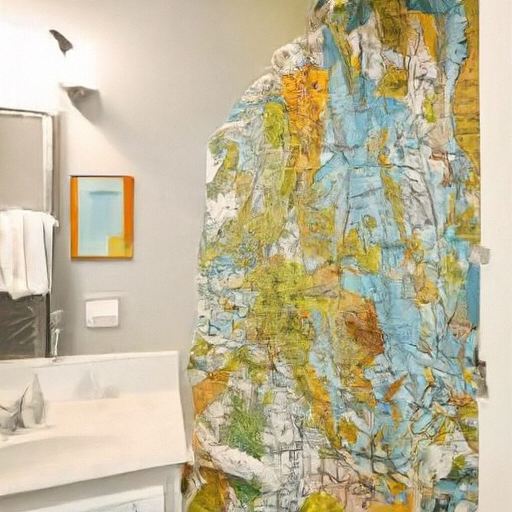}
\end{subfigure}%
\begin{subfigure}[t]{0.185\linewidth}
\includegraphics[width=\linewidth, height=0.95\linewidth]{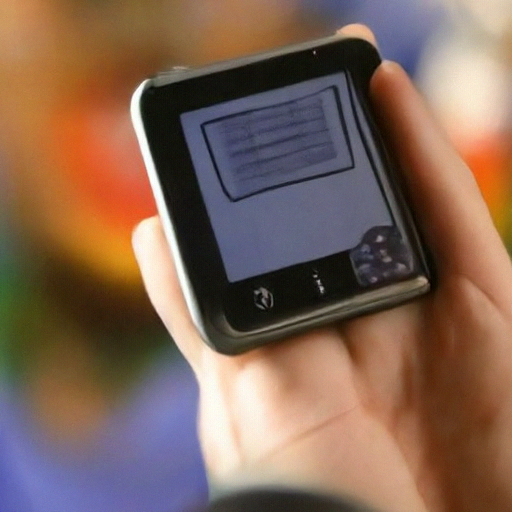}
\end{subfigure}%
\begin{subfigure}[t]{0.185\linewidth}
\includegraphics[width=\linewidth, height=0.95\linewidth]{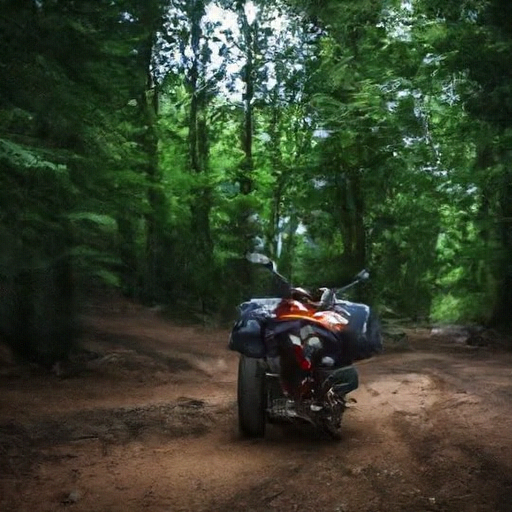}
\end{subfigure}%
\begin{subfigure}[t]{0.185\linewidth}
\includegraphics[width=\linewidth, height=0.95\linewidth]{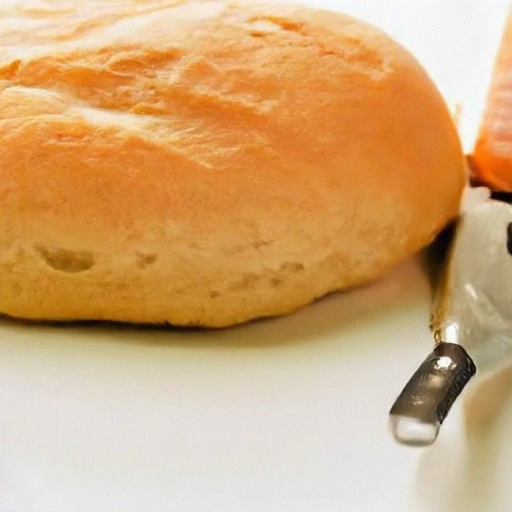}
\end{subfigure}%
\begin{subfigure}[t]{0.185\linewidth}
\includegraphics[width=\linewidth, height=0.95\linewidth]{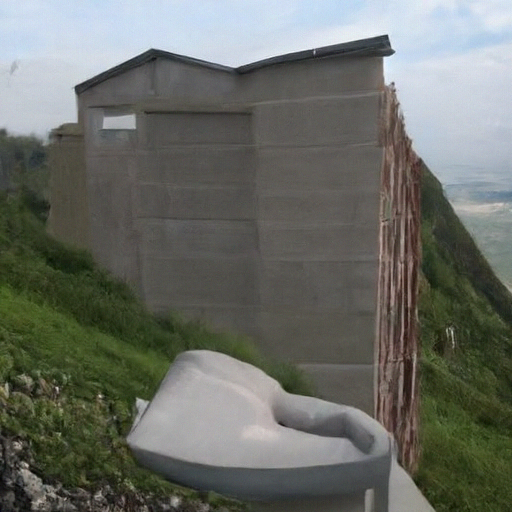}
\end{subfigure}%
\\
\raisebox{+0.1in}{
\subcaptionbox*{}{\rotatebox[origin=b]{90}{{\small FD-Dino}}}}
\begin{subfigure}[t]{0.185\linewidth}
\includegraphics[width=\linewidth, height=0.95\linewidth]{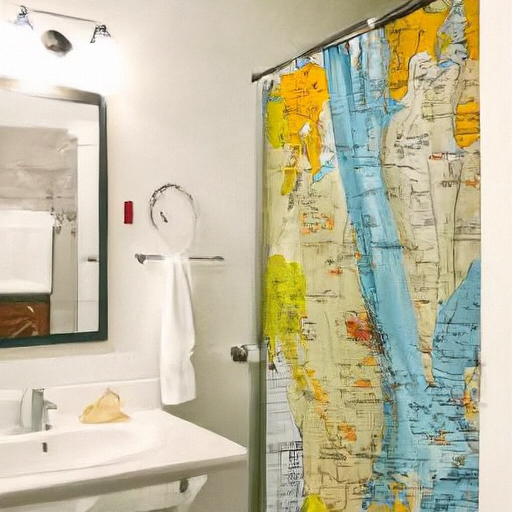}
\end{subfigure}%
\begin{subfigure}[t]{0.185\linewidth}
\includegraphics[width=\linewidth, height=0.95\linewidth]{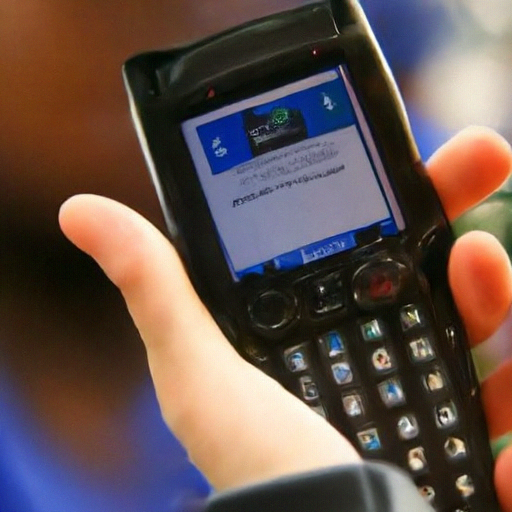}
\end{subfigure}%
\begin{subfigure}[t]{0.185\linewidth}
\includegraphics[width=\linewidth, height=0.95\linewidth]{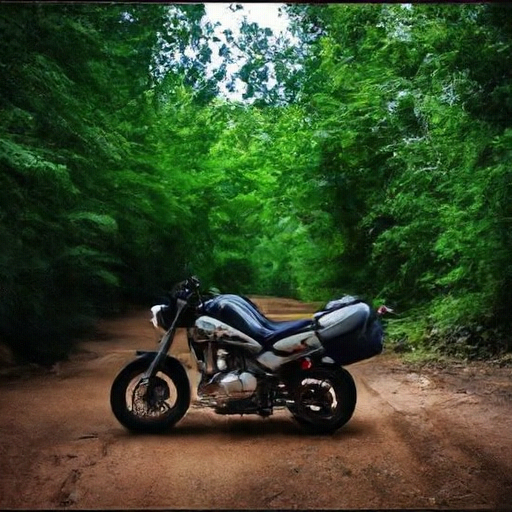}
\end{subfigure}%
\begin{subfigure}[t]{0.185\linewidth}
\includegraphics[width=\linewidth, height=0.95\linewidth]{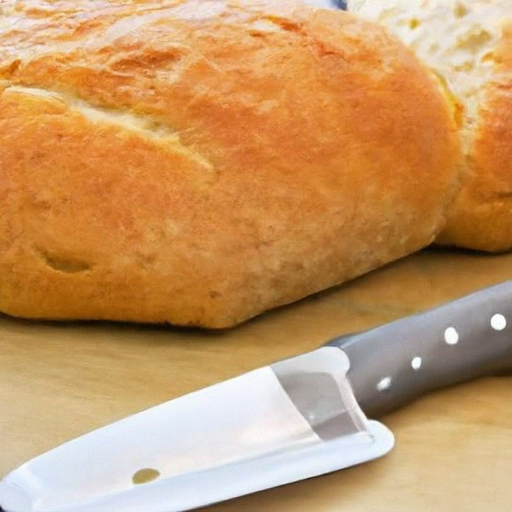}
\end{subfigure}%
\begin{subfigure}[t]{0.185\linewidth}
\includegraphics[width=\linewidth, height=0.95\linewidth]{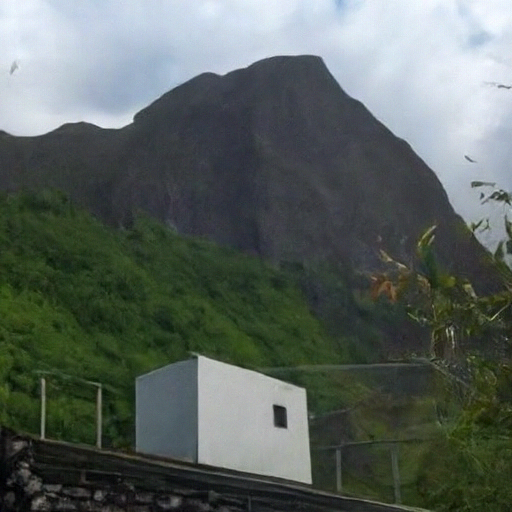}
\end{subfigure}%
\\
\raisebox{+0.1in}{
\subcaptionbox*{}{\rotatebox[origin=b]{90}{{\small CMMD}}}}
\begin{subfigure}[t]{0.185\linewidth}
\includegraphics[width=\linewidth, height=0.95\linewidth]{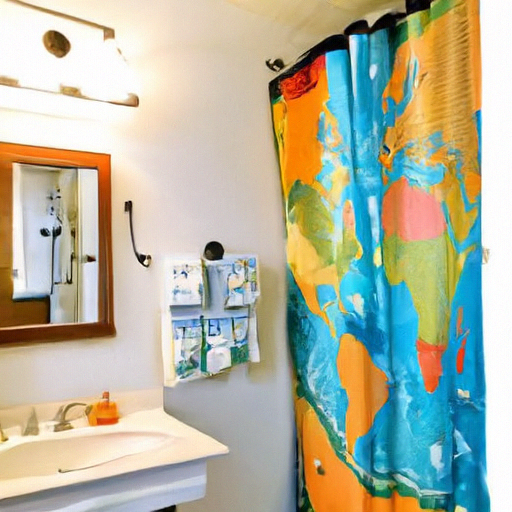}
\end{subfigure}%
\begin{subfigure}[t]{0.185\linewidth}
\includegraphics[width=\linewidth, height=0.95\linewidth]{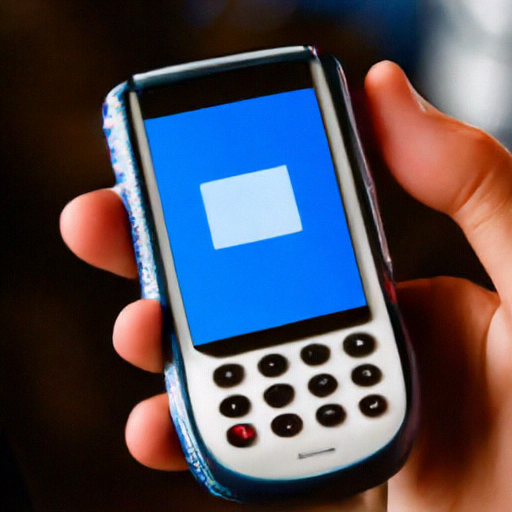}
\end{subfigure}%
\begin{subfigure}[t]{0.185\linewidth}
\includegraphics[width=\linewidth, height=0.95\linewidth]{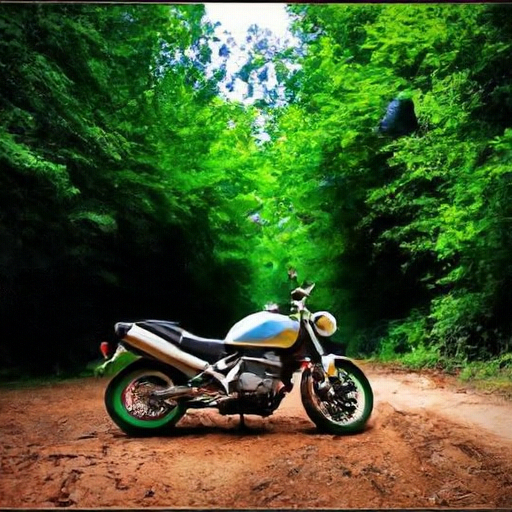}
\end{subfigure}%
\begin{subfigure}[t]{0.185\linewidth}
\includegraphics[width=\linewidth, height=0.95\linewidth]{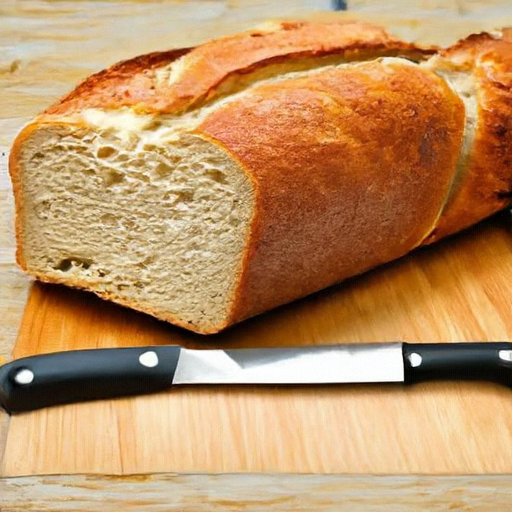}
\end{subfigure}%
\begin{subfigure}[t]{0.185\linewidth}
\includegraphics[width=\linewidth, height=0.95\linewidth]{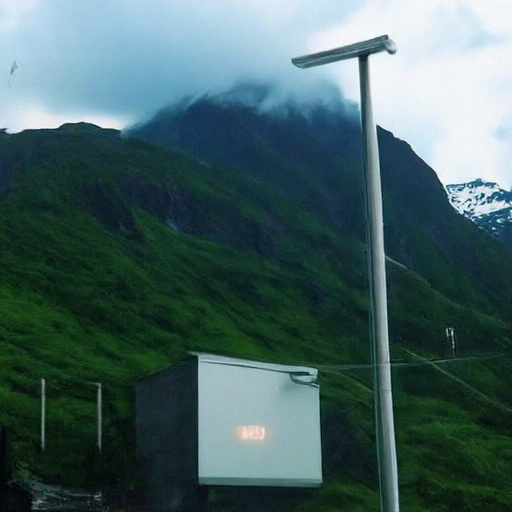}
\end{subfigure}%
\\
\raisebox{+0.1in}{
\subcaptionbox*{}{\rotatebox[origin=b]{90}{{\small CLIP}}}}
\begin{subfigure}[t]{0.185\linewidth}
\includegraphics[width=\linewidth, height=0.95\linewidth]{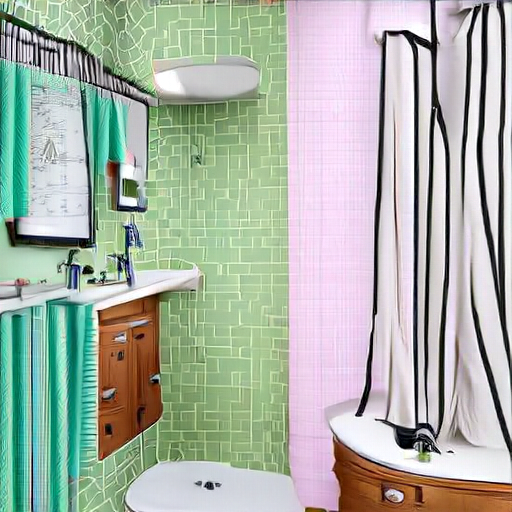}
\end{subfigure}%
\begin{subfigure}[t]{0.185\linewidth}
\includegraphics[width=\linewidth, height=0.95\linewidth]{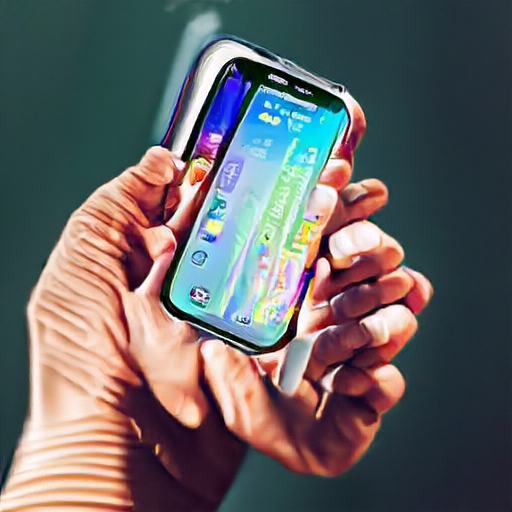}
\end{subfigure}%
\begin{subfigure}[t]{0.185\linewidth}
\includegraphics[width=\linewidth, height=0.95\linewidth]{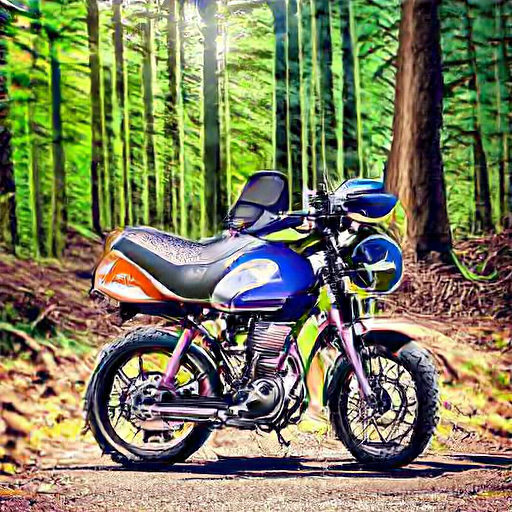}
\end{subfigure}%
\begin{subfigure}[t]{0.185\linewidth}
\includegraphics[width=\linewidth, height=0.95\linewidth]{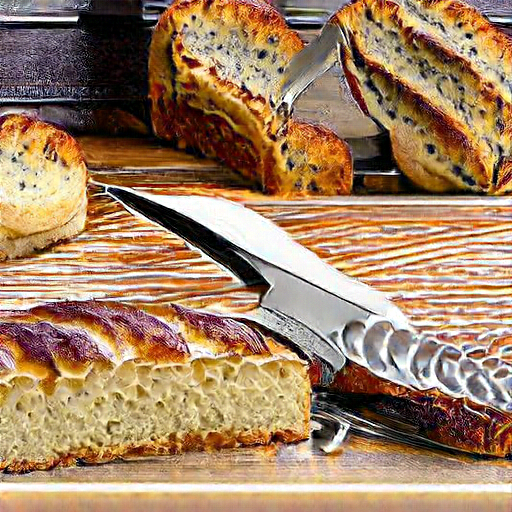}
\end{subfigure}%
\begin{subfigure}[t]{0.185\linewidth}
\includegraphics[width=\linewidth, height=0.95\linewidth]{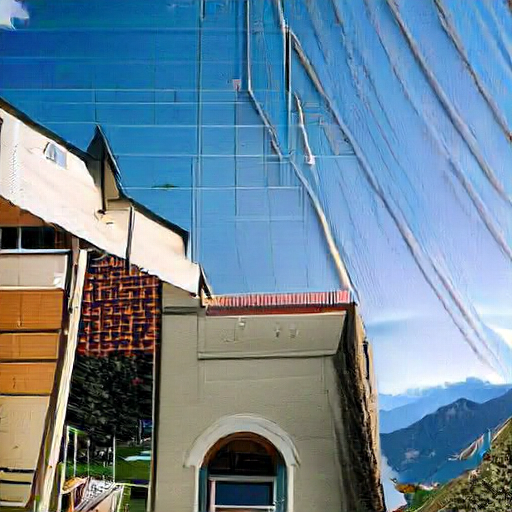}
\end{subfigure}%
\end{subfigure}%
\vspace{-.5em}
\caption{On the choice of the image distribution metric for calibrating guidance. First three rows contain samples from MSCOCO captions by minimizing respectively FID, FD-Dino and CMMD. 
The use of robust features is correlated with better shape and composition of images.
Prompts from MSCOCO caption:
i) {\em A bathroom with a sink and shower curtain with a map print.}
ii) {\em 4 different colored sea horses flying with 4 birds.}
iii) {\em A person holds a flip phone displaying the screen.}
iv) {\em A motorcycle is parked on a dirt road in a forest.}
v) {\em 
A stainless shiny serrated knife sits in front of a sliced loaf.
A restroom hanging off the side of a building over a mountain.}
}
\label{fig:min_fid}}
\vspace{-1em}
\end{figure*}

Diffusion model hyper-parameters affect both training and sampling quality. 
It is a common practice to tune the sampler guidance weights using \FID-\Clip trade-off curves \citep{saharia2022photorealistic,Hoogeboom23,sdxl}.
In doing so one aims to strike a balance between images quality (by minimizing \FID) 
and alignment with the text prompt (maximizing the \Clip score).
%
% Unsurprisingly, the end quality from this balance is affected by the flaws of the metrics chosen. 
That said, it is well known that FID does not correlate particularly well with human perception \citep{stein2023exposing,Human_Evaluation,jayasumana2023rethinking}, and large guidance weights are known to
increase CLIP-Score but tend to produce over-sharpened, high-contrast images and unrealistic objects \citep{ho2021classifierfree,Saharia2022}.
Due to such limitations, despite widespread use of \FID-\Clip scores for performance comparisons, in practice they are adopted as loose measure of performance, and guidance weights are typically set through qualitative inspection. 

Here we explore alternative metrics for hyper-parameters tuning, aiming to better reflect their deployment use, and ultimately human perception.
These include recent measures with alternative feature spaces that exhibit better robustness in classification tasks, and  align somewhat better with human judgements of image quality and alignment. 
More specifically, we investigate the use of FD-Dino and CMMD as alternatives to FID in the calibration of the guidance hyper-parameter. 
\autoref{fig:tradeoff}  plots the response curve of different metrics as a function of guidance weight. They were measured using our UVIT-XHuge frozen model taken over 30k samples from the MSCOCO-caption validation set.
It illustrates that the three image distribution metrics are minimized by 
very different guidance values.
Similar curves are observed on the other models and training modalities, in which the best guidance value for minimizing FID, FD-Dino and CMMD are in increasing order. 
\autoref{fig:min_fid} further illustrates samples obtained at the optimal values for each metric, and also when using the maximum guidance tested (16) for increasing \Clip even further.

A qualitative analysis shows that by minimizing FID, one favors the generation of natural colors and textures, but under closer inspection, it fails to produce realistic object shapes and parts. 
We conjecture that this matches prior observations on the existence of texture vs shape bias by image classifiers \citep{geirhos2018imagenettrained}.
Guidance values minimizing Dino-v2 features, on the other hand, appear to produce natural color distributions and objects with natural  shapes and composition.
We adopt the value at this minimum as our new lower bound. 
Increasing guidance from that value tends to increase color-contrast and sharpening.

Images produced with guidance weights minimizing
CMMD tend to produce images with initial signs of saturated colors and over-sharpening.
Given its use of Clip features for image distribution comparison, this agrees with previous observations on \Clip.
% , given their use of similar embedding for text-image comparison.
But unlike \Clip curves, CMMD curves present an inflection point within the range investigated. We use this inflection point to define a closed range for our search of reasonable guidance weights. 
That is, the range of guidance weights between FD-Dino and CMMD minimums was observed to strike a balance between producing correct shapes and aesthetically pleasing images characterized by enhanced color contrast and sharp edges.

All results presented in this section have their image generated using guidance weights
within the FD-Dino/CMMD trade-off range.
The specific value selected was taken at the intersection of the optimal ranges of models under the same comparison.
Following this approach, our \SU results were obtained with guidance weights fixed at 1.75, and their corresponding UViT models with guidance 4.0.   

\section{A full diffusion pipeline: Vermeer}
\label{sec:vermeer}

% \vspace{-1cm}
\begin{figure*}[tb!]
\centering
\begin{subfigure}[t]{0.92\textwidth}
\setlength{\lineskip}{0pt}
\begin{subfigure}[b]{0.1225\textwidth} 
\setlength{\lineskip}{0pt}
\includegraphics[width=\textwidth,height=\textwidth]{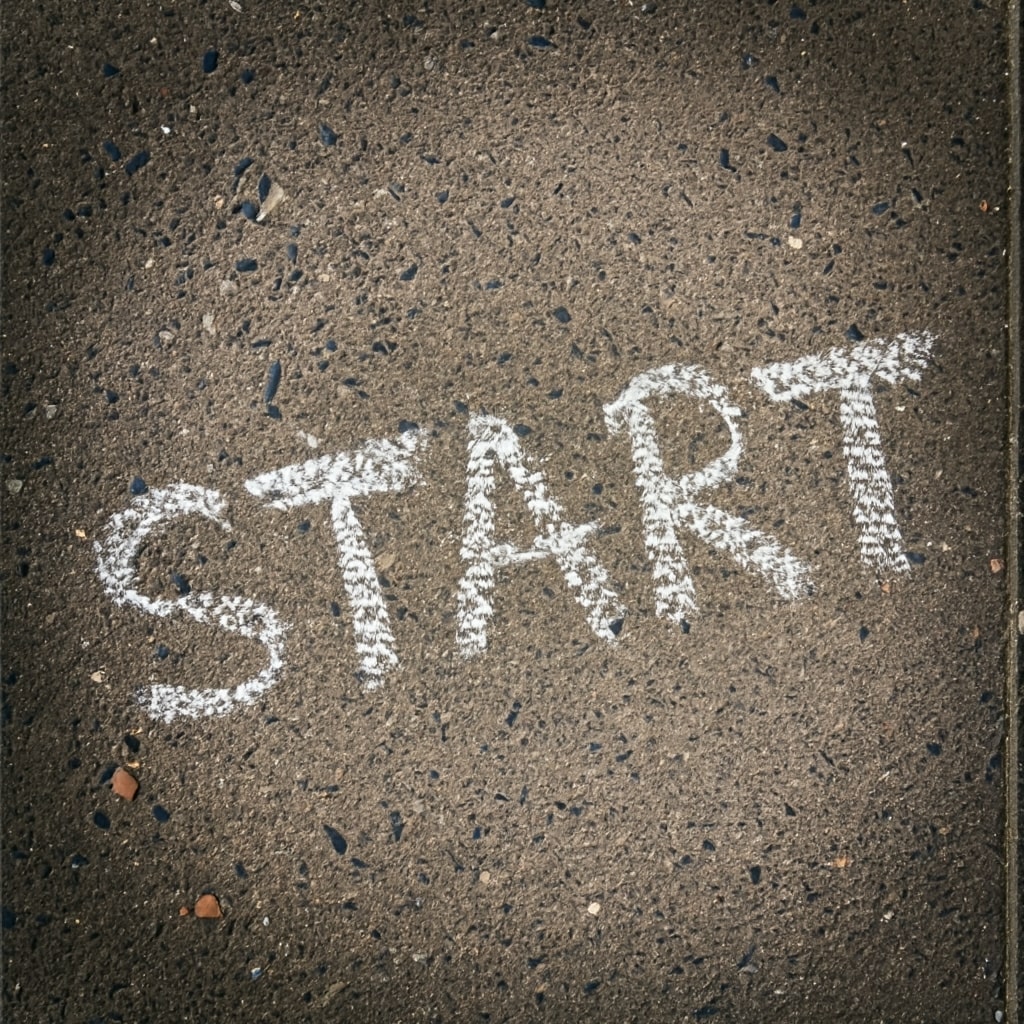}\\
\includegraphics[width=\textwidth,height=\textwidth]{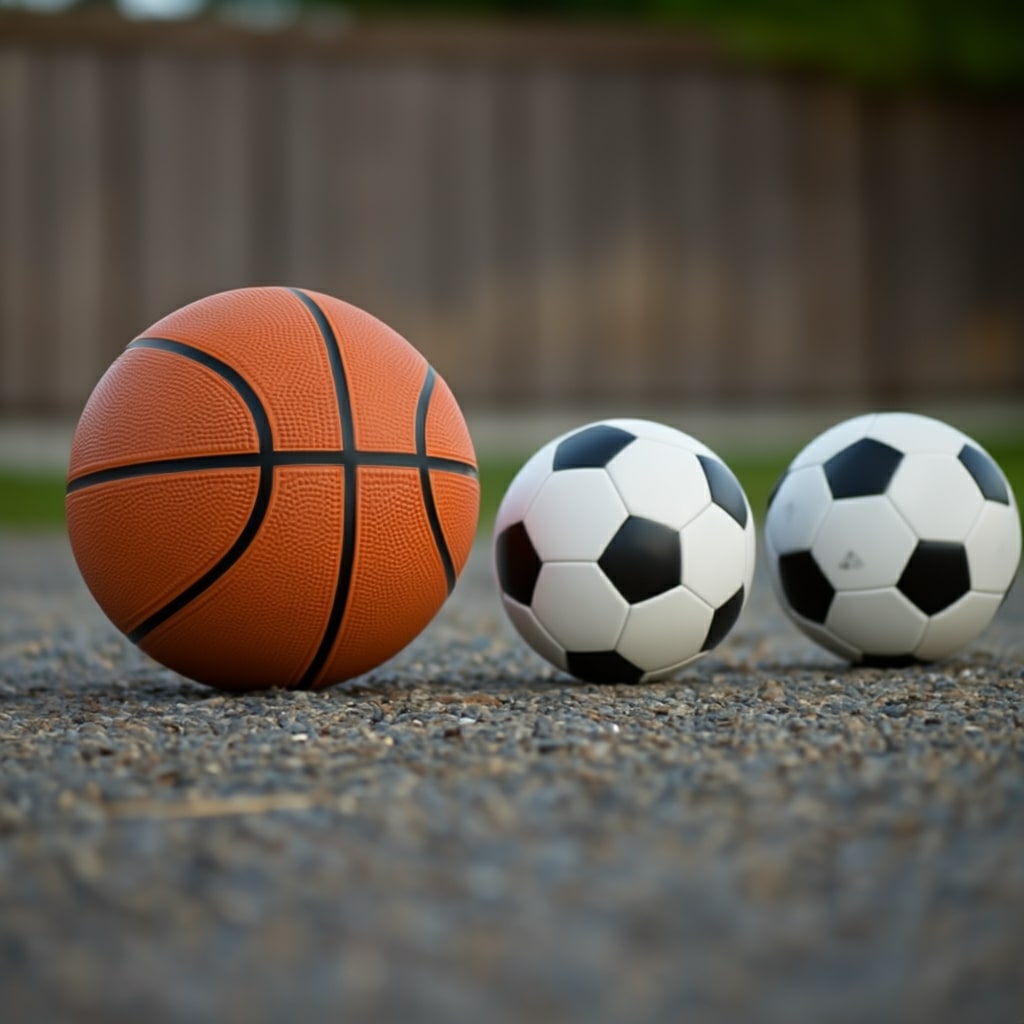}%
\end{subfigure}%
\begin{subfigure}[b]{0.245\textwidth}
\includegraphics[trim={0 1cm 0 0},clip,width=\textwidth,height=\textwidth]{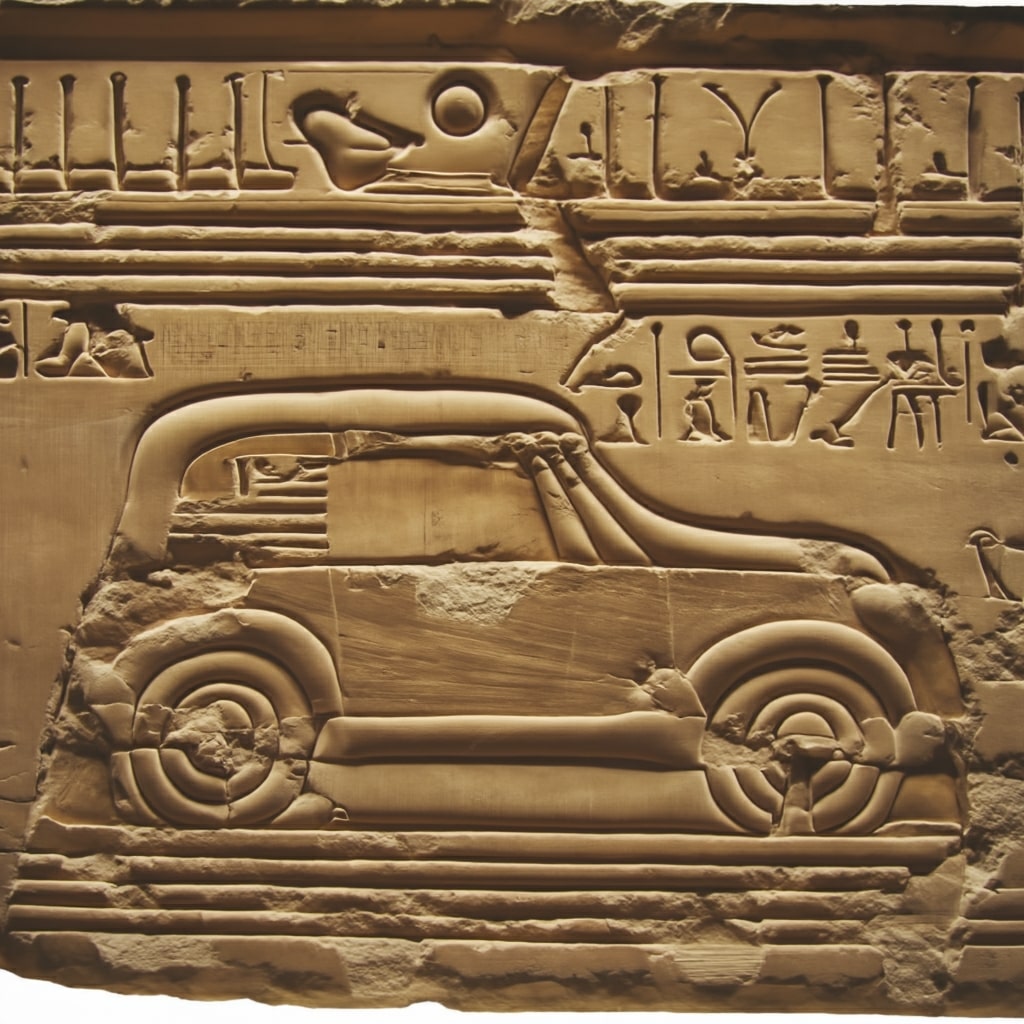} 
\end{subfigure}%
\begin{subfigure}[b]{0.2265\textwidth} 
\setlength{\lineskip}{0pt}
\includegraphics[width=\textwidth,height=0.54\textwidth]{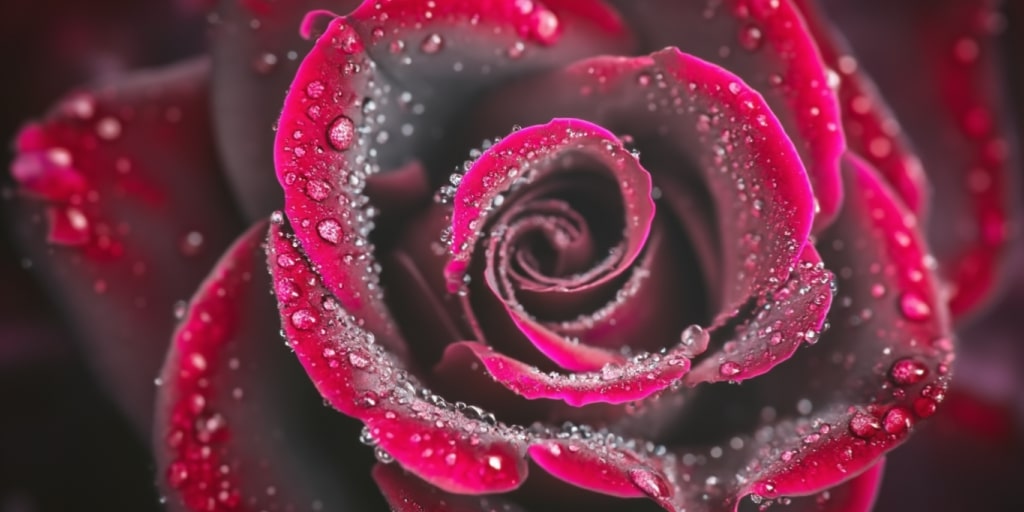}\\
\includegraphics[width=\textwidth,height=0.54\textwidth]{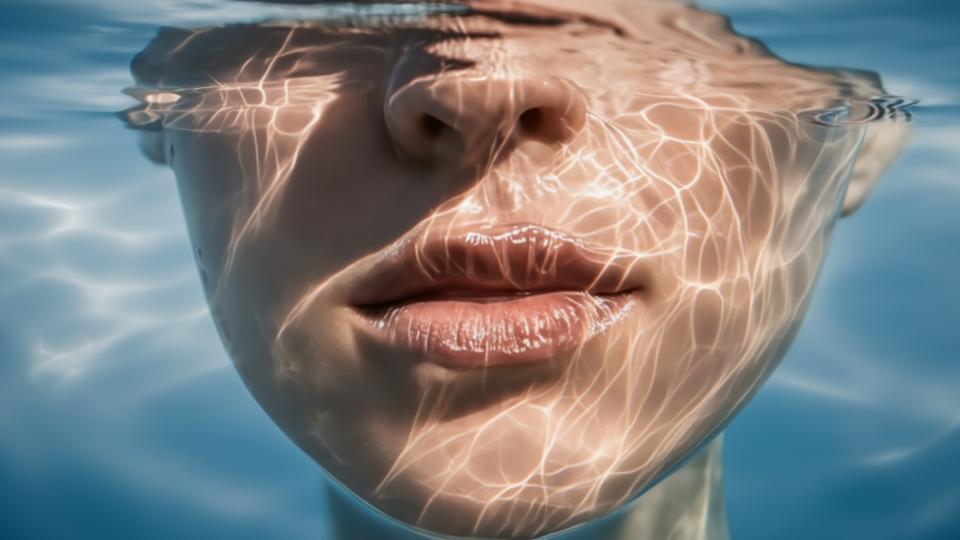}%
\end{subfigure}%
\begin{subfigure}[b]{0.245\textwidth} 
\includegraphics[width=\textwidth,height=\textwidth]{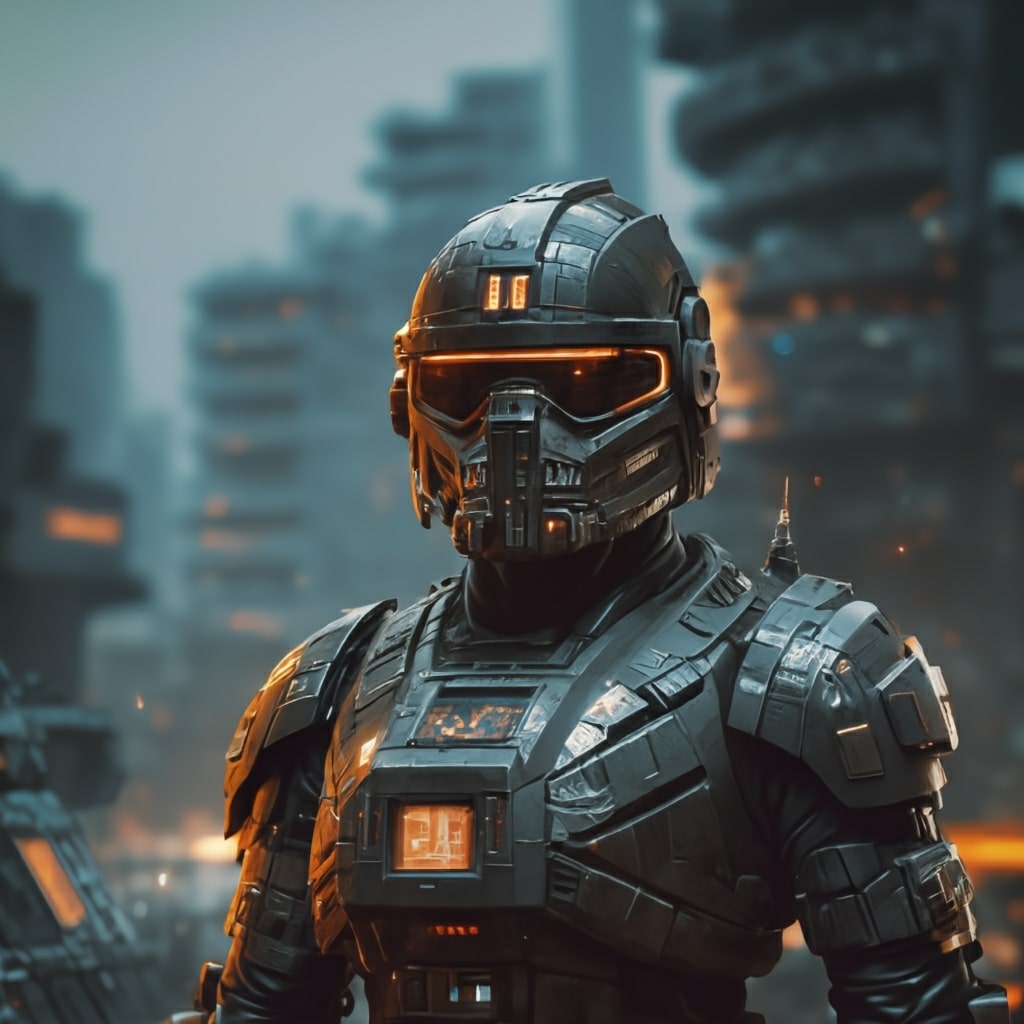}
\end{subfigure}%
\begin{subfigure}[b]{0.1455\textwidth} 
\setlength{\lineskip}{0pt}
\includegraphics[width=\textwidth]{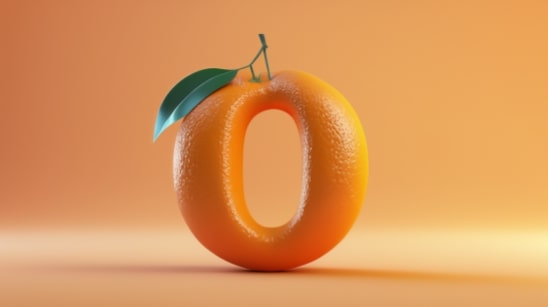}\\
\includegraphics[width=\textwidth]{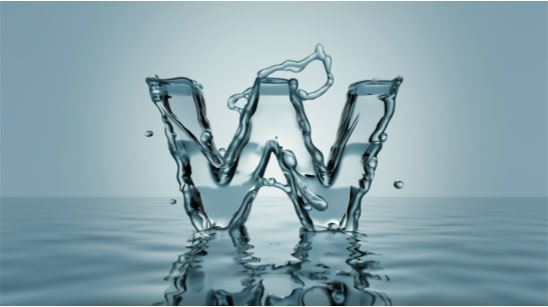}\\
\includegraphics[width=\textwidth]{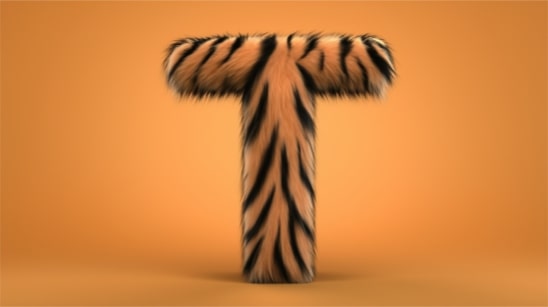}
\end{subfigure}%
\\
\begin{subfigure}[b]{0.1226\textwidth} 
\setlength{\lineskip}{0pt}
\includegraphics[width=\textwidth,height=2.085\textwidth]{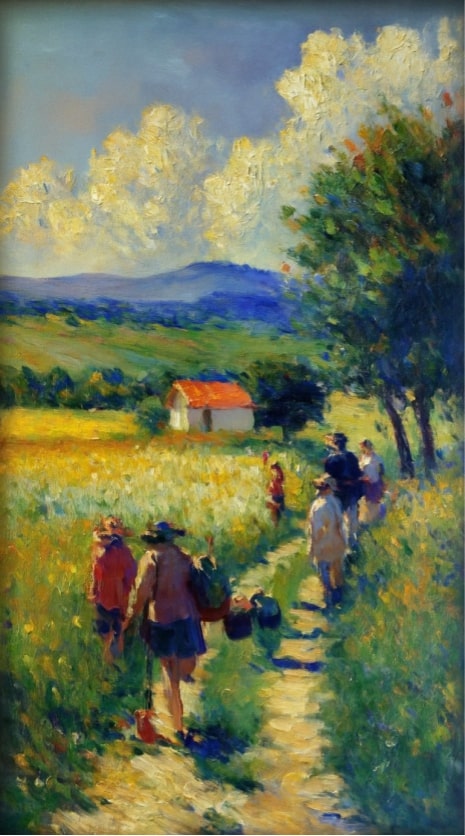}\\
\includegraphics[width=\textwidth,height=2.085\textwidth]{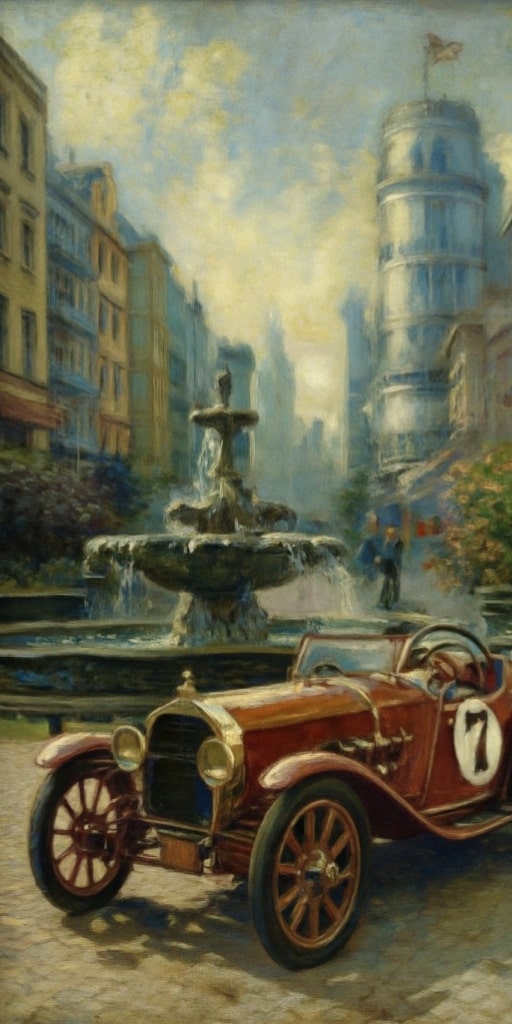}
\end{subfigure}%
\begin{subfigure}[b]{0.246\textwidth} 
\setlength{\lineskip}{0pt}
\includegraphics[width=\textwidth,height=1.04\textwidth]{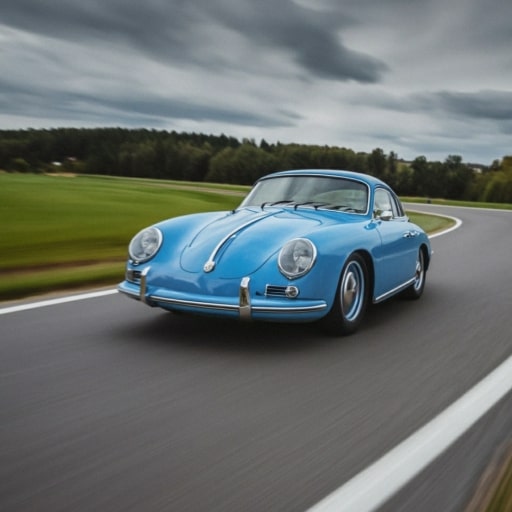}\\
\includegraphics[width=\textwidth,height=1.04\textwidth]{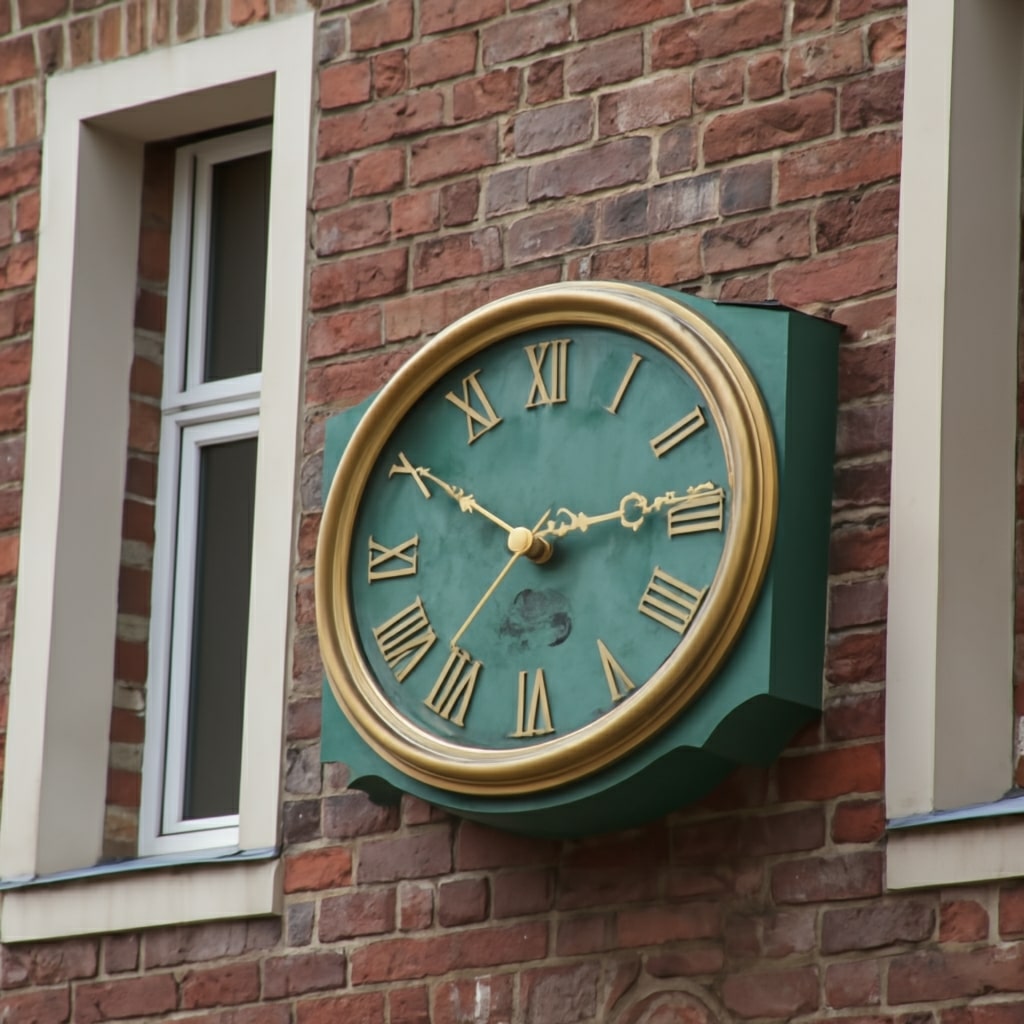}
\end{subfigure}%
\begin{subfigure}[b]{0.2265\textwidth}
\setlength{\lineskip}{0pt}
\includegraphics[width=\textwidth,height=0.565\textwidth]{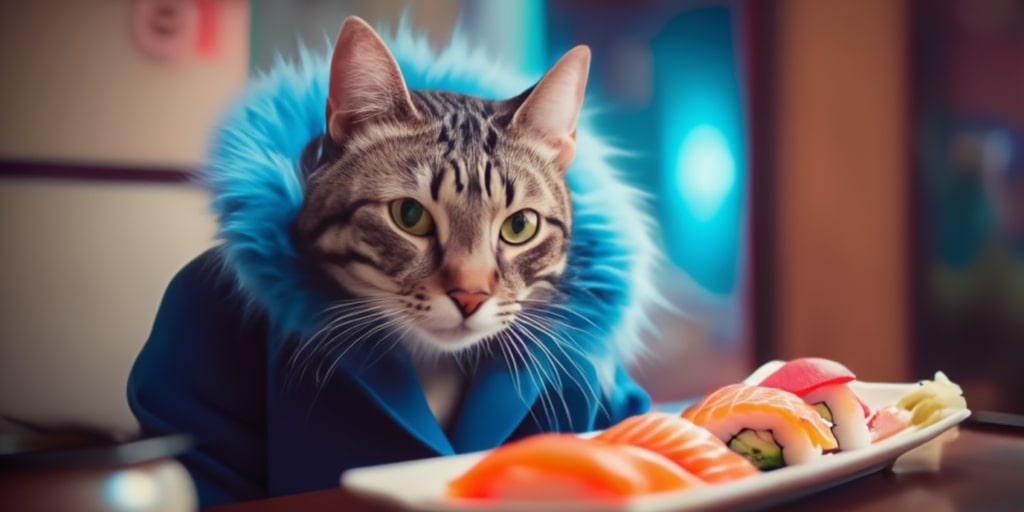}\\
\includegraphics[width=\textwidth,height=0.565\textwidth]{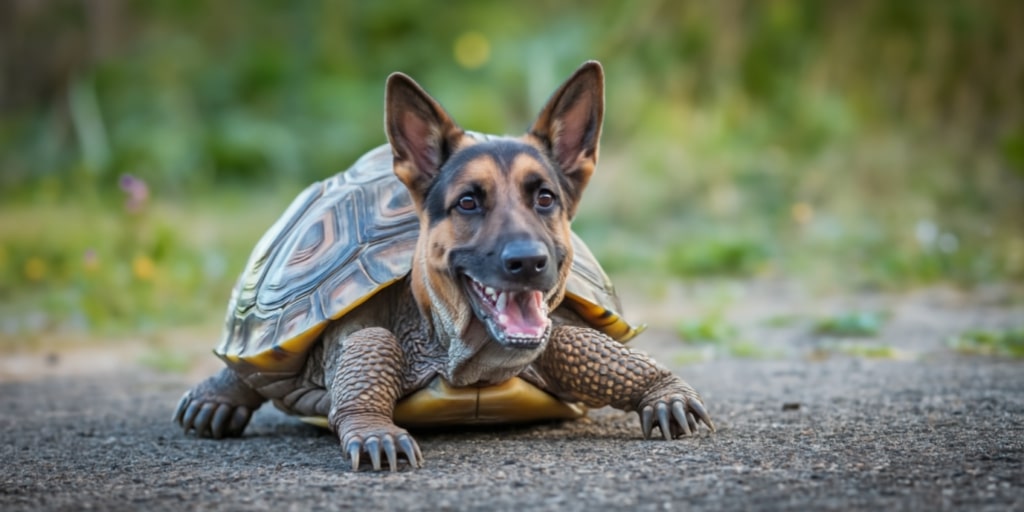}\\
\includegraphics[width=\textwidth,height=0.565\textwidth]{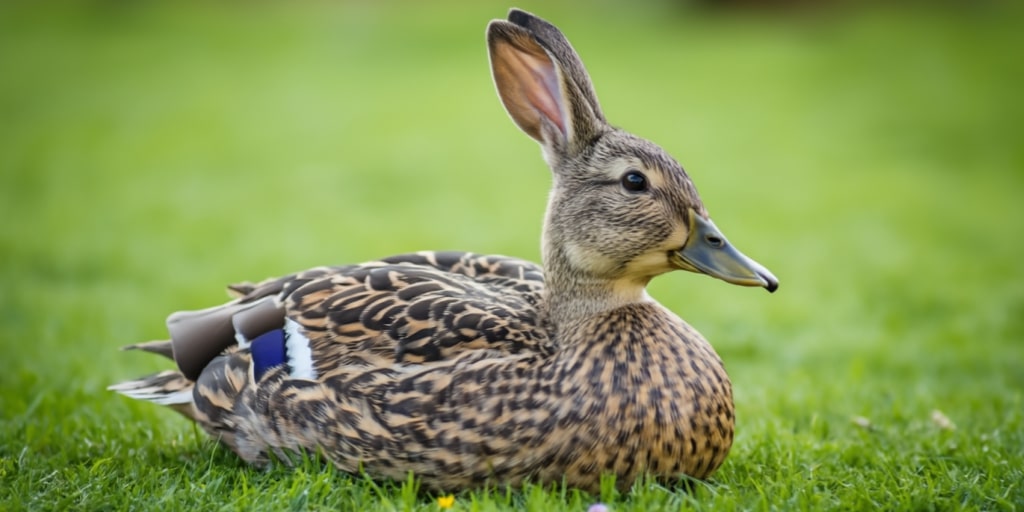}\\
\includegraphics[width=\textwidth,height=0.565\textwidth]{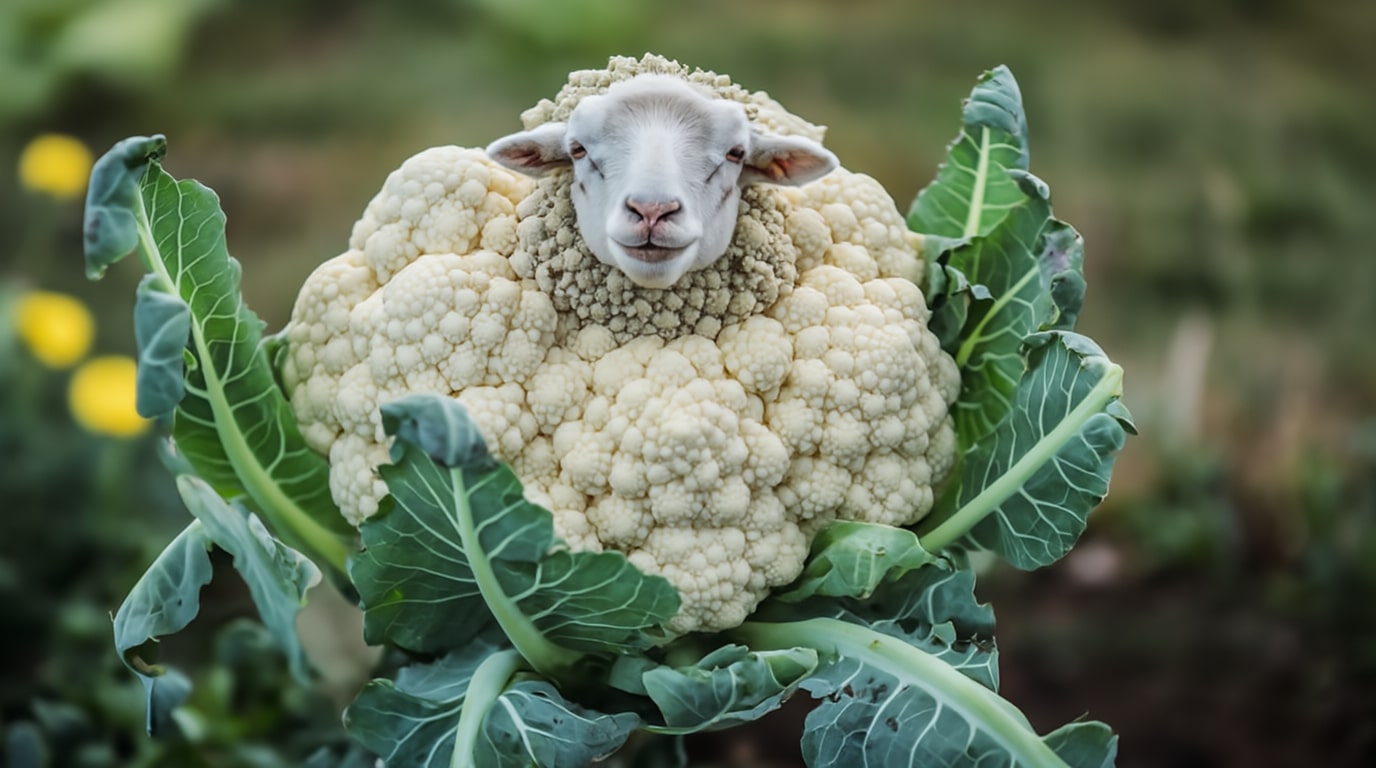}
\end{subfigure}%
\begin{subfigure}[b]{0.245\textwidth}
\setlength{\lineskip}{0pt}
\includegraphics[width=\textwidth,height=0.5425\textwidth]{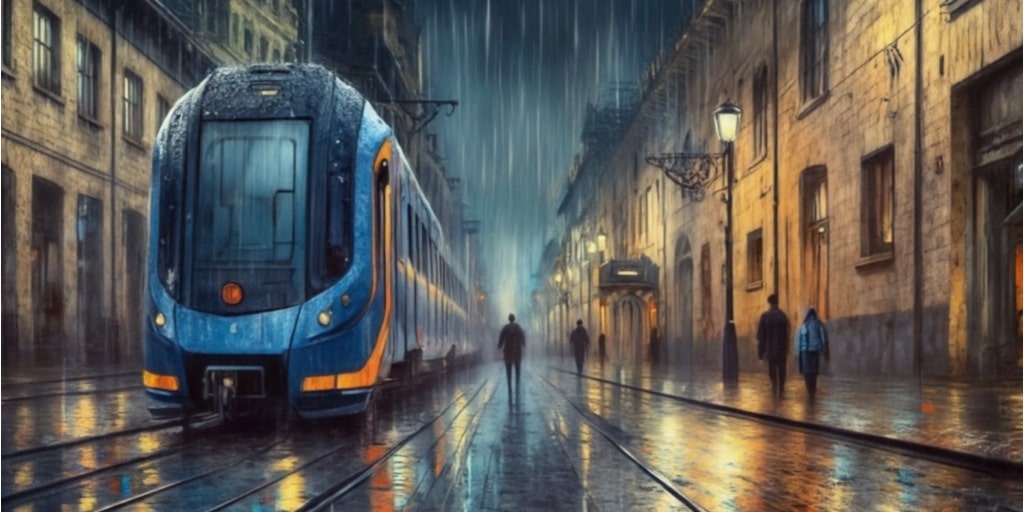}
\includegraphics[width=\textwidth,height=0.985\textwidth]{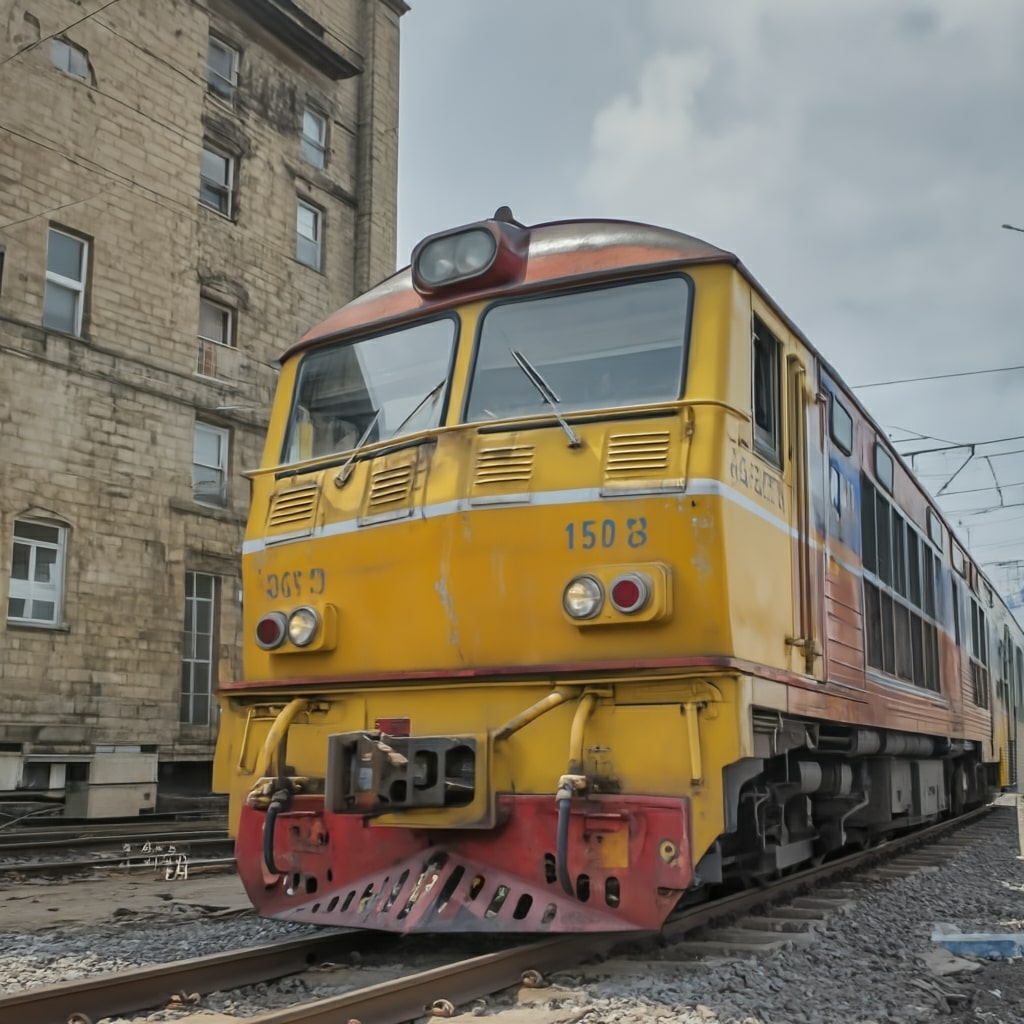} 
\includegraphics[width=\textwidth]{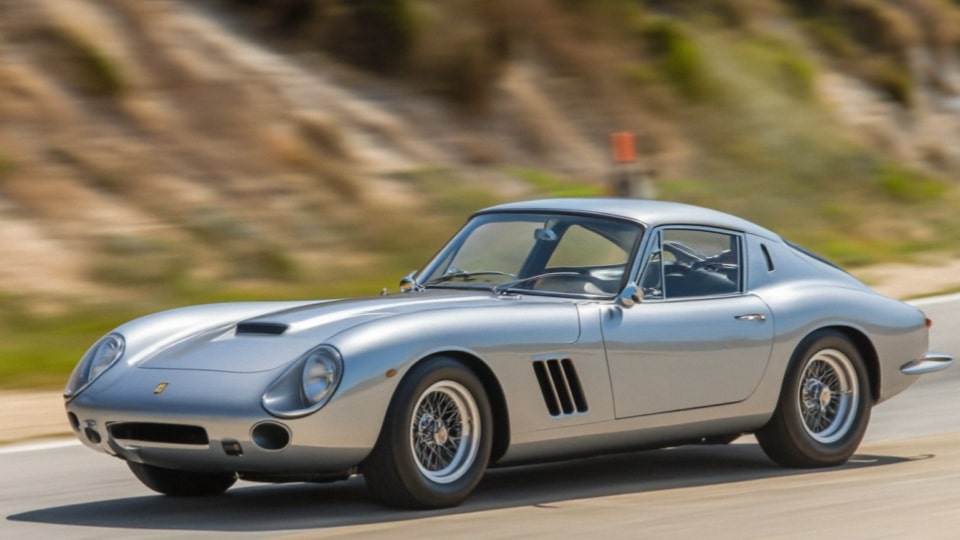} 
\end{subfigure}%
\begin{subfigure}[b]{0.143\textwidth}
\setlength{\lineskip}{0pt}
\includegraphics[width=\textwidth,height=1.79\textwidth]{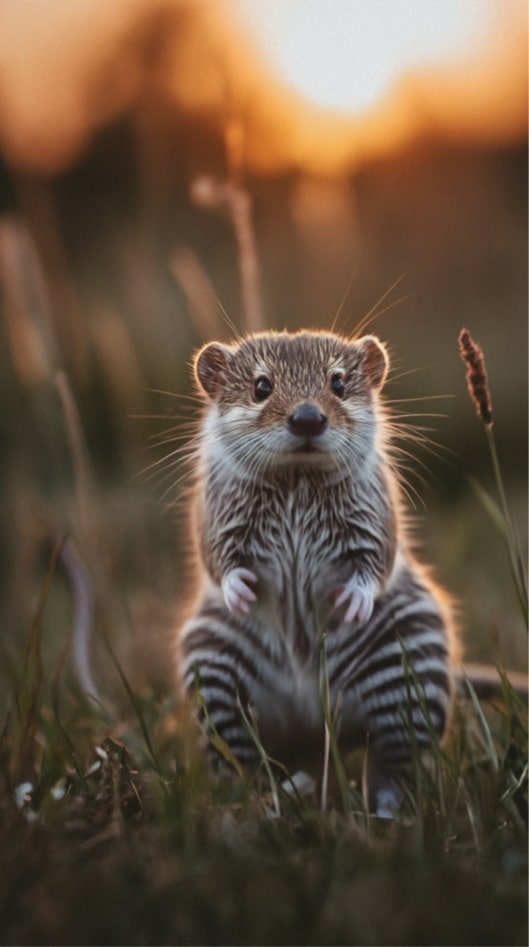}%
\\
\includegraphics[width=\textwidth,height=1.79\textwidth]{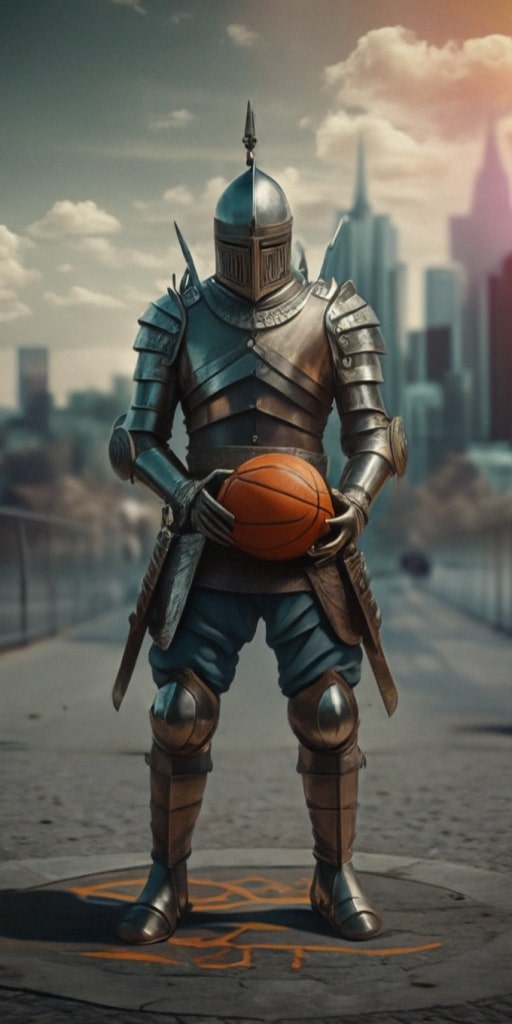}%
\end{subfigure}%
\end{subfigure}
% \vspace*{-0.1cm}
\caption{Images generated with our model \Vermeer. (See Appendix \ref{app:teaser_prompts} for the prompts.)}
\label{fig:teaser_vermeer_images}
% \vspace*{1cm}
\end{figure*}

Vermeer is an 8B parameter model grown from 256 to 1024 pixel resolution. The UViT architecture is similar to our UViT-Huge model (\autoref{tab:uvit_final_params}), except that its bottom layers operate at a grid of $32x32$ and with 32 transformer blocks in total. 
We found that allocating transformer blocks at 32x scale improves details (like small faces).
For Vermeer's text encoding, in addition to T5-XXL~\citep{2020t5} and Clip~\citep{clipradford21a}
embeddings previously mentioned, we also include a ByT5~\citep{xue2022byt5} encoder with 256 sequence length, resulting in a final embedding with sequence length of 461. 

The baseline version (\emph{Vermeer raw model}) is trained with 2k batch size at 256 resolution for 2M iterations, and grown to 1k resolution and finetuned for an additional 1M steps. As illustrated in \autoref{fig:teaser_vermeer_images}, it supports 3 aspect ratios, i.e., $1024\times 1024$, $768 \times 1376$, and $1376 \times 768$ thought aspect ratio bucketing ~\citep{blog}.
Once the \emph{raw model} is trained, we apply the following extra steps to improve the aesthetics of the generated images:
\begin{itemize}
\setlength\itemsep{0em}
    \item Style finetuning.
    We train an image classifier based on images that conform to aesthetic and compositional attributes like those described in \citep{dai2023emu}, and use it to select 3k images from our training data as a fine-tuning set. 
    We then fine-tune for 8K steps with a mixture of the original data and the aesthetic subset. 
    We condition the model on the aesthetic subset by adding a token to the text prompt. We found that finetuning the pixel model with a mixture of pretraining and finetuning data is needed to avoid catastrophic forgetting and to avoid the introduction of additional artifacts.
    
    \item Distillation.
    The vanilla Vermeer model adopts 256-step sampling process, making it computationally expensive for real-world use. We employed the multistep consistency model (MCM)~\citep{heek2024multistep} to distill style-tuned Vermeer to 16 steps, achieving a substantial 16x speedup while maintaining high visual quality. 
\end{itemize}
\subsection{Vermeer results}
\label{sec:results_vermeer}

%%%%%%%%%%%%%%%%%%%%%%%%%%%%%%%%%
%%%%%%%%%%%%%%%%%%%%%%%%%%%%%%%%% No use of Pil resize
%%%%%%%%%%%%%%%%%%%%%%%%%%%%%%%%%
\begin{table*}[tbh!]
\vspace*{-0.1cm}
\centering
\resizebox{0.7\linewidth}{!}{
\begin{tabular}{lcccc|c} 
\hline
 \hline
 model & & FID$_{30k}\downarrow$ & FD-Dino$_{30k}\downarrow$ &  CMMD$_{30k}\downarrow$ &  CLIP$_{score}\uparrow$\\
\hline
\hline
\small{SDXL$_{v1.0}$} & &
%%%%%%%%%%%%%%%%%%%%%%%%%%%%%%%%%
 {\bf 13.19} & 185.57 & 0.898 &  {\bf 0.279}
 \\
% \small{OTHER MODELS} & 
% %%%%%%%%%%%%%%%%%%%%%%%%%%%%%%%%%
%  fid & fd-dino & cmmd
%  \\
% \small{OTHER MODELS} & 
% %%%%%%%%%%%%%%%%%%%%%%%%%%%%%%%%%
%  fid & fd-dino & cmmd
%  \\
%%%%%%%%%%%%%%%%%%%%%%%%%%%%%%%%%
\small{Vermeer} & 
%%%%%%%%%%%%%%%%%%%%%%%%%%%%%%%%%
%%%%%%%%%%%%%%%%%%%%%%%%%%%%%%%%%
\multicolumn{1}{r}{
\emph{raw model}} &
16.26 & {\bf 185.25} & {\bf 0.631}
% clip: 
& 0.270
\\
&\multicolumn{1}{r}{ \emph{+prompt engineering}} &
17.33 & 216.01 & 0.867
% clip: 
&  0.269
\\
&\multicolumn{1}{r}{\emph{+style tuning}} &
24.51 & 336.25 & 1.167
% clip:
& 0.262
\\
%%%%%%%%%%%%%%%%%%%%%%%%%%%%%%%%%
% &\multicolumn{1}{r}{\emph{distilled (old v)}} & 
%  26.31 & 357.25 & 0.991
%  \\
&\multicolumn{1}{r}{\emph{distilled}} & 
25.97 & 347.19 & 0.885 
% clip:
& 0.261
\\
\hline
 \end{tabular}}
\vspace*{0.1cm} 
\caption{Image distribution metrics evaluated on 30k samples of MS-COCO. The raw Vermeer model minimizes  distribution metrics that adopt feature spaces from SOTA models (FD-Dino uses Dino-v2 while CMMD adopts Clip features), while tuning it to produce aesthetically pleasing images intentionally diverges from MSCOCO distribution.
}
\label{tab:vermeer_metrics}
\end{table*}
% imgrid in :
% https://imgrid.googleplex.com/?qq=/cns/oi-d/home/earthsea-dev/olwang/rs=6.3/sxs/imagen_2_5_vermeer_test/98046092/*/.imgrid&kvg=prompt&r=2

\begin{table*}[tbh!]
\vspace*{-0.1cm}
\centering
\resizebox{0.8\linewidth}{!}{
\begin{tabular}{lrcccc|c} 
\hline \hline
&&\multicolumn{5}{c}{DSG$\uparrow$}
\\
model & &Entities & Relations & Attributes & Global 
%& mSpice$_{score}$ 
& DSG \\
\hline 
%%%%%%%%%%%%%%%%%%%%%%%%%%%%%%%%%%%%%%%%%%%%%%%%%%%
%%%%%%%%%%%%%%%%%%%%%%%%%%%%%%%%%%%%%%%%%%%%%%%%%%%
%%%%%%%%%%%%%%%%%%%%%%%%%%%%%%%%%%%%%%%%%%%%%%%%%%%
\footnotesize{SD2.1} &&
 % ms ent, rel, att, glob, overall
 75.44 & 53.06 & 69.66 & 68.49 &
 % 64.73 & 
 71.23
\\

%%%%%%%%%%%%%%%%%%%%%%%%%%%%%%%%%%%%%%%%%%%%%%%%%%%
%%%%%%%%%%%%%%%%%%%%%%%%%%%%%%%%%%%%%%%%%%%%%%%%%%%
%%%%%%%%%%%%%%%%%%%%%%%%%%%%%%%%%%%%%%%%%%%%%%%%%%%
\footnotesize{Muse} &&

 % ms ent, rel, att, glob, overall
 77.65& 60.64& 75.61 & 67.18 &
 % 69.81 &
 73.09
\\

%%%%%%%%%%%%%%%%%%%%%%%%%%%%%%%%%%%%%%%%%%%%%%%%%%%
%%%%%%%%%%%%%%%%%%%%%%%%%%%%%%%%%%%%%%%%%%%%%%%%%%%
%%%%%%%%%%%%%%%%%%%%%%%%%%%%%%%%%%%%%%%%%%%%%%%%%%%
\footnotesize{Imagen Cascade} &&
 % ms ent, rel, att, glob, overall
 79.94 & 62.73 & 75.73 & 69.34 &
 % 71.52 & 
 75.93
\\

%%%%%%%%%%%%%%%%%%%%%%%%%%%%%%%%%%%%%%%%%%%%%%%%%%%
%%%%%%%%%%%%%%%%%%%%%%%%%%%%%%%%%%%%%%%%%%%%%%%%%%%
%%%%%%%%%%%%%%%%%%%%%%%%%%%%%%%%%%%%%%%%%%%%%%%%%%%
\footnotesize{SDXL$_{v1.0}$} &&
 % ms ent, rel, att, glob, overall
 88.04 & 73.00 & 78.48 & 75.19 &
 % 77.35 &
 81.47 
\\
%%%%%%%%%%%%%%%%%%%%%%%%%%%%%%%%%%%%%%%%%%%%%%%%%%%
%%%%%%%%%%%%%%%%%%%%%%%%%%%%%%%%%%%%%%%%%%%%%%%%%%%
%%%%%%%%%%%%%%%%%%%%%%%%%%%%%%%%%%%%%%%%%%%%%%%%%%%

\hline
%%%%%%%%%%%%%%%%%%%%%%%%%%%%%%%%%%%%%%%%%%%%%%%%%%%
%%%%%%%%%%%%%%%%%%%%%%%%%%%%%%%%%%%%%%%%%%%%%%%%%%%
%%%%%%%%%%%%%%%%%%%%%%%%%%%%%%%%%%%%%%%%%%%%%%%%%%%
\footnotesize{Vermeer} &
\multicolumn{1}{r}{\footnotesize{\emph{raw model}}} &
86.92 & 76.36 & 76.48 & 68.49 &
% 76.88 &
80.77
\\
&\multicolumn{1}{r}{\footnotesize{\emph{+promp eng}}} &
87.94 &
74.92 &
76.31 &
67.41 &
80.99 
\\
&\multicolumn{1}{r}{\footnotesize{\emph{+style tunning}}} &
% ms ent, rel, att, glob, overall
88.04 & 74.21 & 77.38 & 69.57 & 
%76.37 &
81.16 
\\
%%%%%%%%%%%%%%%%%%%%%%%%%%%%%%%%%%%%%%%%%%%%%%%%%%%
%%%%%%%%%%%%%%%%%%%%%%%%%%%%%%%%%%%%%%%%%%%%%%%%%%%
%%%%%%%%%%%%%%%%%%%%%%%%%%%%%%%%%%%%%%%%%%%%%%%%%%%
% old: &\multicolumn{1}{r}{\footnotesize{+\emph{distillation. (old)}}} &
 % ms ent, rel, att, glob, overall 85.26 & 69.66 & 72.97 & 65.95 & 72.36 & 77.55
% \\
&\multicolumn{1}{r}{\footnotesize{+\emph{distillation.}}} &
 % ms ent, rel, att, glob, overall
 84.71 & 69.23 & 72.68  & 65.49 &
 % 71.10 & 
 76.88
\\
\hline
\hline
\end{tabular}
}
\vspace*{0.1cm}
\caption{\label{tab:vermeer_vqva}Vermeer. Broad and fine-grained results}
\end{table*}

We ablated four steps of Vermeer's development: 
(i) its raw model resulting from training on a large dataset; 
(ii) the result of applying prompt engineering at inference to the same model, adding words to improve aesthetic image quality, but with no further training;
(iii) the final model, after style finetuning on a curated subset of 3k aesthetically pleasing images; and finally, 
(iv) its distilled, fast inference variation.
\autoref{tab:vermeer_metrics} reports key performance metrics for all four variants, along with Stable Diffusion XL v1.0 (SDXL) \citep{sdxl}.
One can see that the raw model minimizes image distribution metrics that use state of the art feature space, i.e., FD-Dino and CMMD, while CLIP-score suggests a minor drop compared to SDXL.
These metrics also highlight a significant shift away from the distribution of MSCOCO-captions \citep{COCO}, after augmenting the prompts (\emph{+prompt engineering}) that is further increased when combined with the finetuning of the model for aesthetics pleasing image(\emph{+style finetuning}).

The MSCOCO-captions dataset comprises reference image-caption pairs covering a diverse set of object categories and scenes. Thus, it offers an interesting distribution for measuring image quality and text alignment due to the complexity and diversity of the compositions.
At the same time, its use for visual quality preference assessment is spurious as its images were not curated with human aesthetics preferences.  On the contrary, many of the images have relatively poor aesthetic appeal.
Thus, aiming to improve image aesthetics and composition, during Vermeer's prompt engineering and style tuning phases we intentionally move the distribution of images generated by Vermeer away from MSCOCO-caption distribution. 
To validate this we rely on human evaluation (in the next section).

The effect of the changes on the raw model with the CLIP-score and on semantic metrics on the other hand is minimal, aligned with our observation that the consistency of the model is not much affected by these two procedures. 
Semantic VqVa results are presented on \autoref{tab:vermeer_vqva}. 
The references to Imagen \citep{Saharia2022}
and Muse \citep{pmlr-v202-chang23b} models in this table are versions trained on internal data sources thus of similar resources and training pipelines than Vermeer.
It shows that Vermeer presents competitive performance with SDXL, and surpassing the other models, including auto-regressive and cascade models. 

Finally, we also develop a distilled version of our model, 
in order to offer an alternative version with faster inference time that similar to the other models presented in this paper operates as a single, non-cascade end-to-end model at inference time. 
\autoref{fig:teaser_vermeer_images} illustrates Vermeer outputs 
and additional qualitative results including a comparison of samples from the full and distilled versions is presented in Appendix \ref{app:qualitative_vermeer}.
 
\subsection{Human evaluation}
\label{sec:human_eval}

% Prompt following: Which image looks more representative
% to the text shown above and faithfully follows it?
% Visual aesthetics: Given the prompt, which image is of
% higher-quality and aesthetically more pleasing?

% \input{images/humaneval_exp123}

\begin{figure*}[t]
% \vskip 0.1in
\centering
\begin{subfigure}[t]{0.22\linewidth}
\includegraphics[width=\linewidth, trim={0 0 6cm 0},]{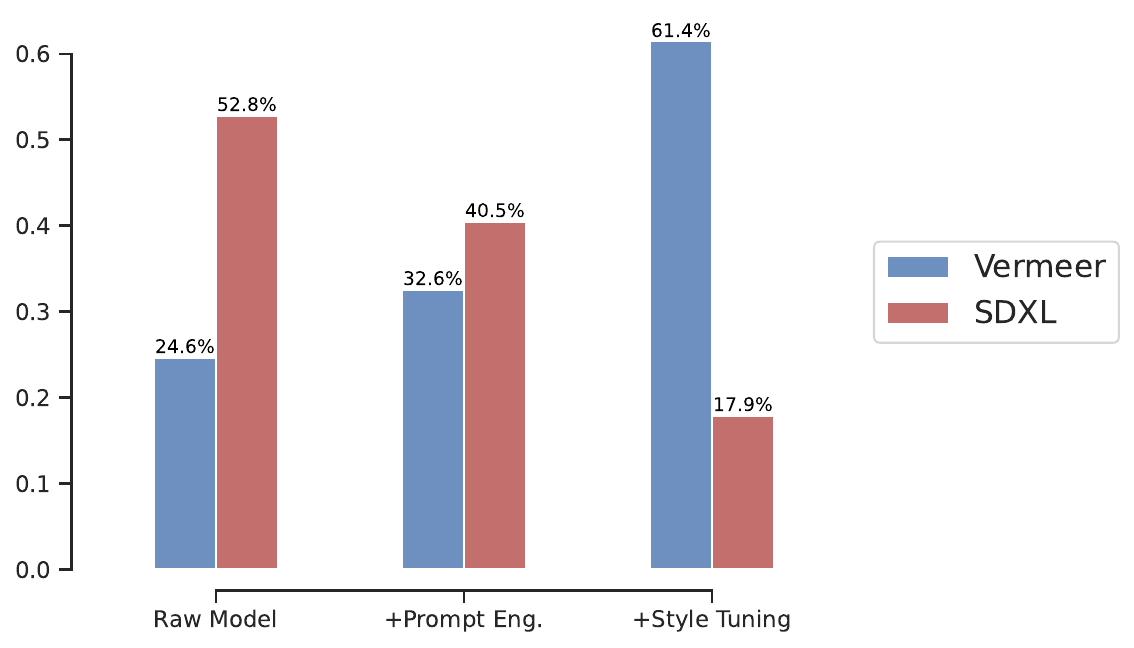}
\caption{{aesthetics}}
\end{subfigure}
\begin{subfigure}[t]{0.22\linewidth}
\includegraphics[width=\linewidth,trim={0 0 6cm 0},clip]{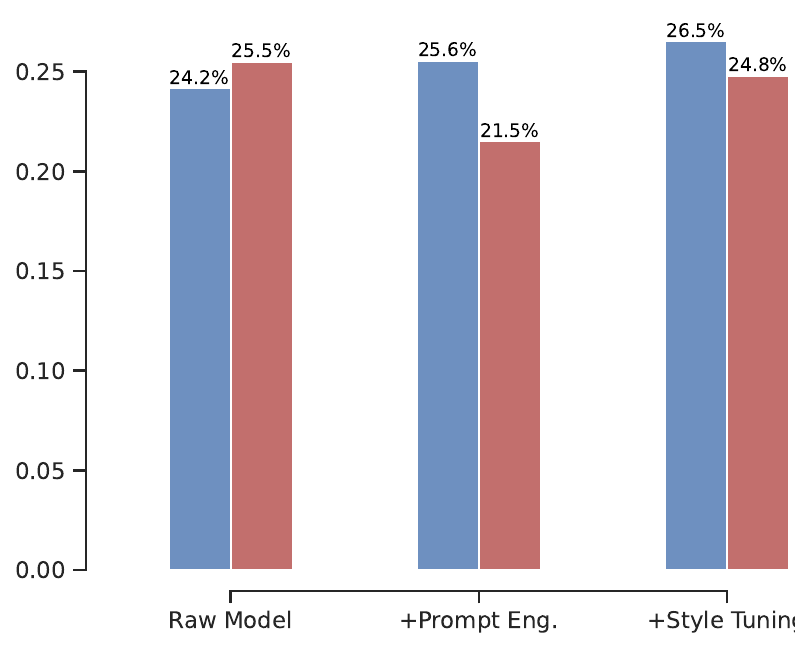}
\caption{{consistency}}
\end{subfigure}
\begin{subfigure}[t]{0.52\linewidth}
\includegraphics[width=\linewidth]{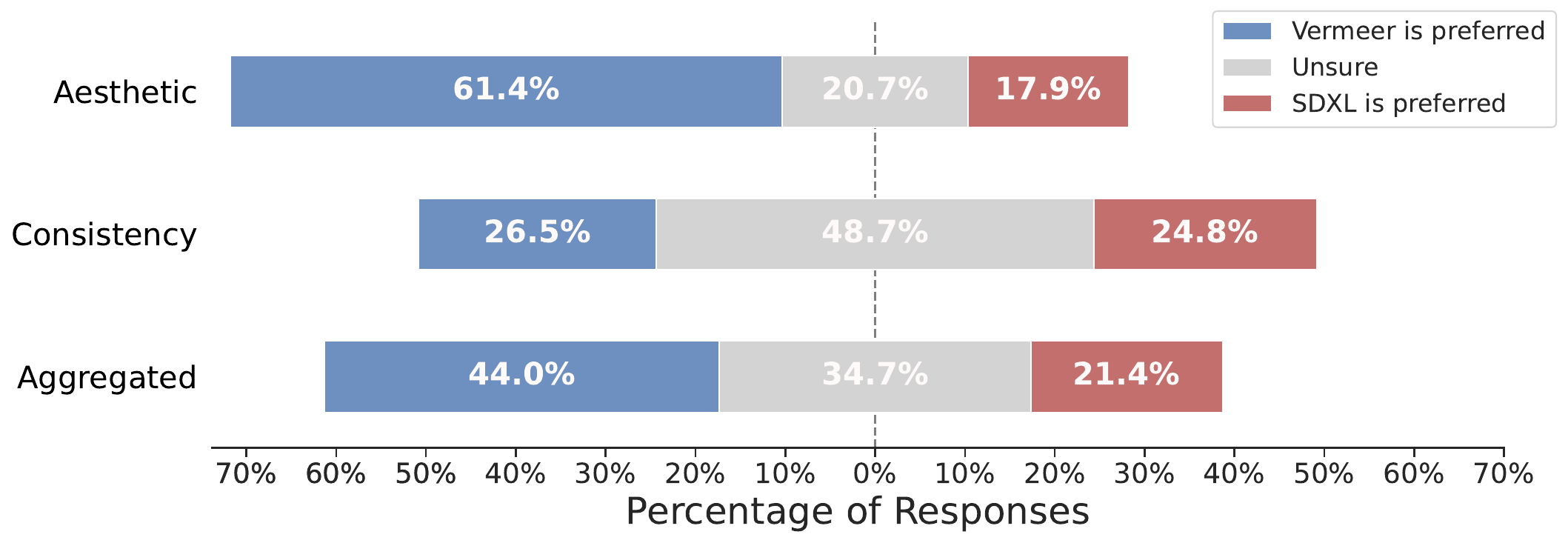}
%vermeer_sxs_likert_plot_vermeer_sdxl_large_scale_exp1_aggregated_v2.pdf}
\caption{{final model 3-point Likert}}
\end{subfigure}
\vskip -0.05in
\caption{\textbf{Human evaluation results}: Likert plot across 495 prompts, two tasks with 13 users each. Vermeer aesthetic is preferred during $61.4\%$ of all comparisons, while its image-text consistency  is marginally preferred. 
Aggregating the 1k annotations, Veermer is preferred during $44.0\%$ of all comparisons, against $21.0\%$ from SDXL. Prompt engineering and style tuning aligned with human preference for visual aesthetics.  
}
\label{fig:sxs_likert_plot_aggregated}
\vskip -0.1in
\end{figure*}

Assessing the performance of text-to-image models, ideally, depends on human evaluation, as this complex cognitive process necessitates a profound understanding of text and image relationships.
Prior research has demonstrated that many recent works rely exclusively on automated metrics, such as the Fréchet Inception Distance (FID). However, it has been observed that the current automated measures are not fully consistent with human perception in assessing the quality of text-to-image samples \citep{Human_Evaluation}.
Thus, to objectively access the quality of images generated by Vermeer,
we conduct a side-by-side human evaluation comparing our model with SDXL \citep{sdxl}.
% , a high-performing open-source text-to-image generation model, focusing on user preference on aesthetic appearance and text consistency.

\vspace{4pt}
\noindent {\bf Setup}. 
In this human evaluation, we ask annotators to evaluate generated images by Vermeer and SDXL based on the same prompt.
For this, we collected 495 prompts
\footnote{We first sampled 510 prompts, and 495 of them were usable after filtering incomplete samples.} 
covering a range of skills: 160 are from TIFA v1.0
targeting measuring the faithfulness of a generated image to its text input covering 12 categories (object, attributes, counting, etc.)\citep{hu2023tifa}; 200 are sampled from the 1600 Parti Prompts \citep{parti}, selecting for both complexity and diversity of challenges; and 150 others are created fresh for, or are sourced from, more recent prompting strategies targeting challenging cases.

We create two tasks in which we instruct annotators to consider either image quality (aesthetics) or fit to the prompt (consistency), and indicate their preferences using 3-point Liker scale: \emph{Vermeer is preferred}, \emph{Unsure}, and \emph{SDXL is preferred} (the model names are anonymized).
The neural response includes cases that both images are equally good and bad.
In the annotation UI, the annotators are shown a prompt along with two images that are randomly shuffled. We collected 13 human ratings per prompt for both aesthetics and consistency (26 ratings per image).

\vspace{4pt}
\noindent {\bf Results}.
Prompt engineering and style tuning are confirmed to have a positive effect on human aesthetics preference (\autoref{fig:sxs_likert_plot_aggregated}, left), and small impact on text consistency (\autoref{fig:sxs_likert_plot_aggregated}, middle). 
They confirm our conjecture that the decrease on Vermeer's performance based on metrics grounded on the appearance of MSCOCO-caption dataset induced by these two steps are in alignment with the ultimate goal of human  preference (\autoref{tab:vermeer_metrics}). 

%  Appendix \ref{app:human_eval} contains the 3-point Likert results on Vermeer intermediary versions.
 
\autoref{fig:sxs_likert_plot_aggregated} (right) plots the Likert scale for our final model in each task (aesthetics or consistency) as well as the aggregated responses (shown in in the bottom bar).
Overall, annotators prefer Vermeer 44\% of the time, while they select SDXL 21.4\% of the time, with relatively fewer \emph{Neutral} responses (34.7\%).
Vermeer is clearly preferred for its aesthetics, with a win rate of $61.4\%$, while the gap in consistency between the two models is small, with a difference in the win rate of just $1.7\%$.
Krippendorff's $\alpha$ for aesthetics and consistency are 0.27 and 0.41, respectively, indicating moderate agreement among annotators.

% For this, we collected 350 prompts that cover a range of skills and challenges: 200 are sampled from the 1600 Parti Prompts \cite{parti}, selecting for both complexity and diversity of challenges, and 150 others are created fresh for, or are sourced from, more recent prompting strategies with recent text-to-image models.
% We instruct annotators to consider both image quality and fit to the prompt, and indicate their preferences using 5-point Liker scale: \emph{A/B is substantially better}, \emph{A/B is marginally better}, and \emph{Neutral}.
% To prevent biased responses, we recruit five annotators who are not familiar with Vermeer's images.
% Each annotator evaluates 350 examples, and subsequently, we verify agreement between the five annotators.
% In the annotation UI, the annotators are shown a prompt along with two images that are randomly shuffled.

% \vspace{4pt}
% \noindent {\bf Results}. \autoref{fig:sxs_likert_plot_per_annotator} plots Likert scale for each annotator as well as the aggregated responses (the bottom bar plot).
% Overall, annotators prefer Vermeer 46\% of the time, while they select SDXL 30\% of the time, with relatively fewer \emph{Neutral} responses (23\%).\footnote{This includes cases that both images are equally good and bad.}
% The bar plots for individual annotators show high agreement; all five annotators selected Vermeer as their majority choice.
% These results demonstrate Vermeer's abilities to produce high-quality images that align with human judgement.

%\input{images/humaneval} 

\section{Conclusion}

We propose a novel recipe for training non-cascaded large scale pixel-space text-to-image diffusion models. It benefits from splitting their training in two phases representing different tasks: learning image-text condition alignment and learning to generate images at high-resolution. 

We identified the model \CC as those responsible for the first task and propose a proxy architecture (\SU) to supports its pretraining. 
The second task is learned with a \emph{greedy growing} algorithm that stacks encoder-decoder layers of the final architecture on top of the pretrained \CC. When learning the second task, our training recipe preserves the \CC representation from the noise introduced by the grown layers and their random initialized weights.

Existing non-cascaded models training recipes struggle with scale, if not supported with large batch size and further regularization like dropout and multi-scale loss. 
Our approach is able to train models up to 8B parameters with small batch size (256) and no further regularization, by pretraining the \CC and preserving it during the second training phase targeting high-resolution generation.

Compared with training from scratch and finetuning, the greedy growing procedure is more stable, and improves performance on a set of different metrics. Qualitative analysis shows that while keeping the \CC representation stable it helps to preserve objects shape and overall structure, improving the definition of body parts. 
Our method allows use of data at different resolutions; the first phase benefits from the larger corpora with minimal requirements on image resolution, while the second phase learn to produce sharp images from the set filtered by the target resolution while reusing the representation learned from the larger set.  We also explore models with increasing size, and show the benefits from scaling under different aspects and metrics. 

In practice, the non-cascaded solution  removes the out-of-distribution shift existent between training and deploying super-resolution phases. Based on that, we present Vermeer, an 8B parameter \emph{Pixel based Text-to-Image Diffusion Model} that produces high-resolution high-quality images using a single non-cascaded model. By training it on a larger dataset, and incorporating a final style tuning phase, Vermeer is able to surpass SDXL v1.0 in human preference study.  

% As future work, we believe non-cascaded pixel-space models will facilitate the exploration of diffusion models into inverse problems, as different from cascaded versions, they are ready to be adopted at the target resolution, and differently from latent based ones, can model the pixel space directly.
% Another direction for future investigation
% is to examine the implicit bias introduced by the pretraining of the model's \CC on low-resolution images. A natural question is how much preserving the features obtained from this data distribution affect model's shape-texture biases.

% Bibliography components
\bibliographystyle{abbrvnat}
\nobibliography*
\bibliography{refs}

\onecolumn

% Some other useful sections you might consider having in your report.
% \section*{Citing this work}
% This is a free, open access paper provided by Google DeepMind.
% The final version of this work was published online (provide venue, date and digital object
% identifier (doi, if available). \textit{Cite as:} \citeas{silver2016mastering}.

% If you add a bibtex entry of your own paper (this paper), you can
% show its full citation inline using \citeas, as above. Note that
% this citation removes the trailing full stop. To make \citeas work,
% you need to load the bibliography data. This can be done in two
% ways:
%
%    1. If you already have a printed bibliography with \bibliography{...},
%       then add the command "\nobibliography*", no arguments, before that.
%    2. If you don't otherwise print a bibliography, add the command
%       \nobibliography{...} at the end of your document.

% \section*{Acknowledgements}
% Any acknowledgements go here.

% \section*{Author Contributions}
% List of author contributions.

% \section*{Funding}
% This research was funded by Google DeepMind.

% \section*{Competing interests}
% The authors declare no competing financial interests. Related
% patent number here if applicable.

% \section*{Data availability}
% The datasets used in the experiments have been made available
% for download at IF AVAILABLE.

% \section*{Supplementary Materials}

\newpage

\newpage 
% \newpage
% \newpage
% \appendix
\section*{Teaser image prompts}
\label{app:teaser_prompts}

\begin{table}[htb!]
\vspace*{-\baselineskip}
\centering
\resizebox{0.35\linewidth}{!}{
\begin{tabular}{|c|c|c|c|c|c|}\hline
\centering
  \multirow{3}{*}{1} & \multirow{6}{*}{~~~~3~~~~} & \multirow{3}{*}{~~~~4~~~~} & \multirow{6}{*}{~~~6~~~~} & \multirow{2}{*}{7} \\ 
  & & & & \\
  \cline{5-5} 
  & & & & \multirow{2}{*}{8} 
  \\
  \cline{1-1}\cline{3-3} \multirow{3}{*}{2} & ~& \multirow{3}{*}{~~5~} &  & \\ 
    \cline{5-5}   & & & & \multirow{2}{*}{9} \\
  & & & & \\\hline
  %%%%%%%%%%%%% second row:
  \multirow{6}{*}{10} & \multirow{6}{*}{11} & \multirow{3}{*}{12} & \multirow{3}{*}{14} & \multirow{6}{*}{16} \\ 
  & & & & \\
  & & & & \\
  \cline{3-4} & ~& \multirow{3}{*}{13} &  \multirow{6}{*}{15}& \\ 
  & & & &\\
  & & & & \\
  %%%%%%%%%%%%% third row:
  \cline{1-3}\cline{5-5}
  \multirow{6}{*}{17} & \multirow{6}{*}{18} & \multirow{3}{*}{19} &  & \multirow{6}{*}{22} \\ 
  & & & & \\
  & & & & \\
  \cline{3-4} & ~& \multirow{3}{*}{20} &  \multirow{3}{*}{21}& \\ 
  & & & &\\
  & & & & \\
  \hline
 \end{tabular}
}
\caption{\label{tab:teaser_map} Map of prompts used to generate Vermeer results illustrated in \autoref{fig:teaser_vermeer_images}}
\end{table}

Next, we list the prompts used for generating images at \autoref{fig:teaser_vermeer_images} using \Vermeer. Their corresponding location is shown in \autoref{tab:teaser_map}).
\begin{enumerate}
\itemsep0em 
\scriptsize{
\item the word 'START' written in chalk on a sidewalk

\item a basketball to the left of two soccer balls on a gravel driveway

\item An Egyptian tablet shows an automobile.

\item Macro photography of rose, centered, mini, dark tones, drops of water, cannon
    
\item photo of a woman's face floating in the water with her eyes closed, you can only see top part of her face above water, reflections, abstract conceptual, realistic reflection, pale sky, scientific photo, high quality fantasy stock photo
    
\item cyberpunk starship troopers cinematic 4d

\item 3-d Letter "O" made from orange fruit, studio shot, pastel orange background, centered
\item 3-d Letter "W" made from transparent water, studio shot, pastel light blue background, centered
\item 3-d Letter "T" made from tiger fur, studio shot, pastel orange background, centered.
%%%second row
\item Many people carry sacks along a trail through a bright field with long grass and flowers and muted tones. Two small cottages. Dark row of trees. Green hills, blue sky, clouds. Pastoral landscape. Ein plein air. Vibrant, saturation, free brush strokes. Impressionism. Oil on canvas by Auguste Renoir.

\item a photograph of a blue porsche 356 coming around a bend in the road

\item photography of a cat sitting at a sushi restaurant, wearing a blue coat and taking sushi from the boat. Neon bright light, high contrast, low vibrance

\item turtle with German Shepherd dog's head growing from it, DSLR

\item A futuristic street train a rainy street at night in an old European city. Painting by David Friedrich, Claude Monet and John Tenniel.

\item building behind train

\item Realistic photograph of a cute otter zebra mouse in a field at sunset, tall grass, macro 35mm film

%%%%%%%%%%%last row:

\item A 1920's race car with number 7 parked near a fountain in a modern city. Painting by David Friedrich, Claude Monet and John Tenniel.
\item The clock on the bricked building is green. The numbers are in roman numerals. The details have gold accents. The bricked building has a window beside the clock.
\item duck with rabbit's head growing from it, DSLR
\item cauliflower with sheep's head growing from it, DSLR
\item Silver 1963 Ferrari 250 GTO in profile racing along a beach front road. Bokeh, high-quality 4k photograph.
\item a photograph of a knight in shining armor holding a basketball
}
\end{enumerate}

\section*{\SU: Vqva detailed categories}
\label{app:vqva_detailed}

\begin{table}[htb!]
\vspace*{-\baselineskip}
\centering
\resizebox{\linewidth}{!}{
\begin{tabular}{lcccccccccccccccc|c} 
\hline
 \hline  &  \multicolumn{15}{c}{VqVa Question types}
 \\
Model &
 \small{whole} &
 \small{part} &
 \small{spatial} &
 \small{shape} &
 \small{color} &
 \small{state} &
 \small{type} &
 \small{count} &
 \small{text rendering} &
 \small{texture} &
 \small{global} &
 \small{material} &
 \small{scale} &
 \small{size} & DSG \\
 \# prompts &
 2851 & 
517 & 
1477 & 
84 & 
432 & 
740 & 
173 & 
196 & 
116 & 
40 & 
649 & 
92 & 
25 & 
11 
 \\
\hline 
\small{\SU Base} &
0.567 & 
0.412 & 
0.333 & 
0.417 & 
0.550 & 
0.409 & 
0.402 & 
0.523 & 
0.487 & 
0.450 & 
0.400 & 
0.272 & 
0.500 & 
0.318 &
48.08
\\
\small{\SU Large} &
0.626 & 
0.451 & 
0.395 & 
0.446 & 
0.579 & 
0.478 & 
0.454 & 
0.554 & 
0.552 & 
0.475 & 
0.437 & 
0.353 & 
0.600 & 
0.364 & 
52.54
\\
\small{\SU Huge} &

0.706 & 
0.617 & 
0.488 & 
0.548 & 
0.624 & 
0.530 & 
0.509 & 
0.587 & 
0.552 & 
0.562 & 
0.433 & 
0.424 & 
0.740 & 
0.409 & 
60.25
\\
\small{\SU XHuge} &
0.724 & 
0.617 & 
0.518 & 
0.577 & 
0.646 & 
0.568 & 
0.540 & 
0.582 & 
0.591 & 
0.562 & 
0.441 & 
0.424 & 
0.820 & 
0.636 & 
61.91
\\
% \small{+ data} &
% 0.728 & 
% 0.584 & 
% 0.505 & 
% 0.506 & 
% 0.726 & 
% 0.556 & 
% 0.526 & 
% 0.612 & 
% 0.582 & 
% 0.637 & 
% 0.421 & 
% 0.397 & 
% 0.720 & 
% 0.636 & 
% 61.58
% \\
\hline
% \hline
% \small{Base 8k} &
% % base 8k
% 0.577 & 
% 0.456 & 
% 0.353 & 
% 0.506 & 
% 0.512 & 
% 0.431 & 
% 0.387 & 
% 0.523 & 
% 0.517 & 
% 0.350 & 
% 0.385 & 
% 0.266 & 
% 0.500 & 
% 0.273 & 
% 48.10 
% \\
% \small{Large 8k} &
% 0.623 & 
% 0.464 & 
% 0.391 & 
% 0.476 & 
% 0.566 & 
% 0.489 & 
% 0.462 & 
% 0.528 & 
% 0.573 & 
% 0.512 & 
% 0.407 & 
% 0.304 & 
% 0.660 & 
% 0.227 & 
% 52.36
% \\
% \small{Huge 8k} &
% 0.711 & 
% 0.567 & 
% 0.479 & 
% 0.494 & 
% 0.647 & 
% 0.539 & 
% 0.509 & 
% 0.505 & 
% 0.647 & 
% 0.588 & 
% 0.422 & 
% 0.380 & 
% 0.560 & 
% 0.364 & 
% 60.16
% \\
% \small{XHuge 8k} &
% 0.715 & 
% 0.598 & 
% 0.490 & 
% 0.548 & 
% 0.641 & 
% 0.533 & 
% 0.529 & 
% 0.574 & 
% 0.552 & 
% 0.438 & 
% 0.437 & 
% 0.386 & 
% 0.780 & 
% 0.409 & 
% 61.14
% \\
% \small{+ data} & 
% 0.738 & 
% 0.624 & 
% 0.529 & 
% 0.506 & 
% 0.698 & 
% 0.563 & 
% 0.572 & 
% 0.633 & 
% 0.595 & 
% 0.650 & 
% 0.445 & 
% 0.397 & 
% 0.780 & 
% 0.500 & 
% 63.10 
% \\
% \small{+ data 1.5M steps} & 
% 0.745 & 
% 0.632 & 
% 0.552 & 
% 0.554 & 
% 0.699 & 
% 0.572 & 
% 0.520 & 
% 0.615 & 
% 0.603 & 
% 0.575 & 
% 0.430 & 
% 0.408 & 
% 0.760 & 
% 0.500 & 
% 64.28
% \\
\hline
 \end{tabular}
 }
\caption{\SU scaling: DSG fine-grained semantic categories. 
DSG: average score accross DS1K images.}
\label{tab:shallow_vqva_detailed}
\end{table}

\begin{table}[htb!]
\vspace*{-\baselineskip}
\centering
\resizebox{\linewidth}{!}{
\begin{tabular}{llcccccccccccccccc|c} 
\hline
 \hline  &  \multicolumn{15}{c}{VqVa Question types}
 \\
Model & &
 \small{whole} &
 \small{part} &
 \small{spatial} &
 \small{shape} &
 \small{color} &
 \small{state} &
 \small{type} &
 \small{count} &
 \small{text rendering} &
 \small{texture} &
 \small{global} &
 \small{material} &
 \small{scale} &
 \small{size} & DSG \\
 &\# prompts &
 2851 & 
517 & 
1477 & 
84 & 
432 & 
740 & 
173 & 
196 & 
116 & 
40 & 
649 & 
92 & 
25 & 
11 
 \\
\hline 
%%%%%%%%%%%%%%%%%%%%%%%%%%%%%%%%%%%%%%%%%%%%%%%%%%%
%%%%%%%%%%%%%%%%%%%%%%%%%%%%%%%%%%%%%%%%%%%%%%%%%%%
%%%%%%%%%%%%%%%%%%%%%%%%%%%%%%%%%%%%%%%%%%%%%%%%%%%
\footnotesize{UVIT-Base} &
 \emph{scratch} & 
0.743 & 
0.671 & 
0.540 & 
0.655 & 
0.635 & 
0.639 & 
0.552 & 
0.564 & 
0.690 & 
0.575 & 
0.555 & 
0.489 & 
0.780 & 
0.545 & 
64.83 
 \\  %%%%%%%%%%%%%%%%%
 & 
 \emph{finetuning} &
 0.723 & 
0.587 & 
0.500 & 
0.560 & 
0.597 & 
0.596 & 
0.590 & 
0.628 & 
0.625 & 
0.525 & 
0.532 & 
0.478 & 
0.700 & 
0.500 & 
62.75 
\\  %%%%%%%%%%%%%%%%%
 & 
 \emph{frozen} &
 0.702 & 
0.666 & 
0.495 & 
0.560 & 
0.544 & 
0.647 & 
0.566 & 
0.584 & 
0.591 & 
0.613 & 
0.534 & 
0.332 & 
0.640 & 
0.409 & 
61.16 
 \\  %%%%%%%%%%%%%%%%% freeze-unfreeze
& \emph{freeze-unfreeze} & 
 0.748 & 
0.657 & 
0.536 & 
0.649 & 
0.650 & 
0.645 & 
0.587 & 
0.640 & 
0.694 & 
0.625 & 
0.569 & 
0.418 & 
0.620 & 
0.455 & 
66.13
\\
 \hline 
 %%%%%%%%%%%%%%%%%%%%%%%%%%%%%%%%%%%%%%%%%%%%%%%%%%%
 %%%%%%%%%%%%%%%%%%%%%%%%%%%%%%%%%%%%%%%%%%%%%%%%%%%
 %%%%%%%%%%%%%%%%%%%%%%%%%%%%%%%%%%%%%%%%%%%%%%%%%%%
\footnotesize{UVIT-Large}  &
 \emph{scratch} & 
0.750 & 
0.642 & 
0.521 & 
0.631 & 
0.642 & 
0.640 & 
0.618 & 
0.643 & 
0.547 & 
0.562 & 
0.580 & 
0.511 & 
0.640 & 
0.591 & 
66.02 
 \\  %%%%%%%%%%%%%%%%%
 & 
 \emph{finetuning} &
 0.761 & 
0.688 & 
0.542 & 
0.601 & 
0.681 & 
0.678 & 
0.604 & 
0.666 & 
0.638 & 
0.650 & 
0.579 & 
0.473 & 
0.780 & 
0.636 & 
67.39 
\\  %%%%%%%%%%%%%%%%%
 & 
 \emph{frozen} &
 0.800 & 
0.730 & 
0.616 & 
0.643 & 
0.738 & 
0.684 & 
0.618 & 
0.648 & 
0.728 & 
0.700 & 
0.614 & 
0.418 & 
0.760 & 
0.500 & 
72.13 
 \\  %%%%%%%%%%%%%%%%% freeze-unfreeze
& \emph{freeze-unfreeze} & 
 0.761 & 
0.668 & 
0.556 & 
0.571 & 
0.646 & 
0.655 & 
0.665 & 
0.717 & 
0.616 & 
0.625 & 
0.588 & 
0.473 & 
0.840 & 
0.545 & 
67.79 \\
 \hline
%%%%%%%%%%%%%%%%%%%%%%%%%%%%%%%%%%%%%%%%%%%%%%%%%%%
%%%%%%%%%%%%%%%%%%%%%%%%%%%%%%%%%%%%%%%%%%%%%%%%%%%
%%%%%%%%%%%%%%%%%%%%%%%%%%%%%%%%%%%%%%%%%%%%%%%%%%%
\footnotesize{UViT-Huge}  &
 \emph{scratch} & 
 0.758 & 
0.662 & 
0.551 & 
0.595 & 
0.634 & 
0.645 & 
0.627 & 
0.607 & 
0.698 & 
0.600 & 
0.586 & 
0.505 & 
0.840 & 
0.591 & 
66.90 
 \\  %%%%%%%%%%%%%%%%%
 & 
 \emph{finetuning} &
 0.773 & 
0.775 & 
0.565 & 
0.548 & 
0.684 & 
0.716 & 
0.705 & 
0.648 & 
0.659 & 
0.650 & 
0.626 & 
0.484 & 
0.640 & 
0.591 & 
69.67 
 \\  %%%%%%%%%%%%%%%%%
 & 
 \emph{frozen} &
 0.827 & 
0.814 & 
0.648 & 
0.649 & 
0.749 & 
0.722 & 
0.685 & 
0.635 & 
0.797 & 
0.688 & 
0.619 & 
0.500 & 
0.820 & 
0.500 & 
75.15 
 \\  %%%%%%%%%%%%%%%%% freeze-unfreeze
& \emph{freeze-unfreeze} & 
0.798 & 
0.748 & 
0.582 & 
0.583 & 
0.675 & 
0.684 & 
0.653 & 
0.666 & 
0.711 & 
0.575 & 
0.609 & 
0.467 & 
0.780 & 
0.500 & 
71.50 
 
\\
\hline
\footnotesize{UVIT-XHuge}  &
 \emph{freeze} & 
\bf{0.840} & 
\bf{0.821} & 
\bf{0.668} & 
\bf{0.637} & 
\bf{0.744} & 
0.720 & 
\bf{0.720} & 
\bf{0.671} & 
\bf{0.668} & 
\bf{0.688} & 
0.629 & 
0.473 & 
\bf{0.860} & 
0.455 & 
\bf{75.70}
\\
& \emph{freeze-unfreeze} & 
0.817 & 
0.783 & 
0.607 & 
0.631 & 
0.722 & 
0.705 & 
0.679 & 
0.681 & 
0.681 & 
0.675 & 
0.602 & 
0.533 & 
0.880 & 
0.727 & 
73.53 \\
\hline
%%%%%%%%%%%%%%%%%%%%%%%%%%%%%%%%%%%%%%%%%%%%%%%%%%%
%%%%%%%%%%%%%%%%%%%%%%%%%%%%%%%%%%%%%%%%%%%%%%%%%%%
%%%%%%%%%%%%%%%%%%%%%%%%%%%%%%%%%%%%%%%%%%%%%%%%%%%
\footnotesize{SD2.1} & &
0.760 & 
0.730 & 
0.530 & 
0.679 & 
0.707 & 
0.729 & 
0.665 & 
0.571 & 
0.655 & 
0.637 & 
0.685 & 
0.495 & 
0.780 & 
0.455 & 
71.23 
\\ 
\hline
\end{tabular}
 }
\caption{End-to-end models: DSG fine-grained semantic categories. 
DSG: average score accross DS1K images.}
\label{tab:uvit_vqva_detailed}
\end{table}
% imgrid in :
% https://imgrid.googleplex.com/?qq=/cns/oi-d/home/earthsea-dev/olwang/rs=6.3/sxs/imagen_2_5_vermeer_test/98046092/*/.imgrid&kvg=prompt&r=2
 
\begin{table}[htb!]
\vspace*{-\baselineskip}
\centering
\resizebox{\linewidth}{!}{
\begin{tabular}{lcccccccccccccccc|c} 
\hline
 \hline  &  \multicolumn{15}{c}{VqVa Question types}
 \\
Model & &
 \small{whole} &
 \small{part} &
 \small{spatial} &
 \small{shape} &
 \small{color} &
 \small{state} &
 \small{type} &
 \small{count} &
 \small{text rendering} &
 \small{texture} &
 \small{global} &
 \small{material} &
 \small{scale} &
 \small{size} & DSG \\
\hline 
%%%%%%%%%%%%%%%%%%%%%%%%%%%%%%%%%%%%%%%%%%%%%%%%%%%
%%%%%%%%%%%%%%%%%%%%%%%%%%%%%%%%%%%%%%%%%%%%%%%%%%%
%%%%%%%%%%%%%%%%%%%%%%%%%%%%%%%%%%%%%%%%%%%%%%%%%%%
\hline 
%%%%%%%%%%%%%%%%%%%%%%%%%%%%%%%%%%%%%%%%%%%%%%%%%%%
%%%%%%%%%%%%%%%%%%%%%%%%%%%%%%%%%%%%%%%%%%%%%%%%%%%
%%%%%%%%%%%%%%%%%%%%%%%%%%%%%%%%%%%%%%%%%%%%%%%%%%%
\footnotesize{Muse} & &
0.780 & 
0.761 & 
0.605 & 
0.714 & 
0.814 & 
0.766 & 
0.668 & 
0.610 & 
0.651 & 
0.838 & 
0.672 & 
0.647 & 
0.780 & 
0.773 & 
73.09
\\
%%%%%%%%%%%%%%%%%%%%%%%%%%%%%%%%%%%%%%%%%%%%%%%%%%%
%%%%%%%%%%%%%%%%%%%%%%%%%%%%%%%%%%%%%%%%%%%%%%%%%%%
%%%%%%%%%%%%%%%%%%%%%%%%%%%%%%%%%%%%%%%%%%%%%%%%%%%
\footnotesize{SD2.1} & &
0.760 & 
0.730 & 
0.530 & 
0.679 & 
0.707 & 
0.729 & 
0.665 & 
0.571 & 
0.655 & 
0.637 & 
0.685 & 
0.495 & 
0.780 & 
0.455 & 
71.23 
\\ 
%
% \footnotesize{SDXL} &
% \\
%%%%%%%%%%%%%%%%%%%%%%%%%%%%%%%%%%%%%%%%%%%%%%%%%%%
%%%%%%%%%%%%%%%%%%%%%%%%%%%%%%%%%%%%%%%%%%%%%%%%%%%
%%%%%%%%%%%%%%%%%%%%%%%%%%%%%%%%%%%%%%%%%%%%%%%%%%%
\footnotesize{Imagen Cascade} & &
0.799 & 
0.806 & 
0.626 & 
0.714 & 
0.806 & 
0.772 & 
0.723 & 
0.673 & 
0.750 & 
0.738 & 
0.693 & 
0.641 & 
0.820 & 
0.636 & 
75.93
\\
%%%%%%%%%%%%%%%%%%%%%%%%%%%%%%%%%%%%%%%%%%%%%%%%%%%
%%%%%%%%%%%%%%%%%%%%%%%%%%%%%%%%%%%%%%%%%%%%%%%%%%%
%%%%%%%%%%%%%%%%%%%%%%%%%%%%%%%%%%%%%%%%%%%%%%%%%%%
\footnotesize{Imagen Vermeer} &
raw &
0.884 & 
0.787 & 
0.765 & 
0.690 & 
0.810 & 
0.737 & 
0.798 & 
0.689 & 
0.789 & 
0.787 & 
0.685 & 
0.701 & 
0.940 & 
0.773 & 
80.77
\\
& + prompt eng. &
0.892 & 
0.810 & 
0.751 & 
0.750 & 
0.840 & 
0.732 & 
0.809 & 
0.679 & 
0.784 & 
0.825 & 
0.674 & 
0.625 & 
0.880 & 
0.591 & 
80.99
\\
&  + style tuning &
0.889 & 
0.833 & 
0.744 & 
0.696 & 
0.836 & 
0.747 & 
0.818 & 
0.714 & 
0.716 & 
0.838 & 
0.696 & 
0.707 & 
0.840 & 
0.591 & 
81.16
\\
%%%%%%%%%%%%%%%%%%%%%%%%%%%%%%%%%%%%%%%%%%%%%%%%%%%
%%%%%%%%%%%%%%%%%%%%%%%%%%%%%%%%%%%%%%%%%%%%%%%%%%%
%%%%%%%%%%%%%%%%%%%%%%%%%%%%%%%%%%%%%%%%%%%%%%%%%%%
%\footnotesize{Imagen Vermeer \emph{dist.}} &
\hline
\hline
\end{tabular}
}
\caption{\label{tab:vermeer_vqva_detailed}Vermeer: DSG fine-grained semantic categories .}
\end{table}

This appendix complements the results on broad categories presented in the main text by providing   the fine grain corresponding results.
\section*{On validating the representation quality improvements from scale by counting}
\label{app:counting}

\begin{figure}[htb!]
\begin{center}
\centerline{\includegraphics[width=0.5\columnwidth]{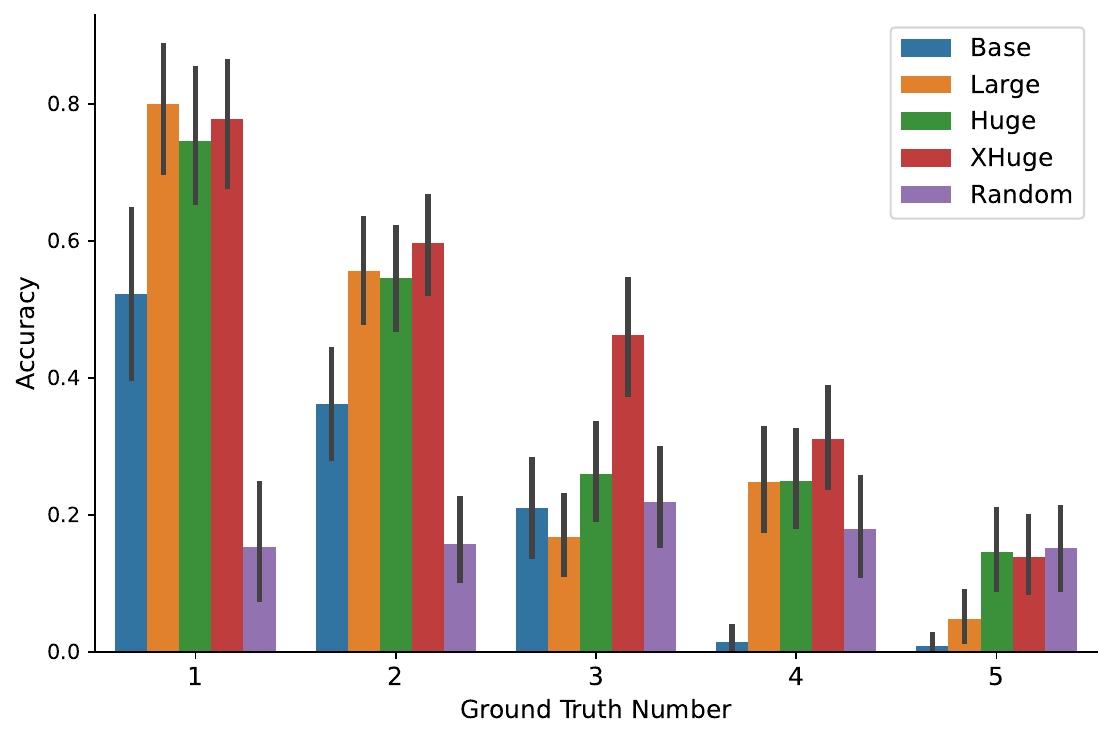}}
\caption{Breakdown of accuracy per number in the original prompt used to generate the image. }
\label{fig:count_per_number}
\end{center}
\end{figure}

Given the importance of counting and other basic numerical skills in biological intelligence~\citep{nieder2009representation}, we expect that competitively performing T2I  show similar behaviour when evaluated on such skills.
Counting requires manipulation of abstract concepts (numbers) and evaluating this ability provides an objective measure of a well-defined skill. As such it is easier to evaluate and interpret the performance of the model on the counting task, in contrast to some other image characteristics such as aesthetics that might depend on an individual's preferences.

To evaluate models' ability to correctly generate an image with an exact number of objects,  we use 59 prompts in the \emph{att/count} category of the Gecko benchmark~\citep{gecko2024}.
The Gecko benchmark aims to comprehensively and systematically probe T2I model alignment along  different skills such as numerical and spatial reasoning, text rendering, depicting of colors and shapes, and many others.

Specifically, our analyses include 48 \emph{simple modifier} prompts and 11 \emph{additive} prompts with numbers between 1 and 5.
\emph{Simple modifier} prompts are of form ``\emph{num noun}'' (eg. ``1 cat''), where \emph{num} is a number represented by a single digit (ie. 1, 2, 3) or a numeral (ie. ``one'', ``two'' or ``three'') and the noun is a word from a common natural semantic categories such as foods, animals and everyday objects. 
\emph{Additive} prompts are compositions of individual {\em simple modifier} prompts as they combine two nouns and two numbers, such as ``1 cat and 3 dogs''.
By using such systematically curated prompts, we are implicitly testing whether models can count, as the ability to correctly generate a number of objects depends on the ability to keep track of objects that were already generated.

To evaluate the correctness of T2I generation of numbers, we recruit human raters through a crowd-sourcing platform to provide the count of objects in every generated image. 
The study design, including remuneration for the work were reviewed and approved by our institution's independent ethical review committee.
We collect 5 annotations per generated image by asking ``How many X are there in the image?'' where X is the object mentioned in the original prompt used to generate that image. We generate three images for each prompt and each model using different seeds.

Figure~\ref{fig:count_per_number} shows the breakdown of accuracy per model type as well as per the ground truth number. The ground truth number is the number in the original prompt used to generate the image. The accuracy is the average number of annotations that match the ground truth label for a question and a given model.
We observe that all models (with the exception of Base) perform comparably well on generating images with only one object, but this deteriorates with higher number, and only XHuge is able to correctly generate number 3 above the chance level. While exact number generation appears to improve with scale, it is unclear whether this pattern saturates for higher numbers. 
\section*{Qualitative comparison of finetuning and frozen e2e models}
\label{app:early_finetune_frozen}

\begin{figure*}[t]
\setlength{\fboxsep}{0pt}%
\setlength{\fboxrule}{1pt}%
\centering
\footnotesize{
\setlength{\lineskip}{1pt}
\begin{subfigure}{\linewidth}
    \begin{subfigure}[ht]{\linewidth}
            \begin{subfigure}[t]{0.09\linewidth}
               \includegraphics[width=\linewidth, height=0.95\linewidth,cfbox=green]{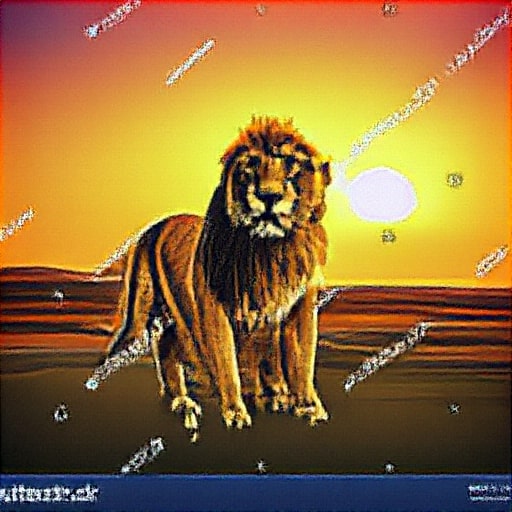}
            \end{subfigure}%            
            \begin{subfigure}[t]{0.09\linewidth}
               \includegraphics[width=\linewidth, height=0.95\linewidth,cfbox=cyan]{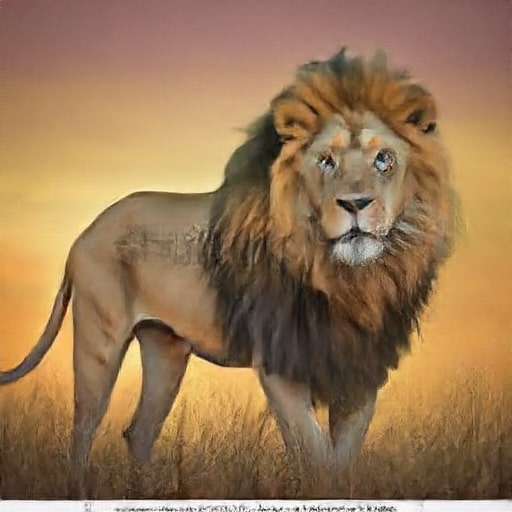}
            \end{subfigure}
            \hspace{2pt}
            \begin{subfigure}[t]{0.09\linewidth}
               \includegraphics[width=\linewidth, height=0.95\linewidth,cfbox=green]{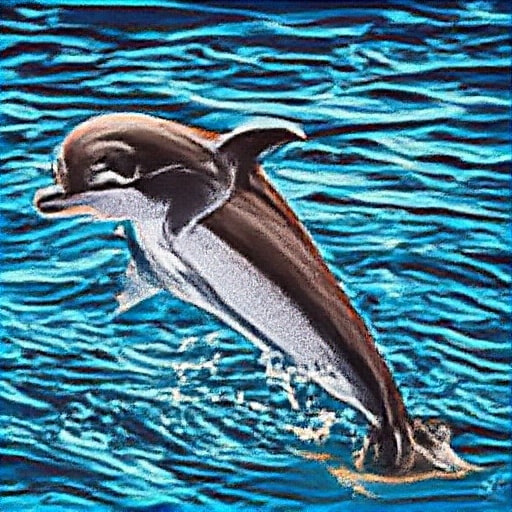}
            \end{subfigure}%            
            \begin{subfigure}[t]{0.09\linewidth}
               \includegraphics[width=\linewidth, height=0.95\linewidth,cfbox=cyan]{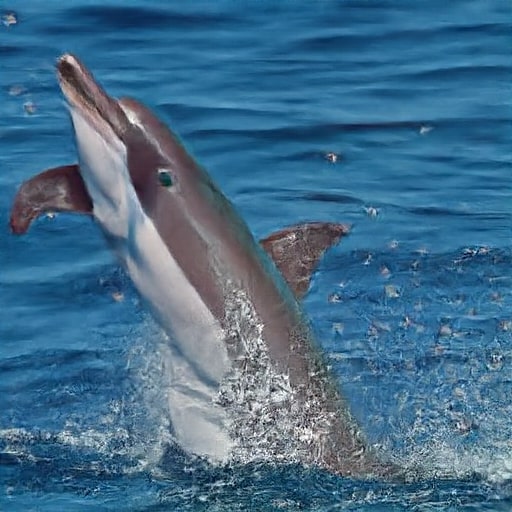}
            \end{subfigure}
            \hspace{2pt}
            \begin{subfigure}[t]{0.09\linewidth}
               \includegraphics[width=\linewidth, height=0.95\linewidth,cfbox=green]{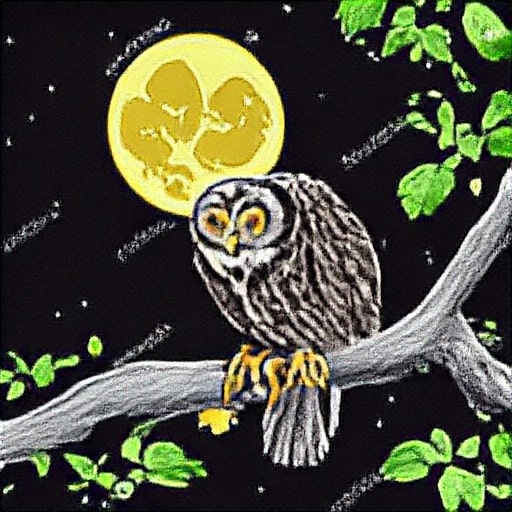}
            \end{subfigure}%            
            \begin{subfigure}[t]{0.09\linewidth}
               \includegraphics[width=\linewidth, height=0.95\linewidth,cfbox=cyan]{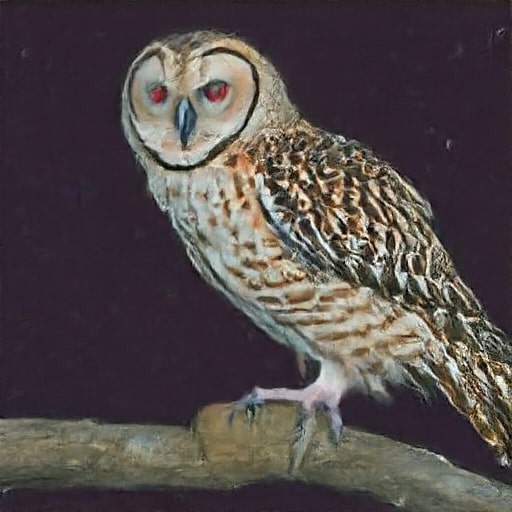}
            \end{subfigure}
            \hspace{2pt}
            \begin{subfigure}[t]{0.09\linewidth}
               \includegraphics[width=\linewidth, height=0.95\linewidth,cfbox=green]{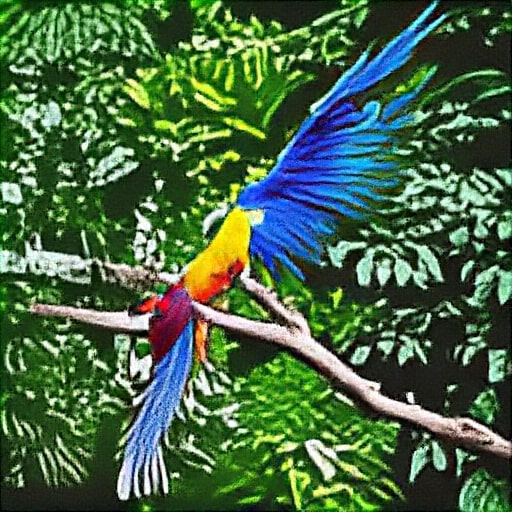}
            \end{subfigure}%            
            \begin{subfigure}[t]{0.09\linewidth}
               \includegraphics[width=\linewidth, height=0.95\linewidth,cfbox=cyan]{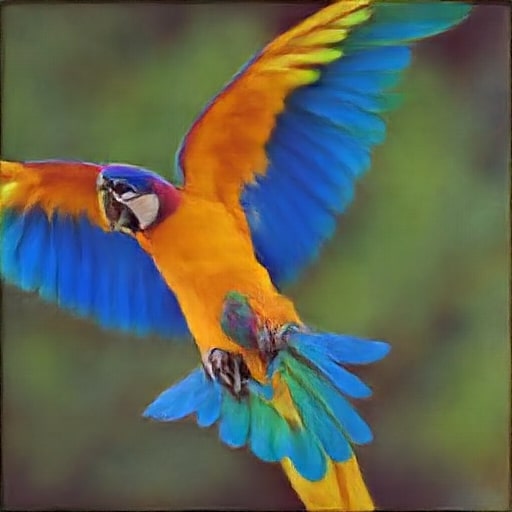}
            \end{subfigure}
            \hspace{2pt}
            \begin{subfigure}[t]{0.09\linewidth}
               \includegraphics[width=\linewidth, height=0.95\linewidth,cfbox=green]{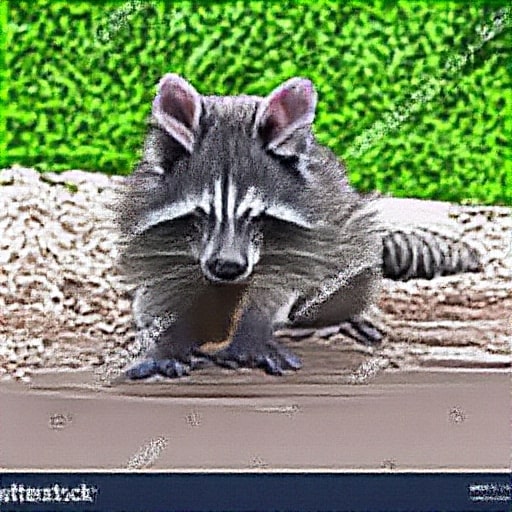}
            \end{subfigure}%            
            \begin{subfigure}[t]{0.09\linewidth}
               \includegraphics[width=\linewidth, height=0.95\linewidth,cfbox=cyan]{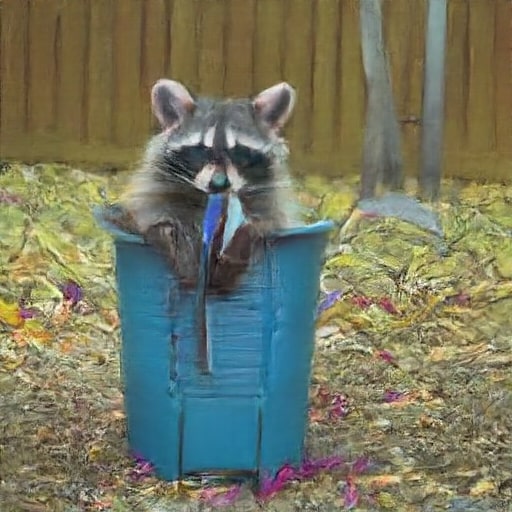}
            \end{subfigure}\\
            \begin{subfigure}[t]{0.09\linewidth}
               \includegraphics[width=\linewidth, height=0.95\linewidth,cfbox=green]{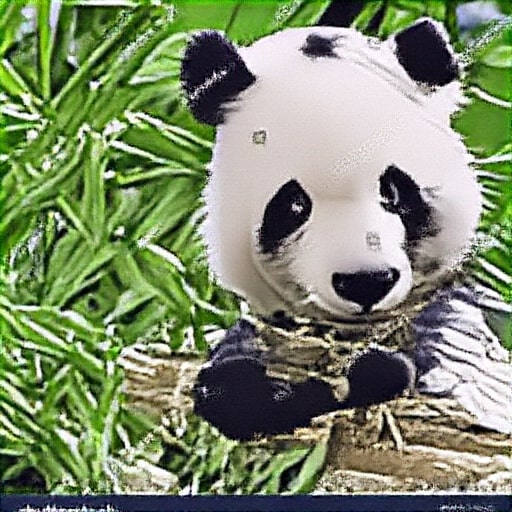}
            \end{subfigure}%            
            \begin{subfigure}[t]{0.09\linewidth}
               \includegraphics[width=\linewidth, height=0.95\linewidth,cfbox=cyan]{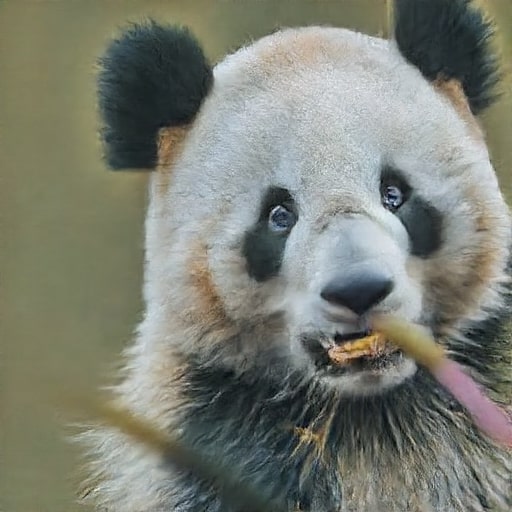}
            \end{subfigure}
            \hspace{2pt}
            \begin{subfigure}[t]{0.09\linewidth}
               \includegraphics[width=\linewidth, height=0.95\linewidth,cfbox=green]{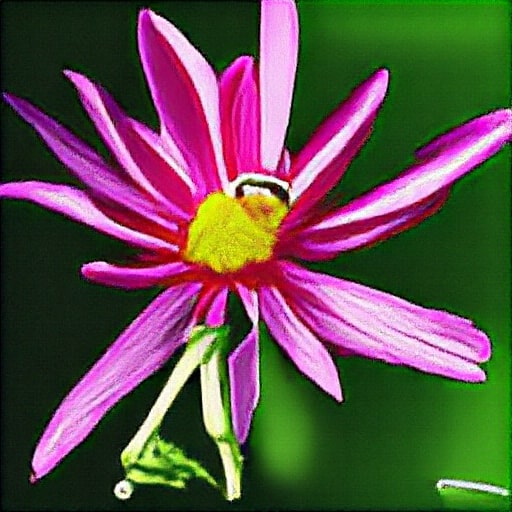}
            \end{subfigure}%            
            \begin{subfigure}[t]{0.09\linewidth}
               \includegraphics[width=\linewidth, height=0.95\linewidth,cfbox=cyan]{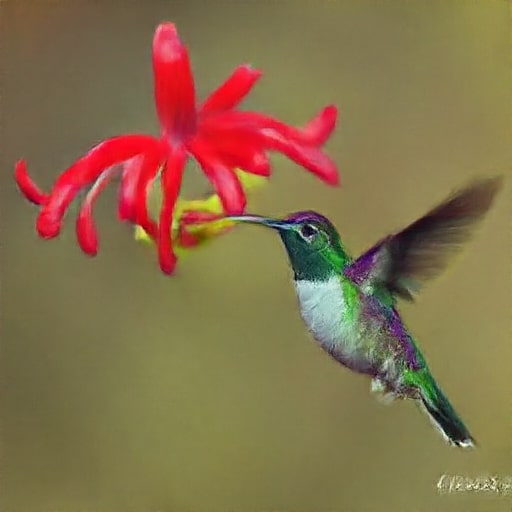}
            \end{subfigure}
            \hspace{2pt}
            \begin{subfigure}[t]{0.09\linewidth}
               \includegraphics[width=\linewidth, height=0.95\linewidth,cfbox=green]{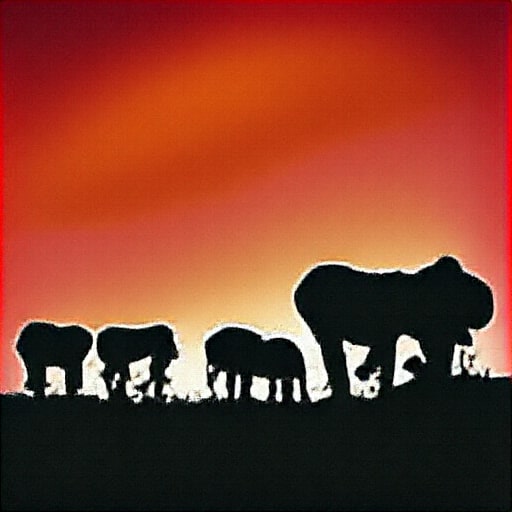}
            \end{subfigure}%            
            \begin{subfigure}[t]{0.09\linewidth}
               \includegraphics[width=\linewidth, height=0.95\linewidth,cfbox=cyan]{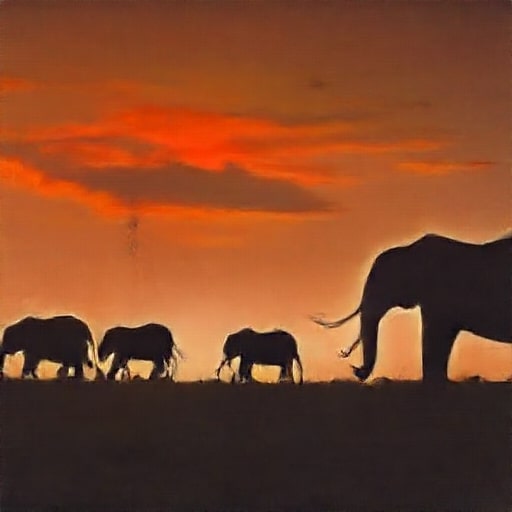}
            \end{subfigure}
            \hspace{2pt}
            \begin{subfigure}[t]{0.09\linewidth}
               \includegraphics[width=\linewidth, height=0.95\linewidth,cfbox=green]{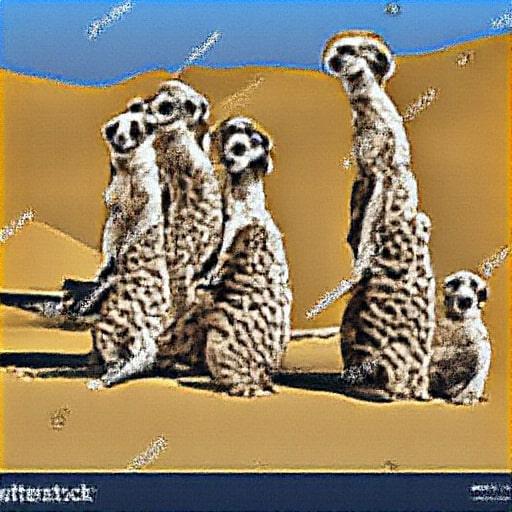}
            \end{subfigure}%            
            \begin{subfigure}[t]{0.09\linewidth}
               \includegraphics[width=\linewidth, height=0.95\linewidth,cfbox=cyan]{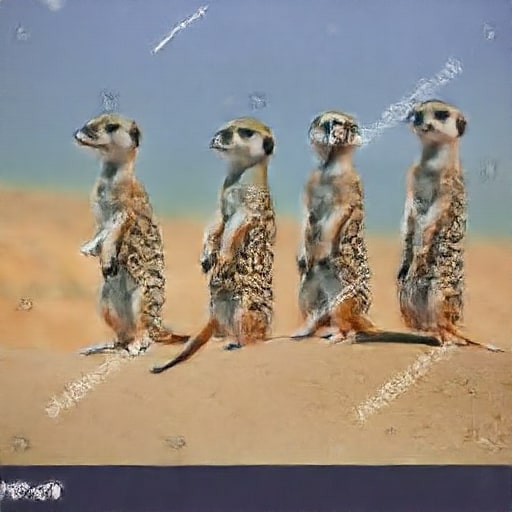}
            \end{subfigure}
            \hspace{2pt}
            \begin{subfigure}[t]{0.09\linewidth}
               \includegraphics[width=\linewidth, height=0.95\linewidth,cfbox=green]{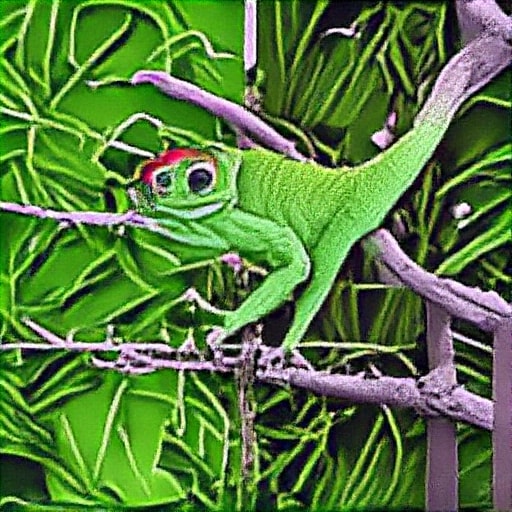}
            \end{subfigure}%            
            \begin{subfigure}[t]{0.09\linewidth}
               \includegraphics[width=\linewidth, height=0.95\linewidth,cfbox=cyan]{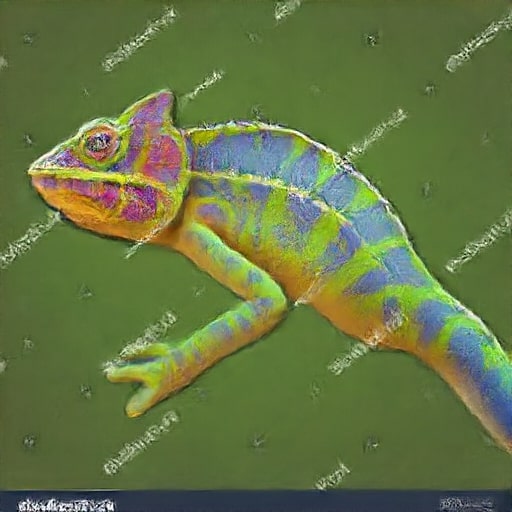}
            \end{subfigure}\\
            \begin{subfigure}[t]{0.09\linewidth}
               \includegraphics[width=\linewidth, height=0.95\linewidth,cfbox=green]{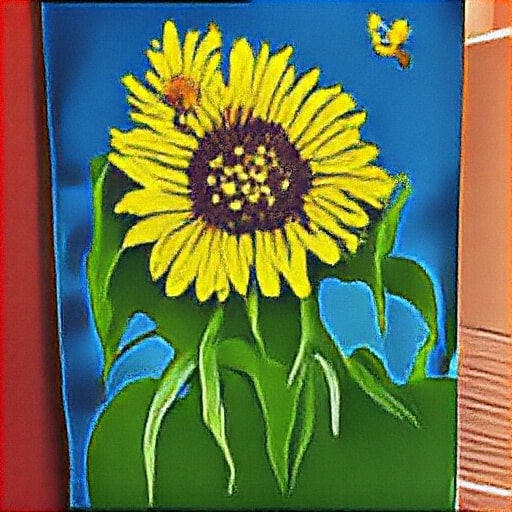}
            \end{subfigure}%            
            \begin{subfigure}[t]{0.09\linewidth}
               \includegraphics[width=\linewidth, height=0.95\linewidth,cfbox=cyan]{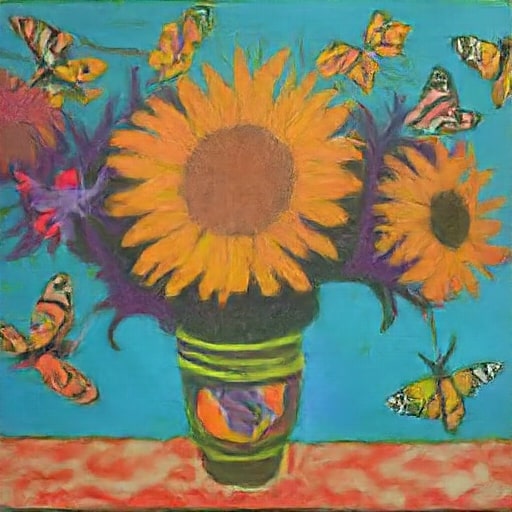}
            \end{subfigure}
            \hspace{2pt}
            \begin{subfigure}[t]{0.09\linewidth}
               \includegraphics[width=\linewidth, height=0.95\linewidth,cfbox=green]{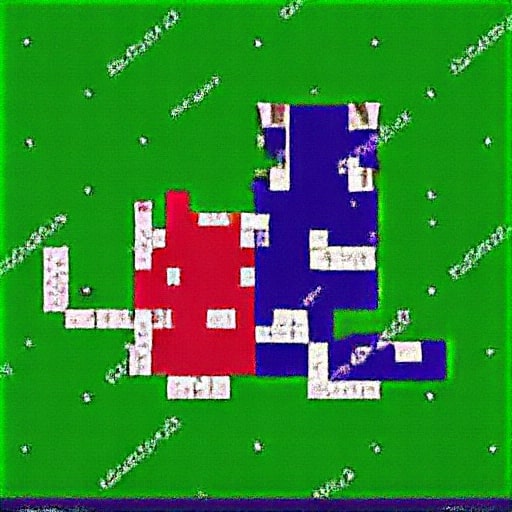}
            \end{subfigure}%            
            \begin{subfigure}[t]{0.09\linewidth}
               \includegraphics[width=\linewidth, height=0.95\linewidth,cfbox=cyan]{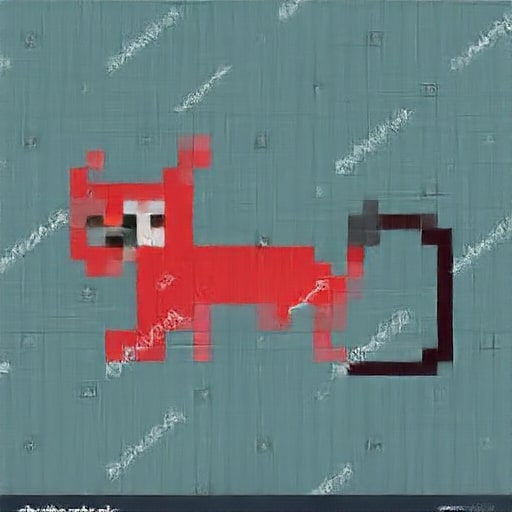}
            \end{subfigure}
            \hspace{2pt}
            \begin{subfigure}[t]{0.09\linewidth}
               \includegraphics[width=\linewidth, height=0.95\linewidth,cfbox=green]{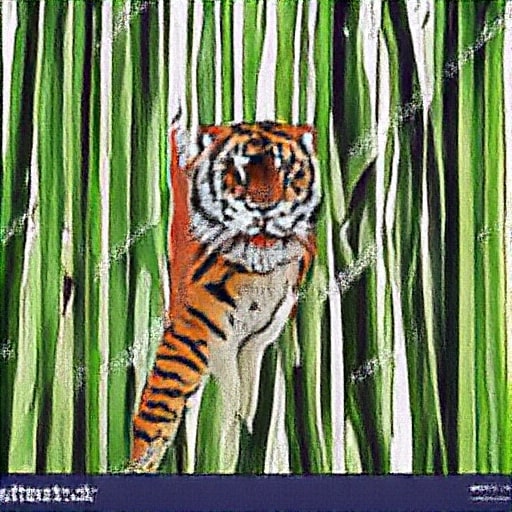}
            \end{subfigure}%            
            \begin{subfigure}[t]{0.09\linewidth}
               \includegraphics[width=\linewidth, height=0.95\linewidth,cfbox=cyan]{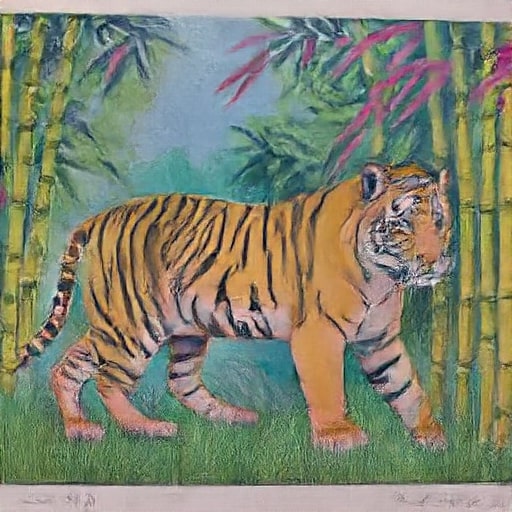}
            \end{subfigure}
            \hspace{2pt}
            \begin{subfigure}[t]{0.09\linewidth}
               \includegraphics[width=\linewidth, height=0.95\linewidth,cfbox=green]{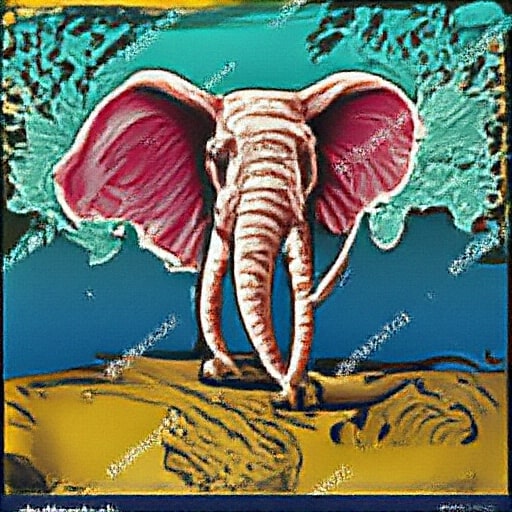}
            \end{subfigure}%            
            \begin{subfigure}[t]{0.09\linewidth}
               \includegraphics[width=\linewidth, height=0.95\linewidth,cfbox=cyan]{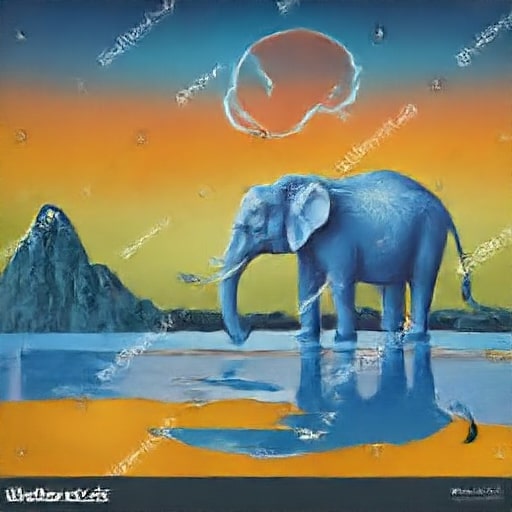}
            \end{subfigure}
            \hspace{2pt}
            \begin{subfigure}[t]{0.09\linewidth}
               \includegraphics[width=\linewidth, height=0.95\linewidth,cfbox=green]{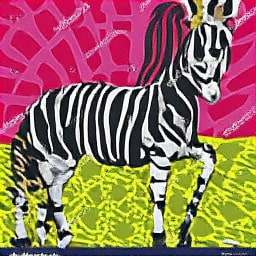}
            \end{subfigure}%            
            \begin{subfigure}[t]{0.09\linewidth}
               \includegraphics[width=\linewidth, height=0.95\linewidth,cfbox=cyan]{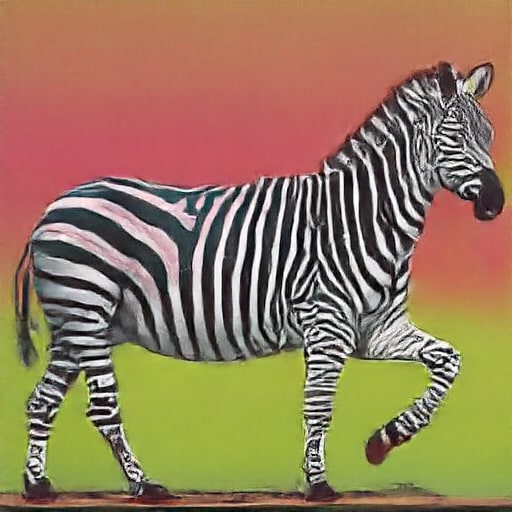}
            \end{subfigure}\\
            \begin{subfigure}[t]{0.09\linewidth}
               \includegraphics[width=\linewidth, height=0.95\linewidth,cfbox=green]{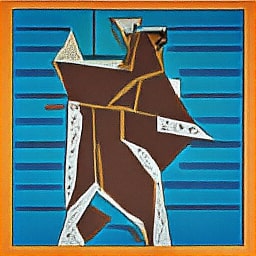}
            \end{subfigure}%            
            \begin{subfigure}[t]{0.09\linewidth}
               \includegraphics[width=\linewidth, height=0.95\linewidth,cfbox=cyan]{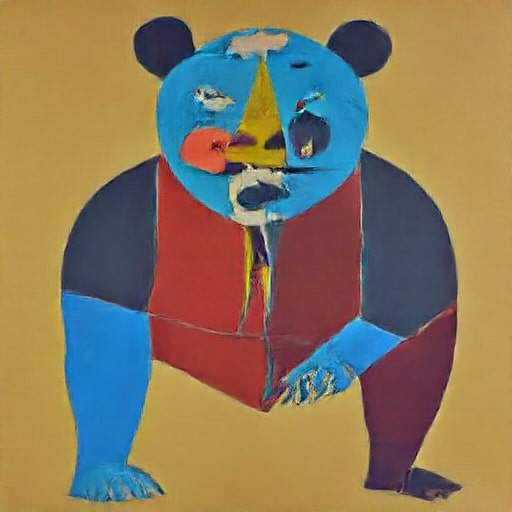}
            \end{subfigure}
            \hspace{2pt}
            \begin{subfigure}[t]{0.09\linewidth}
               \includegraphics[width=\linewidth, height=0.95\linewidth,cfbox=green]{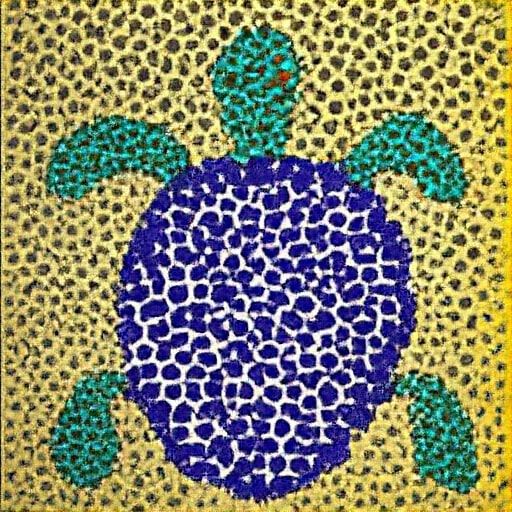}
            \end{subfigure}%            
            \begin{subfigure}[t]{0.09\linewidth}
               \includegraphics[width=\linewidth, height=0.95\linewidth,cfbox=cyan]{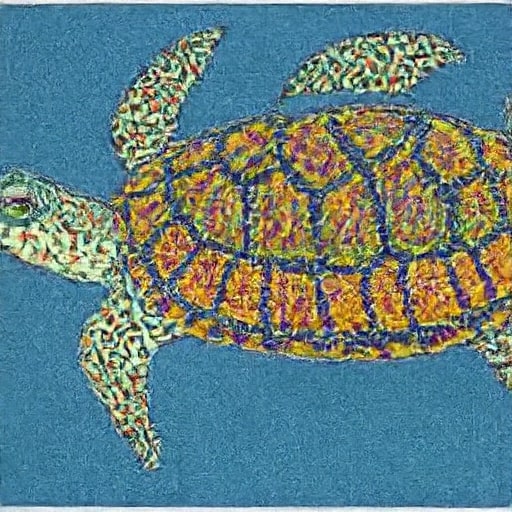}
            \end{subfigure}
            \hspace{2pt}
            \begin{subfigure}[t]{0.09\linewidth}
               \includegraphics[width=\linewidth, height=0.95\linewidth,cfbox=green]{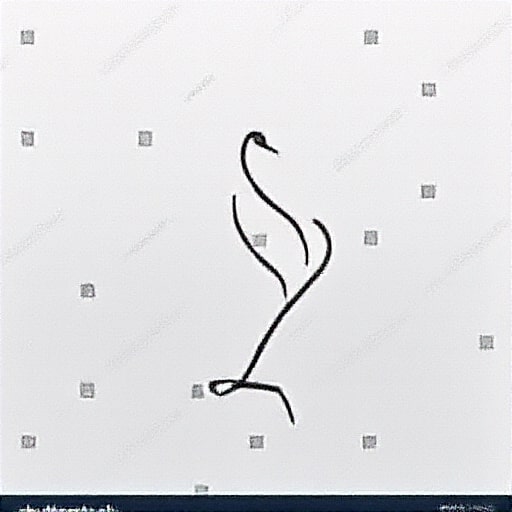}
            \end{subfigure}%            
            \begin{subfigure}[t]{0.09\linewidth}
               \includegraphics[width=\linewidth, height=0.95\linewidth,cfbox=cyan]{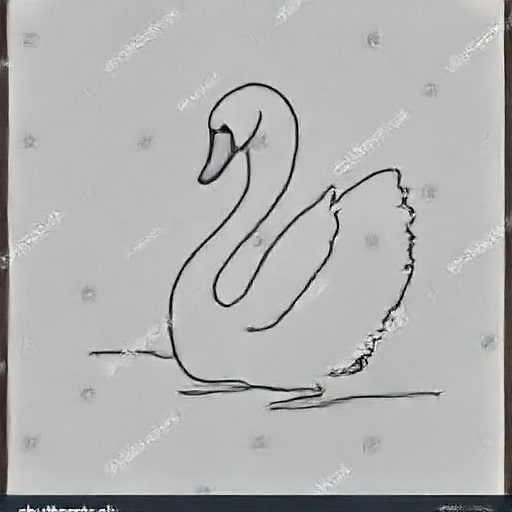}
            \end{subfigure}
            \hspace{2pt}
            \begin{subfigure}[t]{0.09\linewidth}
               \includegraphics[width=\linewidth, height=0.95\linewidth,cfbox=green]{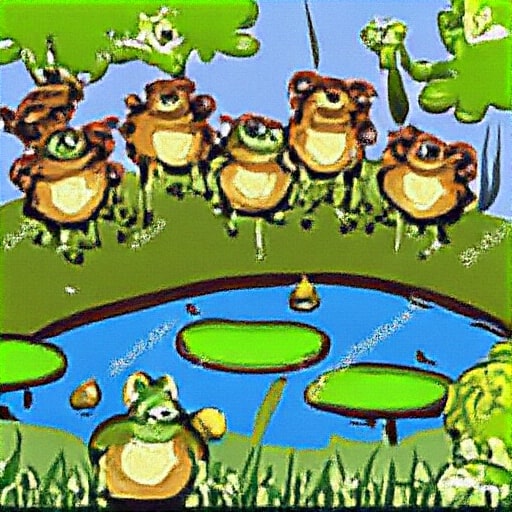}
            \end{subfigure}%            
            \begin{subfigure}[t]{0.09\linewidth}
               \includegraphics[width=\linewidth, height=0.95\linewidth,cfbox=cyan]{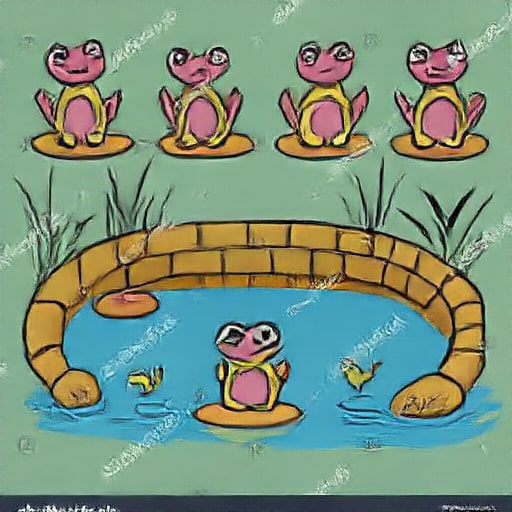}
            \end{subfigure}
            \hspace{2pt}
            \begin{subfigure}[t]{0.09\linewidth}
               \includegraphics[width=\linewidth, height=0.95\linewidth,cfbox=green]{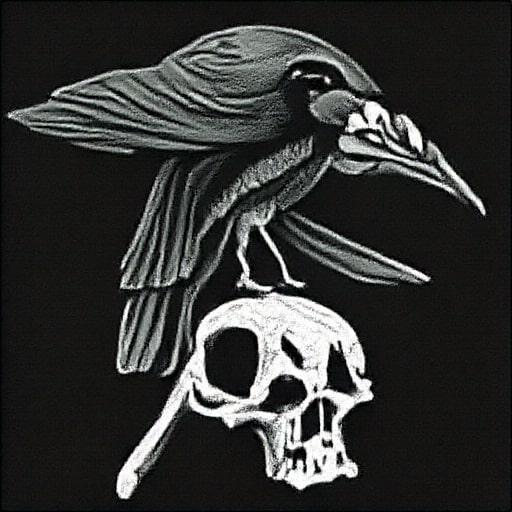}
            \end{subfigure}%            
            \begin{subfigure}[t]{0.09\linewidth}
               \includegraphics[width=\linewidth, height=0.95\linewidth,cfbox=cyan]{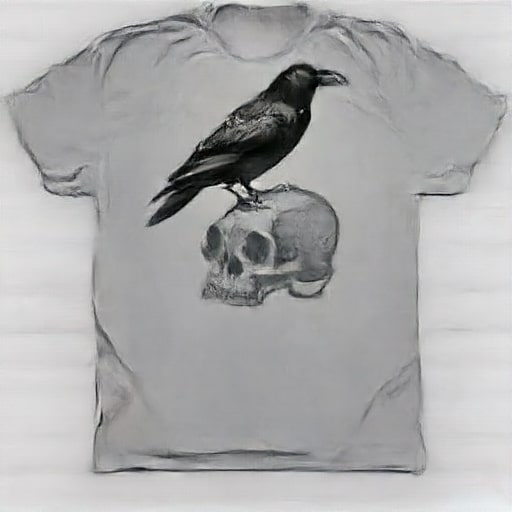}
            \end{subfigure}\\
            \begin{subfigure}[t]{0.09\linewidth}
               \includegraphics[width=\linewidth, height=0.95\linewidth,cfbox=green]{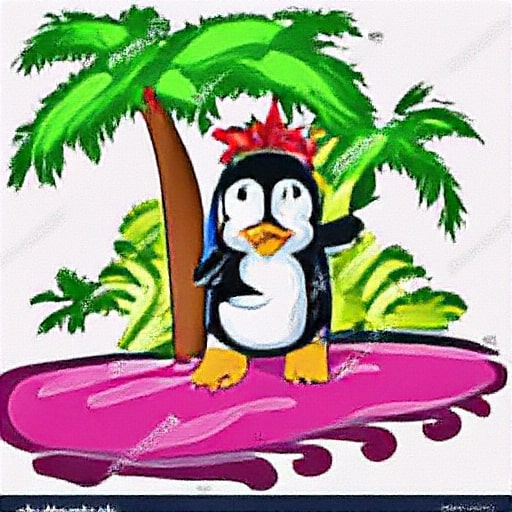}
            \end{subfigure}% 
            \begin{subfigure}[t]{0.09\linewidth}
               \includegraphics[width=\linewidth, height=0.95\linewidth,cfbox=cyan]{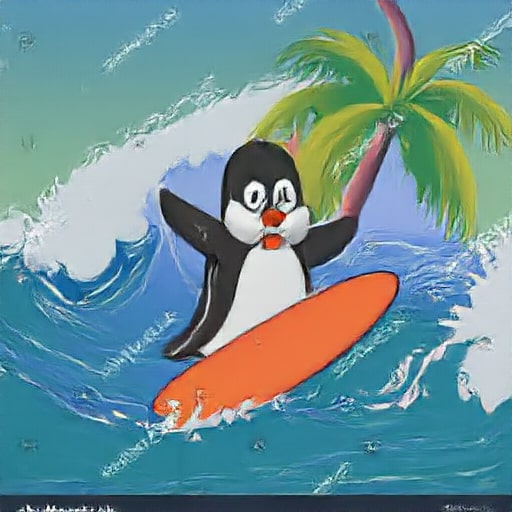}
            \end{subfigure}
            \hspace{2pt}
            \begin{subfigure}[t]{0.09\linewidth}
               \includegraphics[width=\linewidth, height=0.95\linewidth,cfbox=green]{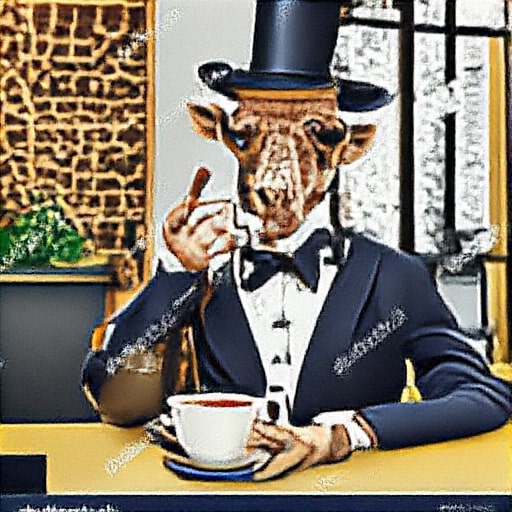}
            \end{subfigure}%            
            \begin{subfigure}[t]{0.09\linewidth}
               \includegraphics[width=\linewidth, height=0.95\linewidth,cfbox=cyan]{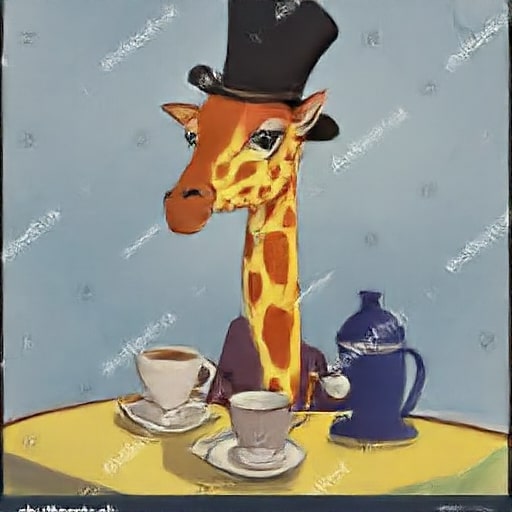}
            \end{subfigure}
            \hspace{2pt}
            \begin{subfigure}[t]{0.09\linewidth}
               \includegraphics[width=\linewidth, height=0.95\linewidth,cfbox=green]{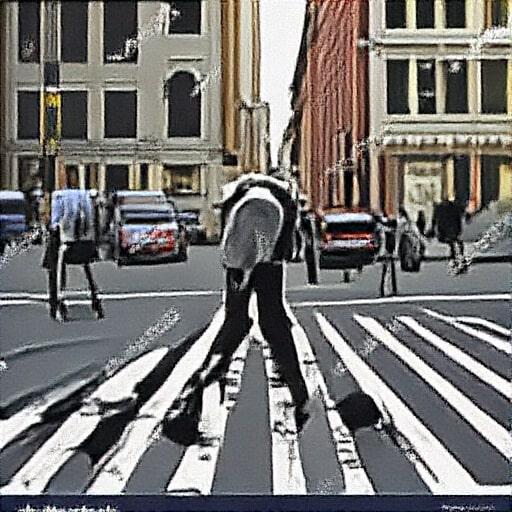}
            \end{subfigure}%            
            \begin{subfigure}[t]{0.09\linewidth}
               \includegraphics[width=\linewidth, height=0.95\linewidth,cfbox=cyan]{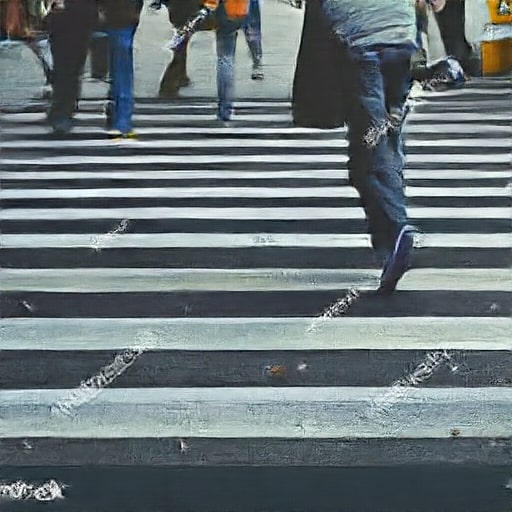}
            \end{subfigure}
            \hspace{2pt}
            \begin{subfigure}[t]{0.09\linewidth}
               \includegraphics[width=\linewidth, height=0.95\linewidth,cfbox=green]{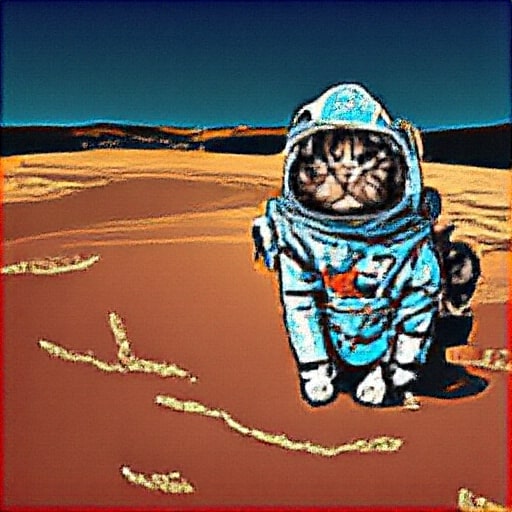}
            \end{subfigure}%            
            \begin{subfigure}[t]{0.09\linewidth}
               \includegraphics[width=\linewidth, height=0.95\linewidth,cfbox=cyan]{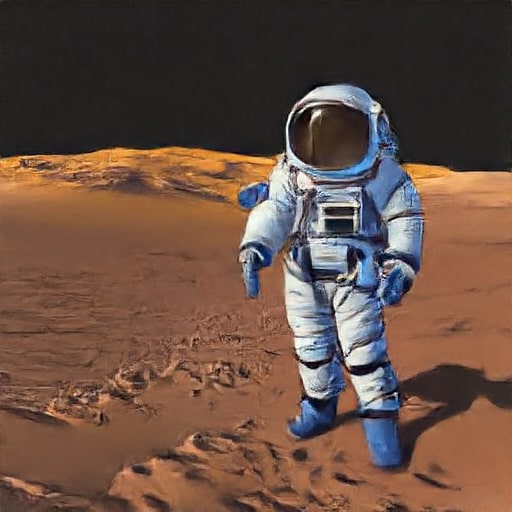}
            \end{subfigure}
            \hspace{2pt}
            \begin{subfigure}[t]{0.09\linewidth}
               \includegraphics[width=\linewidth, height=0.95\linewidth,cfbox=green]{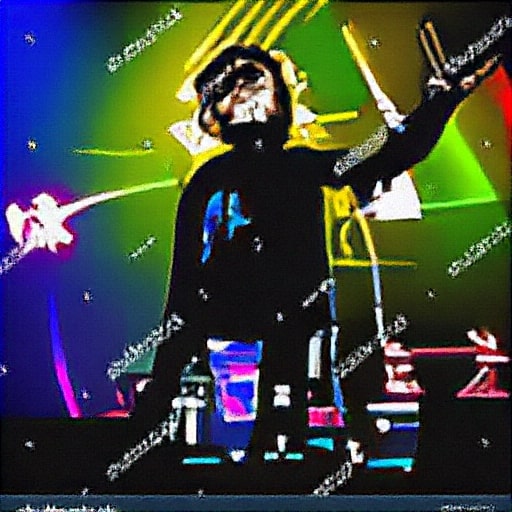}
            \end{subfigure}%            
            \begin{subfigure}[t]{0.09\linewidth}
               \includegraphics[width=\linewidth, height=0.95\linewidth,cfbox=cyan]{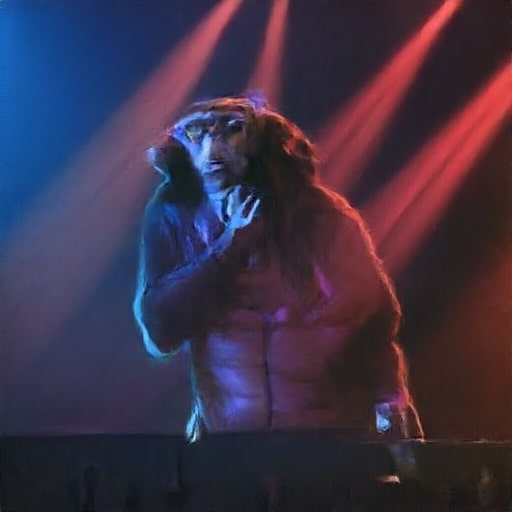}
            \end{subfigure}\\
            \begin{subfigure}[t]{0.09\linewidth}
               \includegraphics[width=\linewidth, height=0.95\linewidth,cfbox=green]{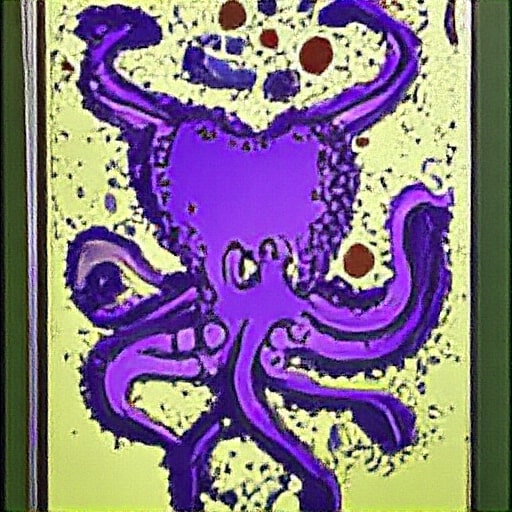}
            \end{subfigure}%            
            \begin{subfigure}[t]{0.09\linewidth}
               \includegraphics[width=\linewidth, height=0.95\linewidth,cfbox=cyan]{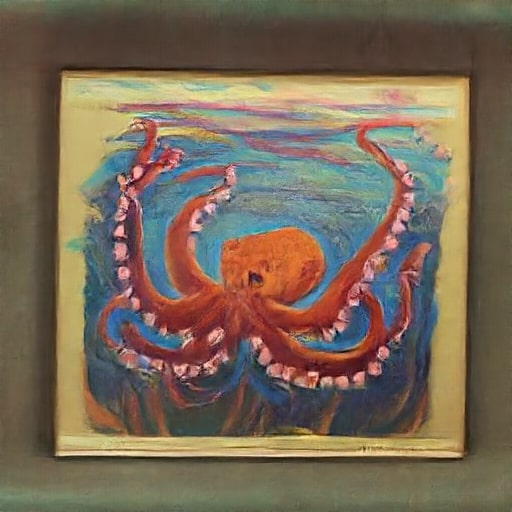}
            \end{subfigure}
            \hspace{2pt}
            \begin{subfigure}[t]{0.09\linewidth}
               \includegraphics[width=\linewidth, height=0.95\linewidth,cfbox=green]{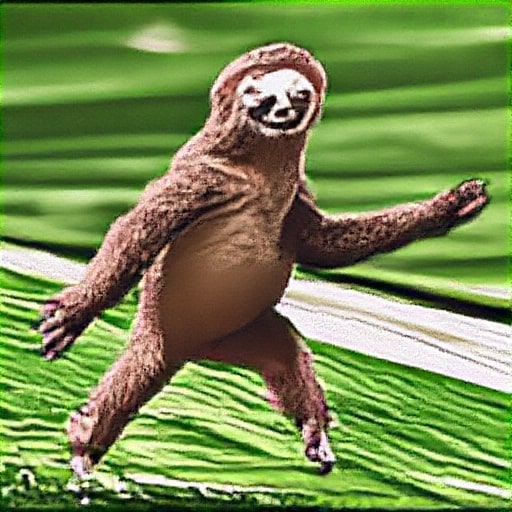}
            \end{subfigure}%            
            \begin{subfigure}[t]{0.09\linewidth}
               \includegraphics[width=\linewidth, height=0.95\linewidth,cfbox=cyan]{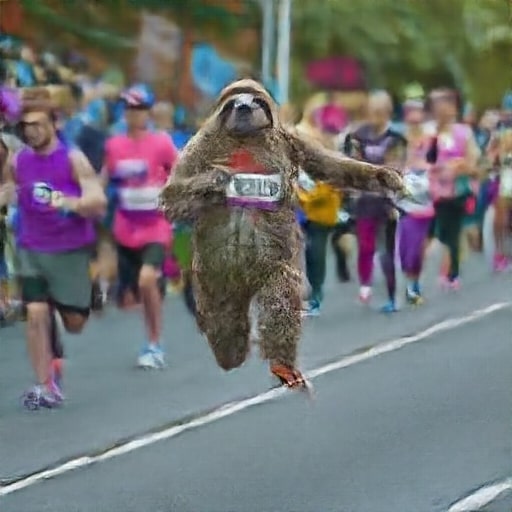}
            \end{subfigure}
            \hspace{2pt}
            \begin{subfigure}[t]{0.09\linewidth}
               \includegraphics[width=\linewidth, height=0.95\linewidth,cfbox=green]{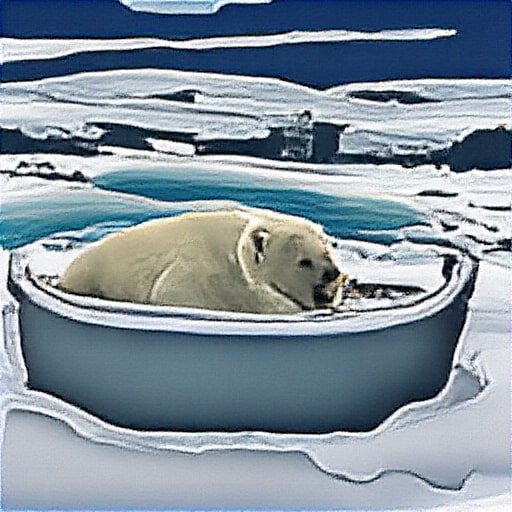}
            \end{subfigure}%            
            \begin{subfigure}[t]{0.09\linewidth}
               \includegraphics[width=\linewidth, height=0.95\linewidth,cfbox=cyan]{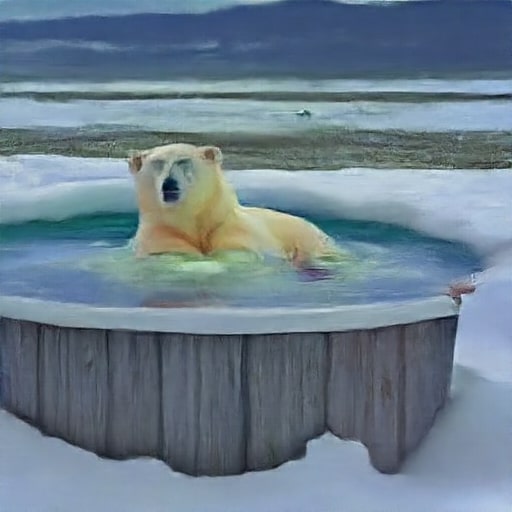}
            \end{subfigure}
            \hspace{2pt}
            \begin{subfigure}[t]{0.09\linewidth}
               \includegraphics[width=\linewidth, height=0.95\linewidth,cfbox=green]{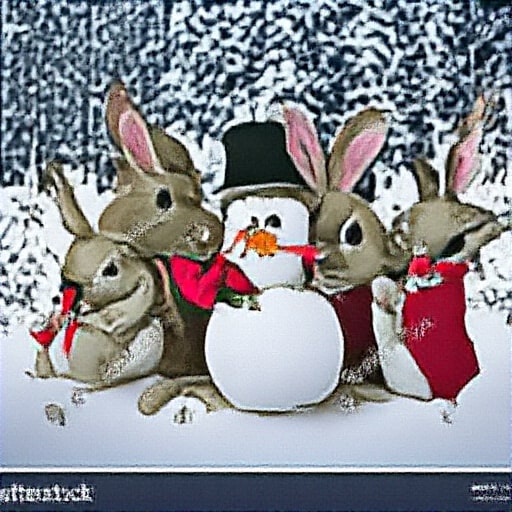}
            \end{subfigure}%            
            \begin{subfigure}[t]{0.09\linewidth}
               \includegraphics[width=\linewidth, height=0.95\linewidth,cfbox=cyan]{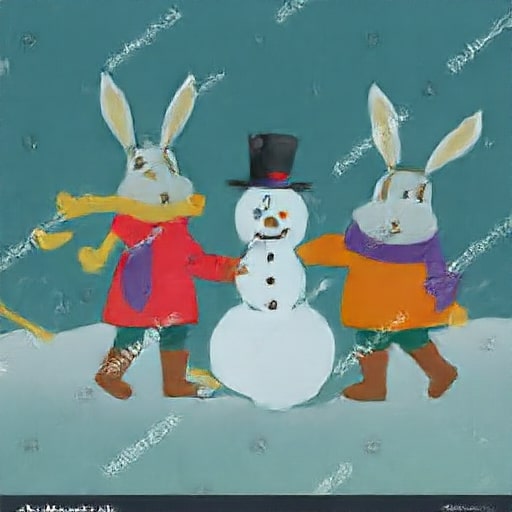}
            \end{subfigure}
            \hspace{2pt}
            \begin{subfigure}[t]{0.09\linewidth}
               \includegraphics[width=\linewidth, height=0.95\linewidth,cfbox=green]{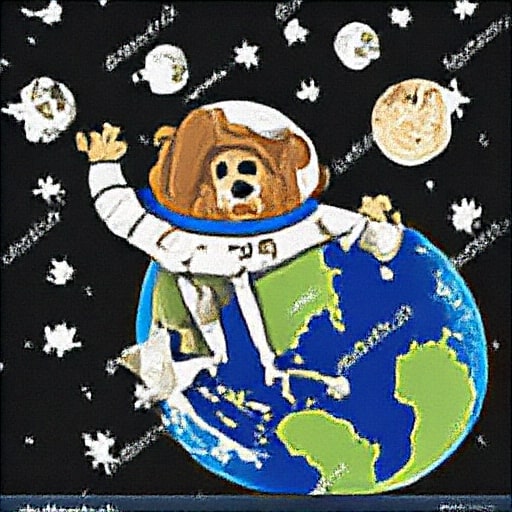}
            \end{subfigure}%            
            \begin{subfigure}[t]{0.09\linewidth}
               \includegraphics[width=\linewidth, height=0.95\linewidth,cfbox=cyan]{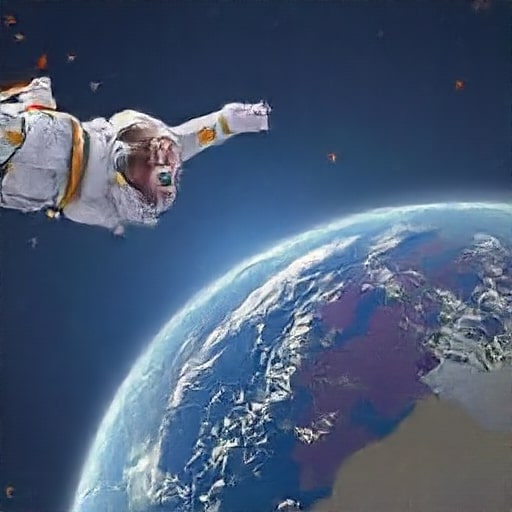}
            \end{subfigure}\\
            \begin{subfigure}[t]{0.09\linewidth}
               \includegraphics[width=\linewidth, height=0.95\linewidth,cfbox=green]{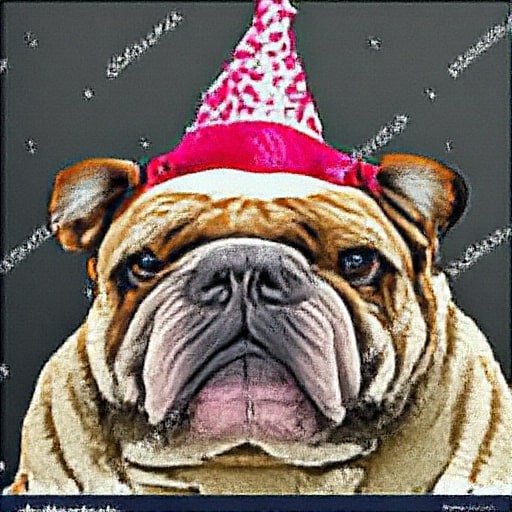}
            \end{subfigure}% 
            \begin{subfigure}[t]{0.09\linewidth}
               \includegraphics[width=\linewidth, height=0.95\linewidth,cfbox=cyan]{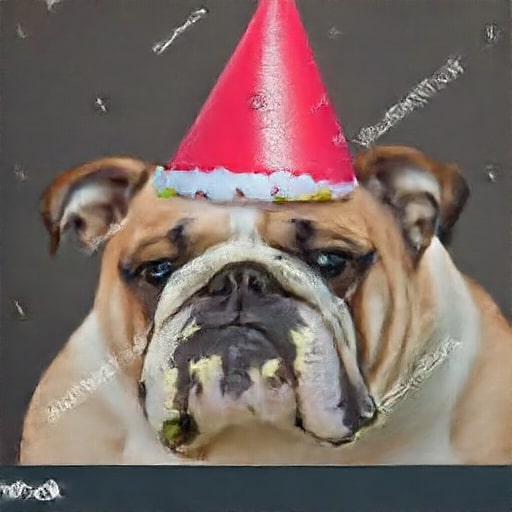}
            \end{subfigure}
            \hspace{2pt}
            \begin{subfigure}[t]{0.09\linewidth}
               \includegraphics[width=\linewidth, height=0.95\linewidth,cfbox=green]{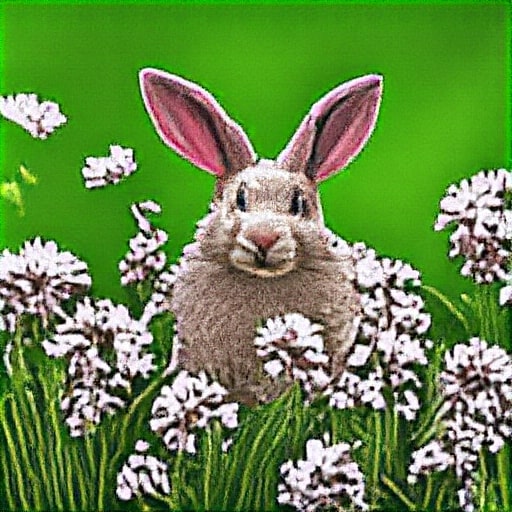}
            \end{subfigure}%            
            \begin{subfigure}[t]{0.09\linewidth}
               \includegraphics[width=\linewidth, height=0.95\linewidth,cfbox=cyan]{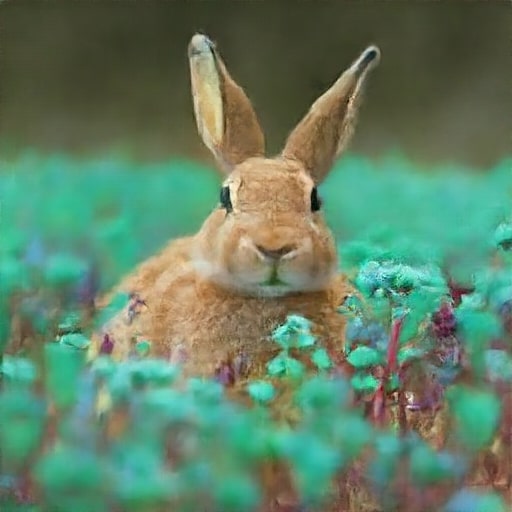}
            \end{subfigure}
            \hspace{2pt}
            \begin{subfigure}[t]{0.09\linewidth}
               \includegraphics[width=\linewidth, height=0.95\linewidth,cfbox=green]{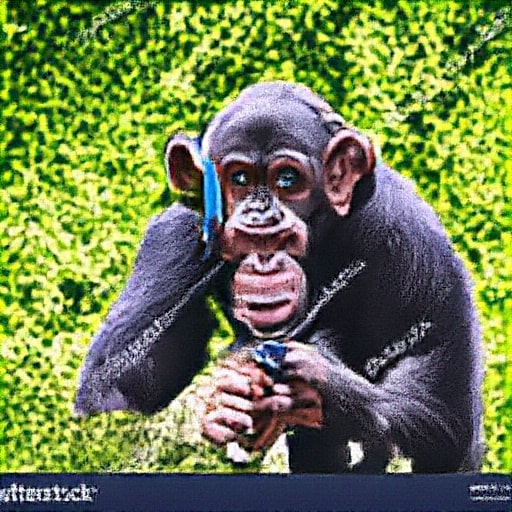}
            \end{subfigure}%            
            \begin{subfigure}[t]{0.09\linewidth}
               \includegraphics[width=\linewidth, height=0.95\linewidth,cfbox=cyan]{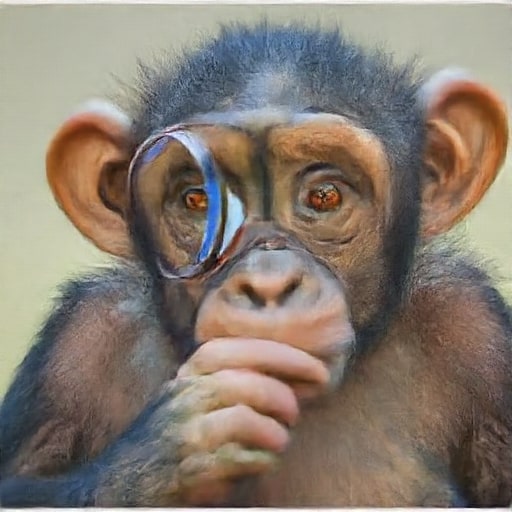}
            \end{subfigure}
            \hspace{2pt}
            \begin{subfigure}[t]{0.09\linewidth}
               \includegraphics[width=\linewidth, height=0.95\linewidth,cfbox=green]{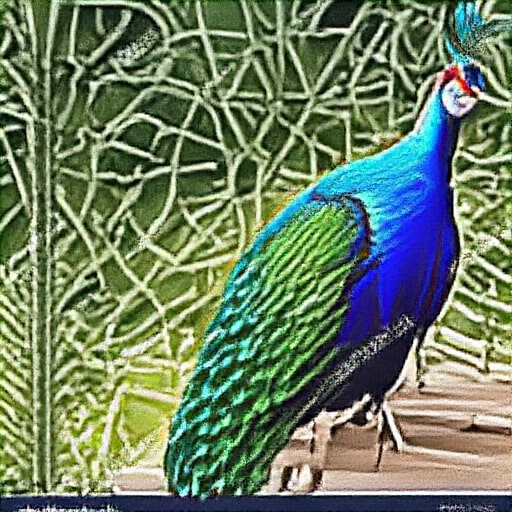}
            \end{subfigure}%            
            \begin{subfigure}[t]{0.09\linewidth}
               \includegraphics[width=\linewidth, height=0.95\linewidth,cfbox=cyan]{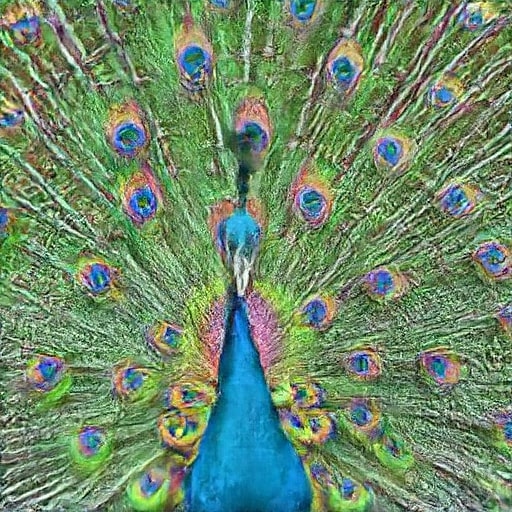}
            \end{subfigure}
            \hspace{2pt}
            \begin{subfigure}[t]{0.09\linewidth}
               \includegraphics[width=\linewidth, height=0.95\linewidth,cfbox=green]{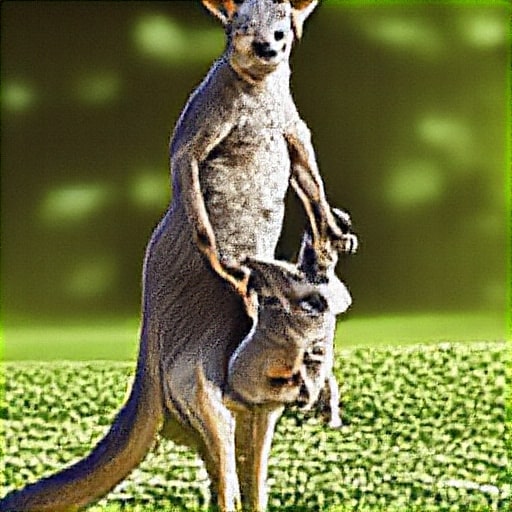}
            \end{subfigure}%            
            \begin{subfigure}[t]{0.09\linewidth}
               \includegraphics[width=\linewidth, height=0.95\linewidth,cfbox=cyan]{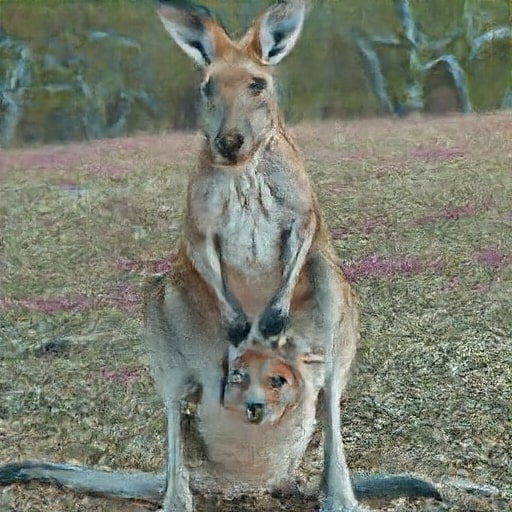}
            \end{subfigure}\\
            \begin{subfigure}[t]{0.09\linewidth}
               \includegraphics[width=\linewidth, height=0.95\linewidth,cfbox=green]{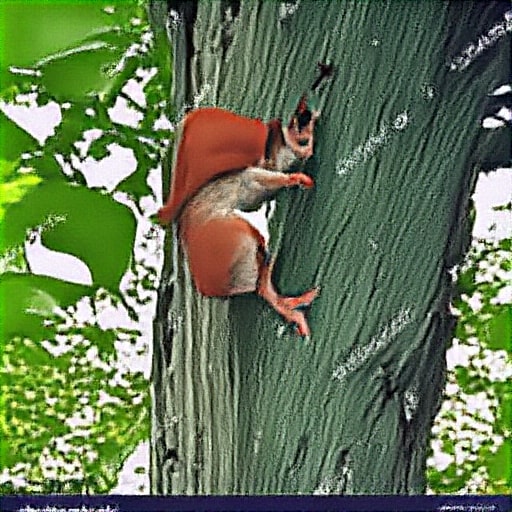}
            \end{subfigure}%            
            \begin{subfigure}[t]{0.09\linewidth}
               \includegraphics[width=\linewidth, height=0.95\linewidth,cfbox=cyan]{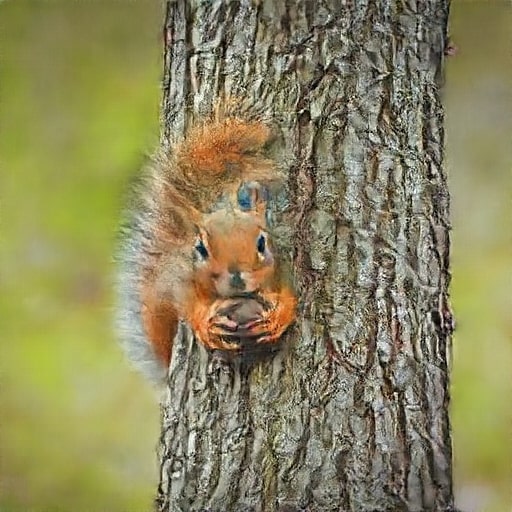}
            \end{subfigure}
            \hspace{2pt}
            \begin{subfigure}[t]{0.09\linewidth}
               \includegraphics[width=\linewidth, height=0.95\linewidth,cfbox=green]{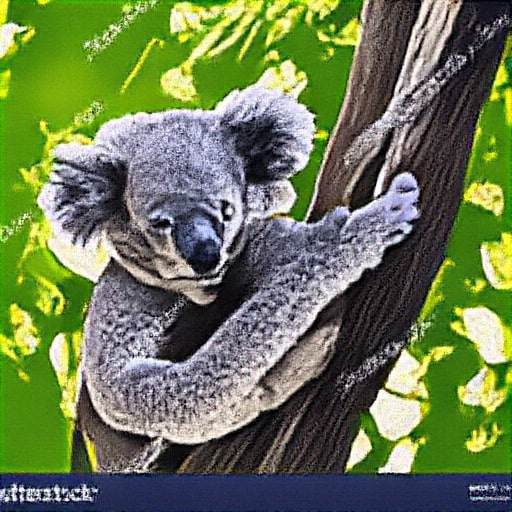}
            \end{subfigure}%            
            \begin{subfigure}[t]{0.09\linewidth}
               \includegraphics[width=\linewidth, height=0.95\linewidth,cfbox=cyan]{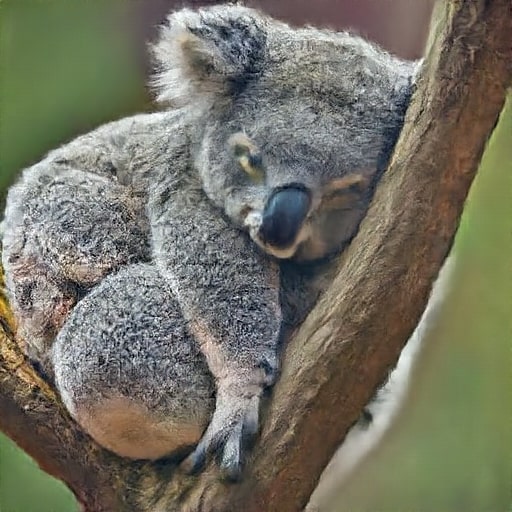}
            \end{subfigure}
            \hspace{2pt}
            \begin{subfigure}[t]{0.09\linewidth}
               \includegraphics[width=\linewidth, height=0.95\linewidth,cfbox=green]{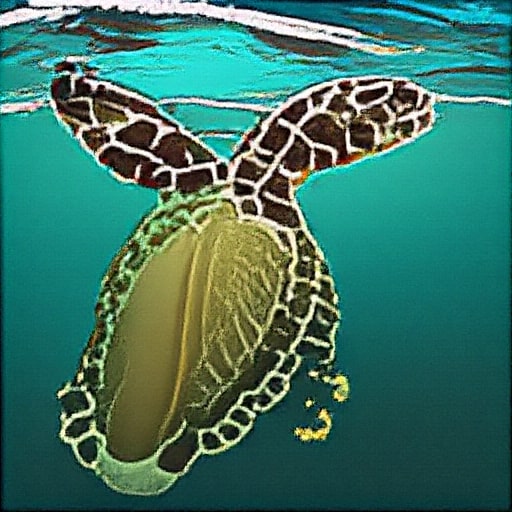}
            \end{subfigure}%            
            \begin{subfigure}[t]{0.09\linewidth}
               \includegraphics[width=\linewidth, height=0.95\linewidth,cfbox=cyan]{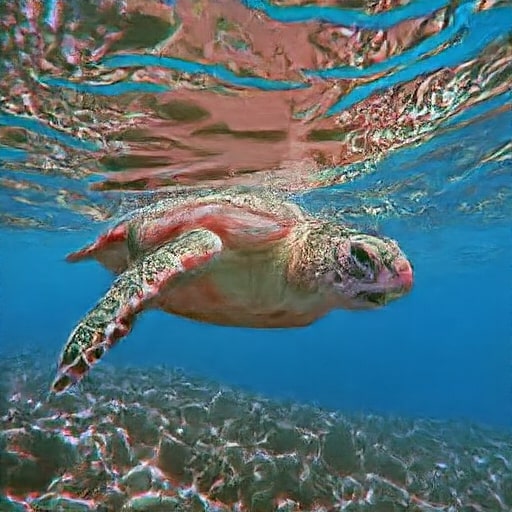}
            \end{subfigure}
            \hspace{2pt}
            \begin{subfigure}[t]{0.09\linewidth}
               \includegraphics[width=\linewidth, height=0.95\linewidth,cfbox=green]{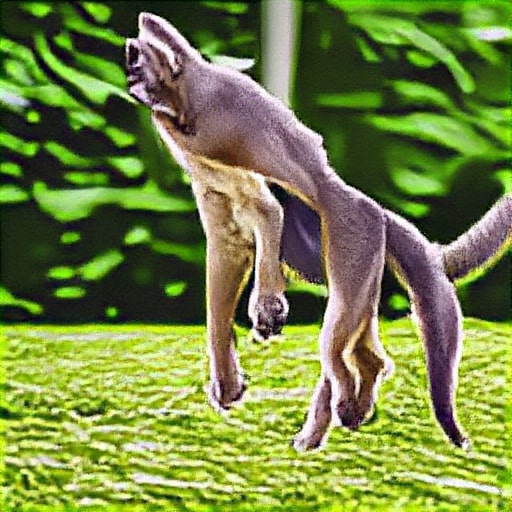}
            \end{subfigure}%            
            \begin{subfigure}[t]{0.09\linewidth}
               \includegraphics[width=\linewidth, height=0.95\linewidth,cfbox=cyan]{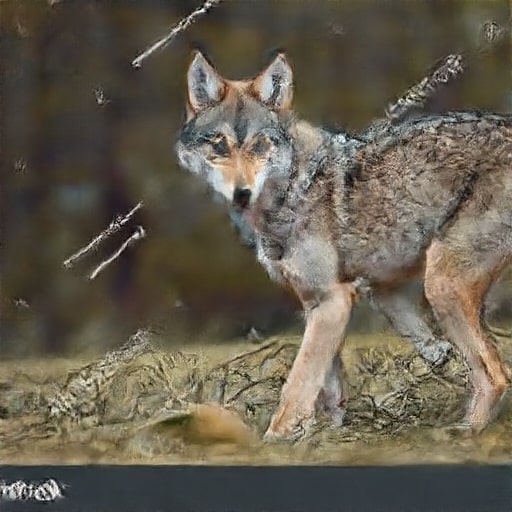}
            \end{subfigure}
            \hspace{2pt}
            \begin{subfigure}[t]{0.09\linewidth}
               \includegraphics[width=\linewidth, height=0.95\linewidth,cfbox=green]{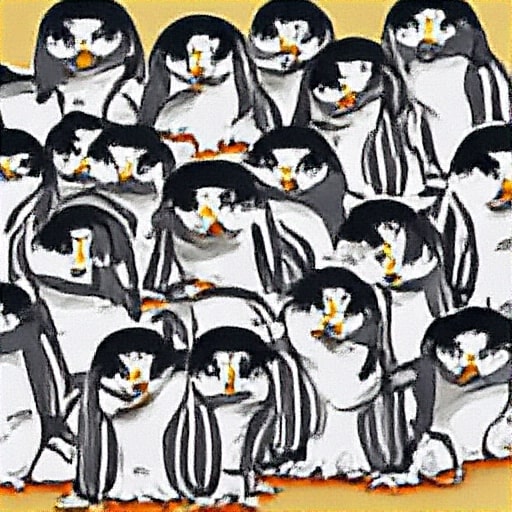}
            \end{subfigure}%            
            \begin{subfigure}[t]{0.09\linewidth}
               \includegraphics[width=\linewidth, height=0.95\linewidth,cfbox=cyan]{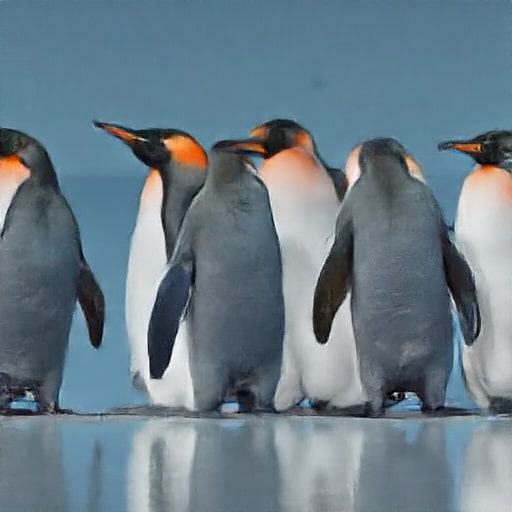}
            \end{subfigure}\\
            \hspace{2pt}
            \begin{subfigure}[t]{0.09\linewidth}
               \includegraphics[width=\linewidth, height=0.95\linewidth,cfbox=green]{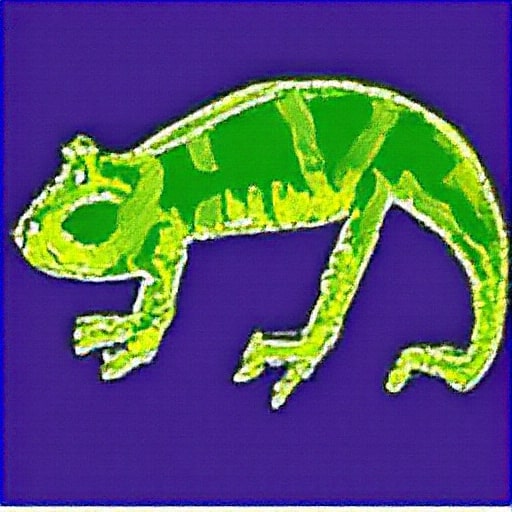}
            \end{subfigure}%            
            \begin{subfigure}[t]{0.09\linewidth}
               \includegraphics[width=\linewidth, height=0.95\linewidth,cfbox=cyan]{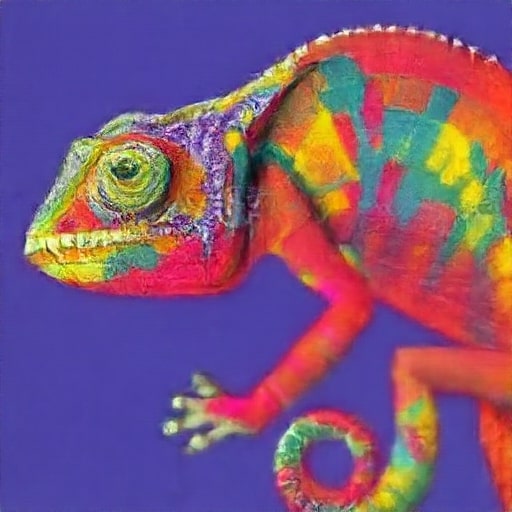}
            \end{subfigure}
            \hspace{2pt}
            \begin{subfigure}[t]{0.09\linewidth}
               \includegraphics[width=\linewidth, height=0.95\linewidth,cfbox=green]{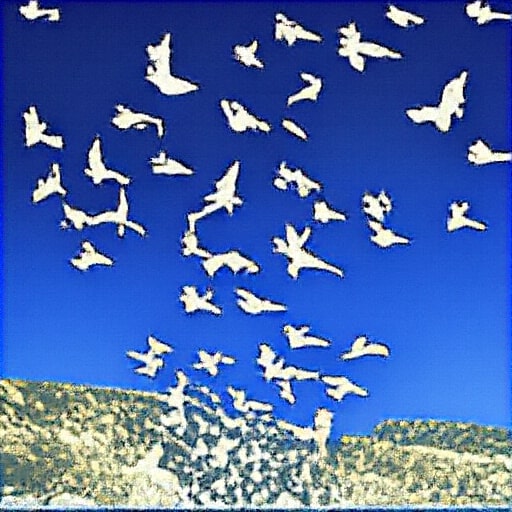}
            \end{subfigure}%            
            \begin{subfigure}[t]{0.09\linewidth}
               \includegraphics[width=\linewidth, height=0.95\linewidth,cfbox=cyan]{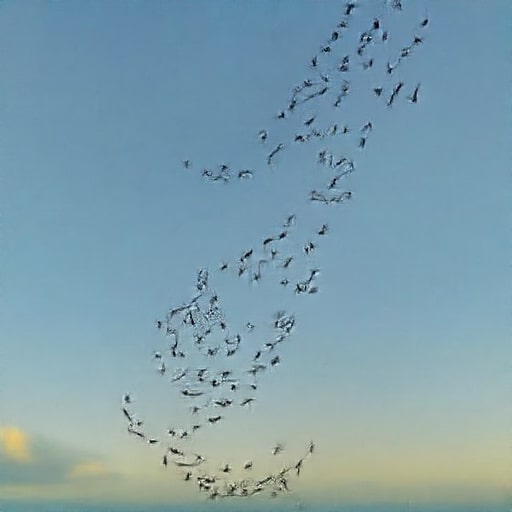}
            \end{subfigure}
            \hspace{2pt}
            \begin{subfigure}[t]{0.09\linewidth}
               \includegraphics[width=\linewidth, height=0.95\linewidth,cfbox=green]{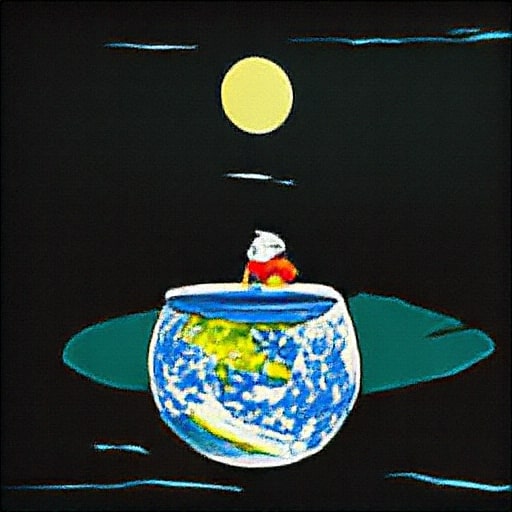}
            \end{subfigure}%            
            \begin{subfigure}[t]{0.09\linewidth}
               \includegraphics[width=\linewidth, height=0.95\linewidth,cfbox=cyan]{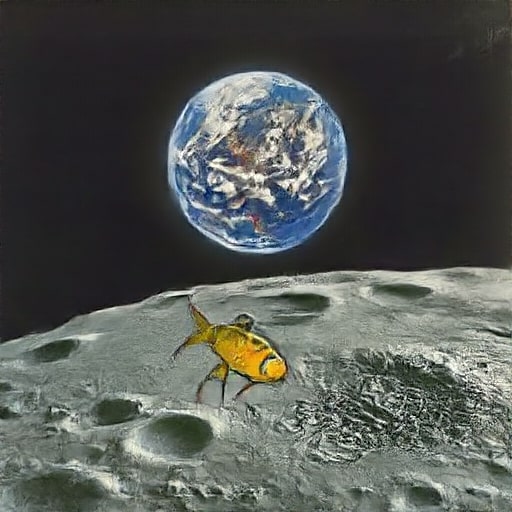}
            \end{subfigure}
            \hspace{2pt}
            \begin{subfigure}[t]{0.09\linewidth}
               \includegraphics[width=\linewidth, height=0.95\linewidth,cfbox=green]{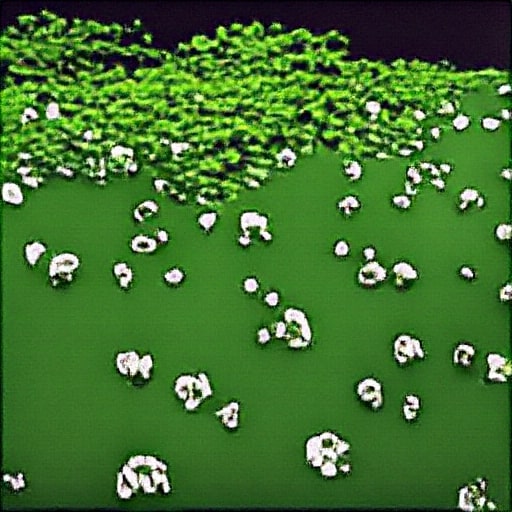}
            \end{subfigure}%            
            \begin{subfigure}[t]{0.09\linewidth}
               \includegraphics[width=\linewidth, height=0.95\linewidth,cfbox=cyan]{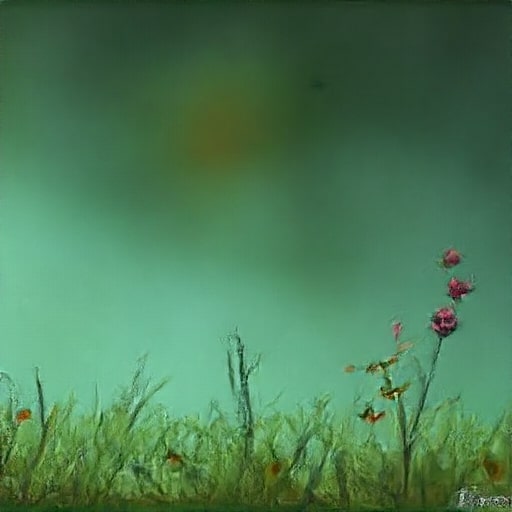}
            \end{subfigure}
            \hspace{2pt}
            \begin{subfigure}[t]{0.09\linewidth}
               \includegraphics[width=\linewidth, height=0.95\linewidth,cfbox=green]{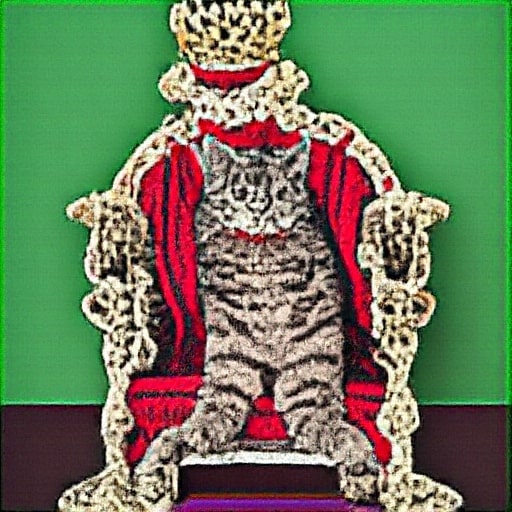}
            \end{subfigure}%            
            \begin{subfigure}[t]{0.09\linewidth}
               \includegraphics[width=\linewidth, height=0.95\linewidth,cfbox=cyan]{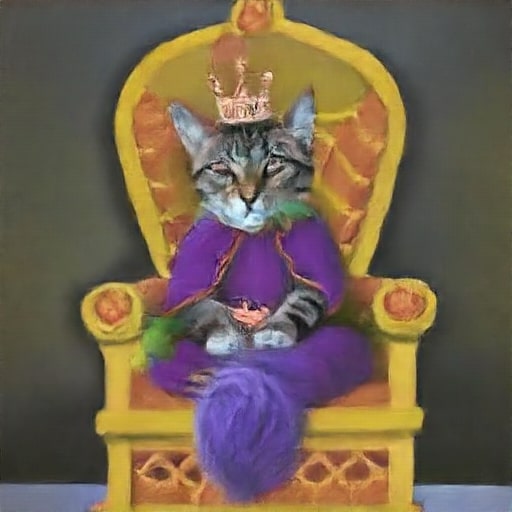}
            \end{subfigure}\\
            \begin{subfigure}[t]{0.09\linewidth}
               \includegraphics[width=\linewidth, height=0.95\linewidth,cfbox=green]{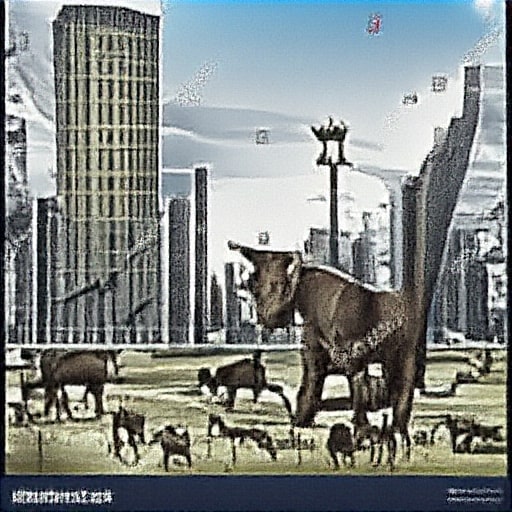}
            \end{subfigure}%            
            \begin{subfigure}[t]{0.09\linewidth}
               \includegraphics[width=\linewidth, height=0.95\linewidth,cfbox=cyan]{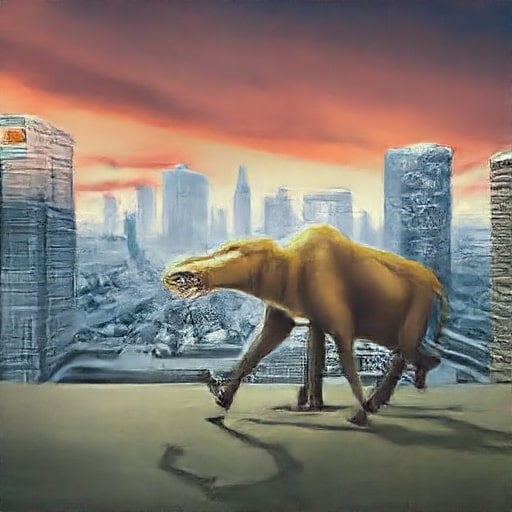}
            \end{subfigure}
            \hspace{2pt}
            \begin{subfigure}[t]{0.09\linewidth}
               \includegraphics[width=\linewidth, height=0.95\linewidth,cfbox=green]{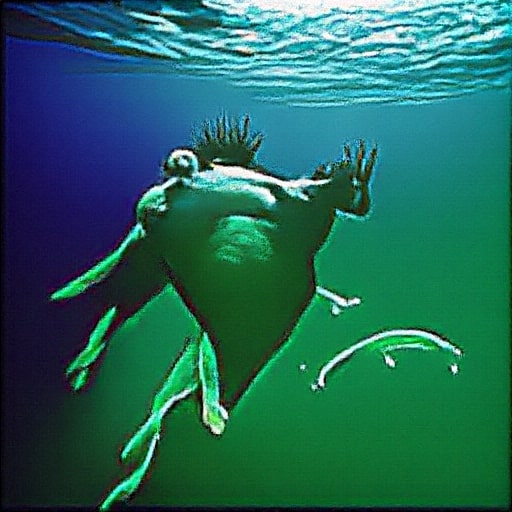}
            \end{subfigure}%            
            \begin{subfigure}[t]{0.09\linewidth}
               \includegraphics[width=\linewidth, height=0.95\linewidth,cfbox=cyan]{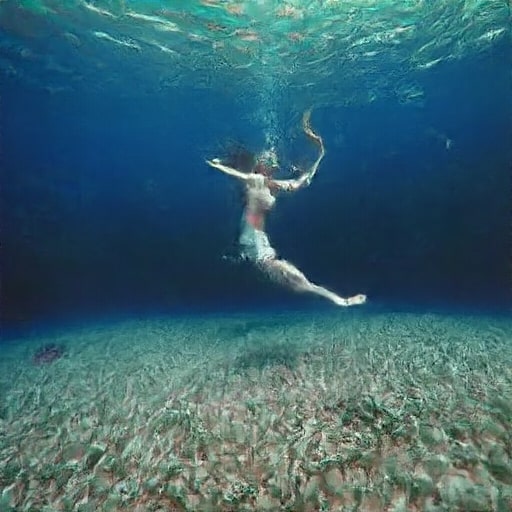}
            \end{subfigure}
            \hspace{2pt}
            \begin{subfigure}[t]{0.09\linewidth}
               \includegraphics[width=\linewidth, height=0.95\linewidth,cfbox=green]{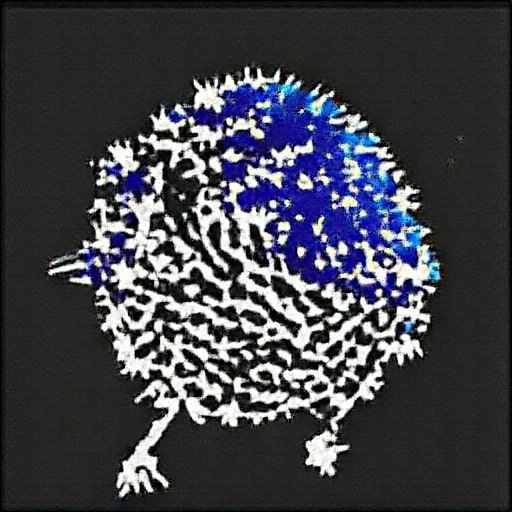}
            \end{subfigure}%            
            \begin{subfigure}[t]{0.09\linewidth}
               \includegraphics[width=\linewidth, height=0.95\linewidth,cfbox=cyan]{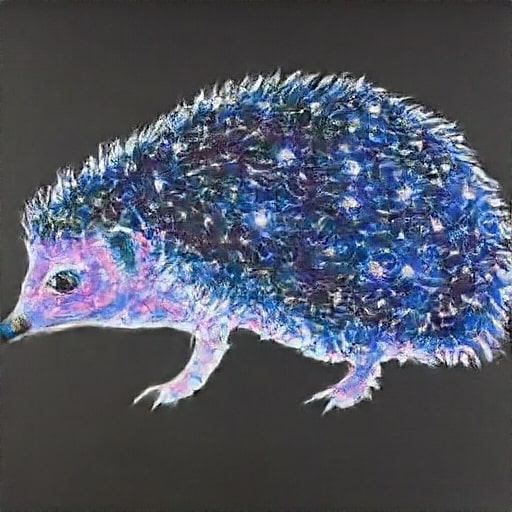}
            \end{subfigure}
            \hspace{2pt}
            \begin{subfigure}[t]{0.09\linewidth}
               \includegraphics[width=\linewidth, height=0.95\linewidth,cfbox=green]{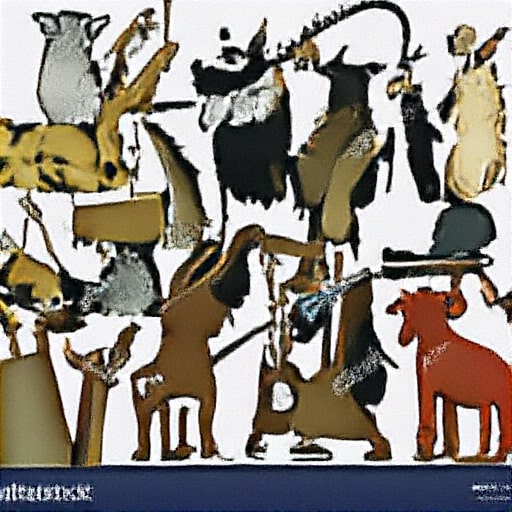}
            \end{subfigure}%            
            \begin{subfigure}[t]{0.09\linewidth}
               \includegraphics[width=\linewidth, height=0.95\linewidth,cfbox=cyan]{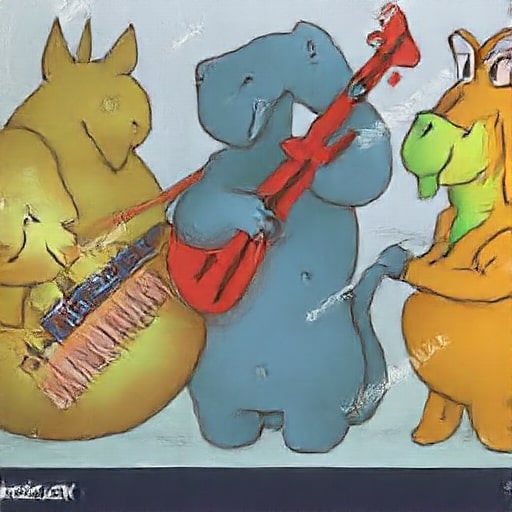}
            \end{subfigure}
            \hspace{2pt}
            \begin{subfigure}[t]{0.09\linewidth}
               \includegraphics[width=\linewidth, height=0.95\linewidth,cfbox=green]{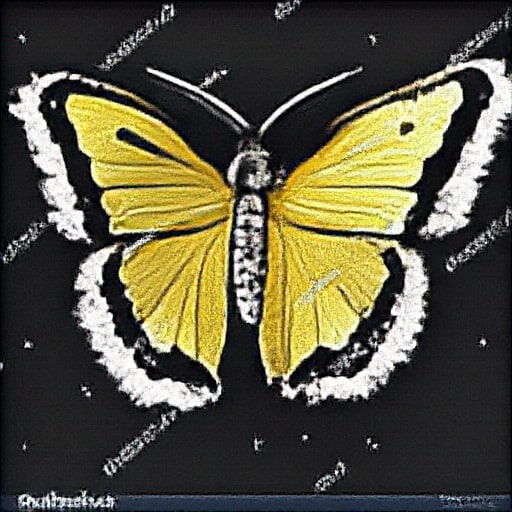}
            \end{subfigure}%            
            \begin{subfigure}[t]{0.09\linewidth}
               \includegraphics[width=\linewidth, height=0.95\linewidth,cfbox=cyan]{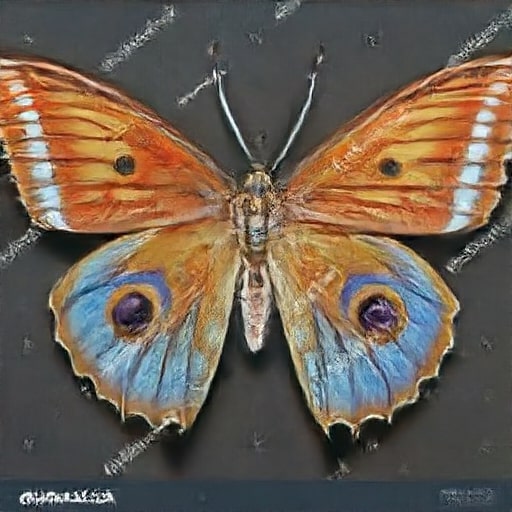}
            \end{subfigure}
    \end{subfigure}
\end{subfigure}
}
\vspace{-.5em}
\caption{On the reuse of the core layers: qualitative results. Finetune (green bounding boxes) $\times$ Frozen results (blue bounding boxes) at $50k$ steps. 
Images at $512\times512 pixels$. Models trained with CC12M. 
Freezing the representation induces objects with better global and parts structure, from the very early steps of training. 
\label{fig:earlysteps_qualitative_50imgs}}
\vspace{-1em}
\end{figure*}

Our qualitative comparison between finetuning and frozen \CC is based on 50 prompts covering different animal species. They are chosen for covering a diverse set of shapes, textures and structures. \autoref{fig:earlysteps_qualitative_50imgs} present a side by side comparison at 50k training steps using the UVIT-Huge model.
Structural elements like legs, wings and trunks are better formed when freezing the pretraing \CC representation. The images were produced with the following list of prompts.

\begin{enumerate}
\itemsep0em 
\scriptsize{
\item "A majestic lion with a flowing mane, basking in the golden African sunset."
\item "A playful dolphin leaping out of the water, glistening with droplets."
\item "A wise old owl perched on a moonlit branch, gazing with piercing yellow eyes."
\item "A colorful macaw soaring through a lush, vibrant rainforest."
\item "A mischievous raccoon rummaging through a trash can in a suburban backyard."
\item "A close-up portrait of a fluffy panda munching on bamboo."
\item "A graceful hummingbird hovering near a bright pink flower."
\item "A herd of elephants silhouetted against a fiery orange sky."
\item "A group of meerkats standing alert in the desert, looking out for danger."
\item "A photorealistic image of a chameleon blending seamlessly with its surroundings."
\item "A Van Gogh-inspired painting of sunflowers with butterflies flitting around them."
\item "A pixel art rendition of a pixelated cat chasing a pixelated mouse."
\item "A watercolor painting of a majestic tiger stalking through a bamboo forest."
\item "A surreal landscape with a melting elephant in the style of Salvador Dalí."
\item "A vibrant pop art image of a zebra with bold stripes and contrasting colors."
\item "A cubist artwork depicting a fragmented and reassembled bear."
\item "A pointillist painting of a turtle, created with tiny dots of color."
\item "A minimalist line drawing of a graceful swan."
\item "A whimsical cartoon illustration of a group of singing frogs in a pond."
\item "A dark and gothic illustration of a raven perched on a skull."
\item "A penguin riding a surfboard on a giant tropical wave."
\item "A giraffe wearing a top hat and monocle, enjoying a cup of tea in a fancy cafe."
\item "A zebra crossing a busy city street at a crosswalk."
\item "A cat wearing a space suit, exploring the surface of Mars."
\item "A monkey DJ mixing beats at a neon-lit dance club."
\item "An octopus painting a self-portrait with its many arms."
\item "A sloth running a marathon, surprisingly outrunning all competitors."
\item "A polar bear relaxing in a hot tub in the middle of the Arctic."
\item "A group of rabbits building a snowman in a winter wonderland."
\item "A dog astronaut floating in space, gazing at the Earth."
\item "A grumpy bulldog wearing a birthday hat and refusing to smile."
\item "A joyful rabbit hopping through a field of wildflowers."
\item "A curious chimpanzee looking intently through a magnifying glass."
\item "A proud peacock displaying its magnificent tail feathers."
\item "A loving mother kangaroo carrying her joey in her pouch."
\item "A mischievous squirrel hiding nuts in a tree trunk."
\item "A sleepy koala clinging to a tree branch, taking a nap."
\item "A determined sea turtle swimming against the ocean current."
\item "A playful wolf pup chasing its own tail."
\item "A group of penguins waddling together in a comical huddle."
\item "A chameleon painted with the vibrant colors of a bustling city skyline." (Imagine a chameleon camouflaged with neon signs and skyscraper patterns.)
\item "A flock of birds forming the shape of a musical note in flight." (Visualize a dynamic dance of birds creating a melody in the sky.)
\item "A fishbowl on the moon, with an astronaut goldfish gazing at Earth." (A whimsical and thought-provoking perspective shift.)
\item"A microscopic landscape teeming with life, where insects are giants and blades of grass are towering trees."
\item "A cat wearing a crown and royal robe, sitting regally on a throne made of yarn balls." (A playful portrait with a touch of humor.)
**\item "A photorealistic image of extinct animals roaming in a modern city landscape." ** (Blend the past and present for a surreal scene.)
\item "An underwater ballet performed by graceful sea creatures." (Capture the beauty and movement of marine life in an artistic way.)
\item "A hedgehog painted as a starry night sky, with its spines representing twinkling stars." (A dreamy fusion of nature and the cosmos.)
\item "Animals playing musical instruments together in a harmonious orchestra." (Imagine the symphony created by a unique animal band.)
\item "A close-up portrait of a butterfly, revealing the intricate patterns and textures on its wings in exquisite detail." (Appreciate the delicate beauty of nature.)
}
\end{enumerate}

\section*{Vermeer distillation: qualitative results}
\label{app:qualitative_vermeer}
% \label{app:distill_qualitative}

\begin{figure*}[t]
\setlength{\fboxsep}{0pt}%
\setlength{\fboxrule}{1pt}%
\centering
\setlength{\lineskip}{1pt}
\begin{subfigure}{0.9\linewidth}
\begin{subfigure}[t]{0.46\linewidth}
   \includegraphics[width=0.5\linewidth, height=0.47\linewidth,cfbox=yellow]{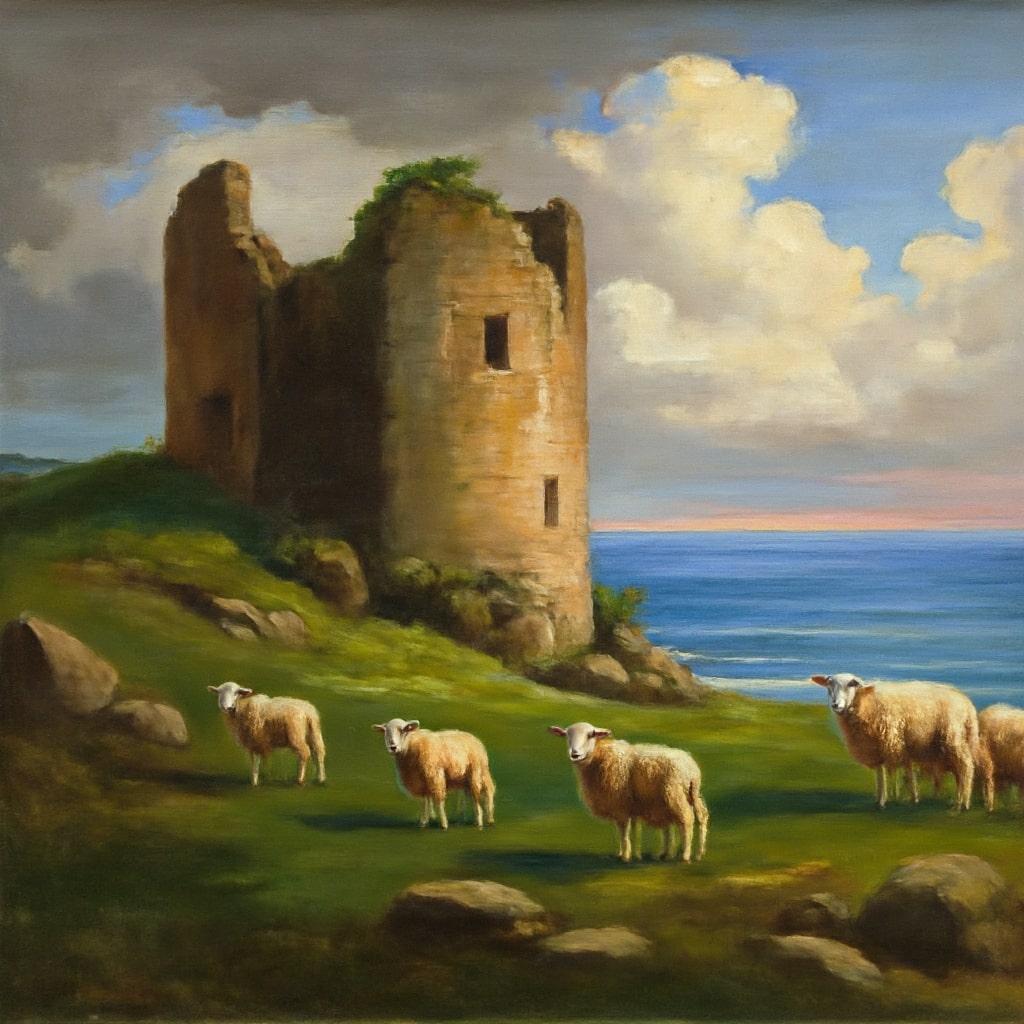}%
   \includegraphics[width=0.5\linewidth, height=0.47\linewidth,cfbox=red]{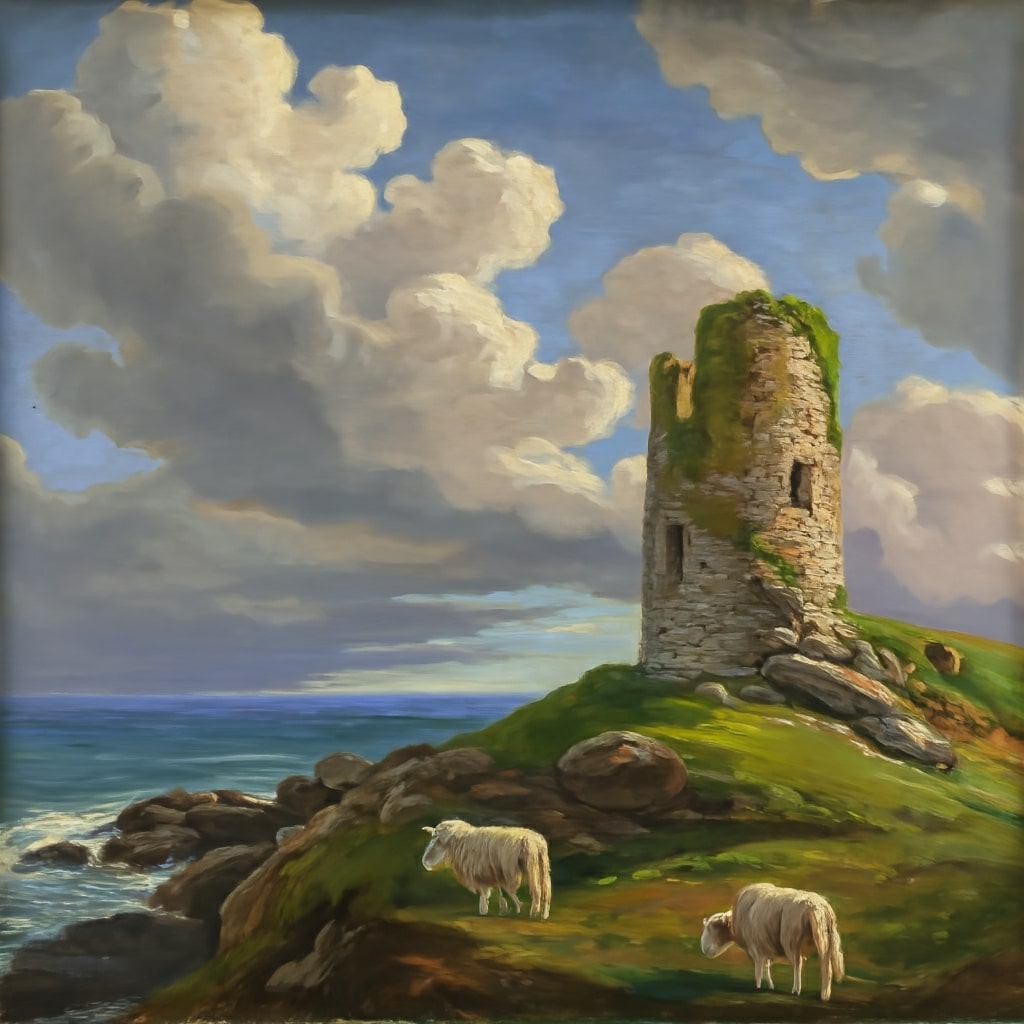}%
\end{subfigure}
\hspace{10pt}
% robot:
\begin{subfigure}[t]{0.46\linewidth}
   \includegraphics[width=0.5\linewidth, height=0.47\linewidth,cfbox=yellow]{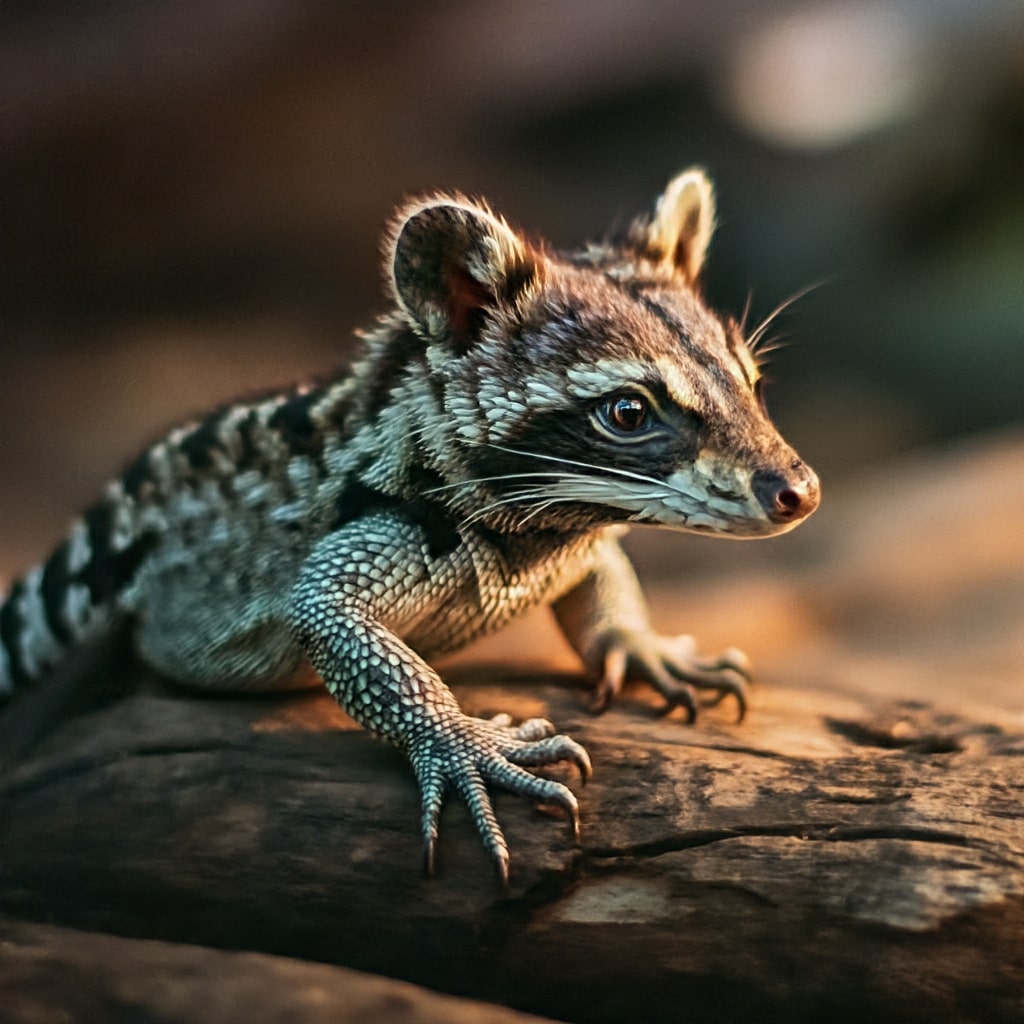}%
   \includegraphics[width=0.5\linewidth, height=0.47\linewidth,cfbox=red]{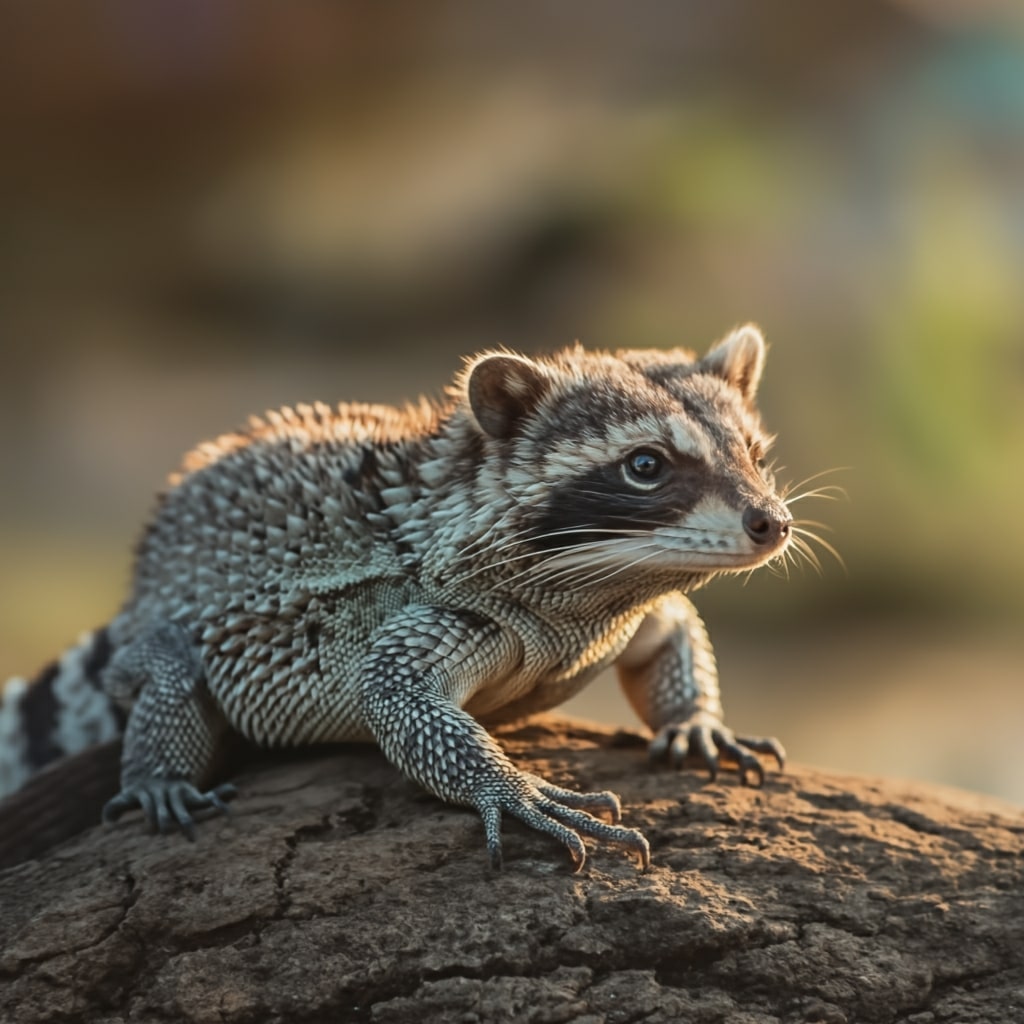}%
\end{subfigure}\\
\begin{subfigure}[t]{0.46\linewidth}%dog:
   \includegraphics[width=0.5\linewidth, height=0.47\linewidth,cfbox=yellow]{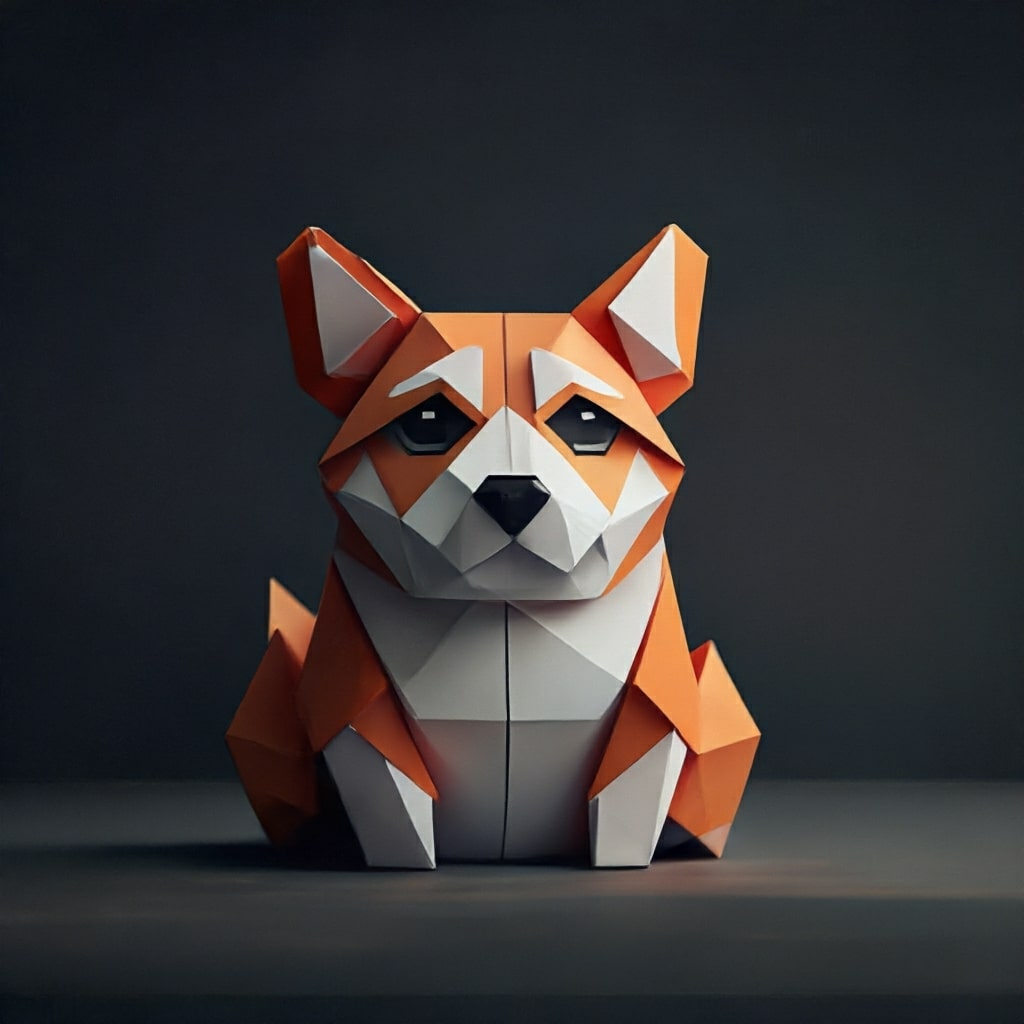}%
   \includegraphics[width=0.5\linewidth, height=0.47\linewidth,cfbox=red]{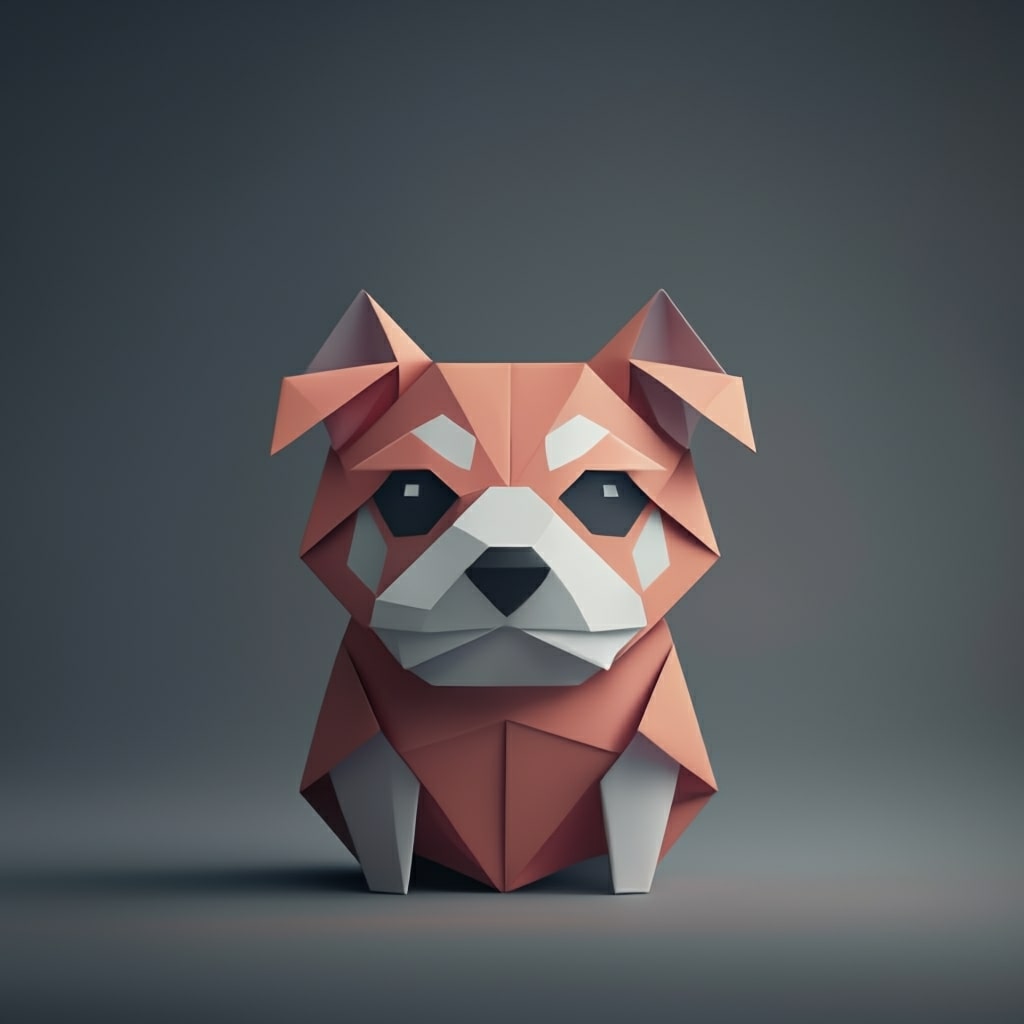}%
\end{subfigure}
\hspace{10pt}
\begin{subfigure}[t]{0.46\linewidth}
   \includegraphics[width=0.5\linewidth, height=0.47\linewidth,cfbox=yellow]{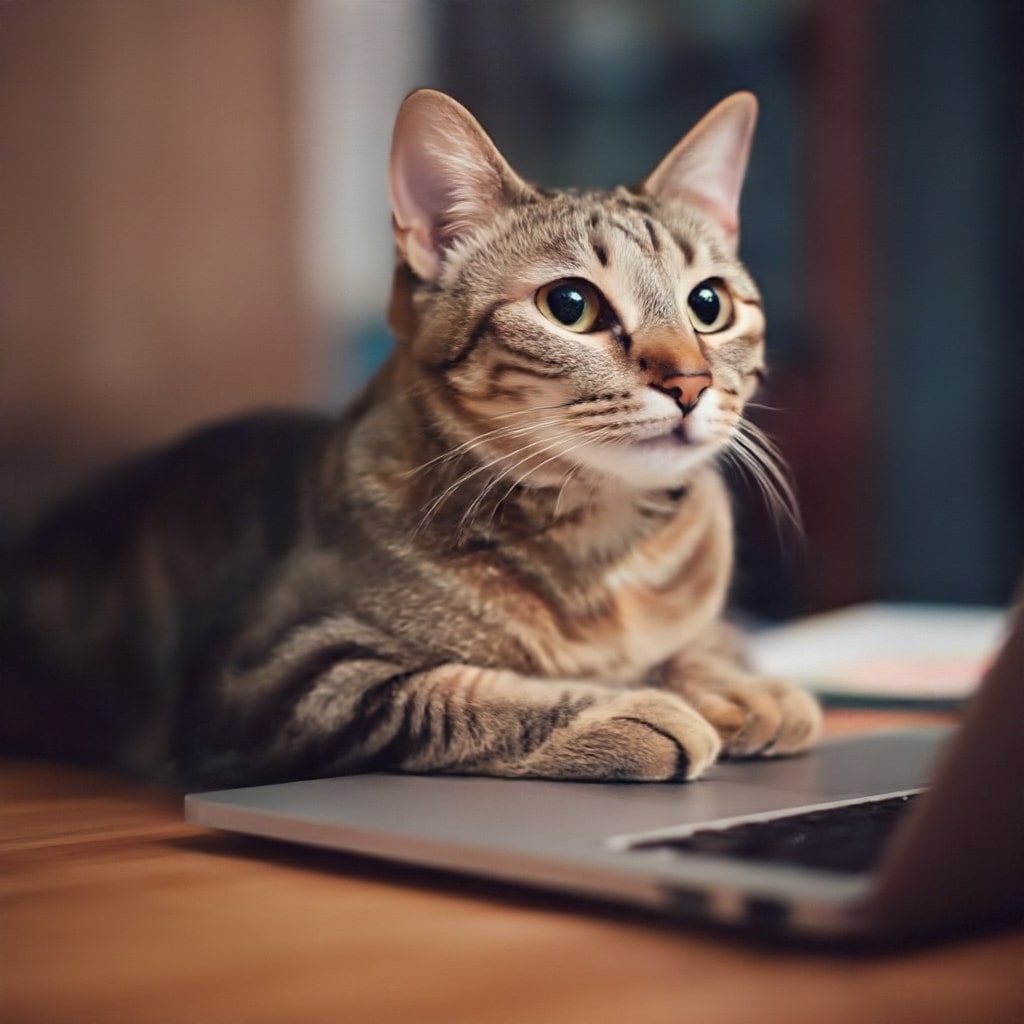}%
   \includegraphics[width=0.5\linewidth, height=0.47\linewidth,cfbox=red]{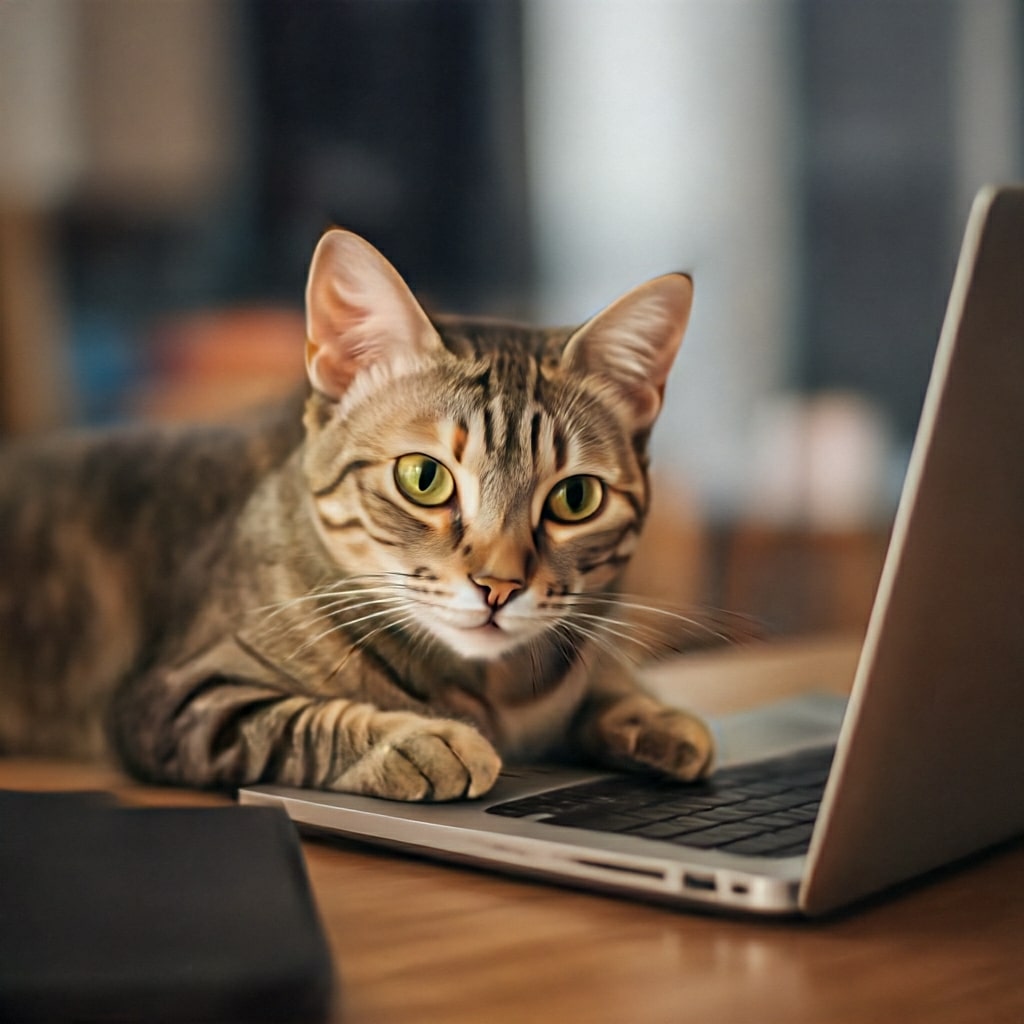}%
\end{subfigure}\\
% coin
% \begin{subfigure}[t]{0.46\linewidth}
%   \includegraphics[width=0.5\linewidth, height=0.47\linewidth,cfbox=yellow]{images/distillation/student/img_0_12.png}%
% \end{subfigure}%            
% \begin{subfigure}[t]{0.46\linewidth}
%   \includegraphics[width=0.5\linewidth, height=0.47\linewidth,cfbox=red]{images/distillation/teacher/img_0_12.png}%
% \end{subfigure}
% \hspace{10pt}
\begin{subfigure}[t]{0.46\linewidth}
   \includegraphics[width=0.5\linewidth, height=0.47\linewidth,cfbox=yellow]{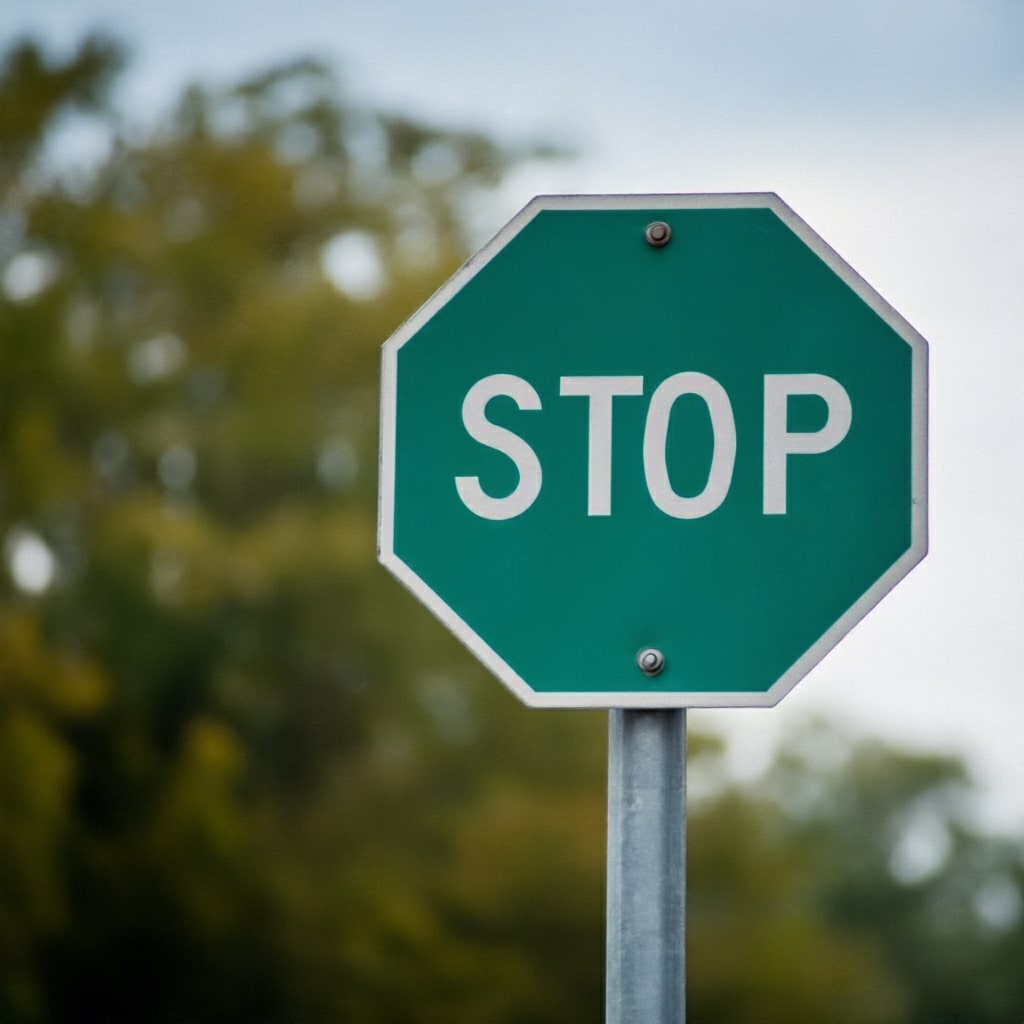}%
   \includegraphics[width=0.5\linewidth, height=0.47\linewidth,cfbox=red]{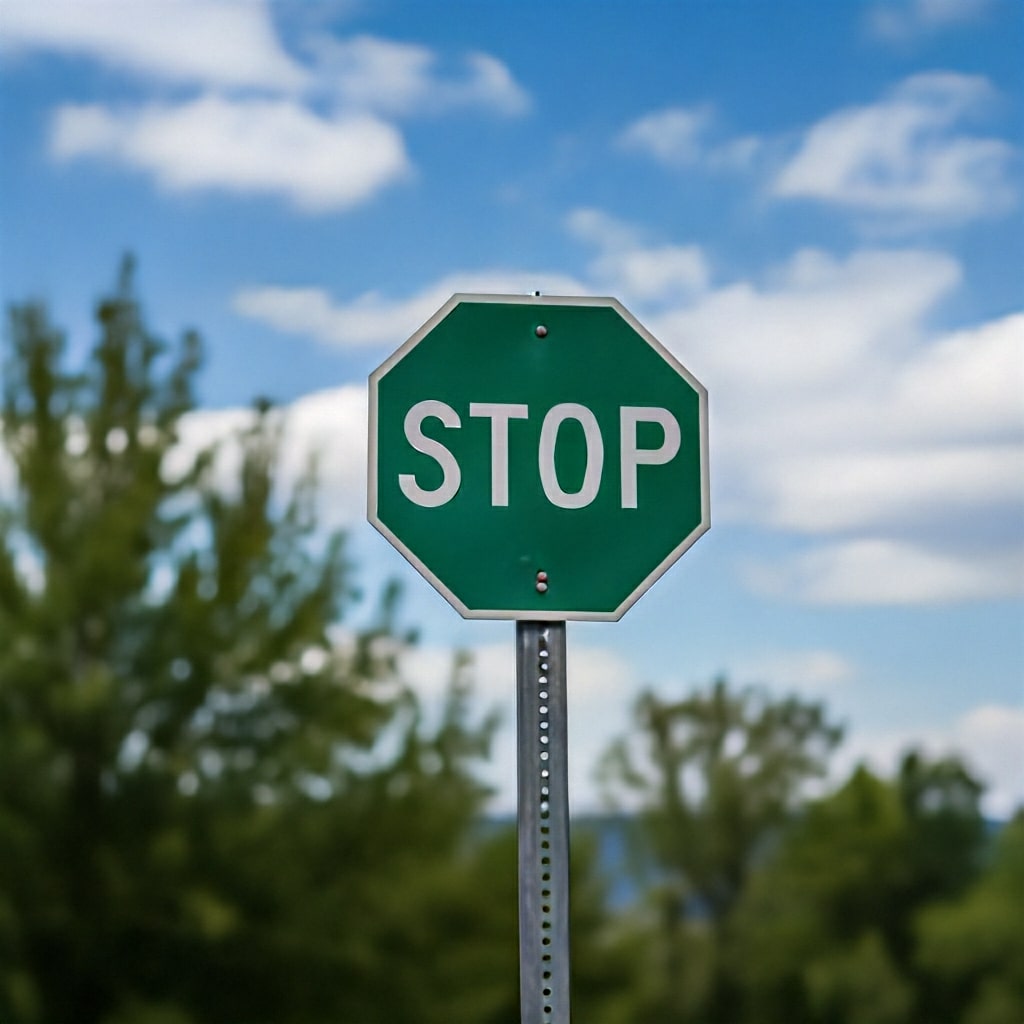}%
\end{subfigure}
\hspace{10pt}
\begin{subfigure}[t]{0.46\linewidth}
   \includegraphics[width=0.5\linewidth, height=0.47\linewidth,cfbox=yellow]{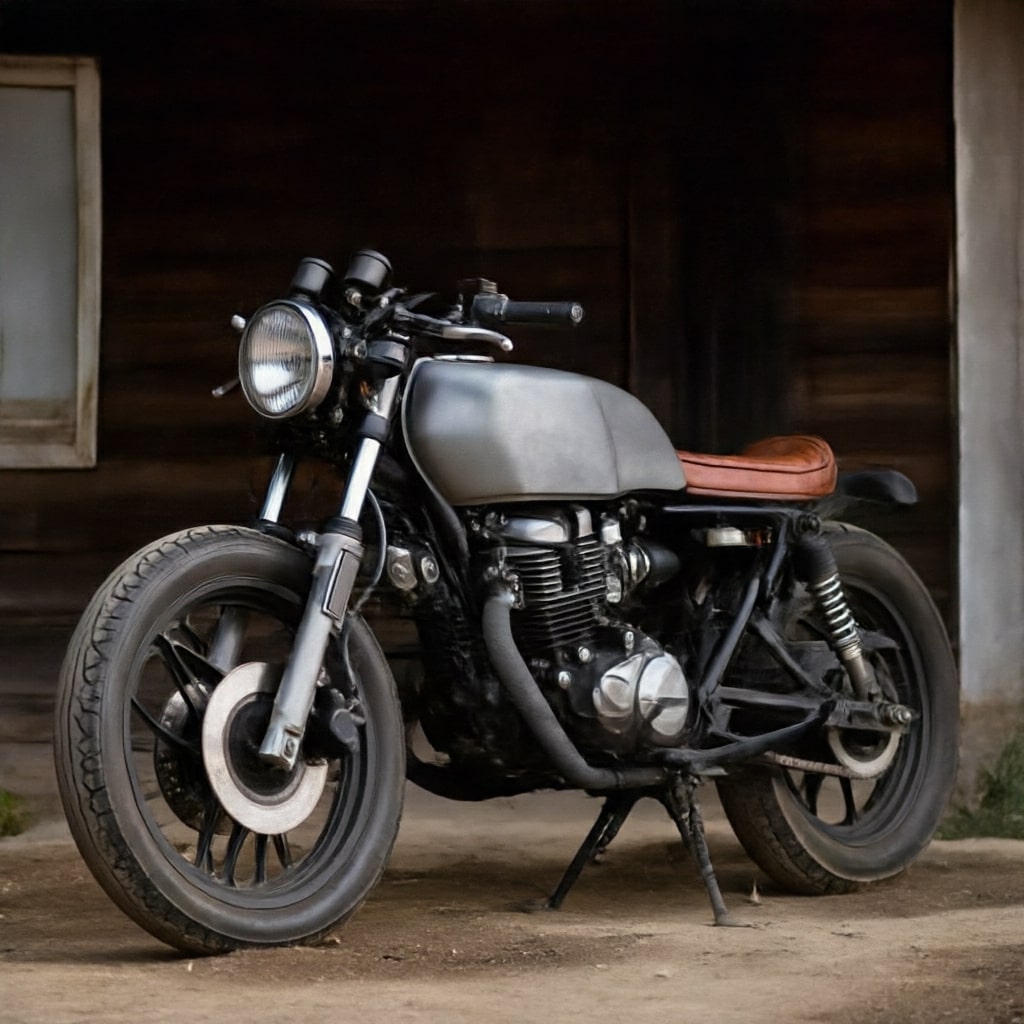}%
   \includegraphics[width=0.5\linewidth, height=0.47\linewidth,cfbox=red]{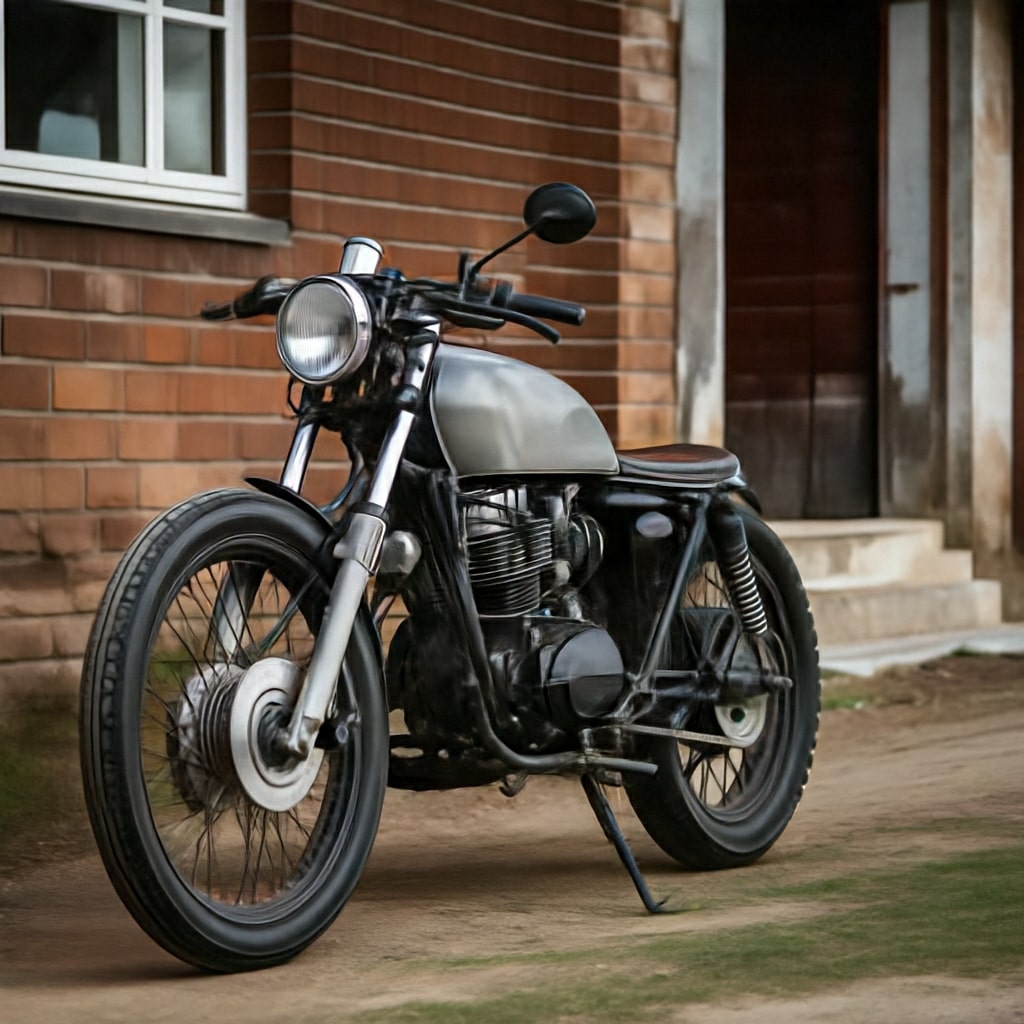}%
\end{subfigure}\\
\begin{subfigure}[t]{0.46\linewidth}
   \includegraphics[width=0.5\linewidth, height=0.47\linewidth,cfbox=yellow]{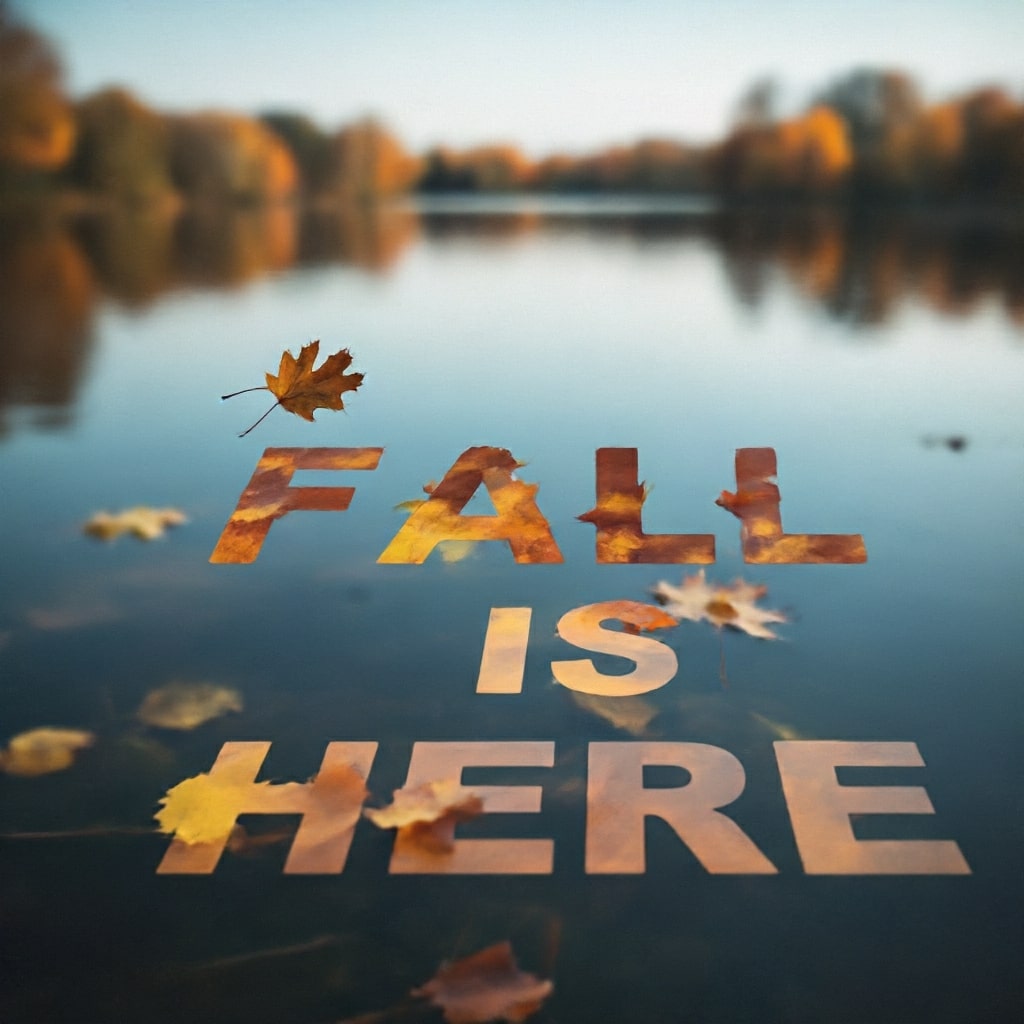}%
   \includegraphics[width=0.5\linewidth, height=0.47\linewidth,cfbox=red]{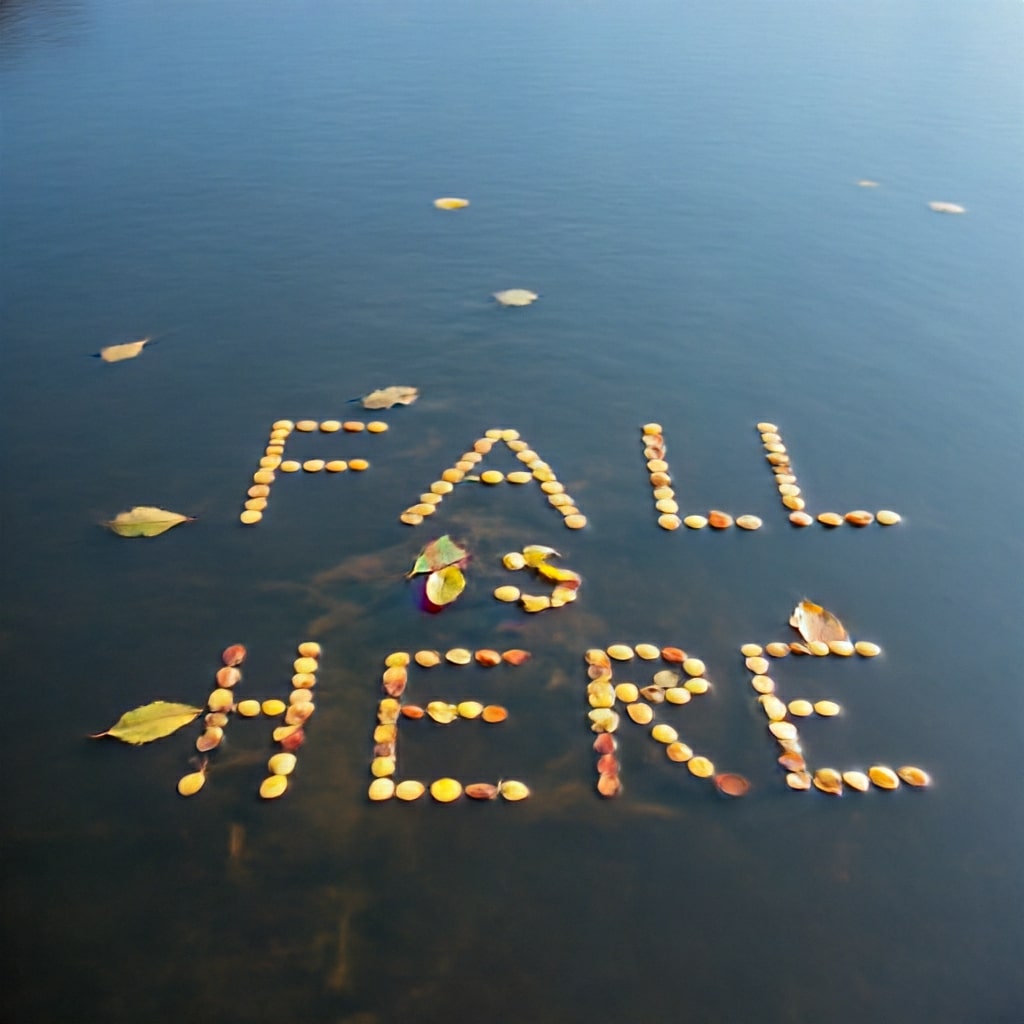}%
\end{subfigure}
\hspace{10pt}
\begin{subfigure}[t]{0.46\linewidth}
   \includegraphics[width=0.5\linewidth, height=0.47\linewidth,cfbox=yellow]{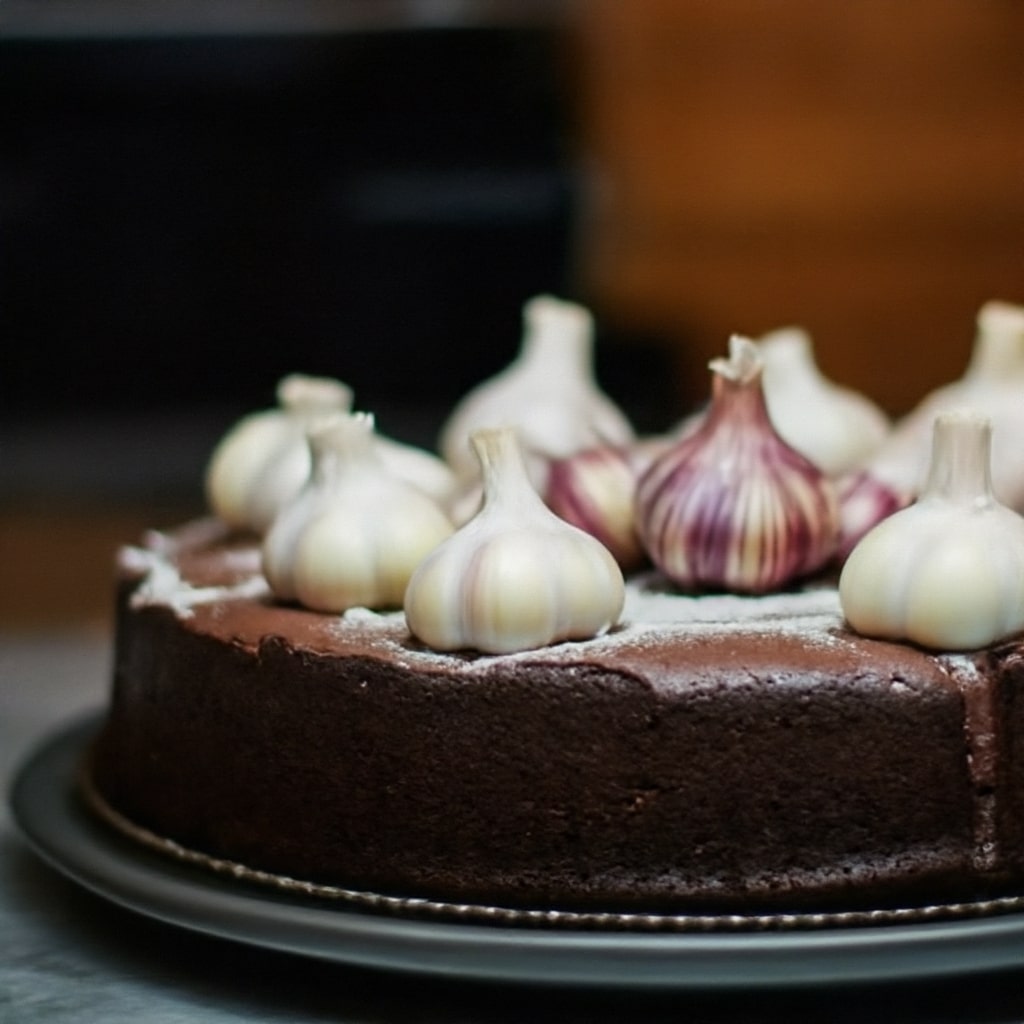}%
   \includegraphics[width=0.5\linewidth, height=0.47\linewidth,cfbox=red]{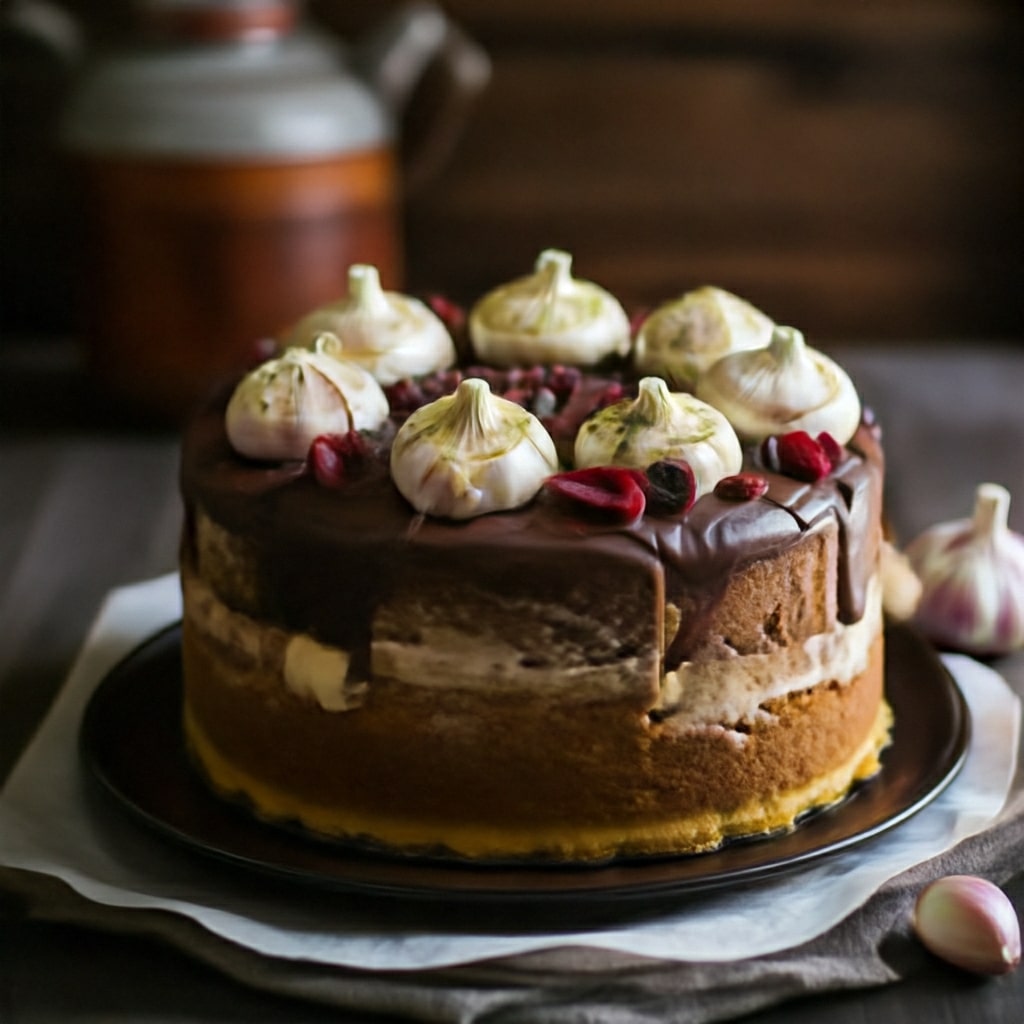}%
\end{subfigure}\\
\begin{subfigure}[t]{0.46\linewidth}
   \includegraphics[width=0.5\linewidth, height=0.47\linewidth,cfbox=yellow]{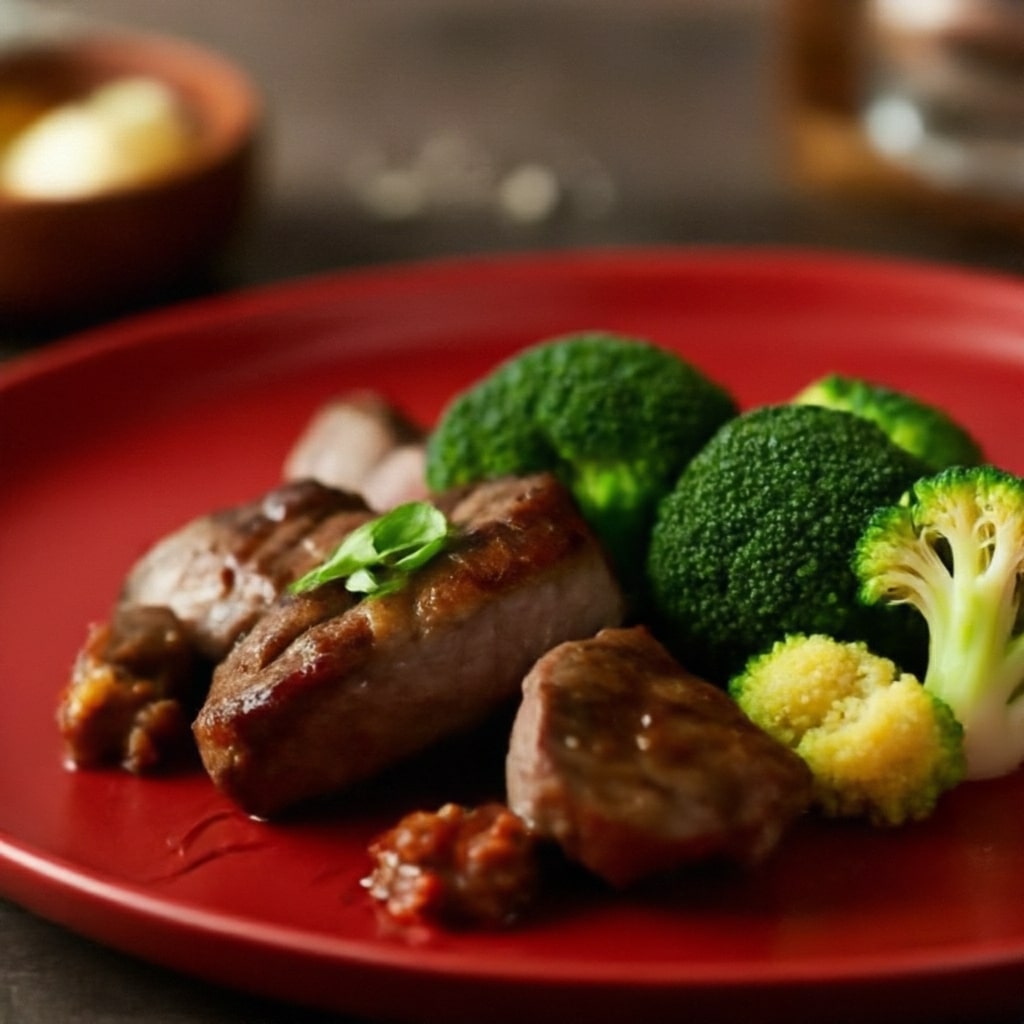}%
   \includegraphics[width=0.5\linewidth, height=0.47\linewidth,cfbox=red]{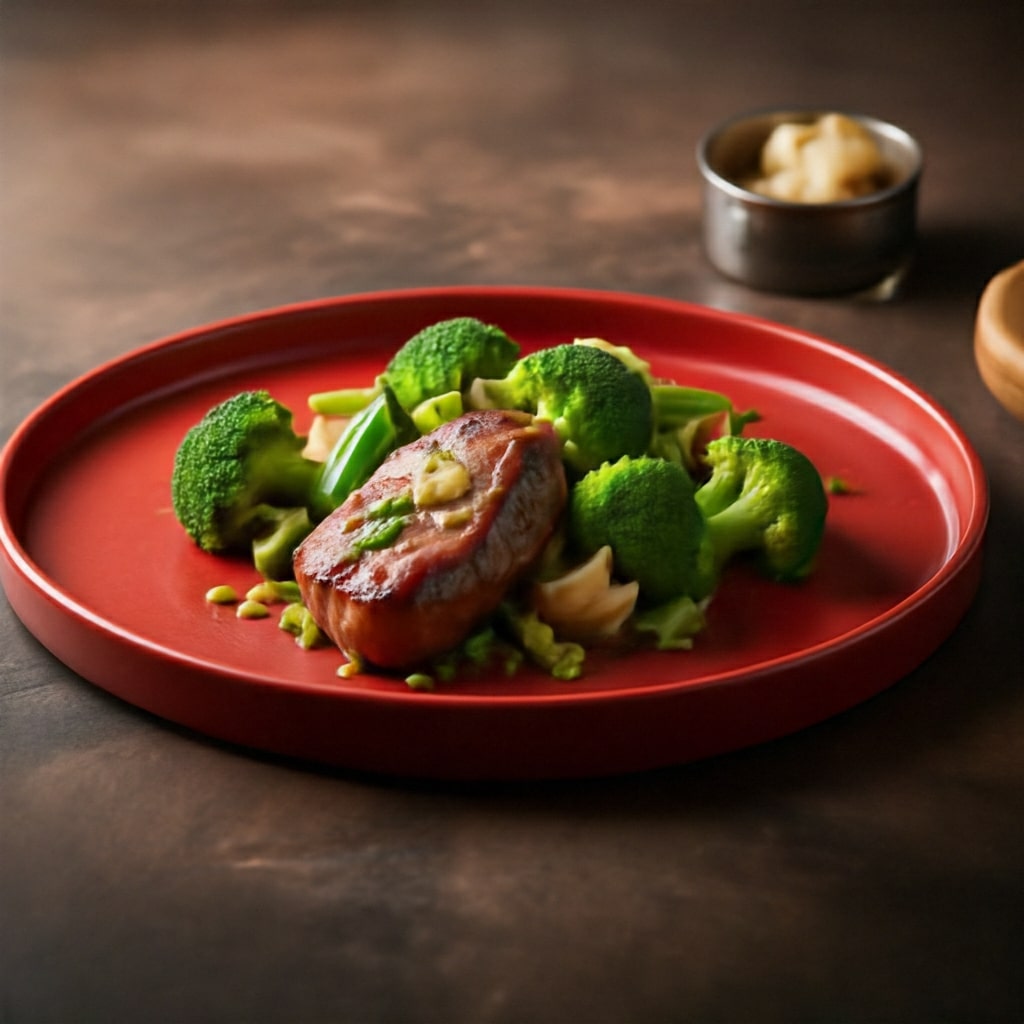}%
\end{subfigure}
\hspace{10pt}
\begin{subfigure}[t]{0.46\linewidth}
   \includegraphics[width=0.5\linewidth, height=0.47\linewidth,cfbox=yellow]{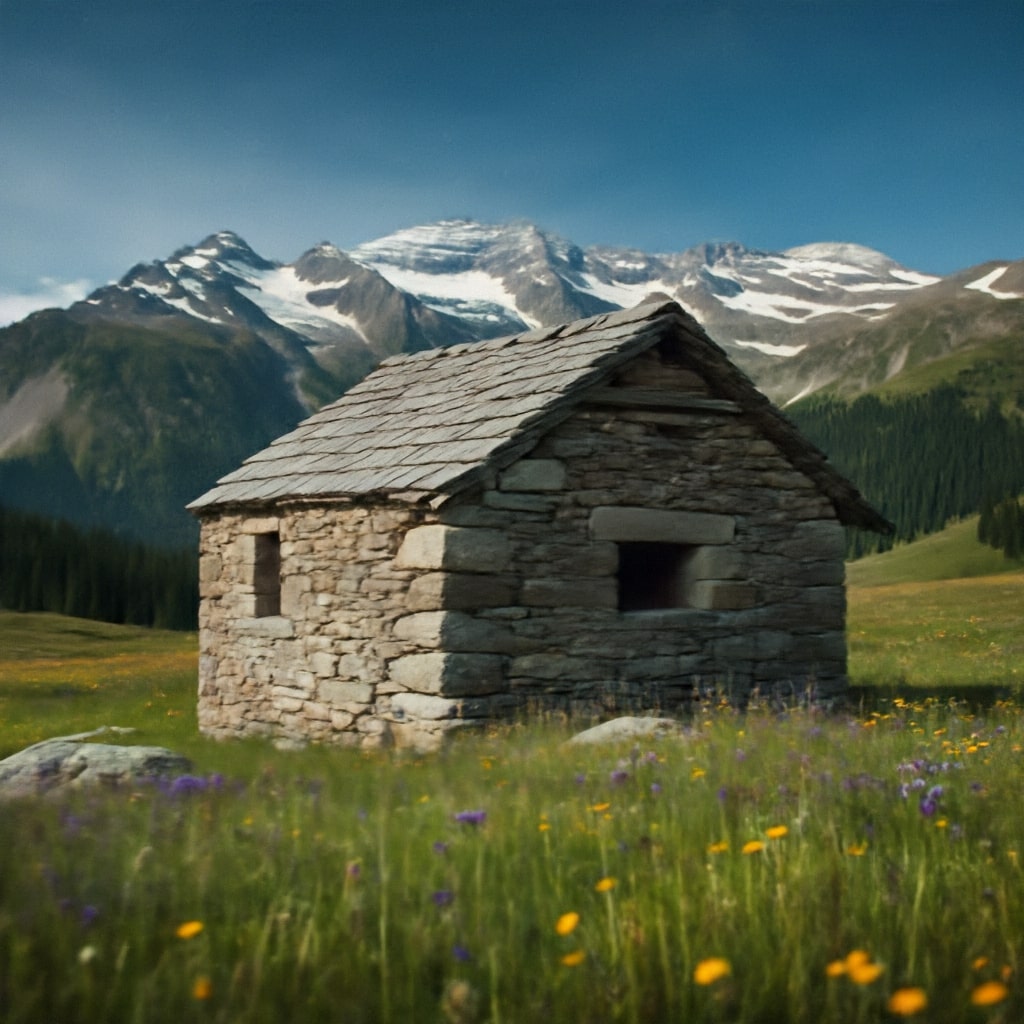}%
   \includegraphics[width=0.5\linewidth, height=0.47\linewidth,cfbox=red]{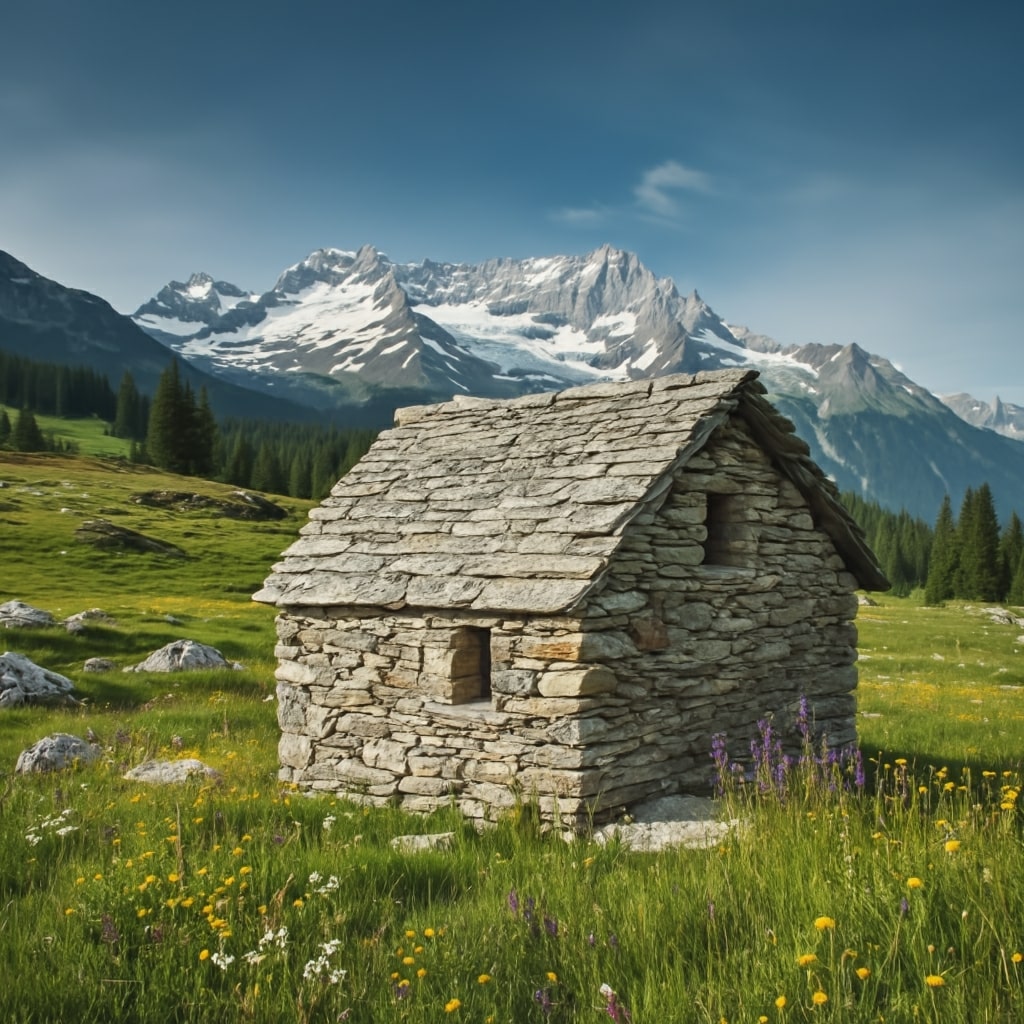}%
\end{subfigure}\\
\begin{subfigure}[t]{0.46\linewidth}
   \includegraphics[width=0.5\linewidth, height=0.47\linewidth,cfbox=yellow]{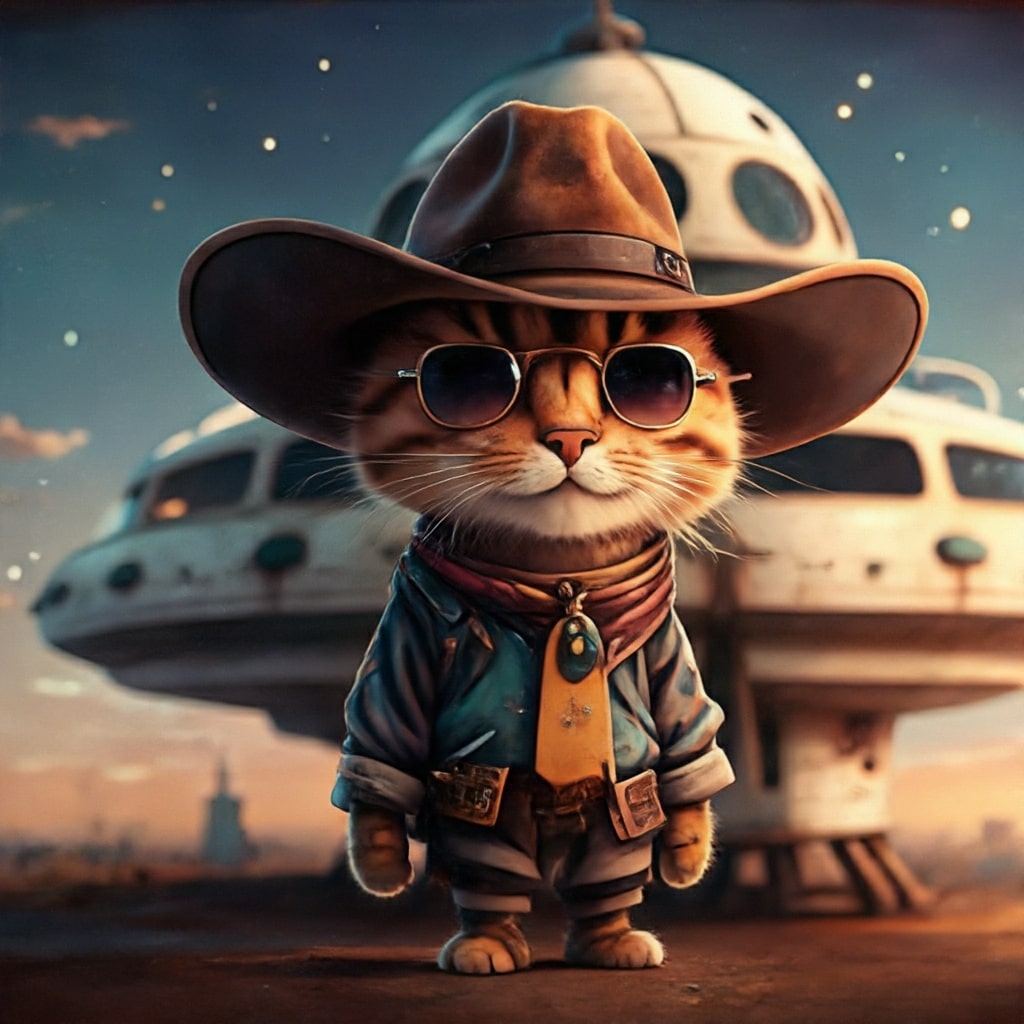}%
   \includegraphics[width=0.5\linewidth, height=0.47\linewidth,cfbox=red]{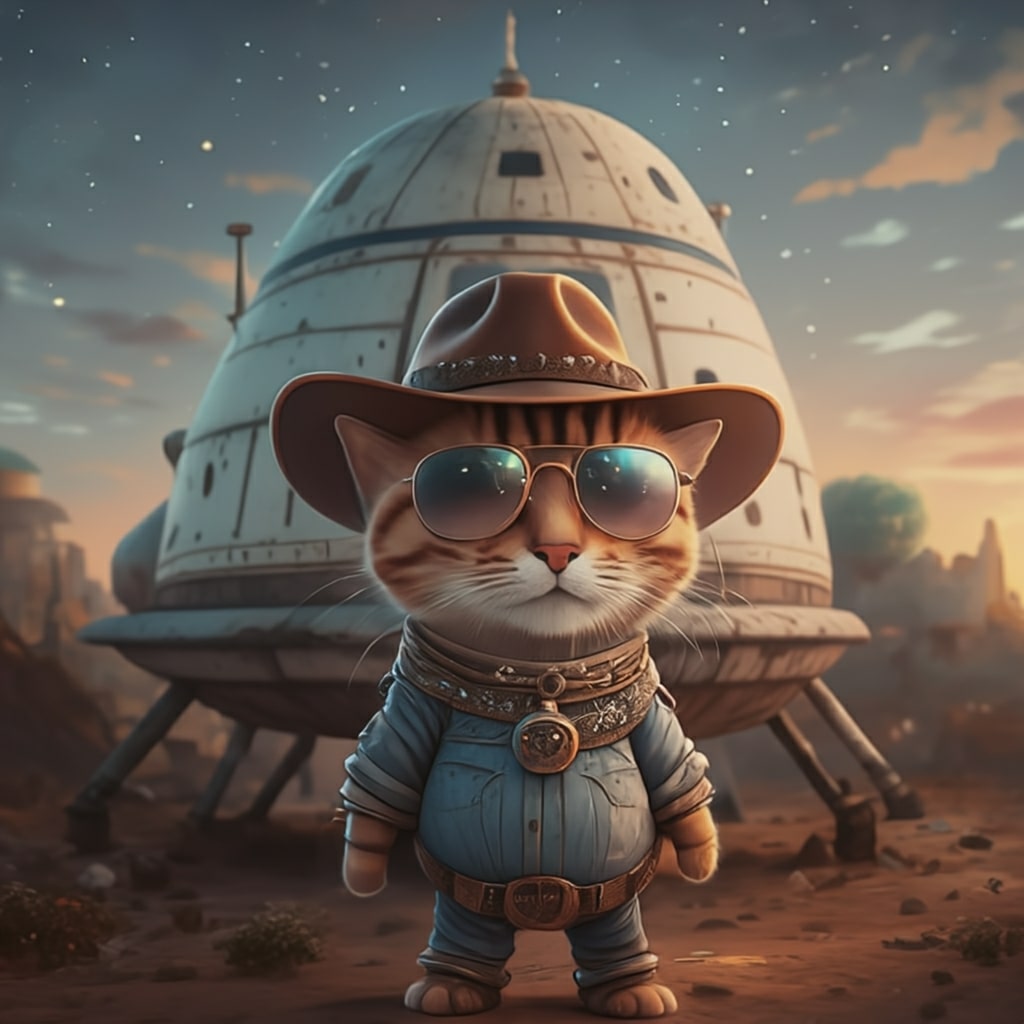}%
\end{subfigure}
\hspace{10pt}
\begin{subfigure}[t]{0.46\linewidth}
  \includegraphics[width=0.5\linewidth, height=0.47\linewidth,cfbox=yellow]{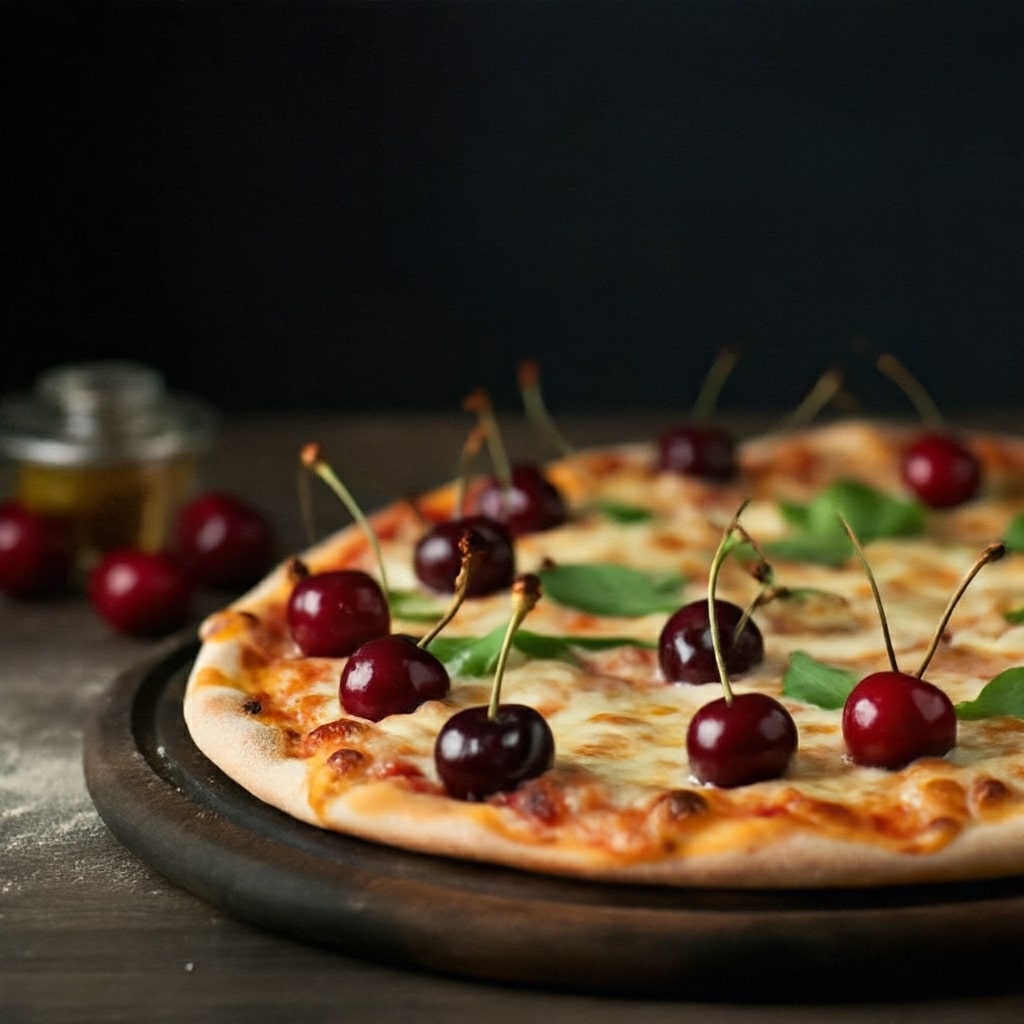}%
  \includegraphics[width=0.5\linewidth, height=0.47\linewidth,cfbox=red]{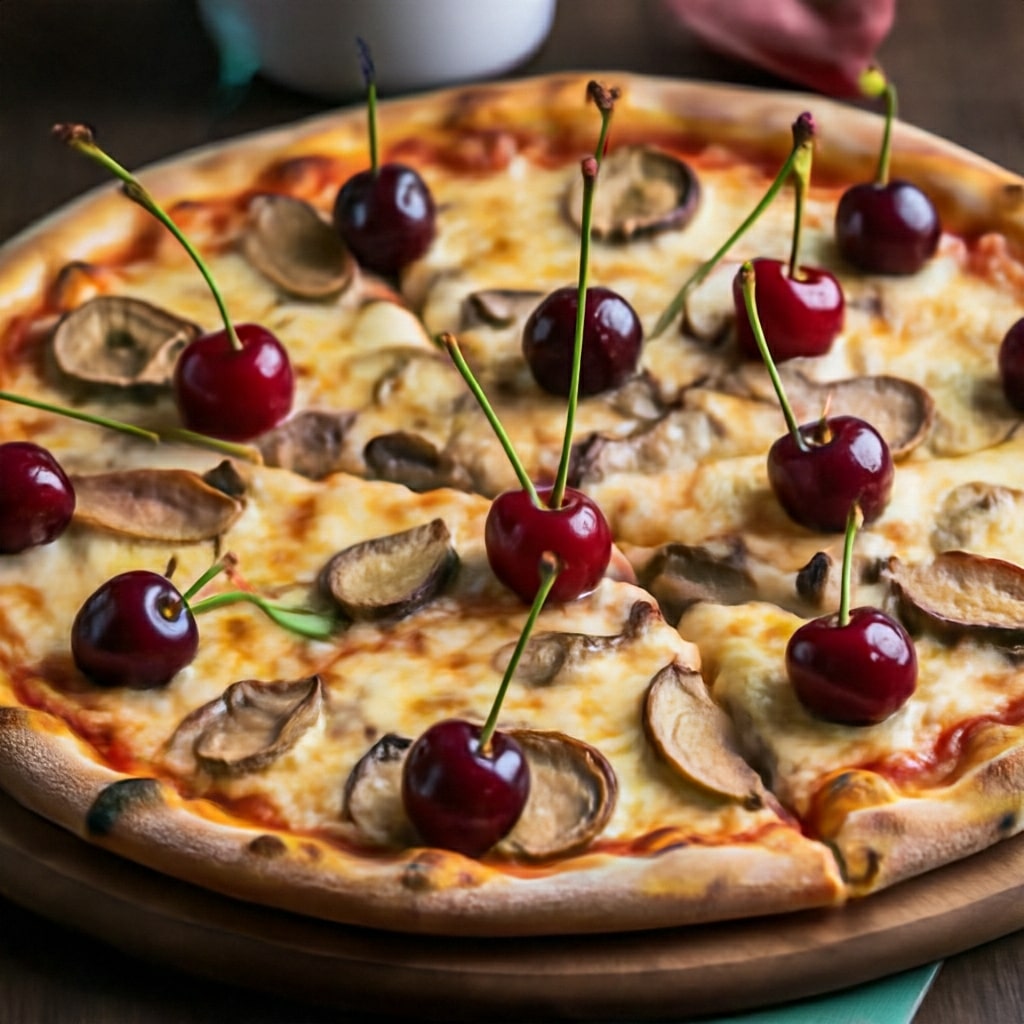}%
\end{subfigure}\\
\end{subfigure}
\vspace{-.5em}
\caption{
\label{fig:vermeer_qualitative}
Qualitative comparison between style-tuned Vermeer using 256 steps (red bounding boxes) and its distilled MCM version using 16 steps (yellow bounding boxes). All images are directly generated at 1024x1024 pixels. }
\vspace{-1em}
\end{figure*}

\autoref{fig:vermeer_qualitative} presents additional qualitative results produced using the Vermeer model and its distilled version.

The images were produced with the following list of prompts.
\begin{enumerate}
\itemsep0em 
\scriptsize{
\item "Ruined circular stone tower on a cliff next to the ocean. Shepherd and sheep on green hillock. Sunrise, big puffy clouds. Naturalistic landscape. Romanticism. Hudson River School. Oil on canvas by Thomas Cole."
\item "Photo of a cute raccoon lizard at sunset, 35mm"
\item "Wallpaper of minimal origami corgi made of multi colored paper, abstract, clean, minimalist, 4K, 8K, soft colors, high definition."
\item "A cat lying a top on the desk on a laptop."
\item "A green stop sign on a pole."
\item "A grey motorcycle on dirt road next to a building."
\item "'Fall is here' written in autumn leaves floating on a lake."
\item "A cake topped with whole bulbs of garlic"
\item "A red plate topped with broccoli, meat and veggies."
\item "A photorealistic image of a chameleon blending seamlessly with its surroundings."
\item "A cat wearing a cowboy hat and sunglasses and standing in front of a rusty old white spaceship at sunrise. Pixar cute. Detailed anime illustration."
\item "A pizza with cherry toppings"
}
\end{enumerate}

\end{document}